\documentclass{article}

\PassOptionsToPackage{numbers, compress}{natbib}

\usepackage{booktabs}       
\usepackage{amsfonts}       
\usepackage{nicefrac}       
\usepackage{microtype}      
\usepackage{subfigure}
\usepackage{graphicx}
\usepackage{amsmath}
\usepackage{multirow}
\usepackage{dblfloatfix}
\usepackage{xcolor}
\usepackage{float}
\usepackage{color}

\usepackage[final]{nips_2018}

\usepackage[utf8]{inputenc} 
\usepackage[T1]{fontenc}    
\usepackage[hidelinks]{hyperref}    
\usepackage{url}            
\usepackage{booktabs}       
\usepackage{amsfonts}       
\usepackage{nicefrac}       
\usepackage{microtype}      

\title{Visualizing the Loss Landscape of Neural Nets}

\newcommand*{\affaddr}[1]{#1}
\newcommand*{\affmark}[1][*]{\textsuperscript{\normalfont#1}}

\author{
  Hao Li\affmark[1], Zheng Xu\affmark[1], Gavin Taylor\affmark[2], Christoph Studer\affmark[3], Tom Goldstein\affmark[1]\\
  \affaddr{\affmark[1]University of Maryland, College Park}
  \affaddr{\affmark[2]United States Naval Academy}
  \affaddr{\affmark[3]Cornell University}\\
  \texttt{\{haoli,xuzh,tomg\}@cs.umd.edu}, \texttt{taylor@usna.edu}, \texttt{studer@cornell.edu}\\
}

\begin{document}

\maketitle

\begin{abstract}
Neural network training relies on our ability to find ``good'' minimizers of highly non-convex loss functions.
It is well-known that certain network architecture designs (e.g., skip connections) produce loss functions that train easier, and well-chosen training parameters (batch size, learning rate, optimizer) produce minimizers that generalize better.
However, the reasons for these differences, and their effects on the underlying loss landscape, are not well understood.
In this paper, we explore the structure of neural loss functions, and the effect of loss landscapes on generalization, using a range of visualization methods.
First, we introduce a simple ``filter normalization'' method that helps us visualize loss function curvature and make meaningful side-by-side comparisons between loss functions.
Then, using a variety of visualizations, we explore how network architecture affects the loss landscape, and how training parameters affect the shape of minimizers.
\end{abstract}

\section{Introduction}
Training neural networks requires minimizing a high-dimensional non-convex
loss function -- a task that is hard in theory, but sometimes easy in
practice.  Despite the NP-hardness of training general neural loss functions
\citep{blum1989training},  simple gradient methods often find global
minimizers (parameter configurations with zero or near-zero training loss), even when data
and labels are randomized before training \citep{zhang2016understanding}.
However, this good behavior is not universal; the trainability of neural nets
is highly dependent on network architecture design choices, the choice of
optimizer, variable initialization, and a variety of other considerations.
Unfortunately, the effect of each of these choices on the structure of the underlying loss surface is
unclear.
Because of the prohibitive cost of loss function evaluations (which requires looping over all the data points in the training set), studies in this field have remained predominantly theoretical.
\begin{figure}[H]
\centering
\subfigure[without skip connections]{\includegraphics[width=0.49\linewidth]{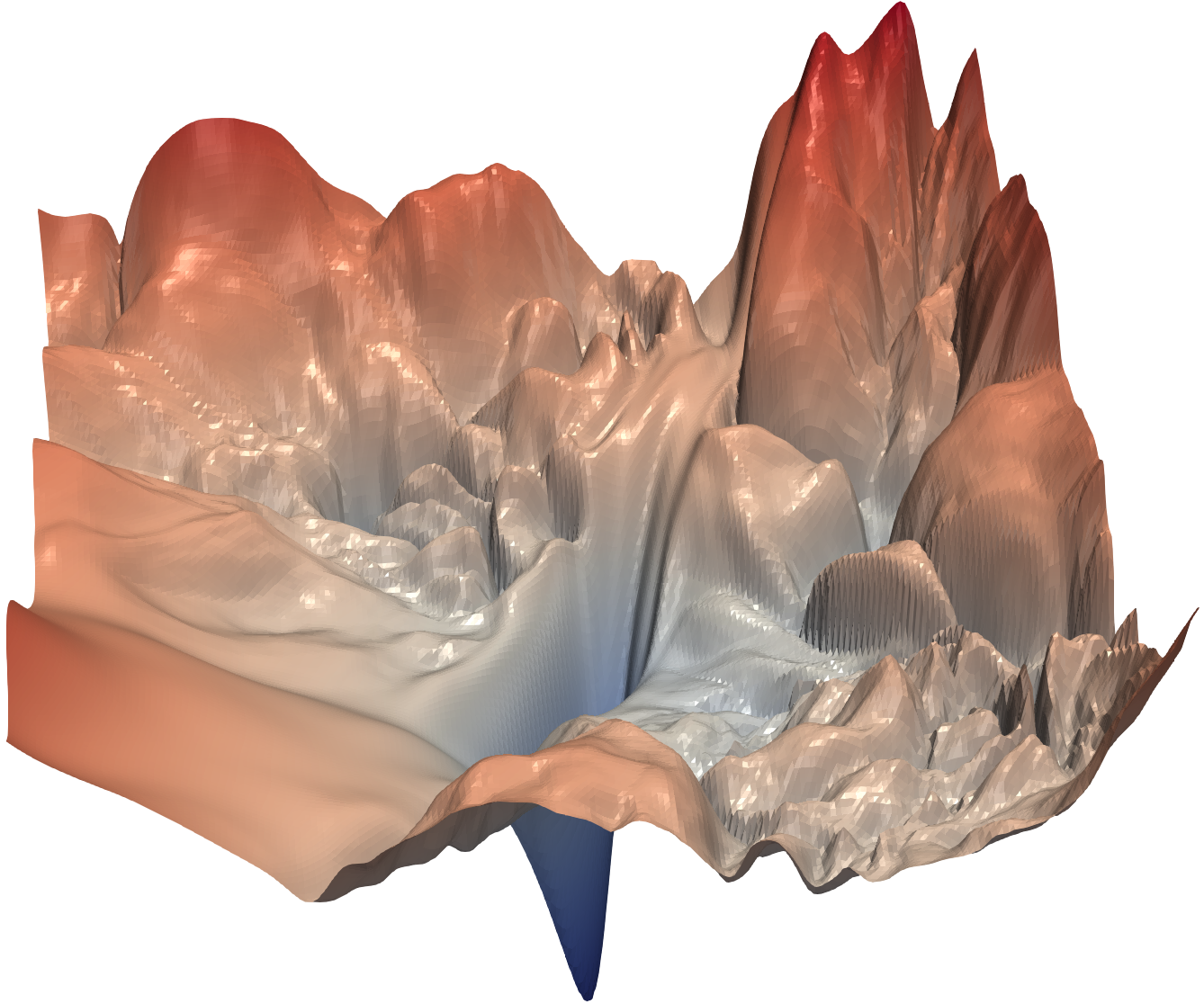}}
\hspace{1.5mm}\subfigure[with skip connections]{\includegraphics[width=0.49\linewidth]{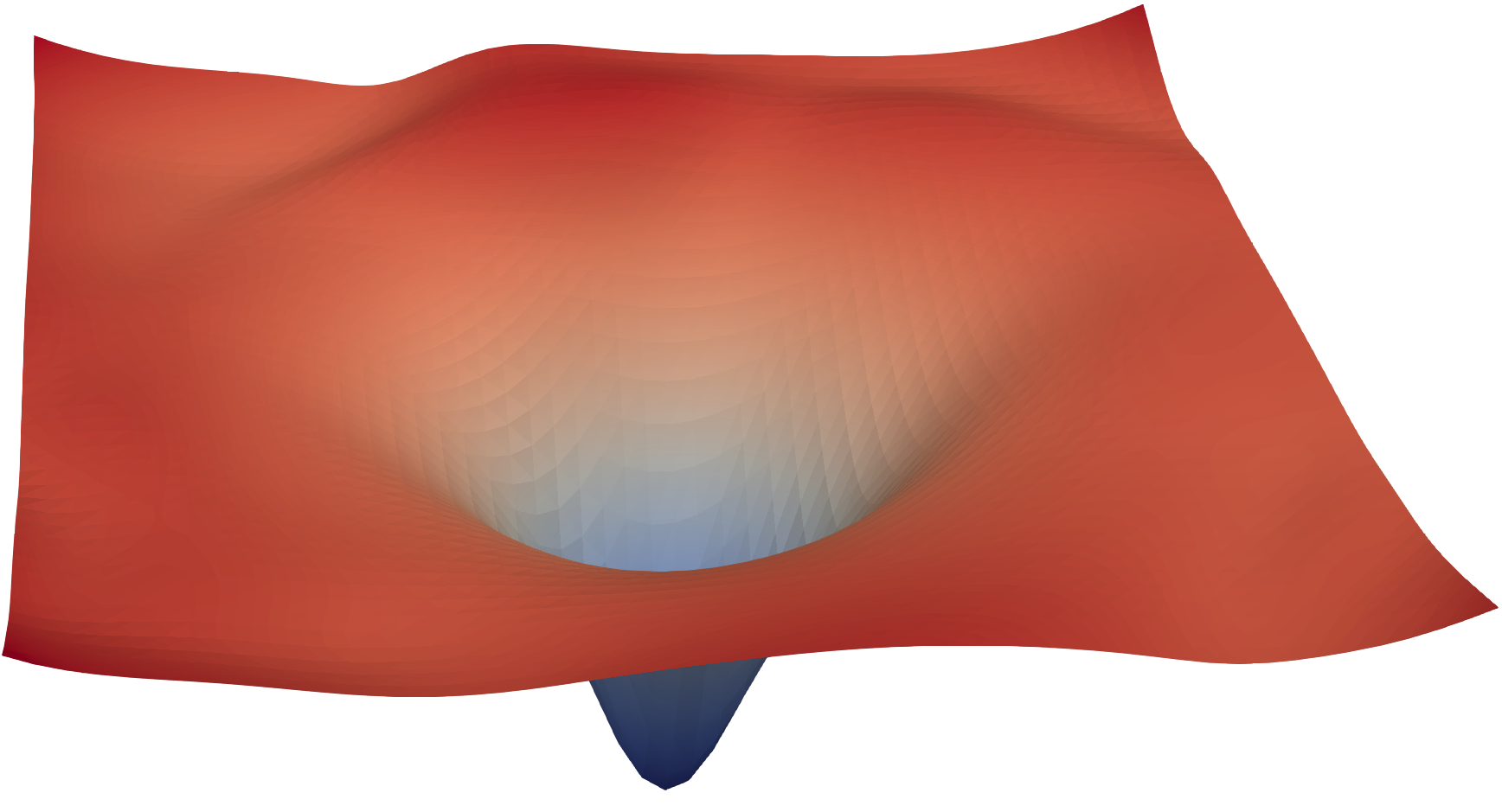}}
\caption{The loss surfaces of ResNet-56 with/without skip connections.
The proposed filter normalization scheme is used to enable comparisons of sharpness/flatness between the two figures.
}
\vspace{-4mm}
\label{fig:surfs}
\end{figure}

Visualizations have the potential to help us answer several important questions about why neural networks work.
In particular, why are we able to minimize highly non-convex neural loss functions?
And why do the resulting minima generalize?
To clarify these questions,
we use high-resolution visualizations to provide an empirical
characterization of neural loss functions, and explore how different
network architecture choices affect the loss landscape.
Furthermore, we explore how the non-convex structure of neural loss functions relates to their
trainability, and how the geometry of neural minimizers (i.e., their
sharpness/flatness, and their surrounding landscape), affects their
generalization properties.

To do this in a meaningful way, we propose a simple ``filter normalization'' scheme that enables us to do side-by-side comparisons of different minima found during training.
We then use visualizations to explore sharpness/flatness of minimizers found by different methods, as well as the effect of network architecture choices (use of skip connections, number of filters, network depth) on the loss landscape.
Our goal is to understand how loss function geometry affects generalization in neural nets.

\subsection{Contributions}
We study methods for producing meaningful loss function visualizations. Then, using these visualization methods, we explore how loss landscape geometry affects generalization error and trainability.  More specifically, we address the following issues:
\begin{itemize}
\item We reveal faults in a number of visualization methods for loss functions, and show that simple visualization strategies fail to accurately capture the local geometry (sharpness or flatness) of loss function minimizers.
\item We present a simple visualization method based on ``filter normalization.''
The sharpness of minimizers correlates well with generalization error when this normalization is used, even when making comparisons across disparate network architectures and training methods.
This enables side-by-side comparisons of different minimizers\footnote{Code and plots are available at \url{https://github.com/tomgoldstein/loss-landscape}}.
\item We observe that, when networks become sufficiently deep, neural loss landscapes quickly transition from being nearly convex to being highly chaotic.  This transition from convex to chaotic behavior
coincides with a dramatic drop in generalization error, and ultimately to a lack of trainability.
\item We observe that skip connections promote flat minimizers and prevent the transition to chaotic behavior, which helps explain why skip connections are necessary for training extremely deep networks.
\item We quantitatively measure non-convexity by calculating the smallest (most negative) eigenvalues of the Hessian around local minima, and visualizing the results as a heat map.
\item We study the visualization of SGD optimization trajectories.  We explain the difficulties that arise when visualizing these trajectories, and show that optimization trajectories lie in an extremely low dimensional space.  This low dimensionality can be explained by the presence of large, nearly convex regions in the loss landscape, such as those observed in our 2-dimensional visualizations.
\end{itemize}

\section{Theoretical Background}
\label{sec:related_work}
Numerous theoretical studies have been done on our ability to optimize neural loss function~\citep{dauphin2014identifying,choromanska2015loss}.
Theoretical results usually make restrictive assumptions about the sample distributions, non-linearity of the architecture, or loss functions~\citep{hardt2016identity,nguyen2017loss,yun2017global,soudry2017exponentially,freeman2016topology,xie2017diverse}.
For restricted network classes, such as those with a single hidden layer, globally optimal or near-optimal solutions can be found by common optimization methods~\citep{soltanolkotabi2017theoretical,li2017convergence,tian2017analytical}.
For networks with specific structures, there likely exists a monotonically decreasing path from an initialization to a global minimum~\citep{safran2016quality, haeffele2017global}.
\citet{swirszcz2016local} show counterexamples that achieve ``bad'' local minima for toy problems.

Several works have addressed the relationship between sharpness/flatness of local minima and their generalization ability.
\citet{hochreiter1997flat} defined ``flatness'' as the size of the connected region around the minimum where the training loss remains low.
\citet{keskar2016large} characterize flatness using eigenvalues of the Hessian, and propose $\epsilon$-sharpness as an approximation, which looks at the maximum loss in a neighborhood of a minimum.
\citet{dinh2017sharp,neyshabur2017exploring} show that these quantitative measure of sharpness are not invariant to symmetries in the network, and are thus not sufficient to determine generalization ability.
\citet{entropysgd} used local entropy as a measure of sharpness, which is invariant to the simple transformation in \cite{dinh2017sharp}, but difficult to accurately compute.
\citet{dziugaite2017computing} connect sharpness to PAC-Bayes bounds for generalization.

\section{The Basics of Loss Function Visualization }
\label{plotMethods}
Neural networks are trained on a corpus of feature vectors (e.g., images) $\{x_i\}$ and accompanying labels $\{y_i\}$ by minimizing a loss of the form
  $L(\theta) = \frac{1}{m}\sum_{i=1}^m \ell(x_i,y_i ; \theta) $,
where $\theta$ denotes the parameters (weights) of the neural network, the function $\ell(x_i,y_i ; \theta)$ measures how well the neural network
with parameters $\theta$ predicts the label of a data sample, and $m$ is the number of data samples.
Neural nets contain many parameters, and so their loss functions live in a very high-dimensional space.
Unfortunately, visualizations are only possible using low-dimensional 1D (line) or 2D (surface) plots.
Several methods exist for closing this dimensionality gap.

\paragraph{1-Dimensional Linear Interpolation}
One simple and lightweight way to plot loss functions is to choose two sets of parameters $\theta$ and $\theta',$ and plot the values of the loss function along the line connecting these two points.   We can parameterize this line by choosing a scalar parameter $\alpha,$ and defining the weighted average
    $\theta(\alpha) = (1-\alpha) \theta + \alpha \theta'.$
Finally, we plot the function $f(\alpha)=L(\theta(\alpha)).$
This strategy was taken by \citet{goodfellow2014qualitatively}, who studied the loss surface along the line between a random initial guess, and a nearby minimizer obtained by stochastic gradient descent.
This method has been widely used to study the  ``sharpness'' and ``flatness'' of different minima, and the dependence of sharpness on batch-size~\citep{keskar2016large,dinh2017sharp}.
\citet{smith2017exploring} use the same technique to show different minima and the ``peaks'' between them, while \citet{im2016empirical} plot the line between minima obtained via different optimizers.

The 1D linear interpolation method suffers from several weaknesses.
First, it is difficult to visualize non-convexities using 1D plots.
Indeed, \citet{goodfellow2014qualitatively} found that loss functions appear to lack local minima along the minimization trajectory.
We will see later, using 2D methods, that some loss functions have extreme non-convexities, and that these non-convexities correlate with the difference in generalization between different network architectures.
Second, this method does not consider batch normalization~\citep{batchnorm} or invariance symmetries in the network.
For this reason, the visual sharpness comparisons produced by 1D interpolation plots may be misleading;
this issue will be explored in depth in Section~\ref{sec:sharp}.

\paragraph{Contour Plots \& Random Directions}
To use this approach, one chooses a center point $\theta^*$ in the graph, and chooses two direction vectors, $\delta$ and $\eta$. One then plots a function of the form $f(\alpha) = L(\theta^*+\alpha \delta)$ in the 1D (line) case, or
\begin{align} \label{plot2d}
f(\alpha,\beta) = L(\theta^*+\alpha \delta+\beta \eta)
\end{align}
in the 2D (surface) case\footnote{When making 2D plots in this paper, batch normalization parameters are held constant, i.e., random directions are not applied to batch normalization parameters.}.
This approach was used in \citep{goodfellow2014qualitatively} to explore the trajectories of different minimization methods.
It was also used in \citep{im2016empirical} to show that different optimization algorithms find different local minima within the 2D projected space.
Because of the computational burden of 2D plotting, these methods generally result in low-resolution plots of small regions that have not captured the complex non-convexity of loss surfaces.  Below, we use high-resolution visualizations over large slices of weight space to visualize how network design affects non-convex structure.

\section{Proposed Visualization:  Filter-Wise Normalization}
\label{sec:norm}
This study relies heavily on plots of the form \eqref{plot2d} produced using random direction vectors, $\delta$ and $\eta,$ each sampled from a random Gaussian distribution with appropriate scaling (described below).
While the ``random directions'' approach to plotting is simple, it fails to capture the intrinsic geometry of loss surfaces, and cannot be used to compare the geometry of two different minimizers or two different networks.
This is because of the {\em scale invariance} in network weights.
When ReLU non-linearities are used, the network remains unchanged if we (for example) multiply the weights in one layer of a network by 10, and divide the next layer by 10.
This invariance is even more prominent when batch normalization is used.
In this case, the size (i.e., norm) of a filter
 is irrelevant because the output of each layer is re-scaled during batch normalization.
For this reason, a network's behavior remains unchanged if we re-scale the weights.
Note, this scale invariance applies only to rectified networks.

Scale invariance prevents us from making meaningful comparisons between plots, unless special precautions are taken.  A neural network with large weights may appear to have a smooth and slowly varying loss function; perturbing the weights by one unit will have very little effect on network performance if the weights live on a scale much larger than one.  However, if the weights are much smaller than one, then that same unit perturbation may have a catastrophic effect, making the loss function appear quite sensitive to weight perturbations.  Keep in mind that neural nets are scale invariant;  if the small-parameter and large-parameter networks in this example are equivalent (because one is simply a rescaling of the other), then any apparent differences in the loss function are merely an artifact of scale invariance.  This scale invariance was exploited by \citet{dinh2017sharp} to build pairs of equivalent networks that have different apparent sharpness.

To remove this scaling effect, we plot loss functions using filter-wise normalized directions.
To obtain such directions for a network with parameters $\theta,$ we begin by producing a random Gaussian direction vector $d$ with dimensions compatible with $\theta$.
Then, we normalize each filter in $d$ to have the same norm of the corresponding filter in $\theta.$ In other words, we make the replacement $d_{i,j} \gets  \frac{d_{i,j}}{\|d_{i,j}\|}\|\theta_{i,j}\|,$
where $d_{i,j}$ represents the $j$th filter (not the $j$th weight) of the $i$th layer of $d$, and $\|\cdot\|$ denotes the Frobenius norm.
Note that the filter-wise normalization is different from that of~\citep{im2016empirical}, which normalize the direction without considering the norm of individual filters.
Note that filter normalization is not limited to convolutional (Conv) layers but also applies to fully connected (FC) layers.
The FC layer is equivalent to a Conv layer with a $1\times1$ output feature map and the filter corresponds to the weights that generate one neuron.

Do contour plots of the form \eqref{plot2d} capture the natural distance scale of loss surfaces when the directions $\delta$ and $\eta$ are filter normalized?  We answer this question to the affirmative in Section \ref{sec:sharp} by showing that the sharpness of filter-normalized plots correlates well with generalization error, while plots without filter normalization can be very misleading.
In Appendix~\ref{sec:compare_norm}, we also compare filter-wise normalization to layer-wise normalization (and no normalization), and show that filter normalization produces superior correlation between sharpness and generalization error.

\section{The Sharp vs Flat Dilemma}
\label{sec:sharp}
Section \ref{sec:norm} introduces the concept of filter normalization, and provides an intuitive justification for its use.  In this section, we address the issue of whether sharp minimizers generalize better than flat minimizers. In doing so, we will see that the sharpness of minimizers correlates well with generalization error when filter normalization is used.  This enables side-by-side comparisons between plots.  In contrast, the sharpness of non-normalized plots may appear distorted and unpredictable.

It is widely thought that small-batch SGD produces ``flat'' minimizers that generalize well, while large batches produce ``sharp'' minima with poor generalization \citep{entropysgd,keskar2016large,hochreiter1997flat}.  This claim is disputed though, with \citet{dinh2017sharp,kawaguchi2017generalization} arguing that generalization is not directly related to the curvature of loss surfaces, and some authors proposing specialized training methods that achieve good performance with large batch sizes \citep{hoffer2017train,goyal2017accurate,de2017automated}.
Here, we explore the difference between sharp and flat minimizers.  We begin by discussing difficulties that arise when performing such a visualization, and how proper normalization can prevent such plots from producing distorted results.

We train a CIFAR-10 classifier using a 9-layer VGG network~\citep{vgg} with batch normalization for a fixed number of epochs.
We use two batch sizes: a large batch size of 8192
(16.4\% of the training data of CIFAR-10), and a small batch size of 128.
Let $\theta^{s}$ and $\theta^{l}$ indicate the solutions obtained by running SGD using small and large batch sizes, respectively\footnote{In this section, we consider the ``running mean'' and ``running variance'' as trainable parameters and include them in $\theta$. Note that the original study by \citet{goodfellow2014qualitatively} does not consider batch normalization. These parameters are not included in $\theta$ in future sections, as they are only needed when interpolating between two minimizers.}.
Using the linear interpolation approach \citep{goodfellow2014qualitatively}, we plot the loss values on both training and testing data sets of CIFAR-10, along a direction containing the two solutions, i.e., $f(\alpha) = L(\theta^s + \alpha (\theta^l - \theta^s))$.

\begin{figure}[t]
\vspace{-2mm}
\begin{tabular}{c}
\begin{minipage}[t]{0.32\linewidth}
\vspace{-2mm}
\subfigure[7.37\%~~~~~~~~~~ 11.07\%]{\includegraphics[height=0.7\linewidth]{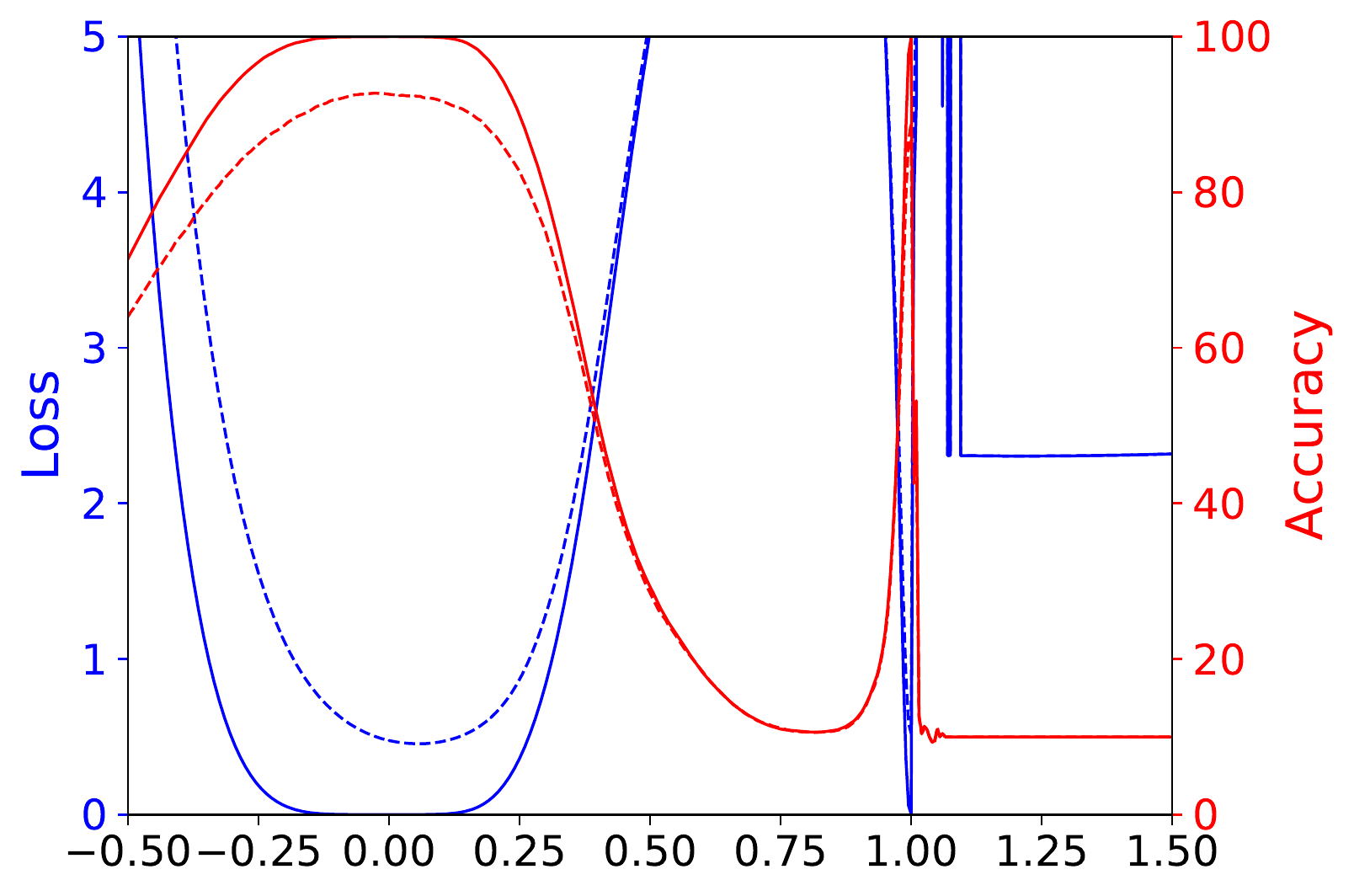}}
\end{minipage}
\begin{minipage}[t]{0.34\linewidth}
\centering
\vspace{-2mm}
\subfigure[$\|\theta\|_2$, WD=0]{\includegraphics[height=0.65\linewidth]{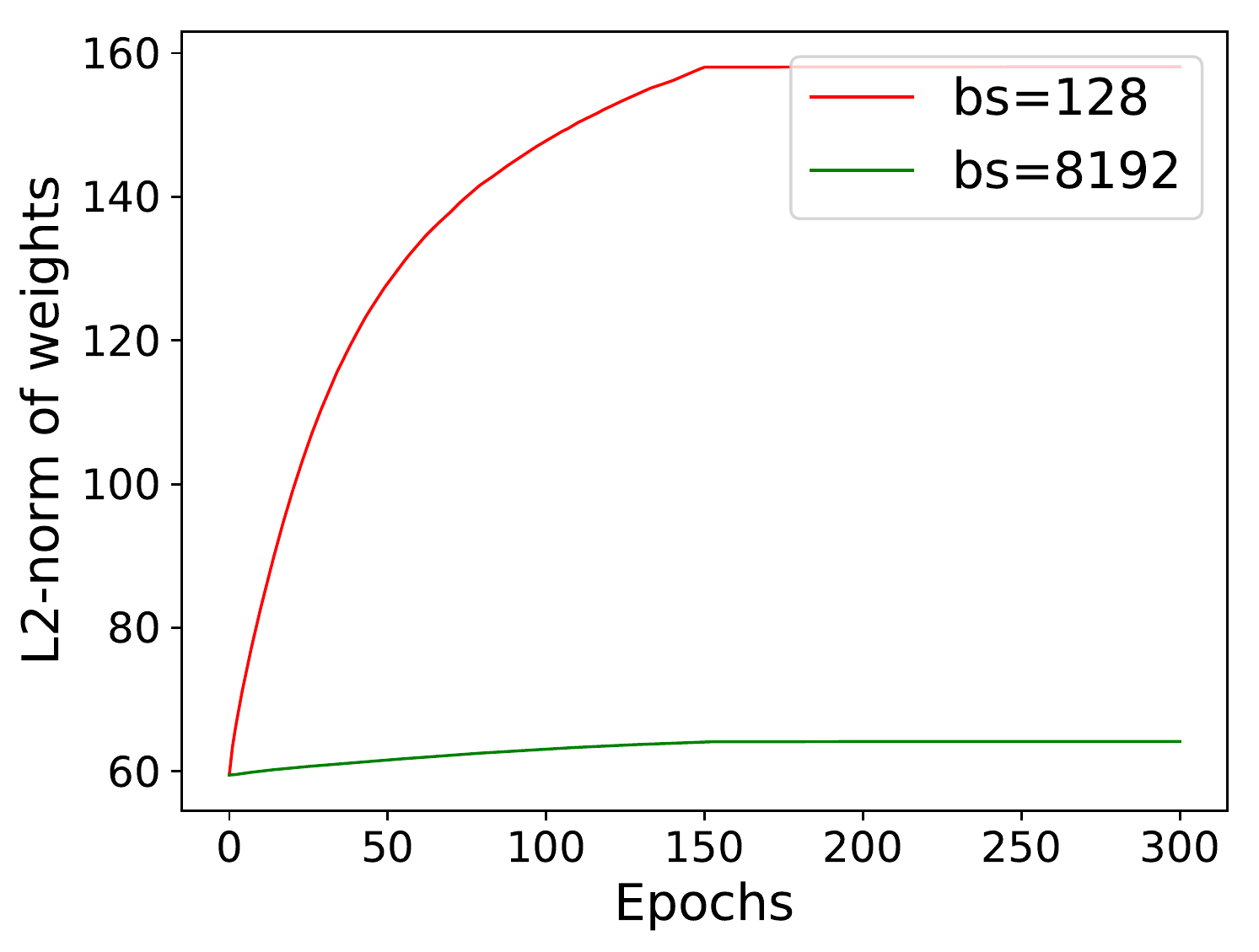}}
\end{minipage}
\begin{minipage}[t]{0.32\linewidth}
\vspace{-2mm}
\subfigure[WD=0]{\includegraphics[height=0.7\linewidth]{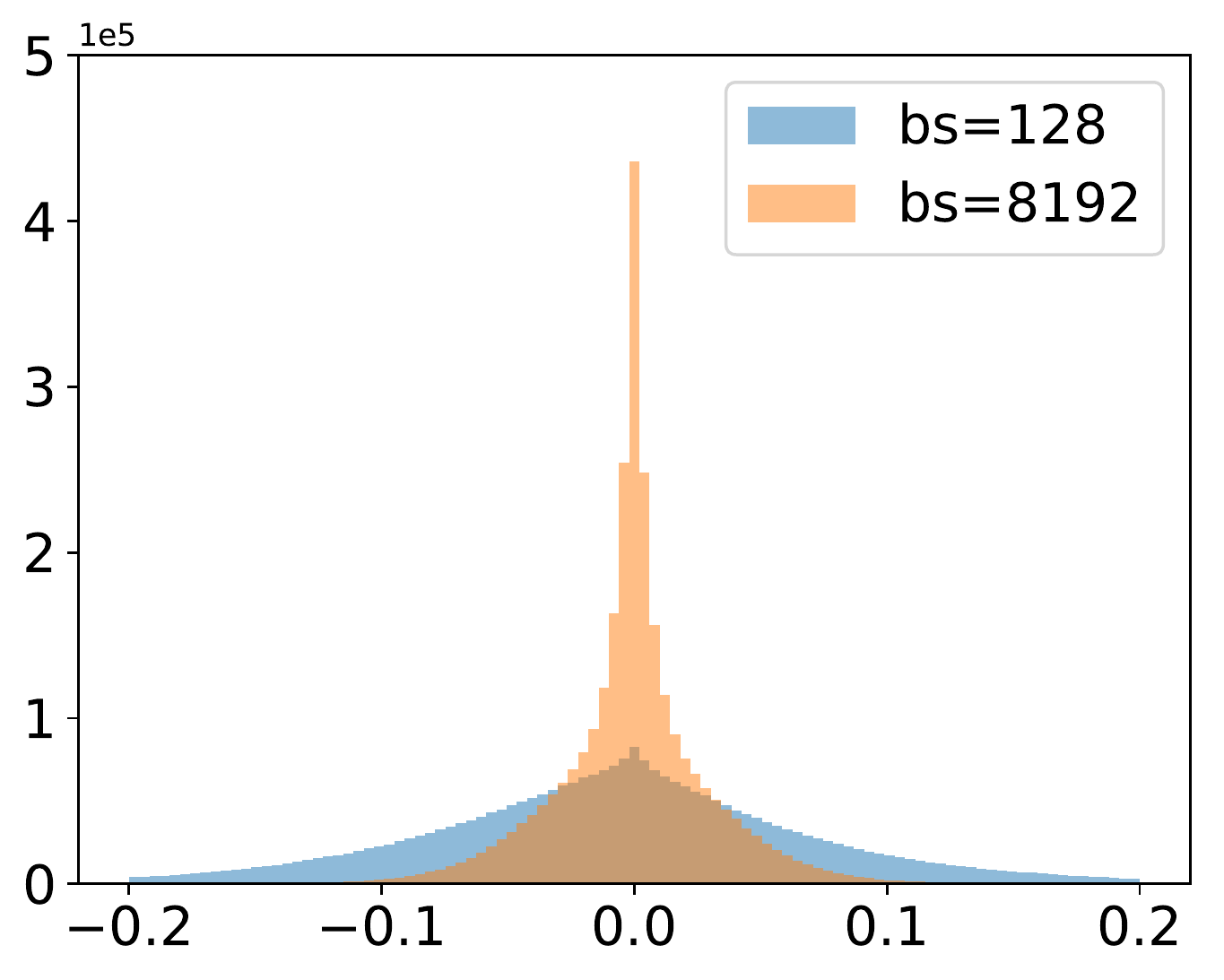}}
\end{minipage}\\
\begin{minipage}[t]{0.32\linewidth}
\vspace{-2mm}
\subfigure[6.0\%~~~~~~~~~~~~10.19\%]{\includegraphics[height=0.7\linewidth]{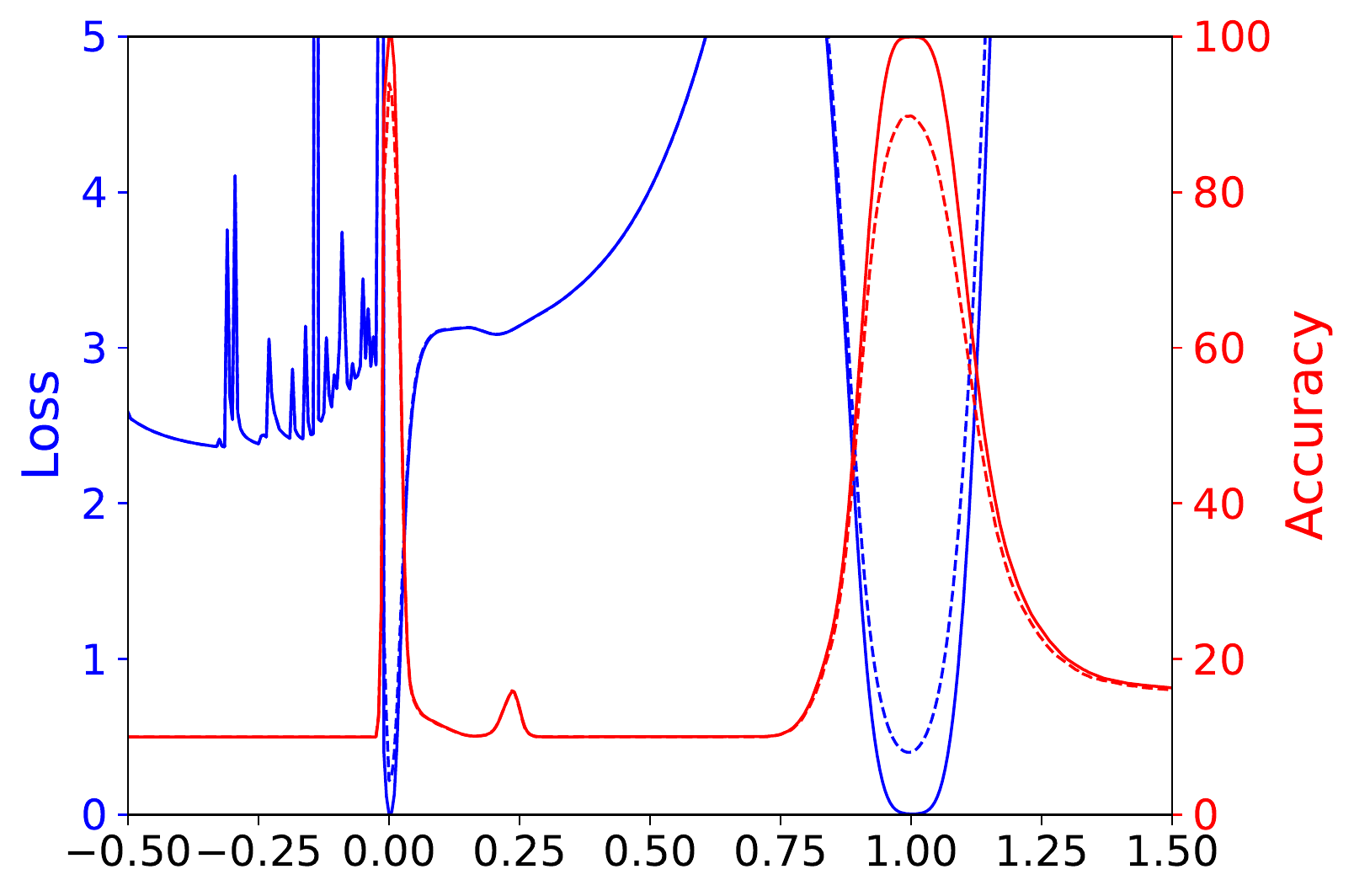}}
\end{minipage}
\begin{minipage}[t]{0.34\linewidth}
\centering
\vspace{-2mm}
\subfigure[$\|\theta\|_2$, WD=5e-4]{\includegraphics[height=0.65\linewidth]{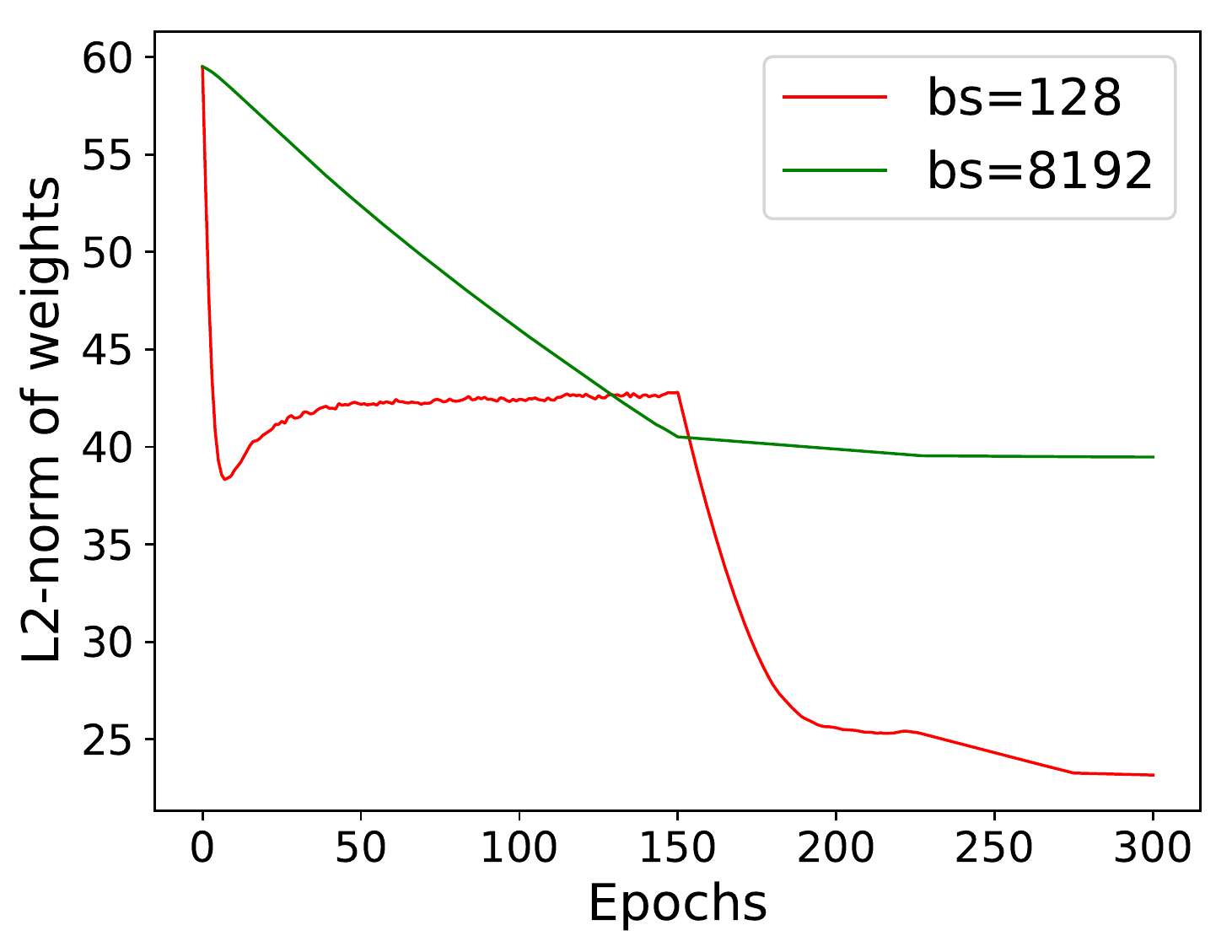}}
\end{minipage}
\begin{minipage}[t]{0.32\linewidth}
\vspace{-2mm}
\subfigure[WD=5e-4]{\includegraphics[height=0.7\linewidth]{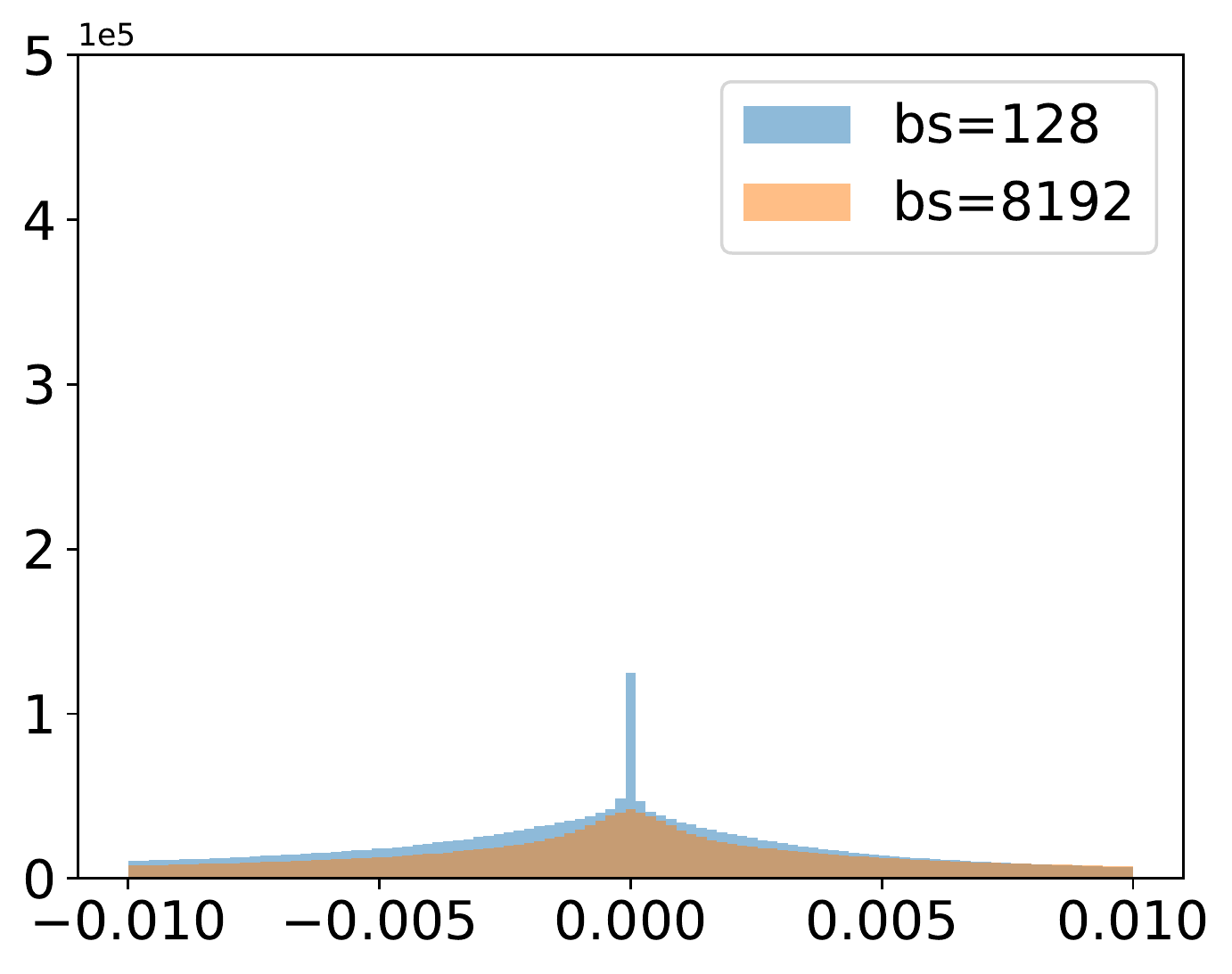}}
\end{minipage}
\end{tabular}
\caption{(a) and (d) are the 1D linear interpolation of VGG-9 solutions obtained by small-batch and large-batch training methods.
The blue lines are loss values and the red lines are accuracies.
The solid lines are training curves and the dashed lines are for testing.
Small batch is at abscissa 0, and large batch is at abscissa~1.
The corresponding test errors are shown below.
(b) and (e) shows the change of weights norm $\|\theta\|_2$ during training.
When weight decay is disabled, the weight norm grows steadily during training without constraints
(c) and (f) are the weight histograms, which verify that small-batch methods produce more large weights with zero weight decay and more small weights with non-zero weight decay.
}
\label{fig:weight_decay_batchsize}
\vspace{-3mm}
\end{figure}

Figure \ref{fig:weight_decay_batchsize}(a) shows linear interpolation plots with $\theta^s$ at $x$-axis location 0, and $\theta^l$ at location 1.
As observed by \cite{keskar2016large}, we can clearly see that the small-batch solution is quite wide, while the large-batch solution is sharp.
However, this sharpness balance can be flipped simply by turning on weight decay~\citep{weight_decay}.
Figure \ref{fig:weight_decay_batchsize}(d) show results of the same experiment, except this time with a non-zero weight decay parameter.
This time, the large batch minimizer is considerably flatter than the sharp small batch minimizer.    However, we see that small batches generalize better in all experiments; there is no apparent correlation between sharpness and generalization.
We will see that these sharpness comparisons are extremely misleading, and fail to capture the endogenous properties of the minima.

The apparent differences in sharpness can be explained by examining the weights of each minimizer.
Histograms of the network weights are shown for each experiment in Figure \ref{fig:weight_decay_batchsize}(c) and (f).
We see that, when a large batch is used with zero weight decay, the resulting weights tend to be smaller than in the small batch case.
We reverse this effect by adding weight decay; in this case the large batch minimizer has much larger weights than the small batch minimizer.
This difference in scale occurs for a simple reason:
A smaller batch size results in more weight updates per epoch than a large batch size, and so the shrinking effect of weight decay (which imposes a penalty on the norm of the weights) is more pronounced.
The evolution of the weight norms during training is depicted in Figure \ref{fig:weight_decay_batchsize}(b) and (e).
Figure \ref{fig:weight_decay_batchsize} is not visualizing the endogenous sharpness of minimizers, but rather just the (irrelevant) weight scaling.
The scaling of weights in these networks is irrelevant because batch normalization re-scales the outputs to have unit variance.
However, small weights still appear more sensitive to perturbations, and produce sharper looking minimizers.

\begin{figure*}
\vspace{-1mm}
\centering
\begin{tabular}{l}
\hspace{-.3cm}
\subfigure[0.0, 128, 7.37\%]{
\includegraphics[width=0.24\linewidth]{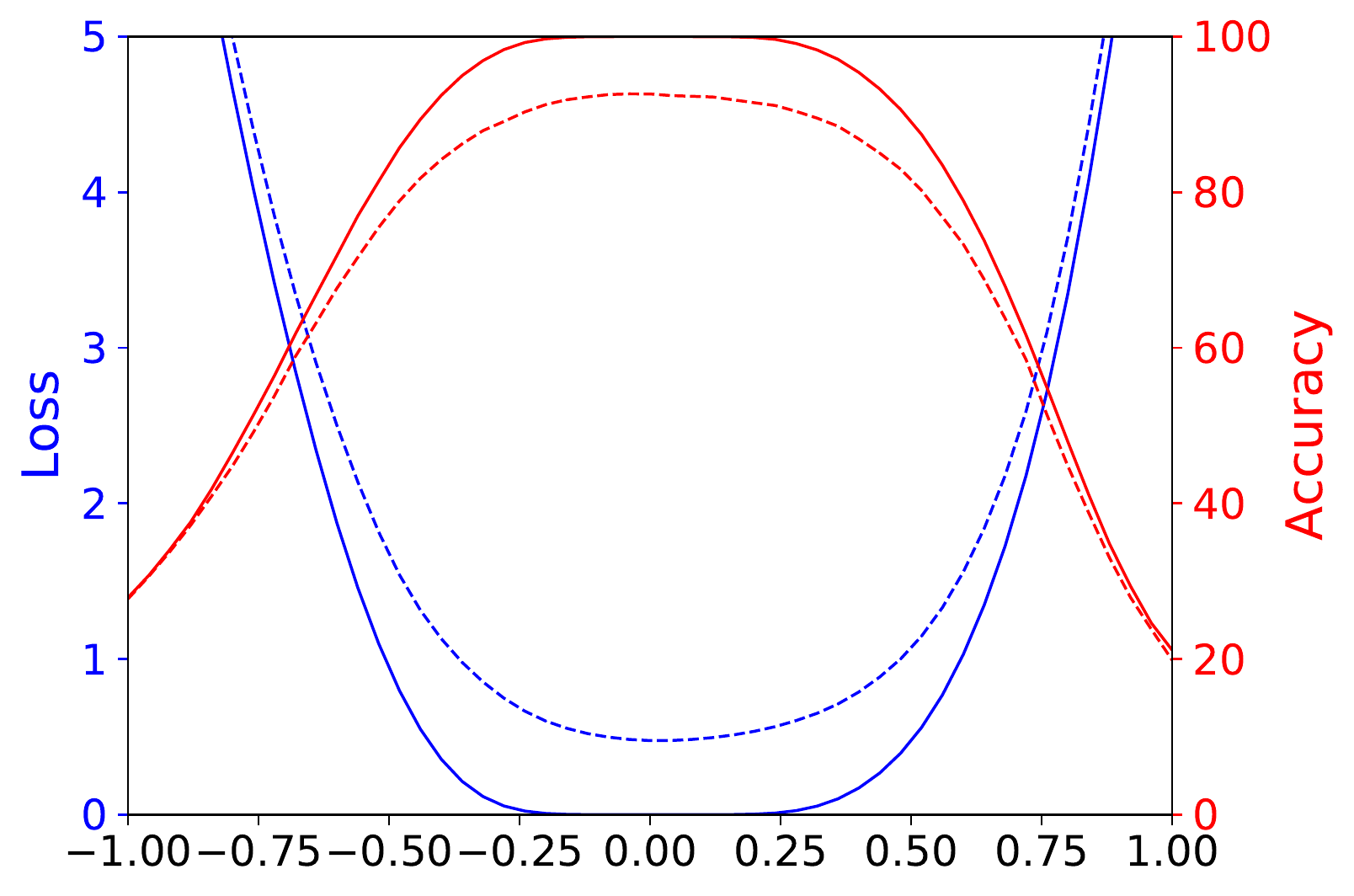}}
\subfigure[0.0, 8192, 11.07\%]{
\includegraphics[width=0.24\linewidth]{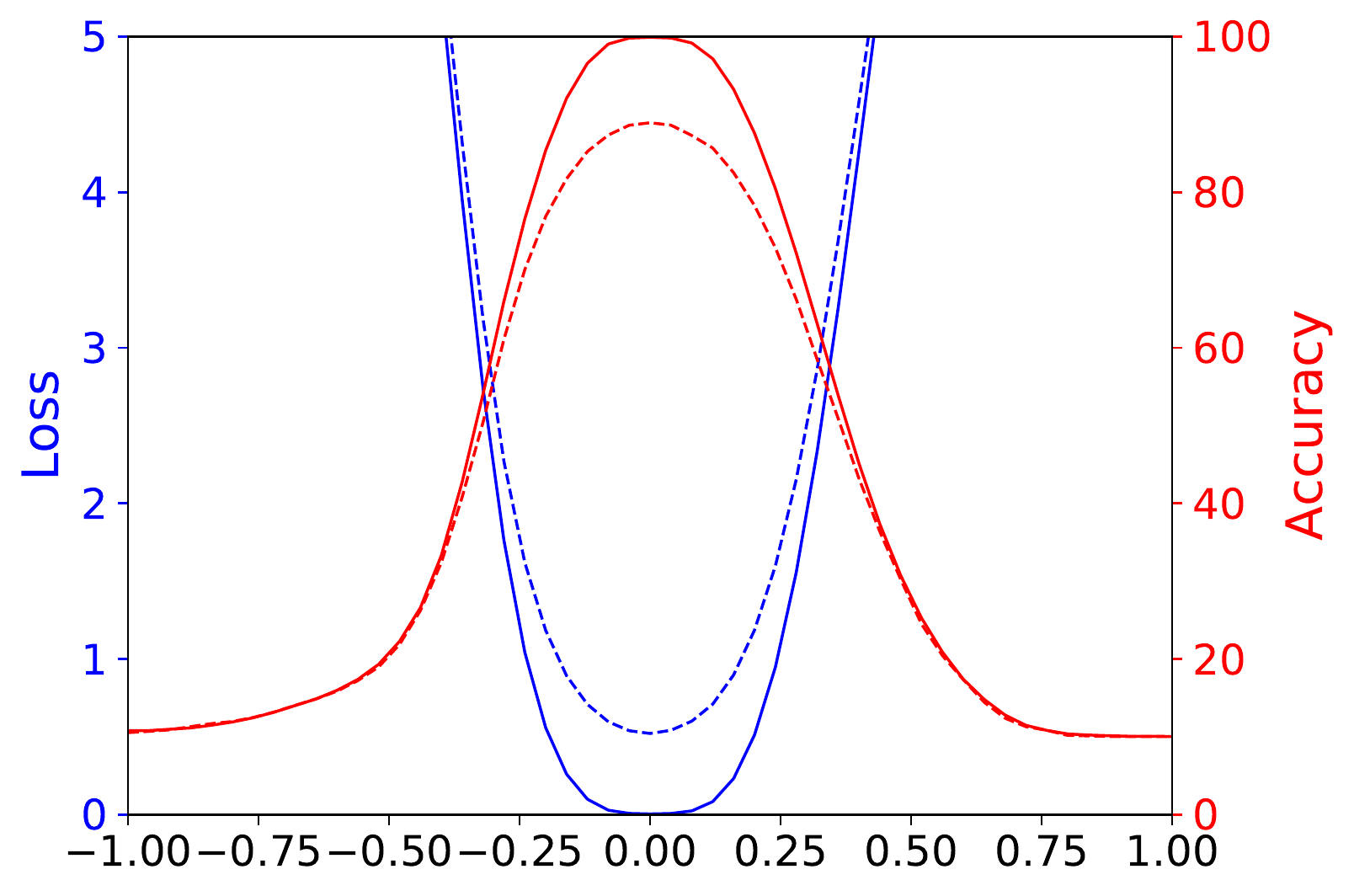}}
\subfigure[5e-4, 128, 6.00\%]{\includegraphics[width=0.24\linewidth]{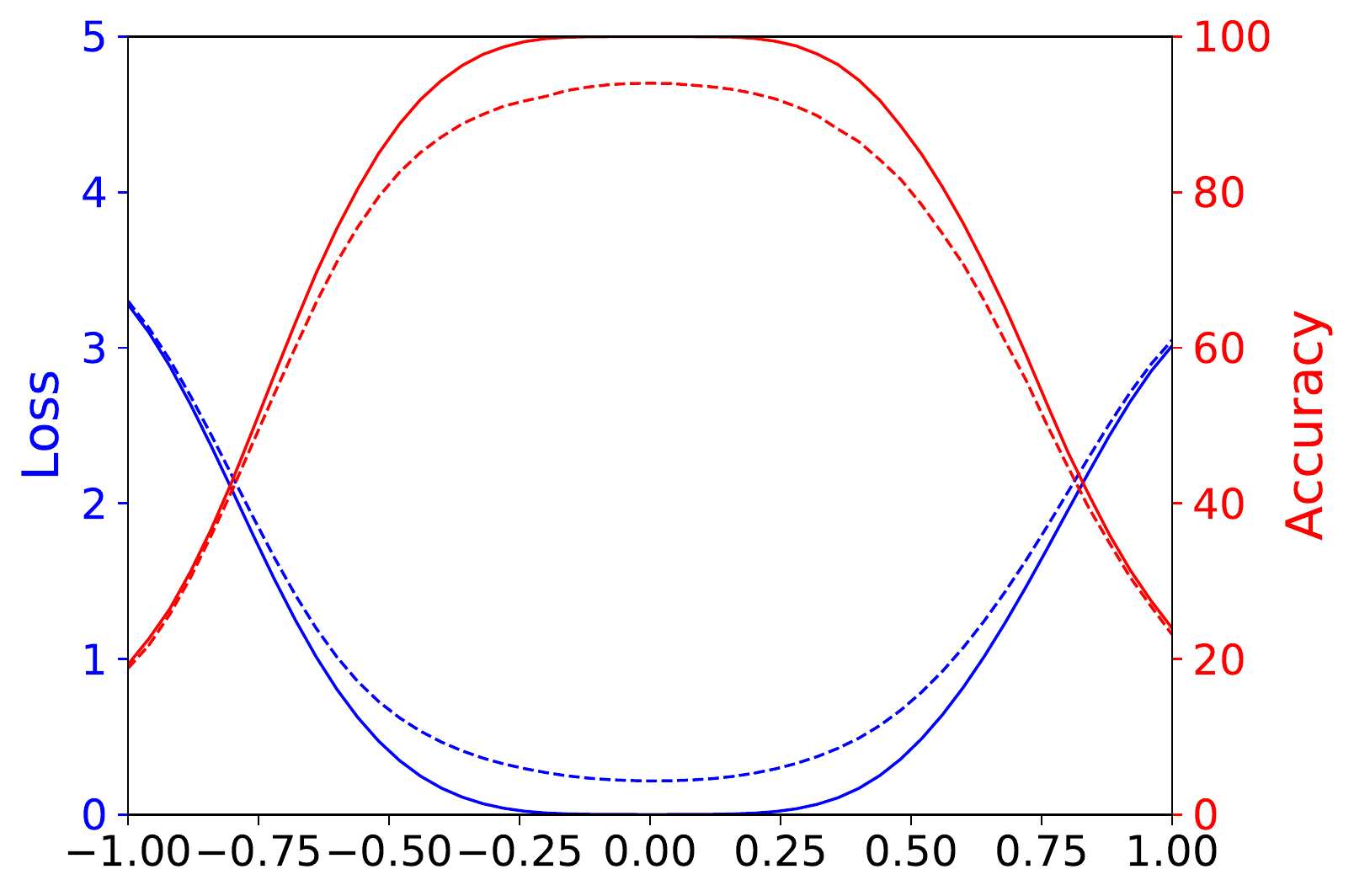}}
\subfigure[5e-4, 8192, 10.19\%]{\includegraphics[width=0.24\linewidth]{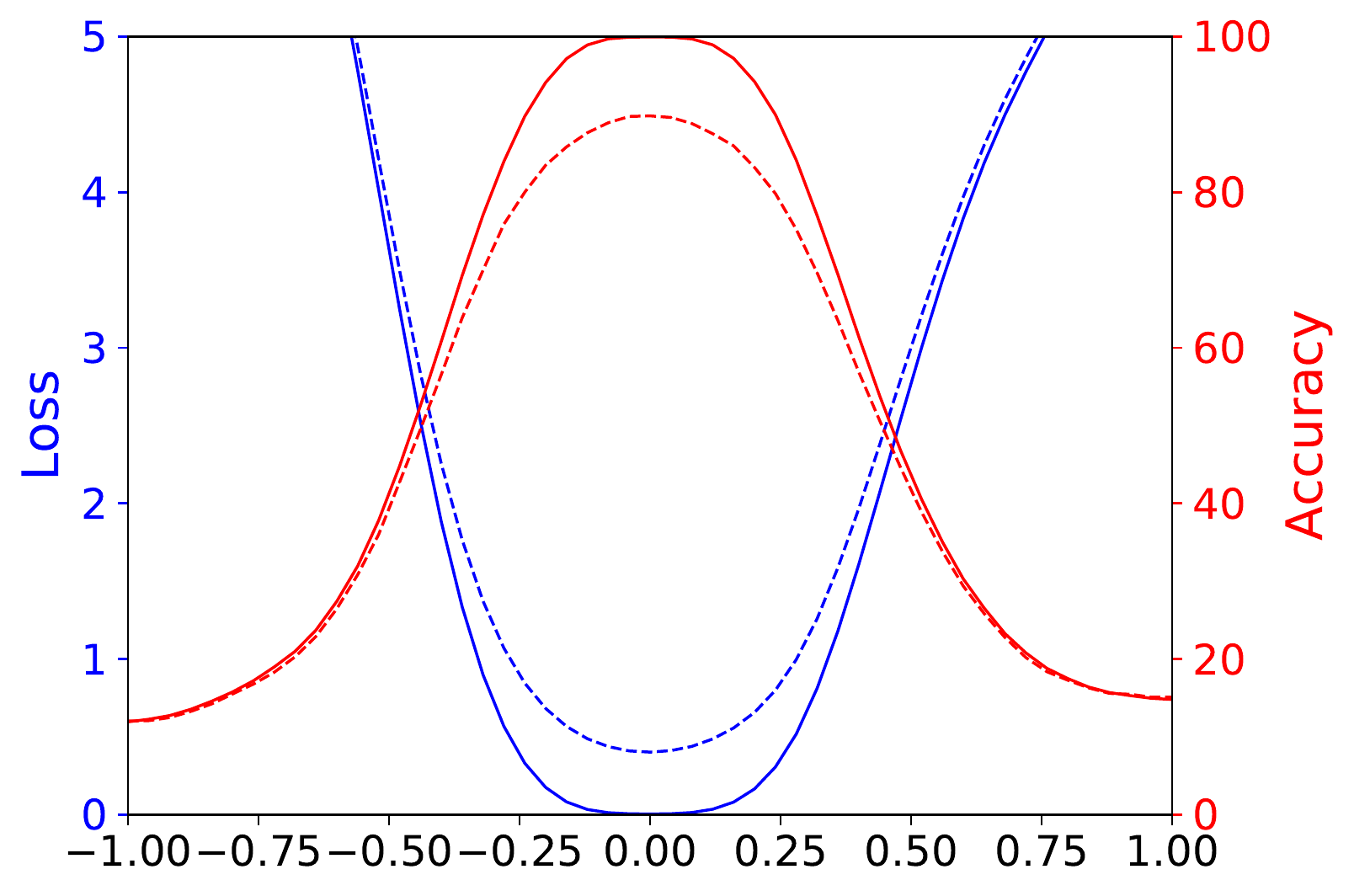}}\\
\hspace{-.3cm}
\subfigure[0.0, 128, 7.37\%]{\includegraphics[width=0.25\linewidth]{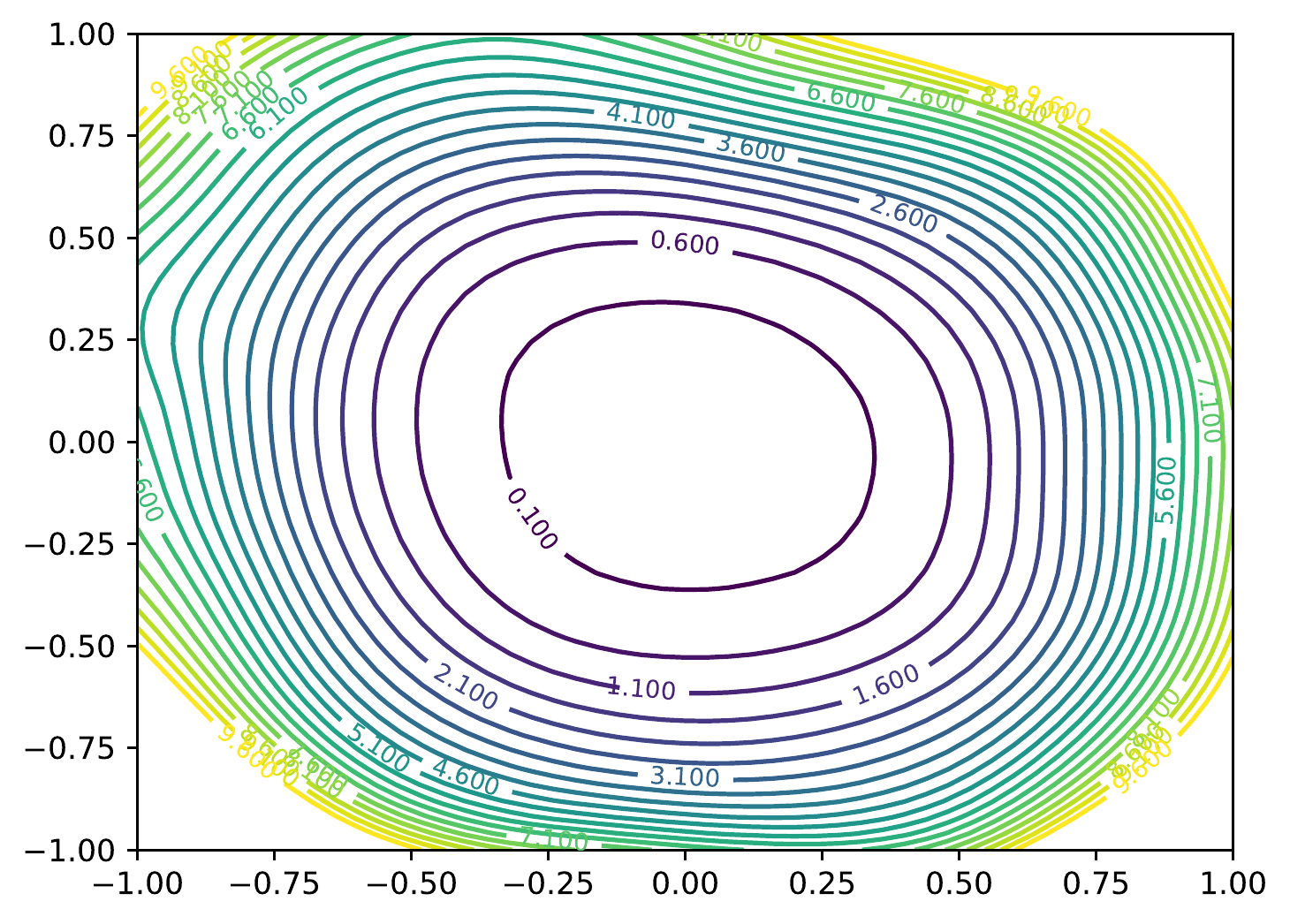}}
\subfigure[0.0, 8192, 11.07\%]{\includegraphics[width=0.24\linewidth]{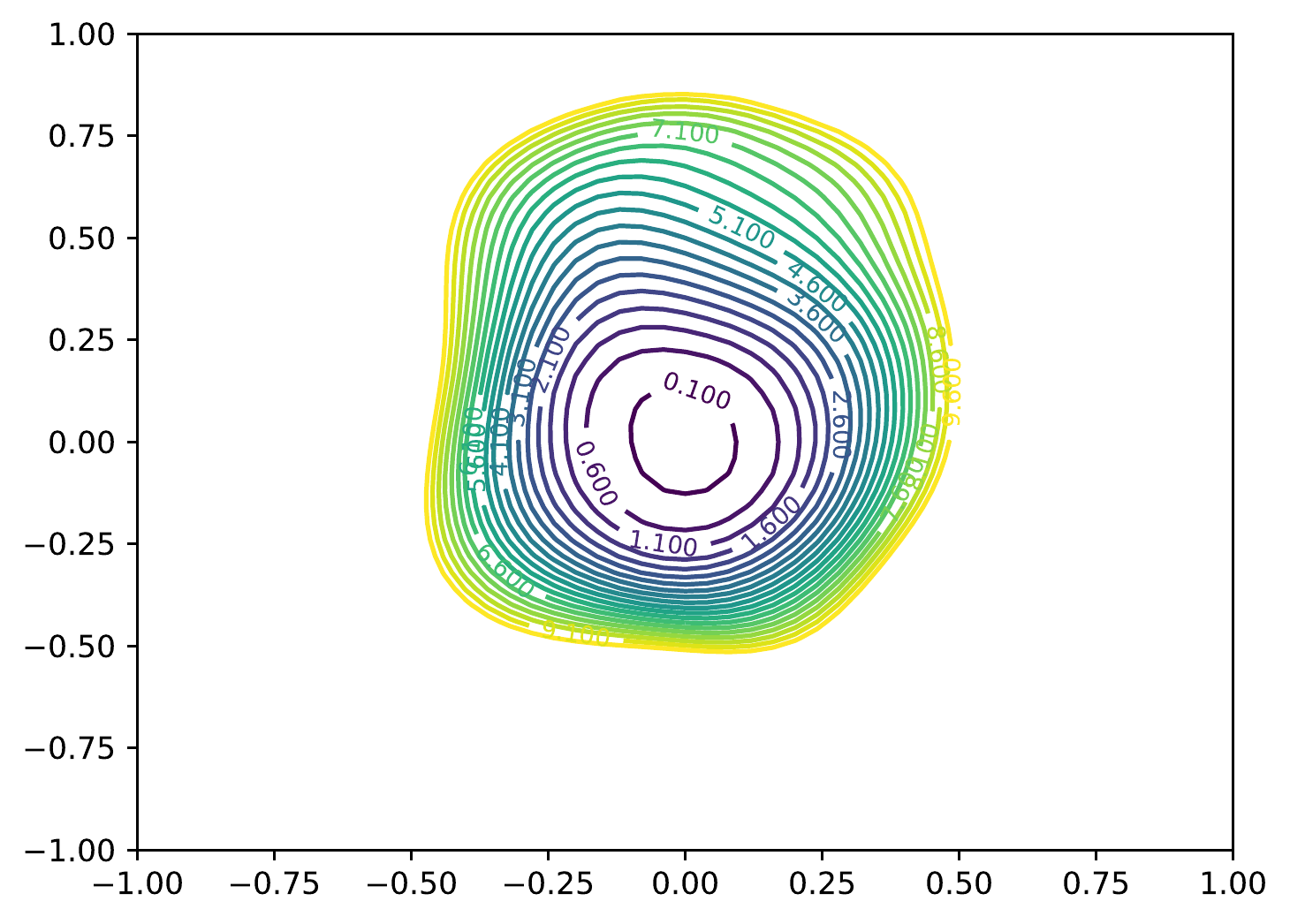}}
\subfigure[5e-4, 128, 6.00\%]{\includegraphics[width=0.24\linewidth]{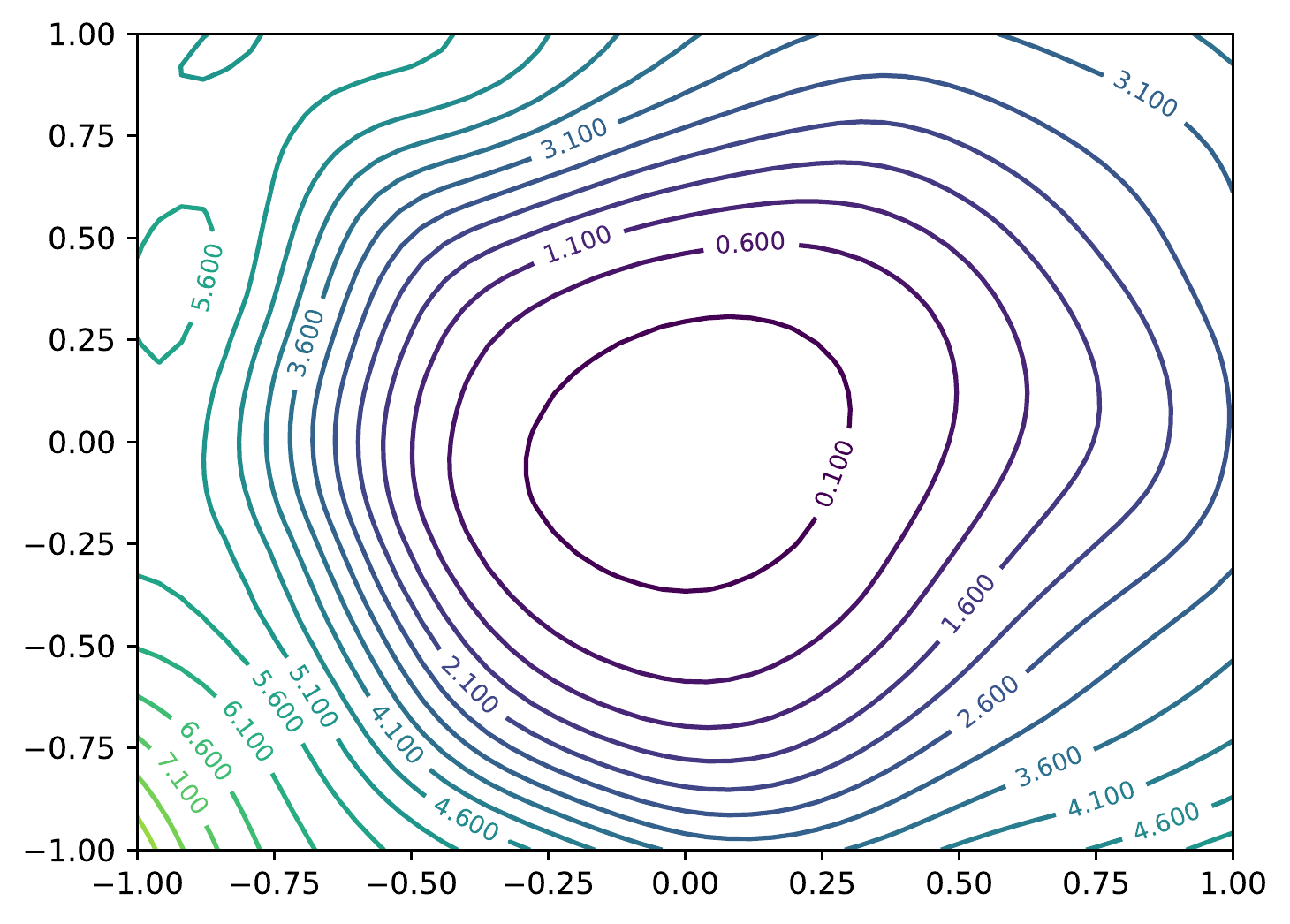}}
\subfigure[5e-4, 8192, 10.19\%]{\includegraphics[width=0.24\linewidth]{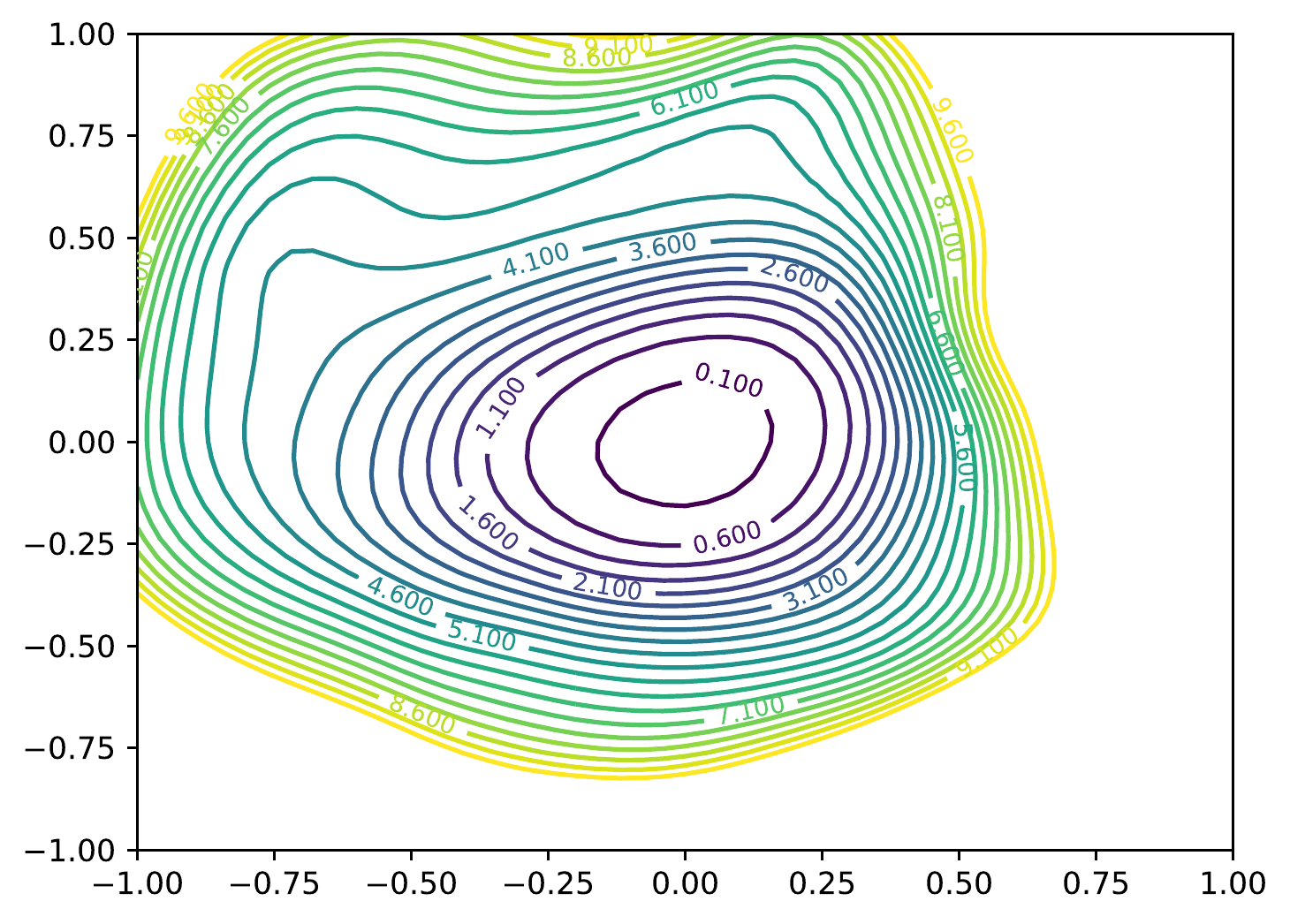}}
\end{tabular}
\caption{The 1D and 2D visualization of solutions obtained using SGD with different weight decay and batch size.
The title of each subfigure contains the weight decay, batch size, and test error.
\vspace{-4mm}
}
\label{fig:noramlized_shape_batchsize}
\end{figure*}

\paragraph{Filter Normalized Plots}
We repeat the experiment in Figure \ref{fig:weight_decay_batchsize}, but this time we plot the loss function near each minimizer separately using random filter-normalized  directions.
This removes the apparent differences in geometry caused by the scaling depicted in Figure
\ref{fig:weight_decay_batchsize}(c) and (f).
The results, presented in Figure~\ref{fig:noramlized_shape_batchsize}, still show differences in sharpness between small batch and large batch minima, however these differences are much more subtle than it would appear in the un-normalized plots.
For comparison, sample un-normalized plots and layer-normalized plots are shown in Section~\ref{sec:compare_norm} of the Appendix.
We also visualize these results using two random directions and contour plots.
The weights obtained with small batch size and non-zero weight decay have wider contours than the sharper large batch minimizers.
Results for ResNet-56 appear in Figure \ref{fig:noramlized_shape_batchsize_resnet56} of the Appendix.
Using the filter-normalized plots in Figure \ref{fig:noramlized_shape_batchsize}, we can make side-by-side comparisons between minimizers, and we see that now sharpness correlates well with generalization error.
Large batches produced visually sharper minima (although not dramatically so) with higher test error.

\section{What Makes Neural Networks Trainable? Insights on the (Non)Convexity
Structure of Loss Surfaces }
\label{sec:exp_different_networks}

Our ability to find global minimizers to neural loss functions is not universal;  it seems that some neural architectures are easier to minimize than others.  For example, using skip connections, the authors of \cite{resnet}  trained extremely deep architectures, while comparable architectures without skip connections are not trainable.  Furthermore, our ability to train seems to depend strongly on the initial parameters from which training starts.
Using visualization methods, we do an empirical study of neural architectures to explore why the non-convexity of loss functions seems to be problematic in some situations, but not in others.
We aim to provide insight into the following questions:  Do loss functions have significant non-convexity at all?  If prominent non-convexities exist, why are they not problematic in all situations?   Why are some architectures easy to train, and why are results so sensitive to the initialization?   We will see that different architectures have extreme differences in non-convexity structure that answer these questions, and that these differences correlate with generalization error.

\begin{figure*}[!b]
\centering
\subfigure[ResNet-110, no skip connections]{
\includegraphics[width=0.45\linewidth]{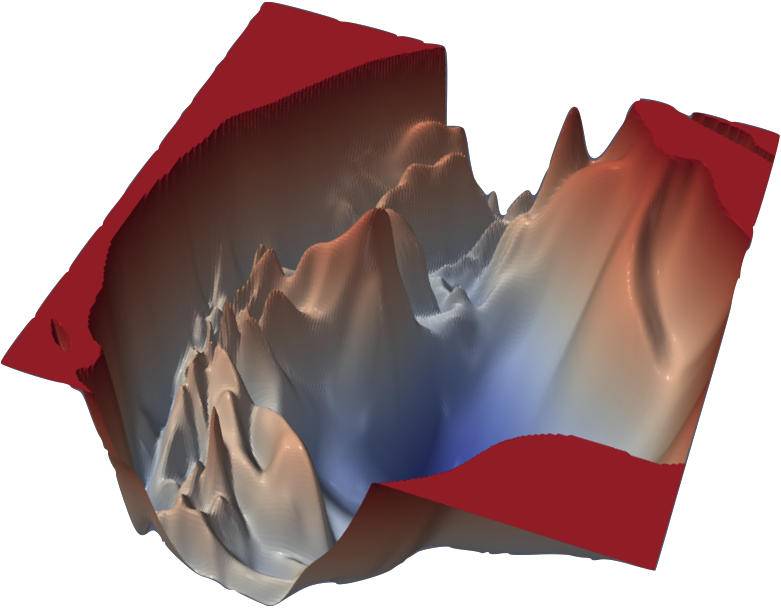}}
\subfigure[DenseNet, 121 layers]{
\includegraphics[width=0.45\linewidth]{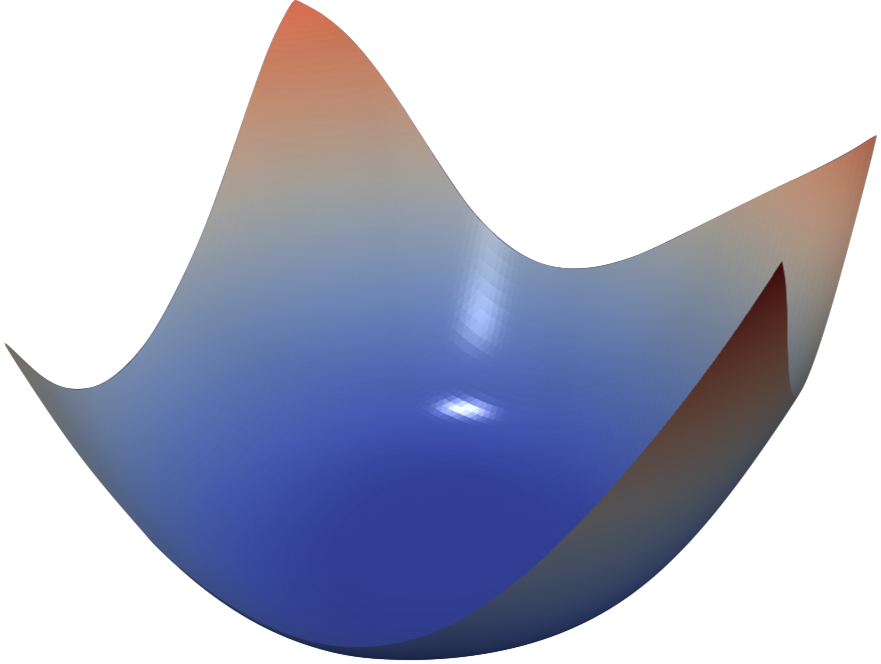}}
\vspace{-2mm}
\caption{The loss surfaces of ResNet-110-noshort and DenseNet for CIFAR-10.}
\label{fig:moresurfs}
\end{figure*}

\begin{figure*}[t]
\centering
\begin{tabular}{l}
\hspace{-.5cm}
 \subfigure[ResNet-20, 7.37\%]{\includegraphics[width=0.25\linewidth]{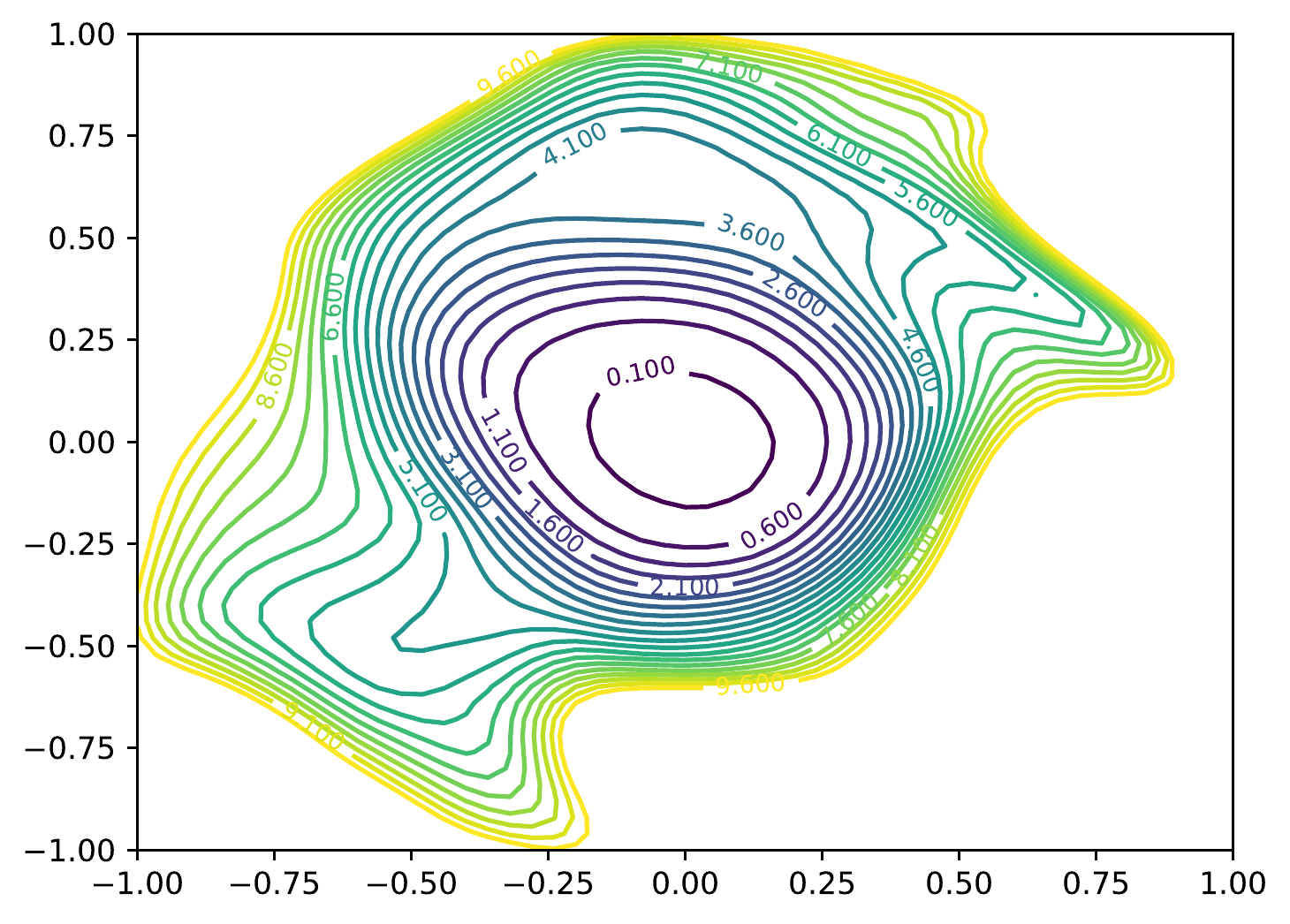}}
 \subfigure[ResNet-56, 5.89\%]{\includegraphics[width=0.25\linewidth]{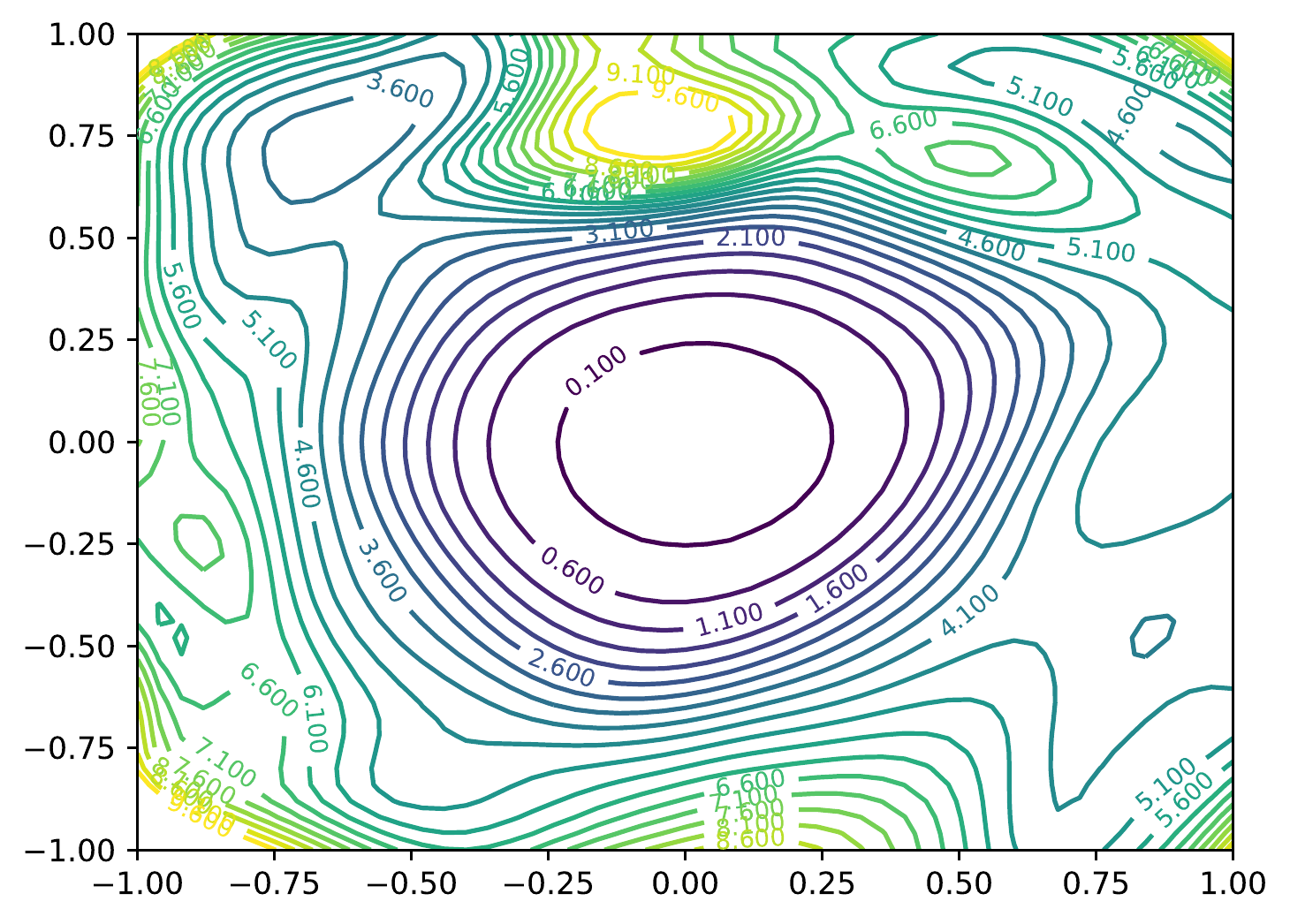}}
 \subfigure[ResNet-110, 5.79\%]{\includegraphics[width=0.25\linewidth]{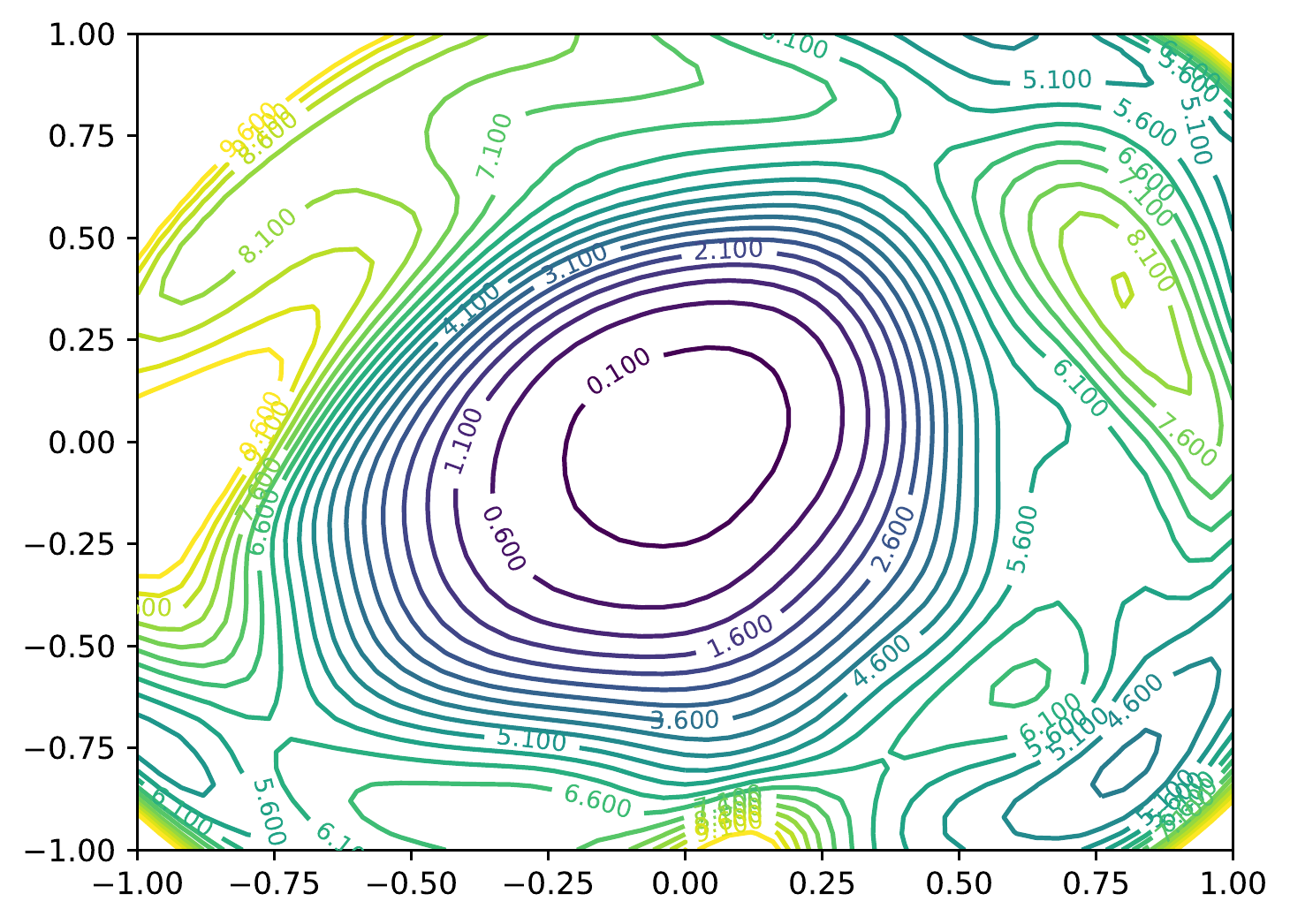}}\\
 \hspace{-.5cm}
 \subfigure[ResNet-20-NS, 8.18\%]{\includegraphics[width=0.25\linewidth]{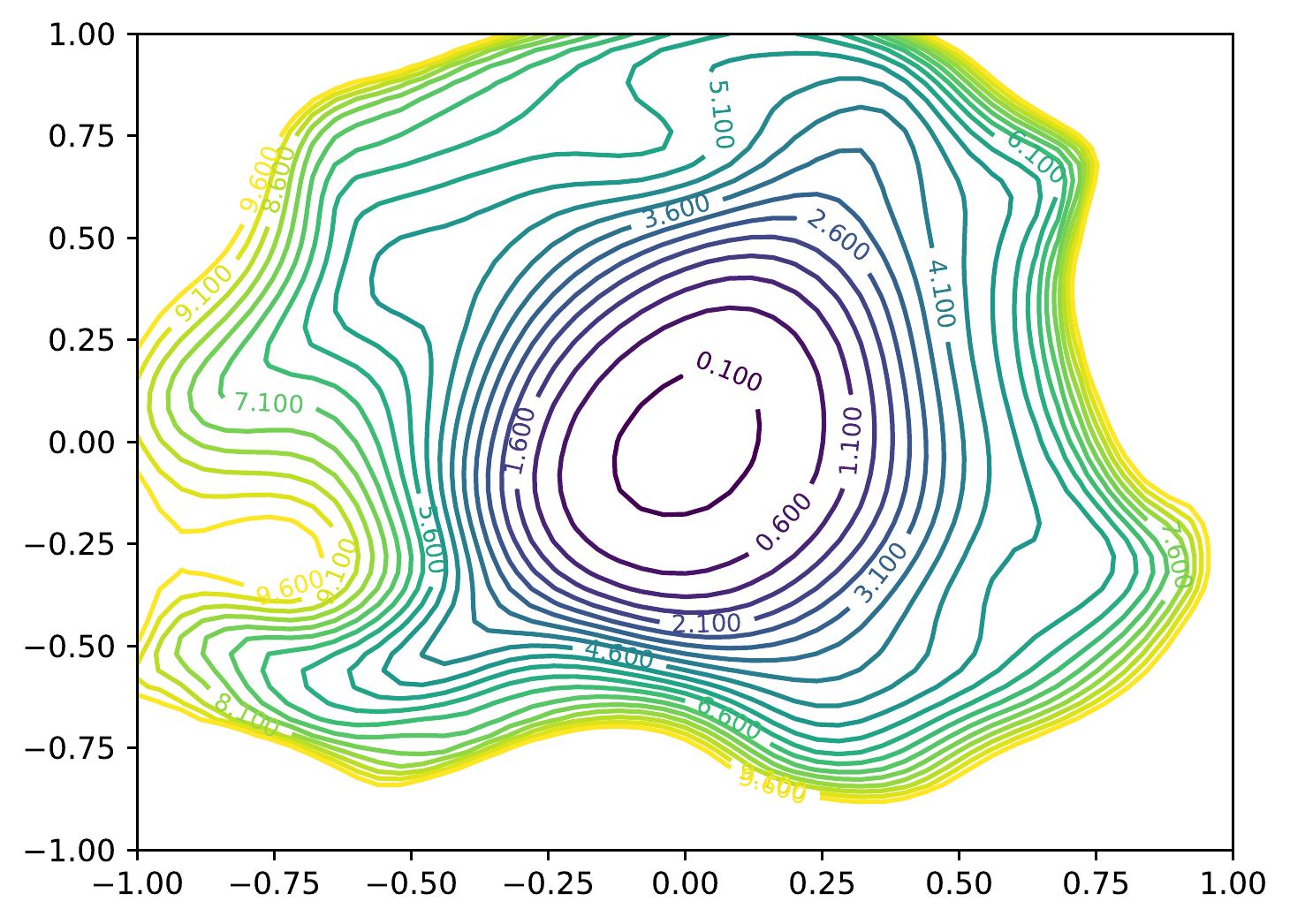}}
 \subfigure[ResNet-56-NS, 13.31\%]{\includegraphics[width=0.25\linewidth]{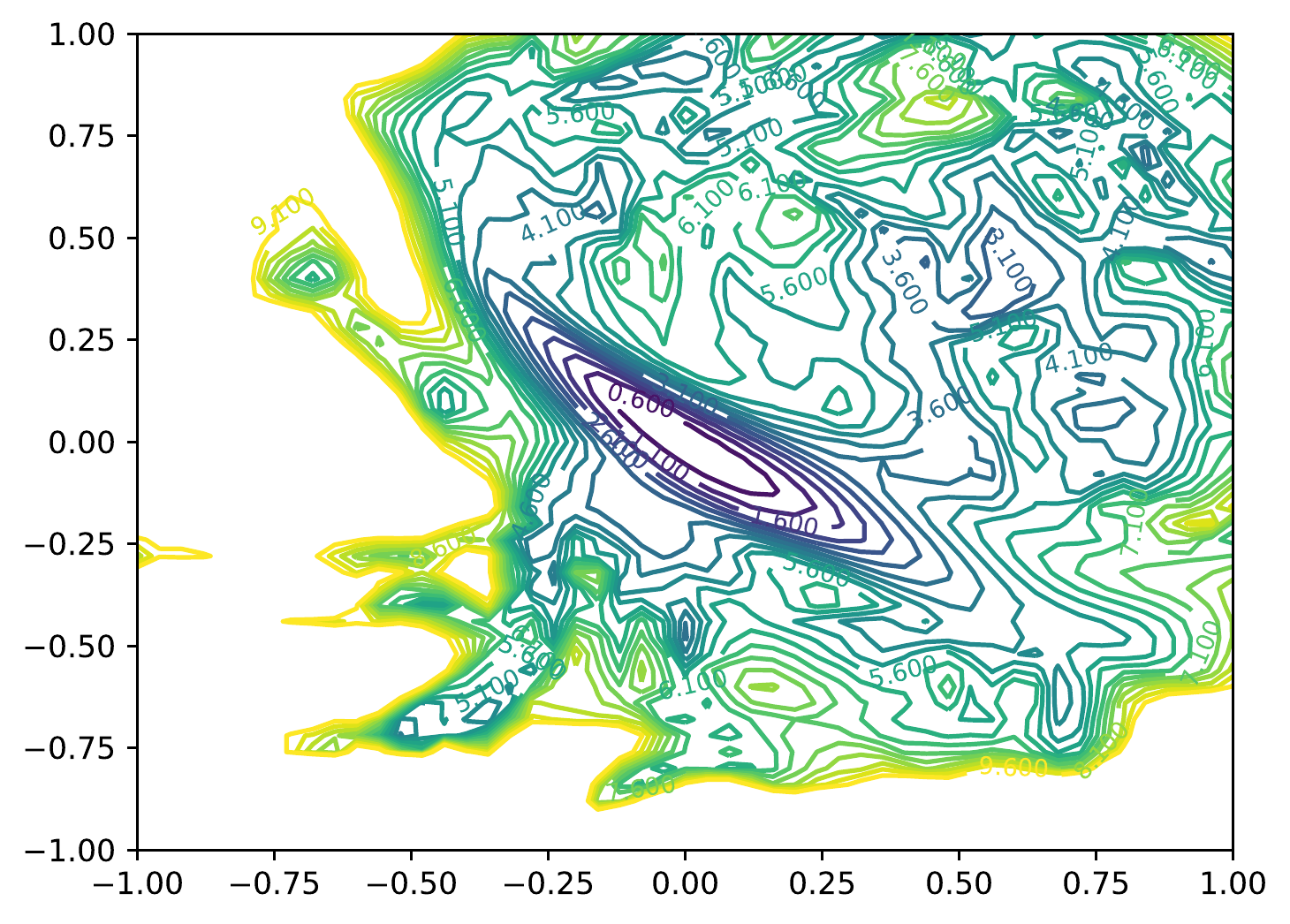}}
 \subfigure[ResNet-110-NS, 16.44\%]{\includegraphics[width=0.25\linewidth]{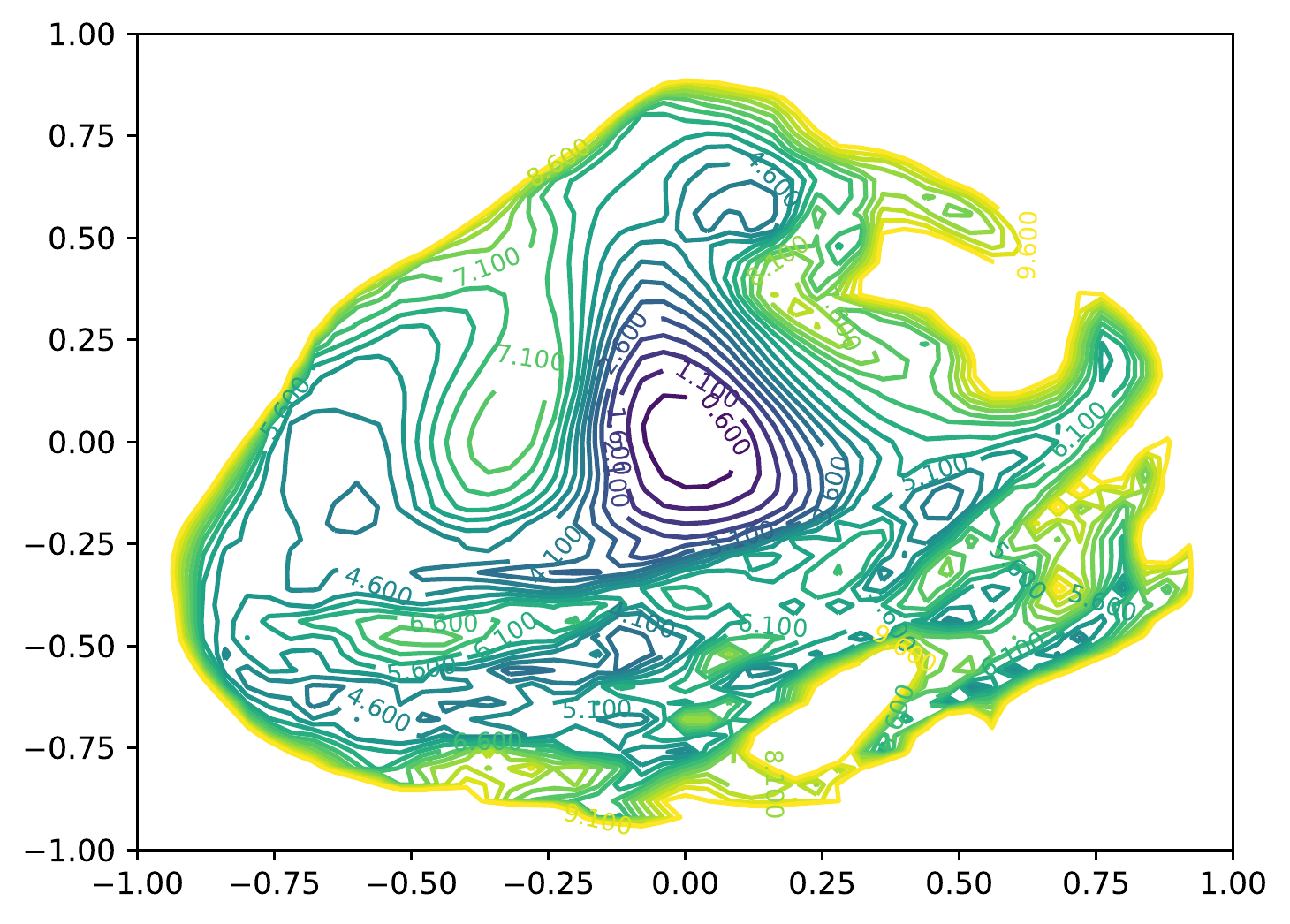}}\\
 \hspace{-.5cm}
\end{tabular}
\vspace{-5mm}
\caption{2D visualization of the loss surface of ResNet and ResNet-noshort with different depth.
}
\label{fig:depth}
\vspace{-4mm}
\end{figure*}

\begin{figure*}[!t]
\centering
\begin{tabular}{l}
\hspace{-0.3cm}
\subfigure[$k=1$, 5.89\%]{\includegraphics[width=0.23\linewidth]{figures/{resnet56_sgd_lr=0.1_bs=128_wd=0.0005/resnet56_random_-1.0,1.0x-1.0,1.0.h5_2dcontour}.pdf}}
\subfigure[$k=2$, 5.07\%]{\includegraphics[width=0.23\linewidth]{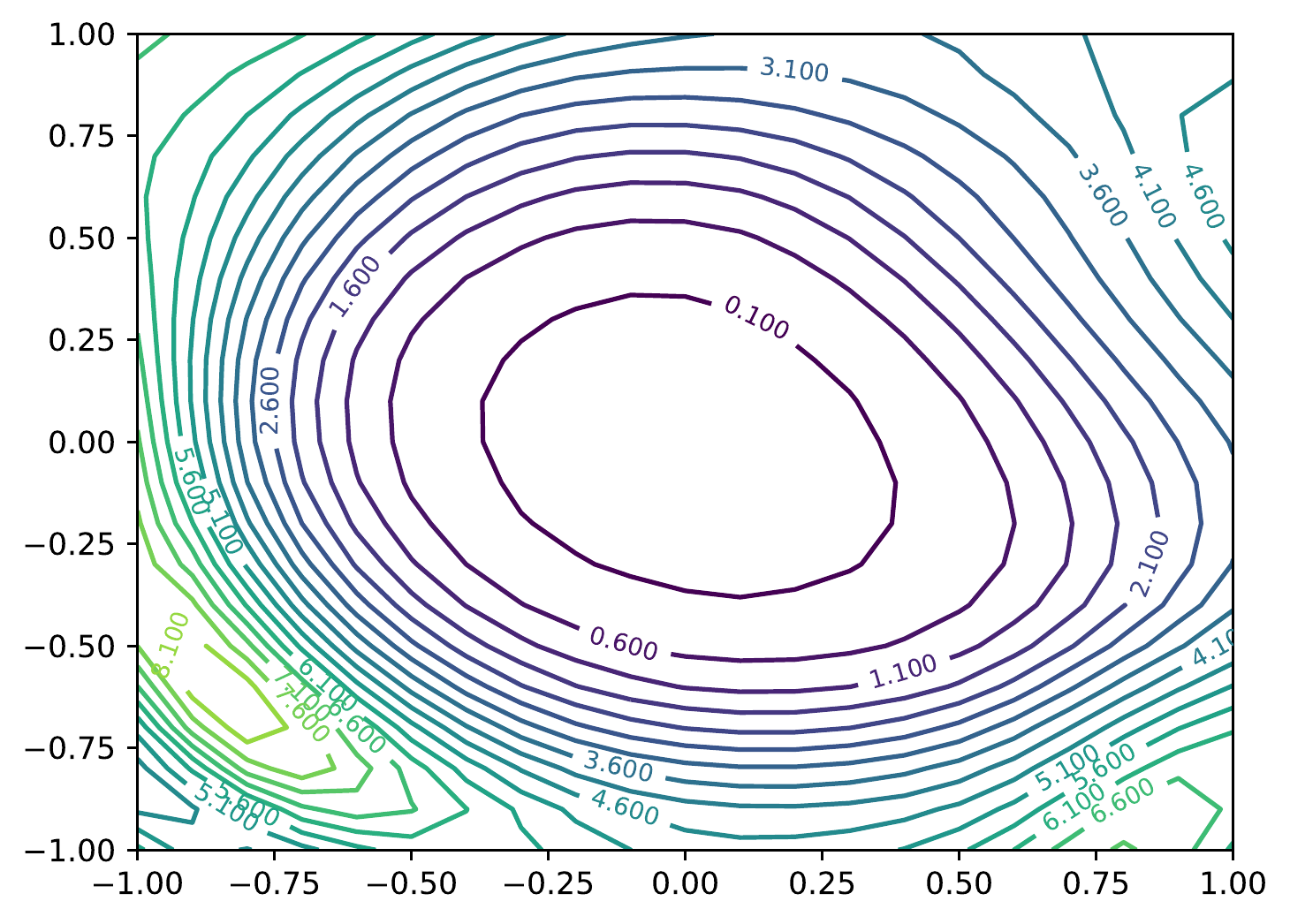}}
\subfigure[$k=4$, 4.34\%]{\includegraphics[width=0.23\linewidth]{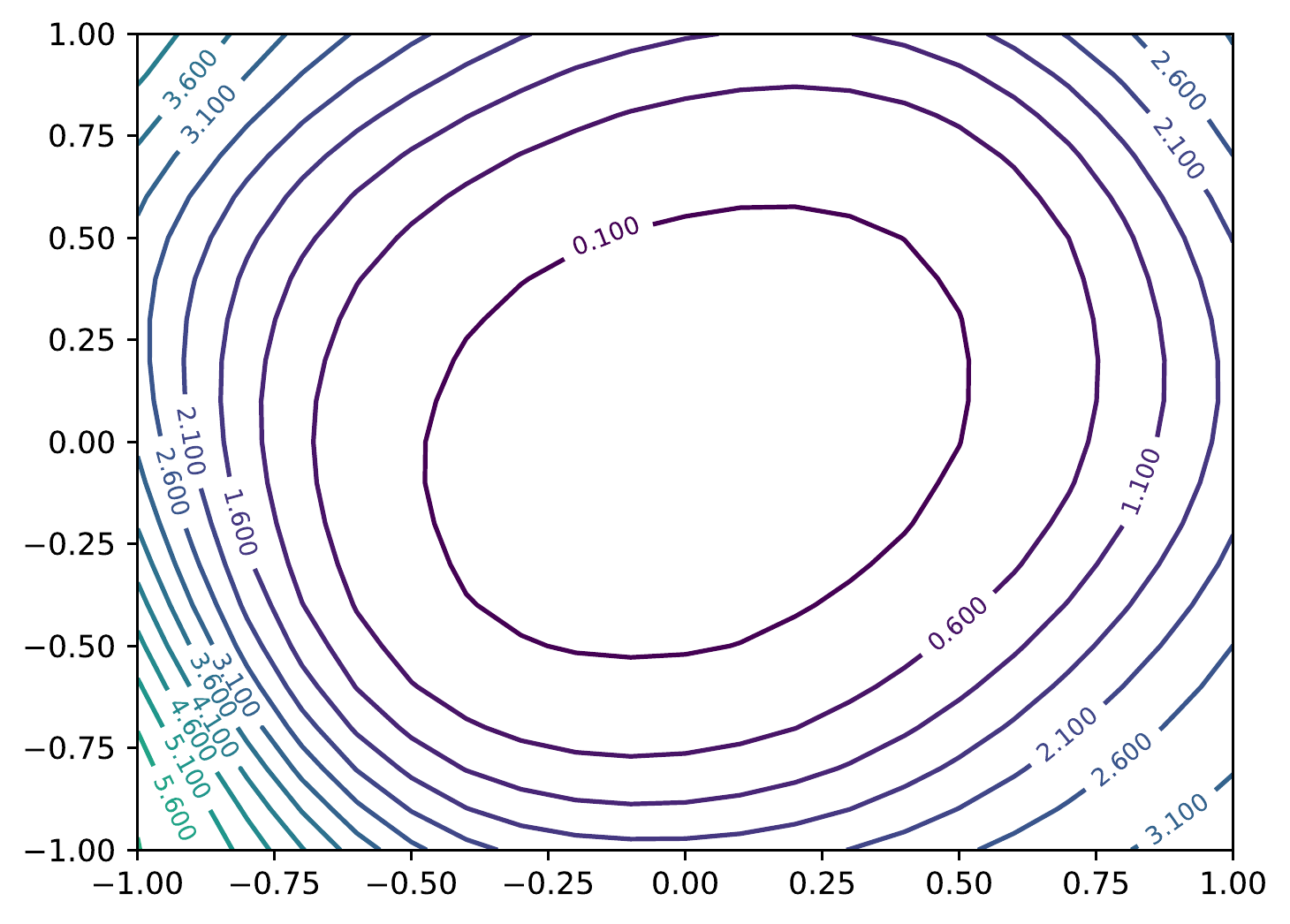}}
\subfigure[$k=8$, 3.93\%]{\includegraphics[width=0.23\linewidth]{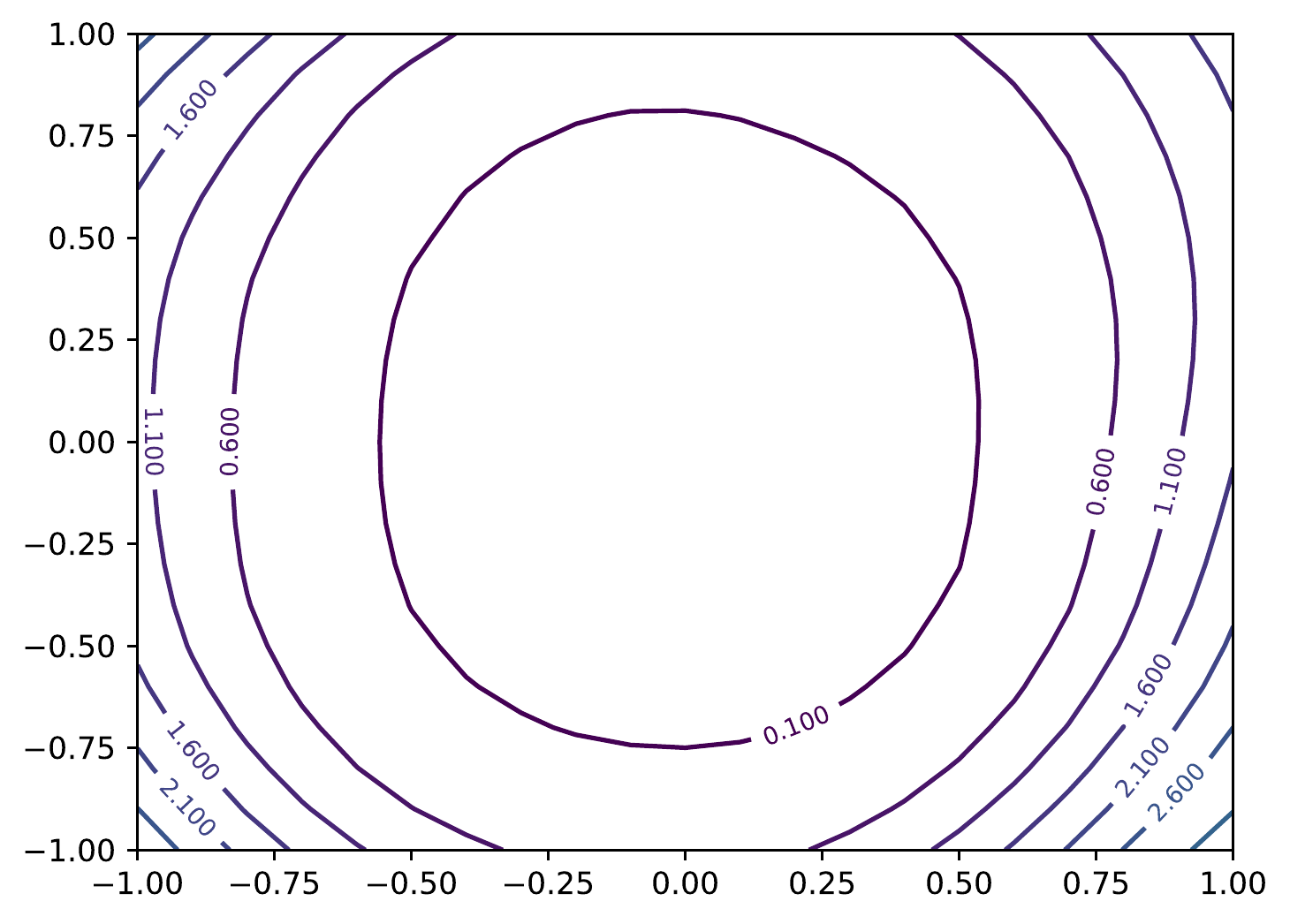}}\\
\hspace{-0.3cm}
\subfigure[$k=1$, 13.31\%]{\includegraphics[width=0.23\linewidth]{figures/{resnet56_noshort_sgd_lr=0.1_bs=128_wd=0.0005/resnet56_noshort_random_-1.0,1.0x-1.0,1.0.h5_2dcontour}.pdf}}
\subfigure[$k=2$, 10.26\%]{\includegraphics[width=0.23\linewidth]{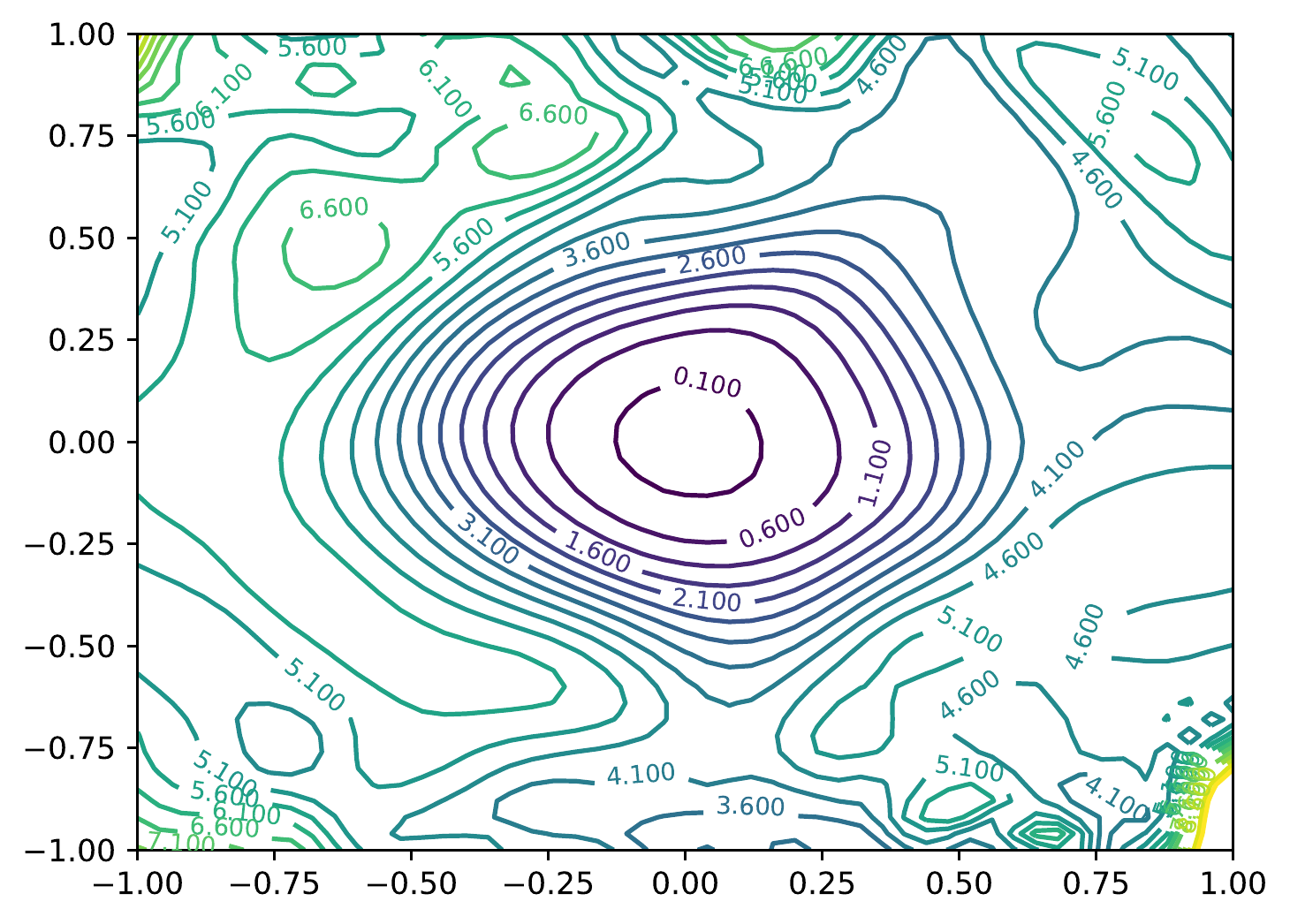}}
\subfigure[$k=4$, 9.69\%]{\includegraphics[width=0.23\linewidth]{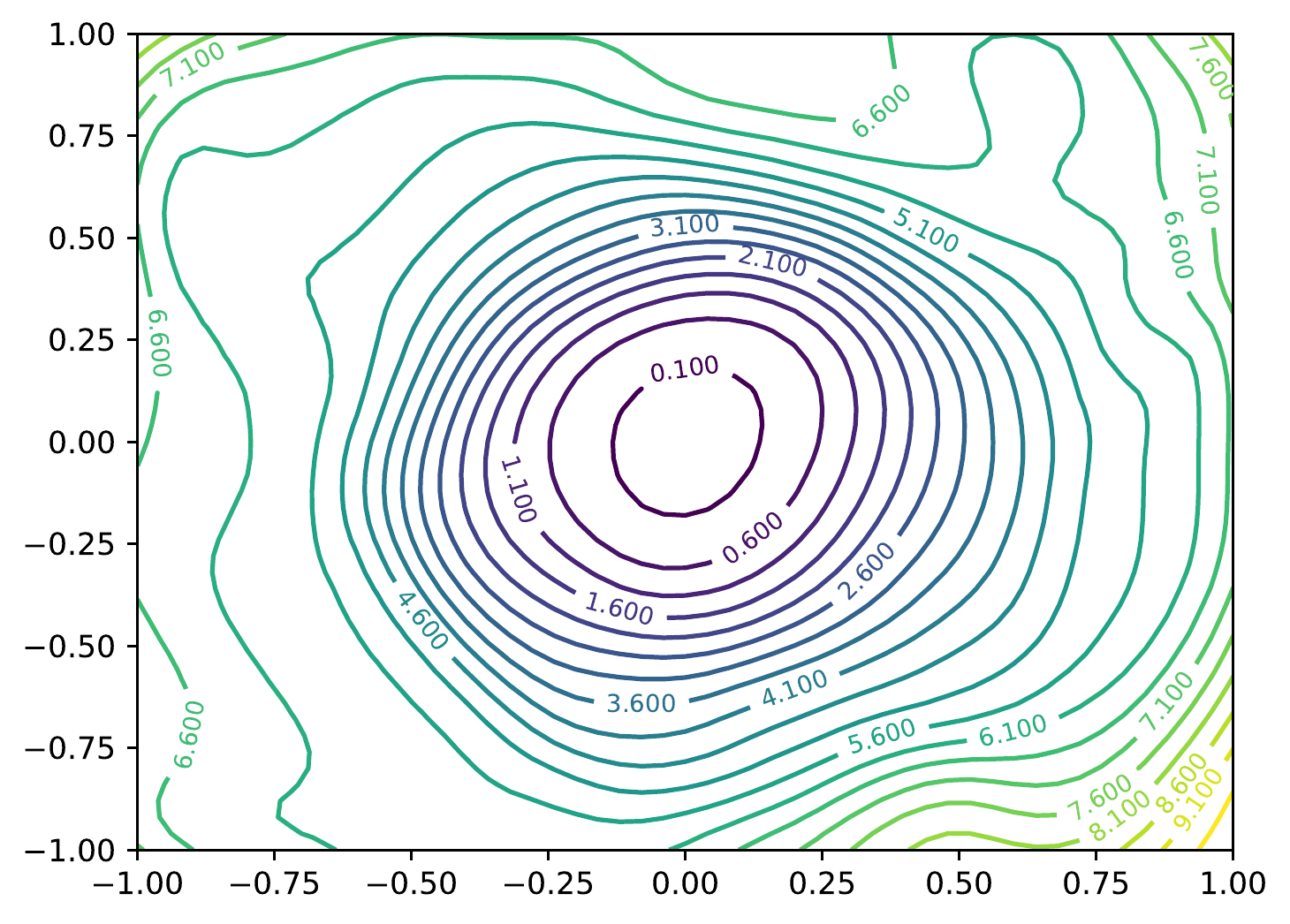}}
\subfigure[$k=8$, 8.70\%]{\includegraphics[width=0.23\linewidth]{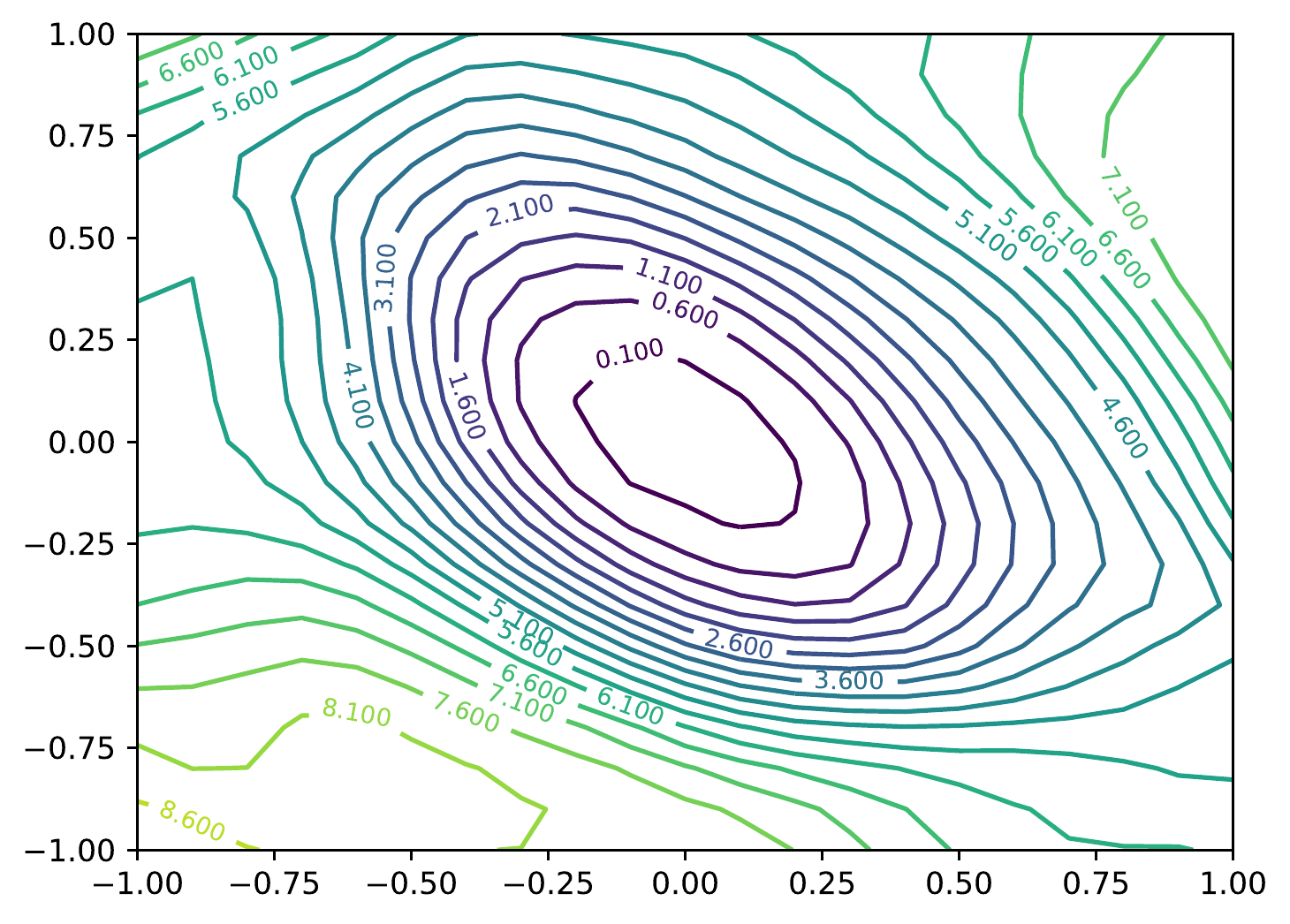}}
\end{tabular}
\caption{Wide-ResNet-56 on CIFAR-10 both with shortcut connections (top) and without (bottom).
The label $k=2$ means twice as many filters per layer. 
Test error is reported below each figure. }
\label{fig:width}
\vspace{-5mm}
\end{figure*}

\paragraph{Experimental Setup}
To understand the effects of network architecture on non-convexity, we trained a number of networks, and plotted the landscape around the obtained minimizers using the filter-normalized random direction method described in Section \ref{sec:norm}.
We consider three classes of neural networks:
1) ResNets~\citep{resnet} that are optimized for performance on CIFAR-10. We consider ResNet-20/56/110, where each name is labeled with the number of layers it has.
2) ``VGG-like'' networks that do not contain shortcut/skip connections.  We produced these networks simply by removing the shortcut connections from ResNets.  We call these networks ResNet-20/56/110-noshort.
3) ``Wide'' ResNets that have more filters per layer than the CIFAR-10 optimized networks.
All models are trained on the CIFAR-10 dataset using SGD with Nesterov momentum, batch-size 128, and 0.0005 weight decay for 300 epochs.
The learning rate was initialized at 0.1, and decreased by a factor of 10 at epochs 150, 225 and 275.
Deeper experimental VGG-like networks (e.g., ResNet-56-noshort, as described below) required a smaller initial learning rate of $0.01$.
High resolution 2D plots of the minimizers for different neural networks are shown in Figure \ref{fig:depth} and Figure \ref{fig:width}.
Results are shown as contour plots rather than surface plots because this makes it extremely easy to see non-convex structures and evaluate sharpness.  For surface plots of ResNet-56, see Figure \ref{fig:surfs}. Note that the center of each plot corresponds to the minimizer, and the two axes parameterize two random directions with filter-wise normalization as in \eqref{plot2d}.
We make several observations below about how architecture affects the loss landscape.

\paragraph{The Effect of Network Depth}
From Figure \ref{fig:depth}, we see that network depth has a dramatic effect
on the loss surfaces of neural networks when skip connections are not used.
The network ResNet-20-noshort has a fairly benign landscape dominated by a
region with convex contours in the center, and no dramatic non-convexity. This
isn't too surprising:  the original VGG networks for ImageNet had 19 layers
and could be trained effectively~\citep{vgg}.
However, as network depth increases, the loss surface of the VGG-like nets spontaneously transitions from (nearly) convex to chaotic.  ResNet-56-noshort has dramatic non-convexities and large regions where the gradient directions (which are normal to the contours depicted in the plots) do not point towards the minimizer at the center.  Also, the loss function becomes extremely large as we move in some directions. ResNet-110-noshort displays even more dramatic non-convexities, and becomes extremely steep as we move in all directions shown in the plot.  Furthermore, note that the minimizers at the center of the deep VGG-like nets seem to be fairly sharp.  In the case of ResNet-56-noshort, the minimizer is also fairly ill-conditioned, as the contours near the minimizer have significant eccentricity.

\paragraph{Shortcut Connections to the Rescue}
Shortcut connections have a dramatic effect of the geometry of the loss functions.   In Figure \ref{fig:depth}, we see that residual connections prevent the transition to chaotic behavior as depth increases.  In fact, the width and shape of the 0.1-level contour is almost identical for the 20- and 110-layer networks.
Interestingly, the effect of skip connections seems to be most important for deep networks.  For the more shallow networks (ResNet-20 and ResNet-20-noshort), the effect of skip connections is fairly unnoticeable.  However residual connections prevent the explosion of non-convexity that occurs when networks get deep.  This effect seems to apply to other kinds of skip connections as well; Figure~\ref{fig:moresurfs}
show the loss landscape of DenseNet \citep{huang2016densely}, which shows no noticeable non-convexity.

\paragraph{Wide Models vs Thin Models}
To see the effect of the number of convolutional filters per layer, we compare the narrow CIFAR-optimized ResNets (ResNet-56) with Wide-ResNets~\cite{zagoruyko2016wide} by multiplying the number of filters per layer by $k=2,4,$ and $8$.
From Figure~\ref{fig:width}, we see that the wider models have loss landscapes with no noticeable chaotic behavior.
Increased network width resulted in flat minima and wide regions of apparent convexity.
We see that increased width prevents chaotic behavior, and skip connections dramatically widen minimizers.
Finally, note that sharpness correlates extremely well with test error.

\paragraph{Implications for Network Initialization}
One interesting property seen in Figure \ref{fig:depth} is that loss landscapes for all the networks considered seem to be partitioned into a well-defined region of low loss value and convex contours, surrounded by a well-defined region of high loss value and non-convex contours.
This partitioning of chaotic and convex regions may explain the importance of good initialization strategies, and also the easy training behavior of ``good'' architectures.
When using normalized random initialization strategies such as those proposed by \citet{glorot2010understanding}, typical neural networks attain an initial loss value less than 2.5.
The well behaved loss landscapes in Figure \ref{fig:depth} (ResNets, and shallow VGG-like nets) are dominated by large, flat, nearly convex attractors that rise to a loss value of 4 or greater.
For such landscapes, a random initialization will likely lie in the ``well- behaved'' loss region, and the optimization algorithm might never ``see'' the pathological non-convexities that occur on the high-loss chaotic plateaus.
Chaotic loss landscapes (ResNet-56/110-noshort) have shallower regions of convexity that rise to lower loss values.
For sufficiently deep networks with shallow enough attractors, the initial iterate will likely lie in the chaotic region where the gradients are uninformative.
In this case, the gradients ``shatter'' \cite{balduzzi2017shattered}, and training is impossible.
SGD was unable to train a 156 layer network without skip connections (even with very low learning rates), which adds weight to this hypothesis.

\paragraph{Landscape Geometry Affects Generalization}
Both Figures \ref{fig:depth} and \ref{fig:width} show that landscape geometry has a dramatic effect on generalization. First, note that visually flatter minimizers consistently correspond to lower test error, which further strengthens our assertion that filter normalization is a natural way to visualize loss function geometry.
Second, we notice that chaotic landscapes (deep networks without skip connections) result in worse training and test error, while more convex landscapes have lower error values.
In fact, the most convex landscapes (Wide-ResNets in the top row of Figure \ref{fig:width}), generalize the best of all, and show no noticeable chaotic behavior.

\paragraph{A note of caution:  Are we really seeing convexity?}
We are viewing the loss surface under a dramatic dimensionality reduction, and we need to be careful how we interpret these plots.
One way to measure the level of convexity in a loss function is to compute the {\em principle curvatures}, which are simply eigenvalues of the Hessian.
A truly convex function has non-negative curvatures (a positive semi-definite Hessian), while a non-convex function has negative curvatures.
It can be shown that the principle curvatures of a dimensionality reduced plot (with random Gaussian directions) are weighted averages of the principle curvatures of the full-dimensional surface (where the weights are Chi-square random variables).

This has several consequences.
First of all, if non-convexity is present in the dimensionality reduced plot, then non-convexity must be present in the full-dimensional surface as well.
However, apparent convexity in the low-dimensional surface does not mean the high-dimensional function is truly convex.
Rather it means that the positive curvatures are dominant (more formally, the {\em mean} curvature, or average eigenvalue, is positive).

\begin{figure*}[h]
\centering
\begin{tabular}{l}
\hspace{-5mm}
\subfigure[Resnet-56]{
\includegraphics[width=0.33\linewidth]{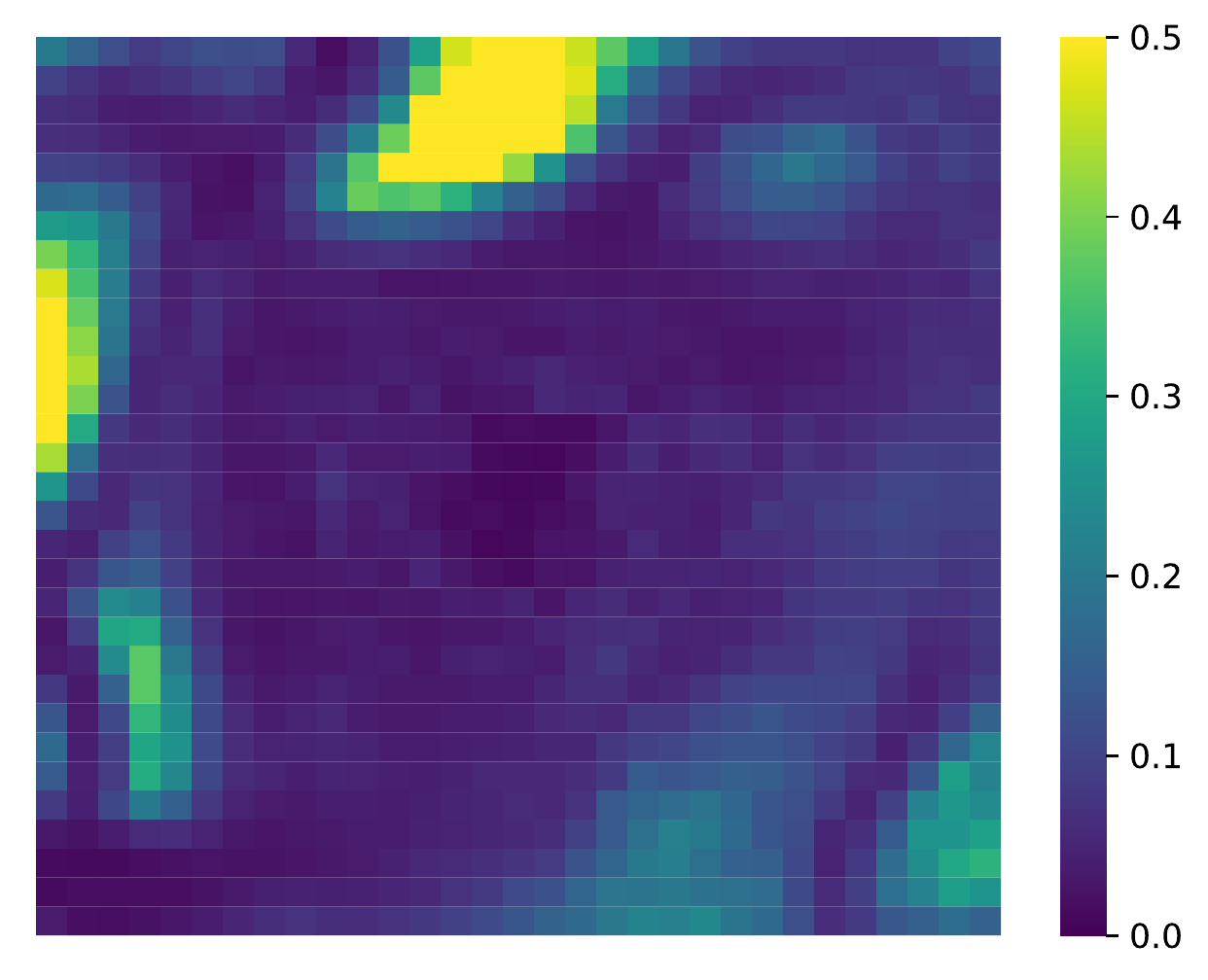}}
\subfigure[Resnet-56-noshort]{
\includegraphics[width=0.33\linewidth]{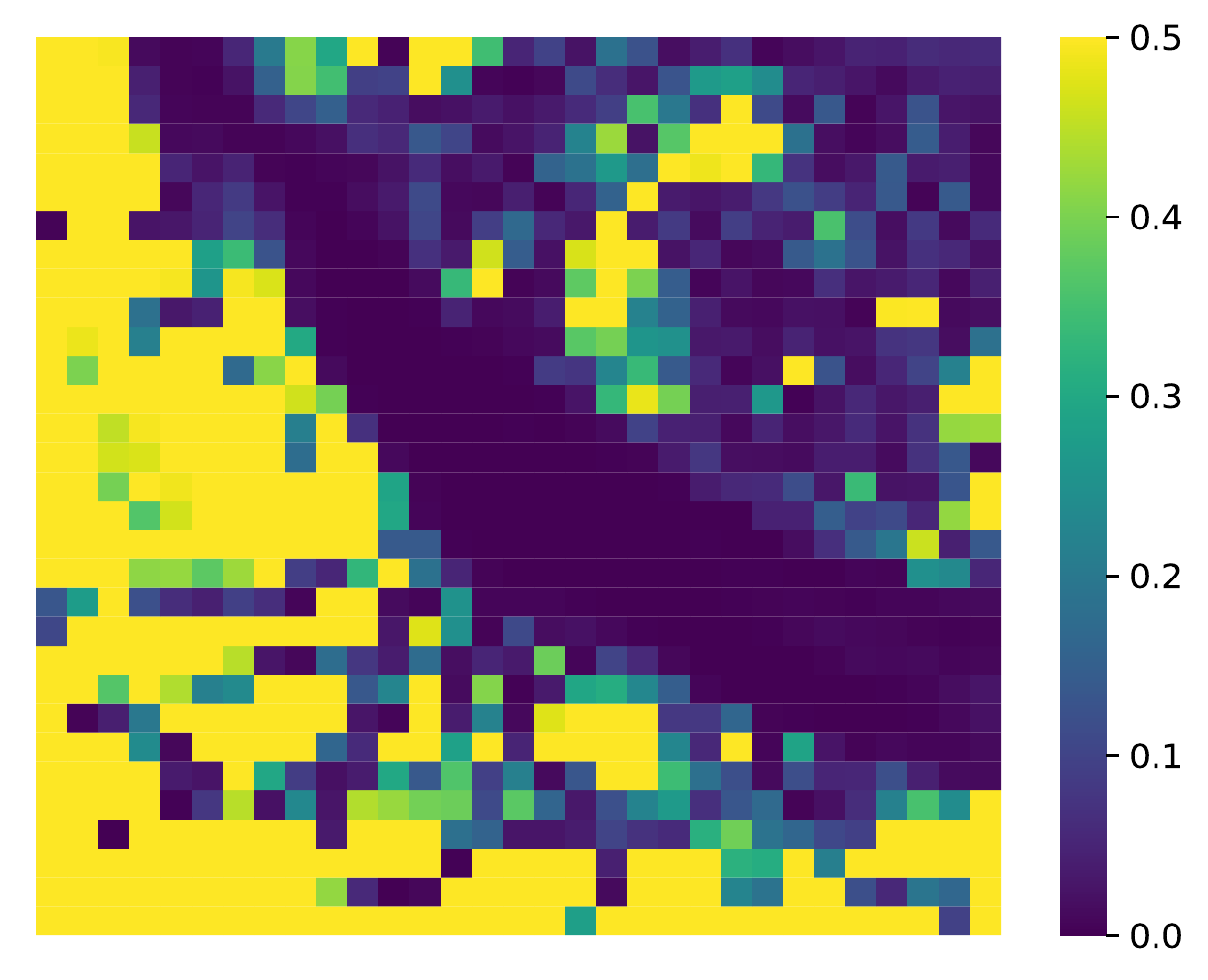}}
\subfigure[DenseNet-121]{
\includegraphics[width=0.33\linewidth]{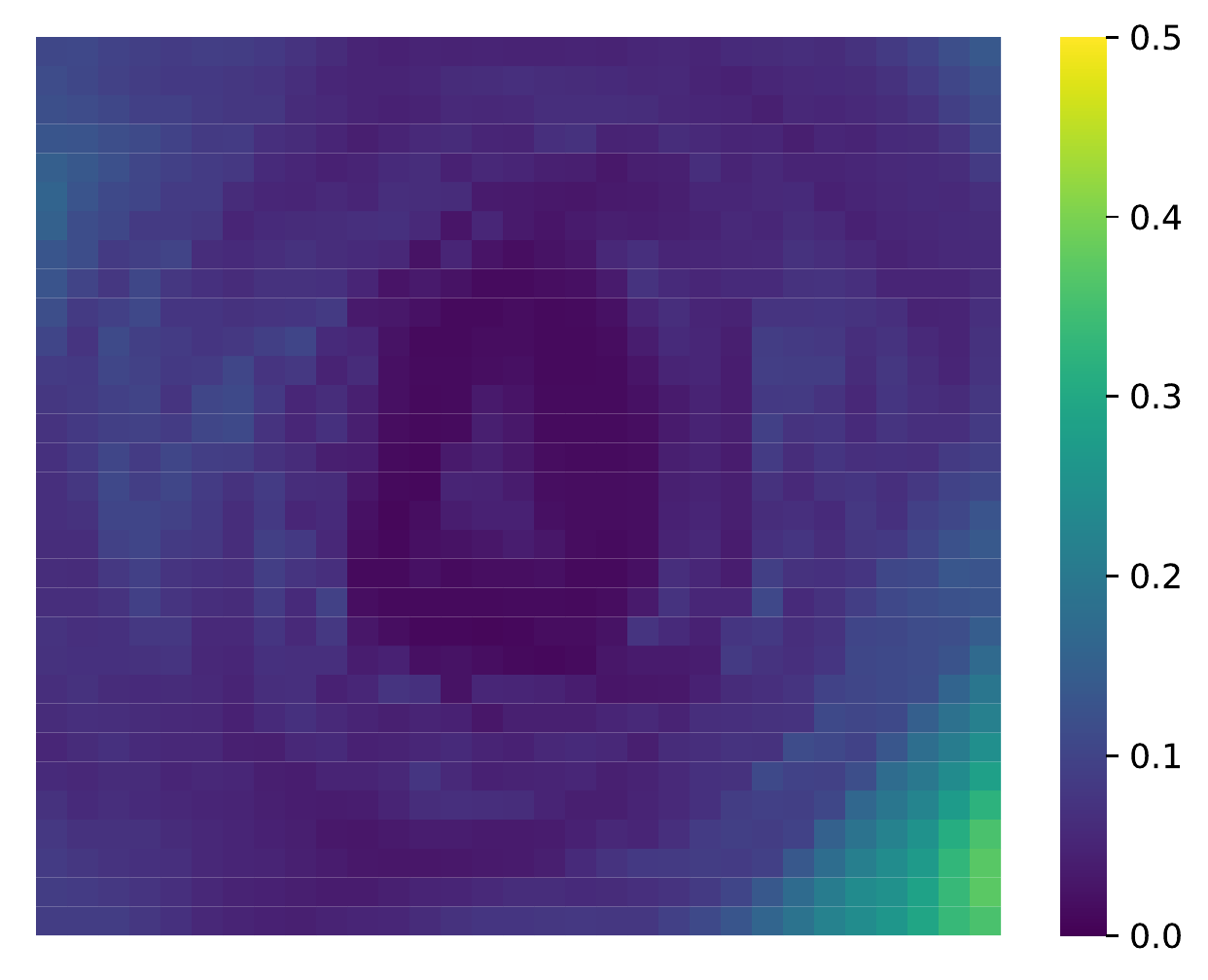}}
\end{tabular}
\vspace{-2mm}
\caption{For each point in the filter-normalized surface plots, we calculate the maximum and minimum eigenvalue of the Hessian, and map the ratio of these two.
}
\label{fig:hessian_eig}
\vspace{-2mm}
\end{figure*}

While this analysis is reassuring, one may still wonder if there is significant ``hidden'' non-convexity that these visualizations fail to capture.
To answer this question, we calculate the {\em minimum} and {\em maximum} eigenvalues of the Hessian, $\lambda_{min}$ and $\lambda_{max}.$\footnote{We compute these using an implicitly restarted Lanczos method that requires only Hessian-vector products (which are calculated directly using automatic differentiation), and does not require an explicit representation of the Hessian or its factorization.}
Figure \ref{fig:hessian_eig} maps the ratio $|\lambda_{min}/\lambda_{max}|$ across the loss surfaces studied above (using the same minimizer and the same random directions).
Blue color indicates a more convex region (near-zero negative eigenvalues relative to the positive eigenvalues), while yellow indicates significant levels of negative curvature.
We see that the convex-looking regions in our surface plots do indeed correspond to regions with insignificant negative eigenvalues (i.e., there are not major non-convex features that the plot missed), while chaotic regions contain large negative curvatures.
For convex-looking surfaces like DenseNet, the negative eigenvalues remain extremely small (less than 1\% the size of the positive curvatures) over a large region of the plot.

\section{Visualizing Optimization Paths}
\label{sec:trajectory}
Finally, we explore methods for visualizing the trajectories of different optimizers.  For this application, random directions are ineffective.  We will provide a theoretical explanation for why random directions fail, and explore methods for effectively plotting trajectories on top of loss function contours.

\begin{figure*}[h]
\centering
\begin{tabular}{l}
\hspace{-.5cm}
 \subfigure[Two random directions]{
 \includegraphics[height=0.23\linewidth]{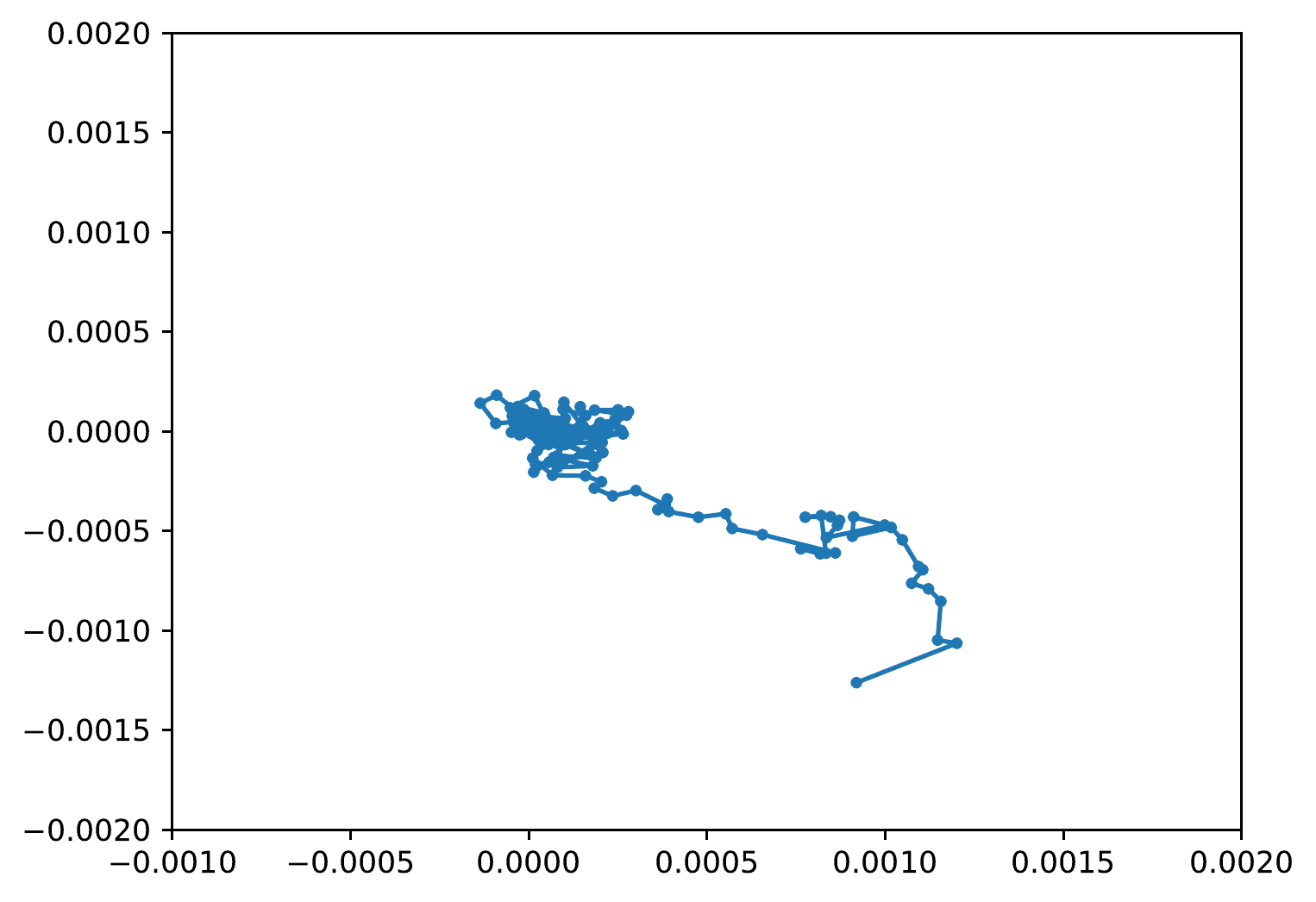}
 \label{fig:other_2d_visulization:random}
 }
 \subfigure[Random direction for y-axis]{
 \includegraphics[height=0.23\linewidth]{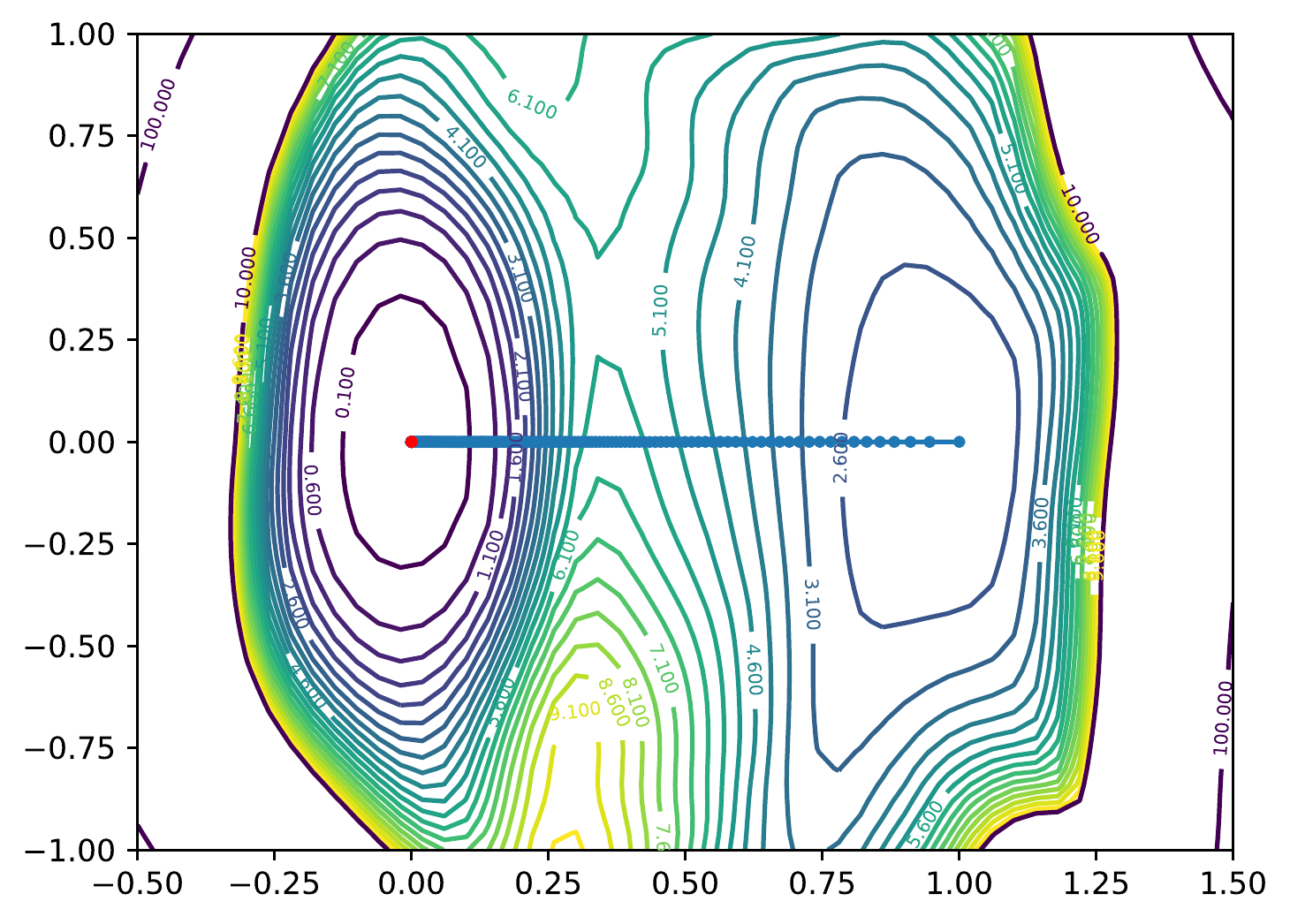}
 \label{fig:other_2d_visulization:normal}
 }
 \subfigure[Enlarged version (b)]{
 \includegraphics[height=0.23\linewidth]{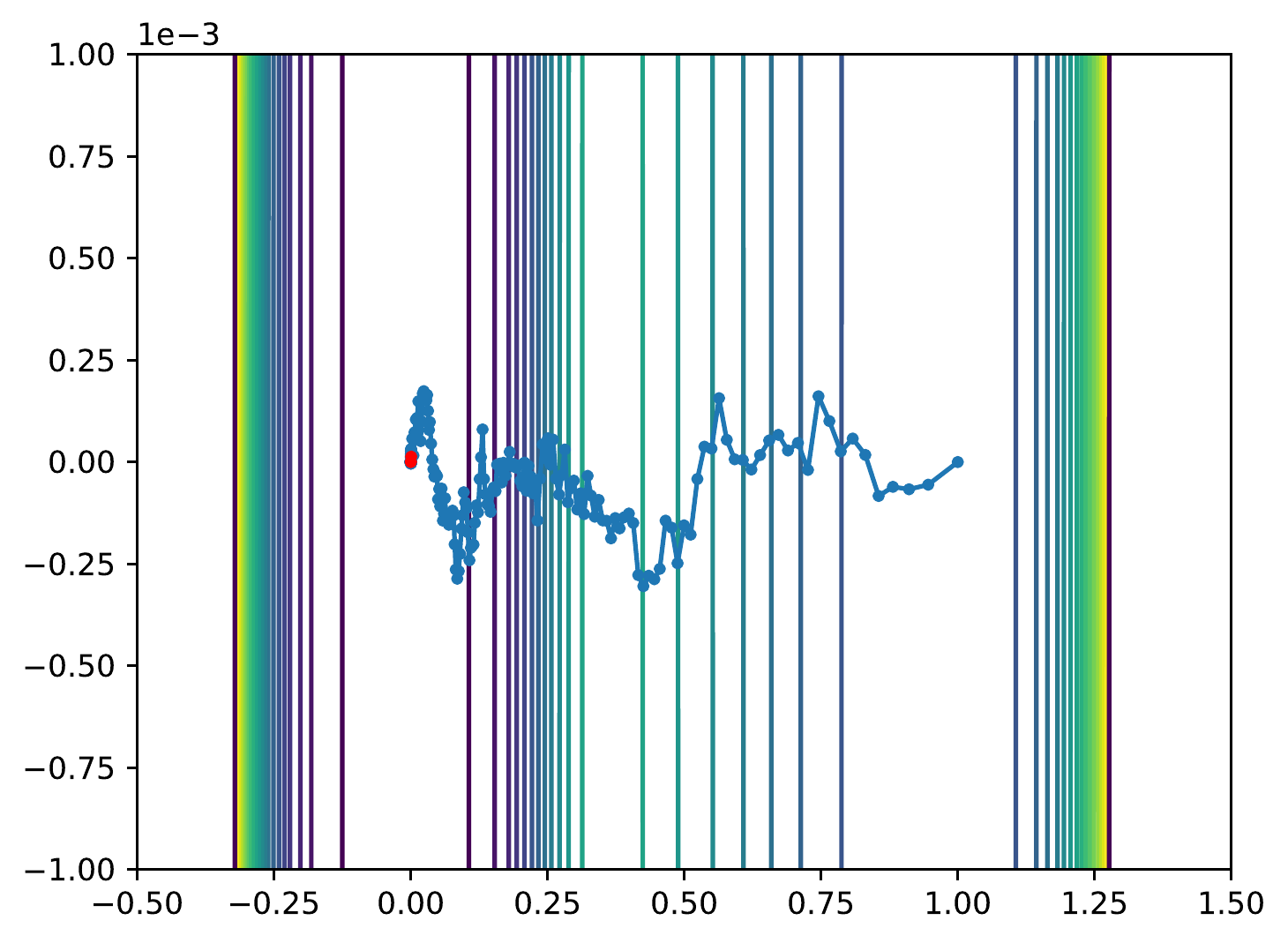}
 \label{fig:other_2d_visulization:enlarge}
 }
\end{tabular}
\caption{Ineffective visualizations of optimizer trajectories.  These visualizations suffer from the orthogonality of random directions in high dimensions.}
\label{fig:bad_trajectories}
\end{figure*}

 Several authors have observed that random directions fail to capture the variation in optimization trajectories, including  \cite{gallagher2003visualization,viztrajectory,lipton2016stuck,liao2017theory}.
 Example failed visualizations are depicted in Figure~\ref{fig:bad_trajectories}.
 In Figure~\ref{fig:other_2d_visulization:random}, we see the iterates of SGD projected onto the plane defined by two random directions.
 Almost none of the motion is captured (notice the super-zoomed-in axes and the seemingly random walk).
 This problem was noticed by \cite{goodfellow2014qualitatively}, who then visualized trajectories using one direction that points from initialization to solution, and one random direction.
 This approach is shown in Figure~\ref{fig:other_2d_visulization:normal}.
 As seen in Figure~\ref{fig:other_2d_visulization:enlarge}, the random axis captures almost no variation, leading to the (misleading) appearance of a straight line path.

\subsection{Why Random Directions Fail:  Low-Dimensional Optimization Trajectories}
It is well-known that two random vectors in a high dimensional space will be nearly orthogonal with high probability. In fact, the expected cosine similarity between Gaussian random vectors in $n$ dimensions is roughly $\sqrt{2/(\pi n)}$ (\cite{goldstein2016phasemax}, Lemma 5).

This is problematic when optimization trajectories lie in extremely low dimensional spaces.  In this case, a randomly chosen vector will lie orthogonal to the low-rank space containing the optimization path, and a projection onto a random direction will capture almost no variation. Figure~\ref{fig:other_2d_visulization:normal} suggests that optimization trajectories are low-dimensional because the random direction captures orders of magnitude less variation than the vector that points along the optimization path.   Below, we use PCA directions to directly validate this low dimensionality, and also to produce effective visualizations.

\subsection{Effective Trajectory Plotting using PCA Directions}
To capture variation in trajectories, we need to use non-random (and carefully chosen) directions.
Here, we suggest an approach based on PCA that allows us to measure how much variation we've captured; we also provide plots of these trajectories along the contours of the loss surface.

\begin{figure*}[b!]
\centering
\begin{tabular}{l}
\hspace{-.5cm}
 \subfigure[SGD,WD=5e-4]{
 \includegraphics[width=0.25\linewidth]{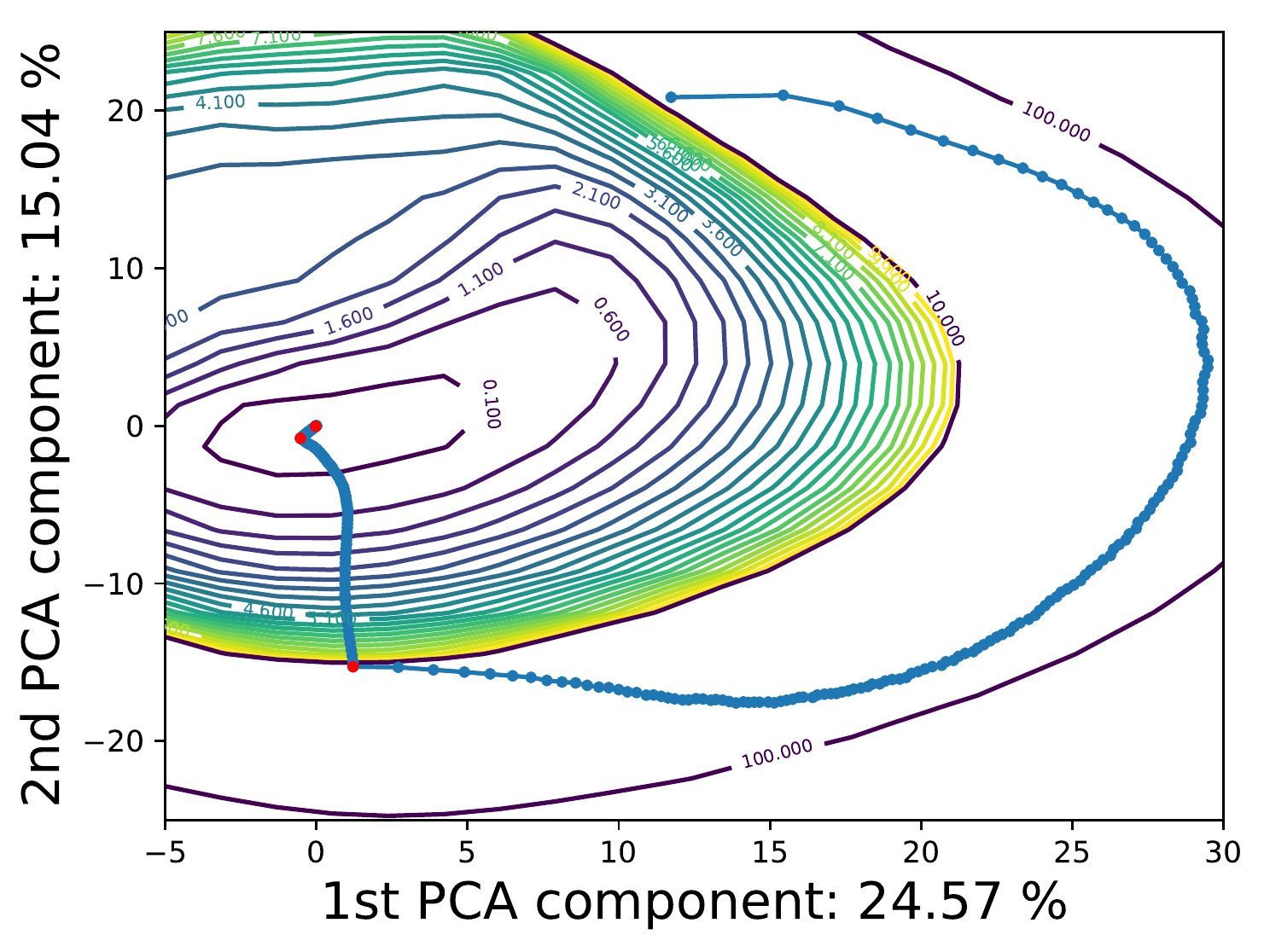}
 \includegraphics[width=0.25\linewidth]{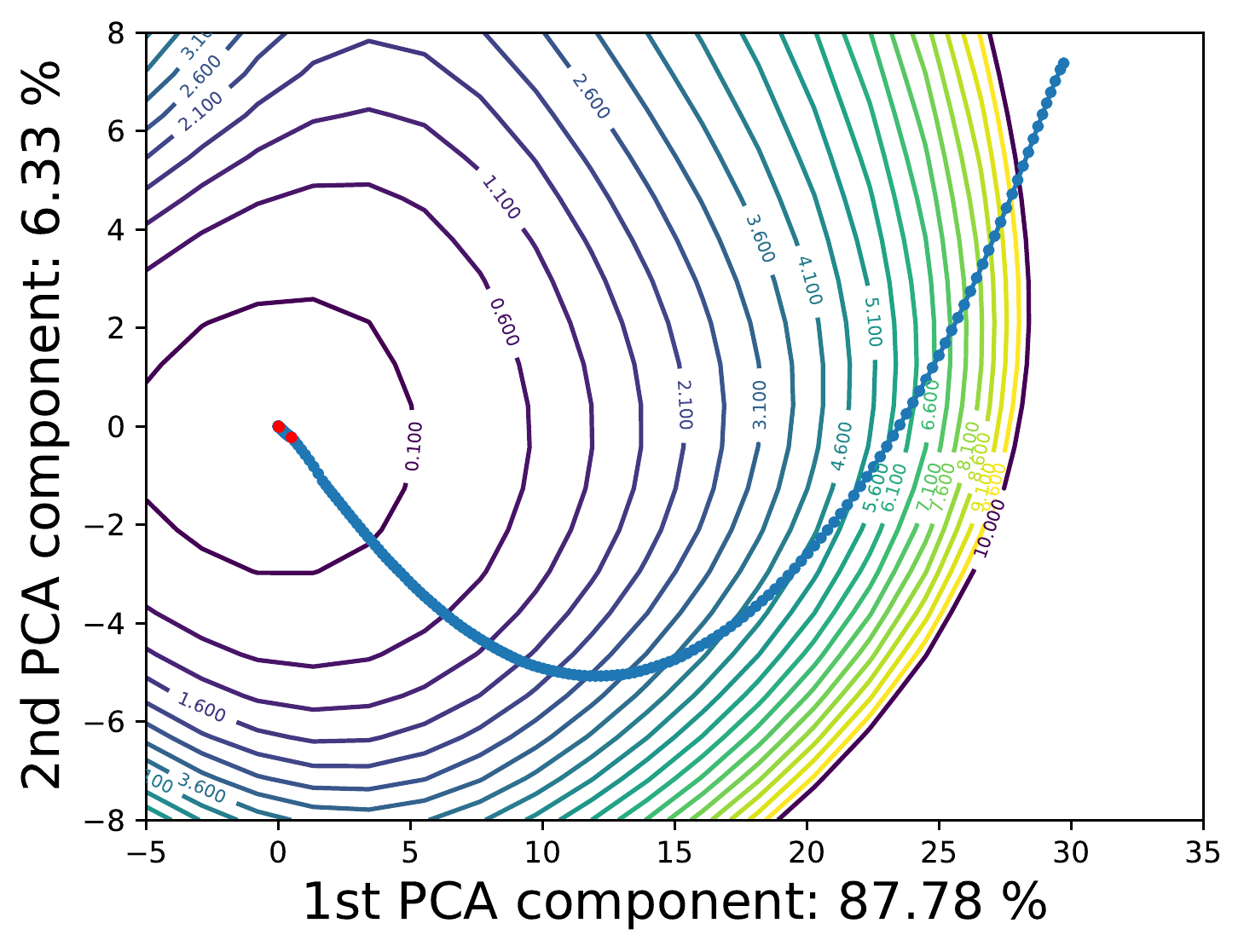}}
 \subfigure[Adam, WD=5e-4]{
 \includegraphics[width=0.25\linewidth]{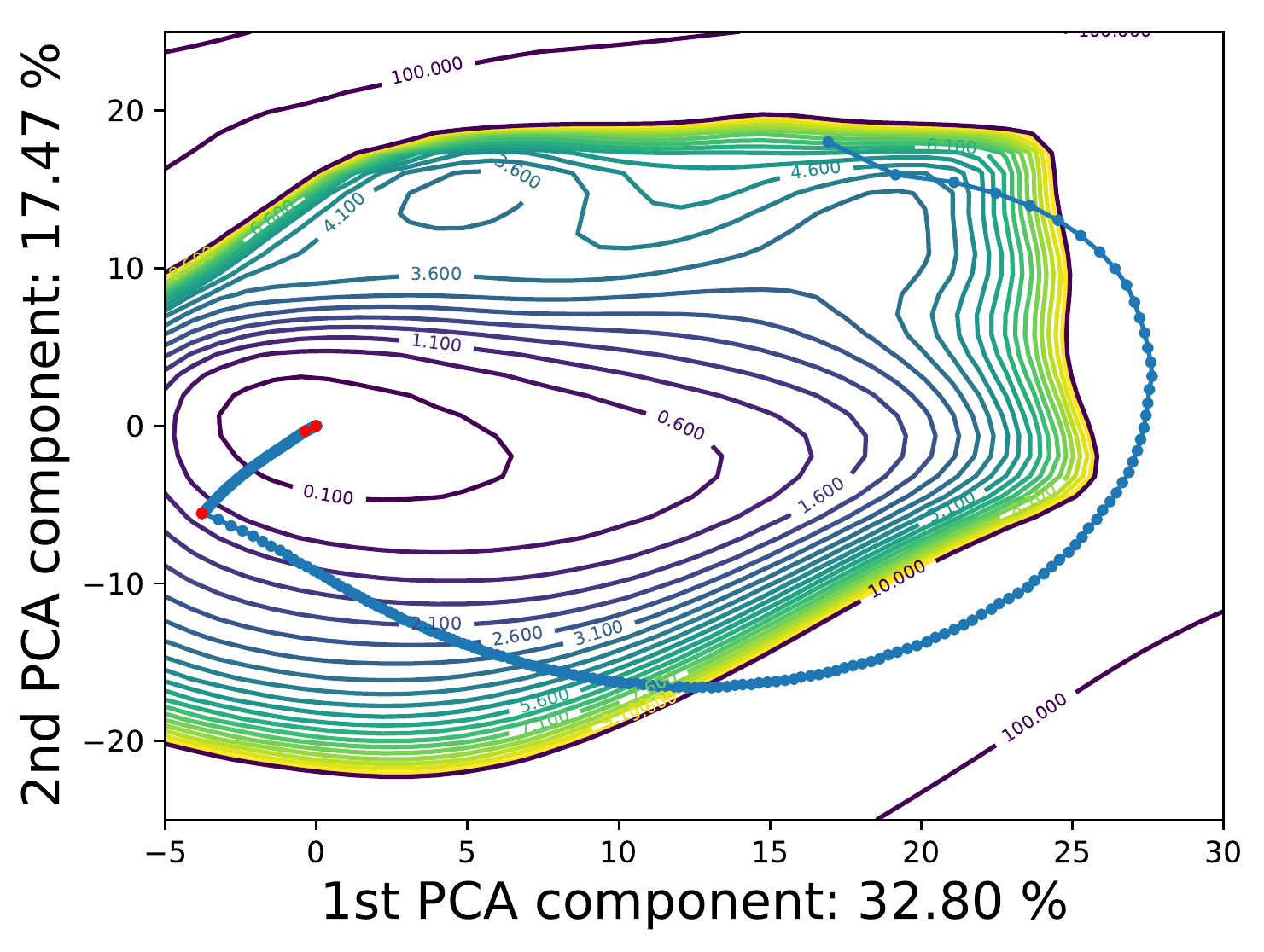}
 \includegraphics[width=0.25\linewidth]{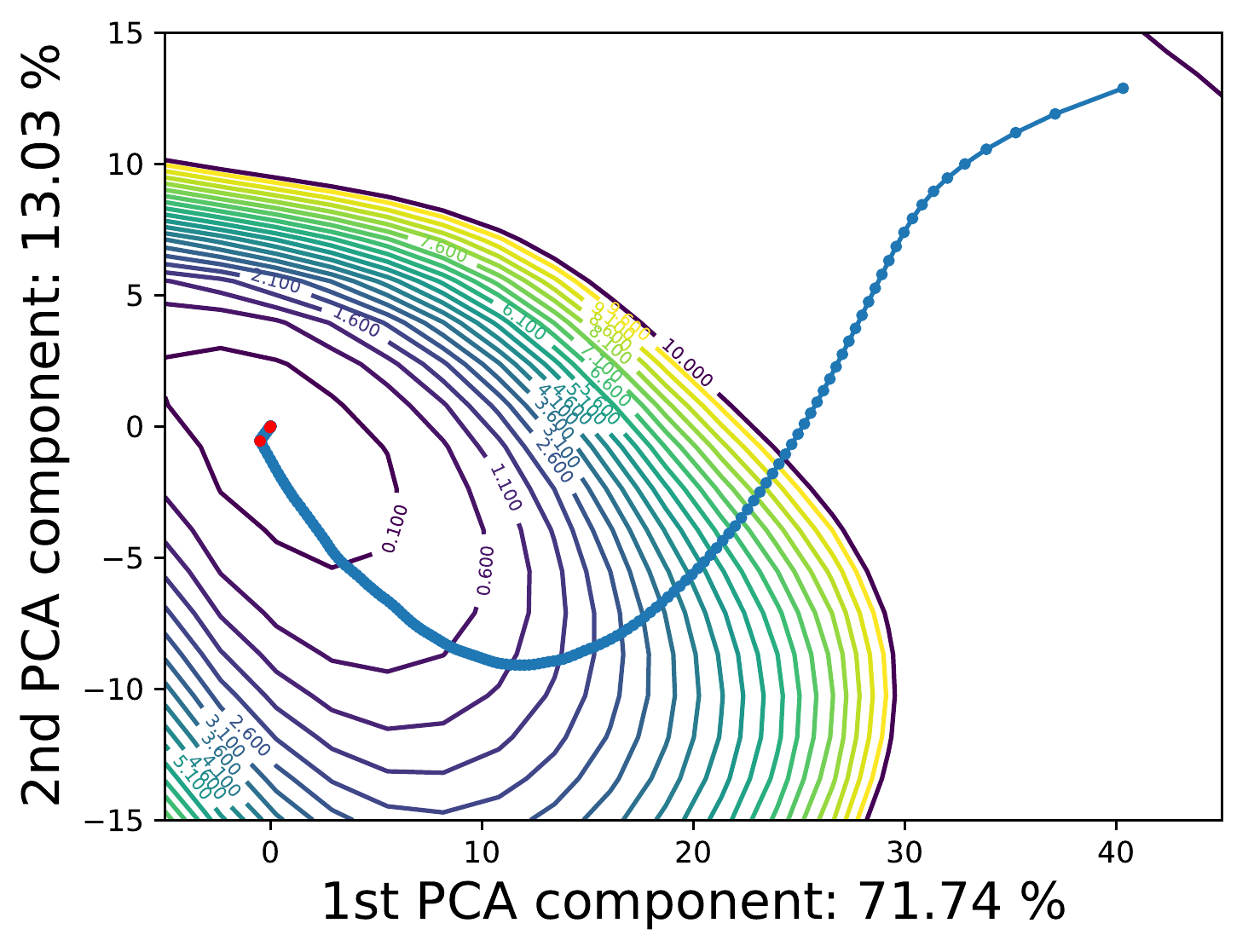}
 }\\
 \hspace{-.5cm}
 \subfigure[SGD, WD=0]{
 \includegraphics[width=0.25\linewidth]{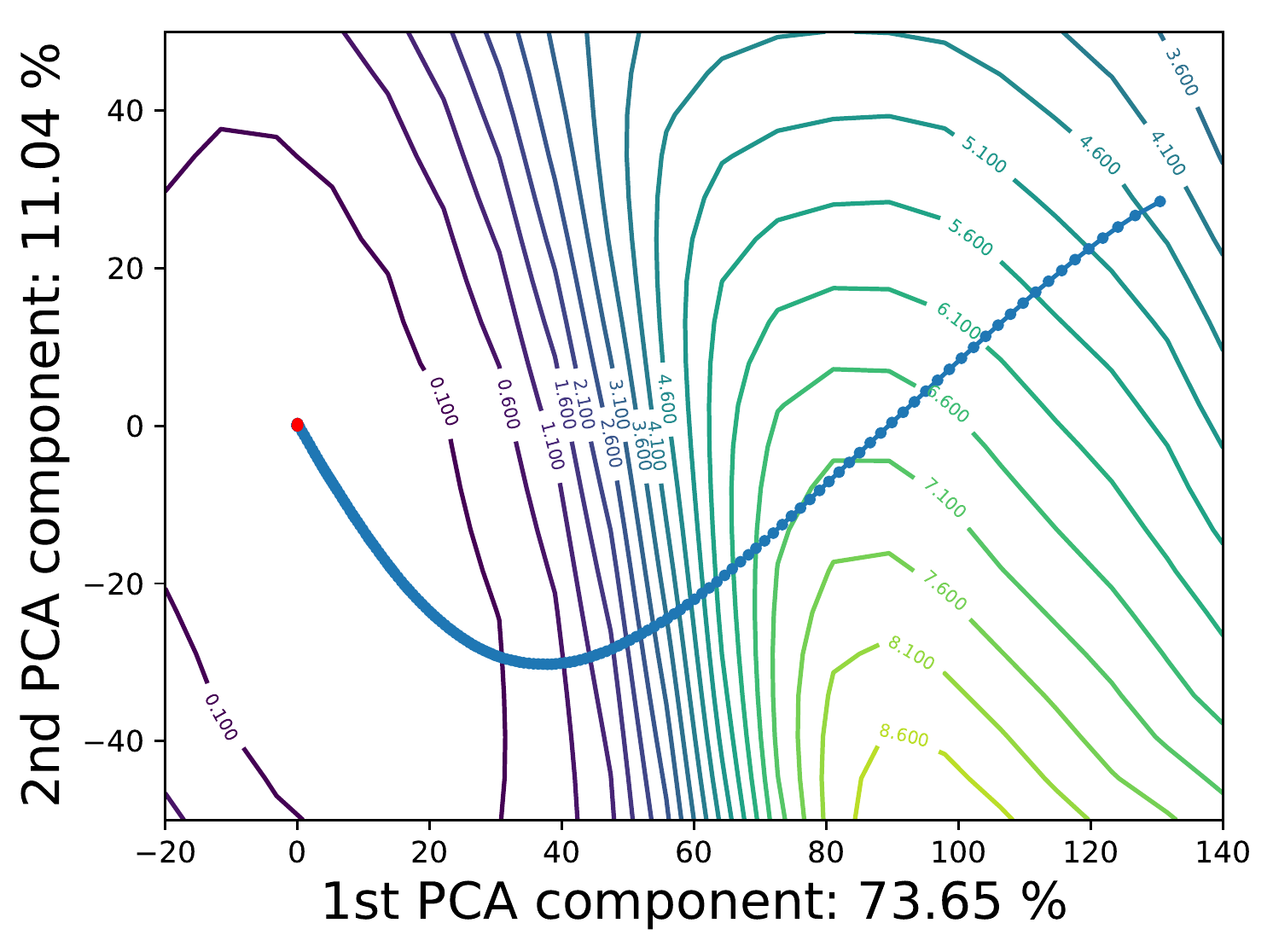}
 \includegraphics[width=0.25\linewidth]{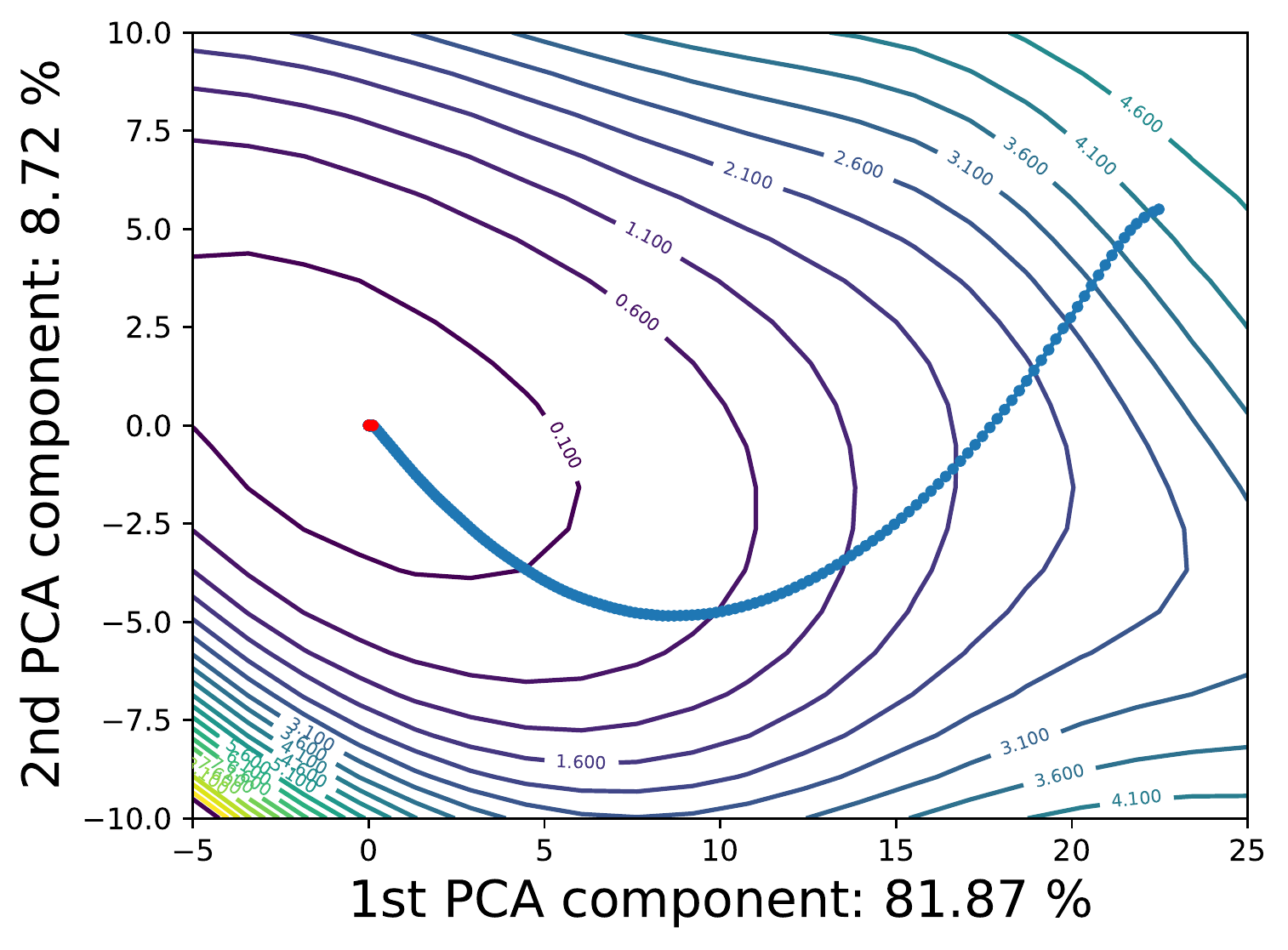}}
 \subfigure[Adam,WD=0]{
 \includegraphics[width=0.25\linewidth]{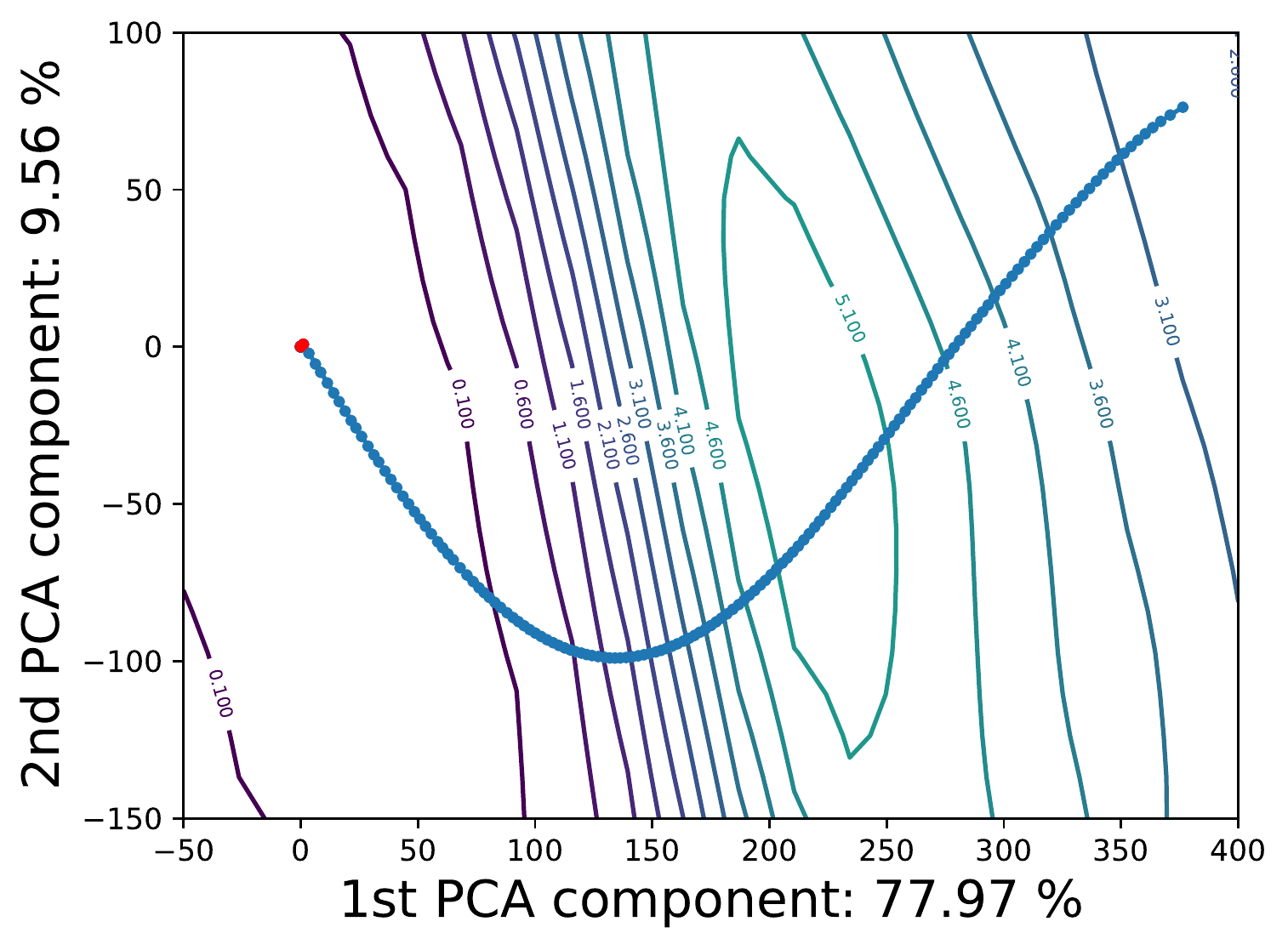}
 \includegraphics[width=0.25\linewidth]{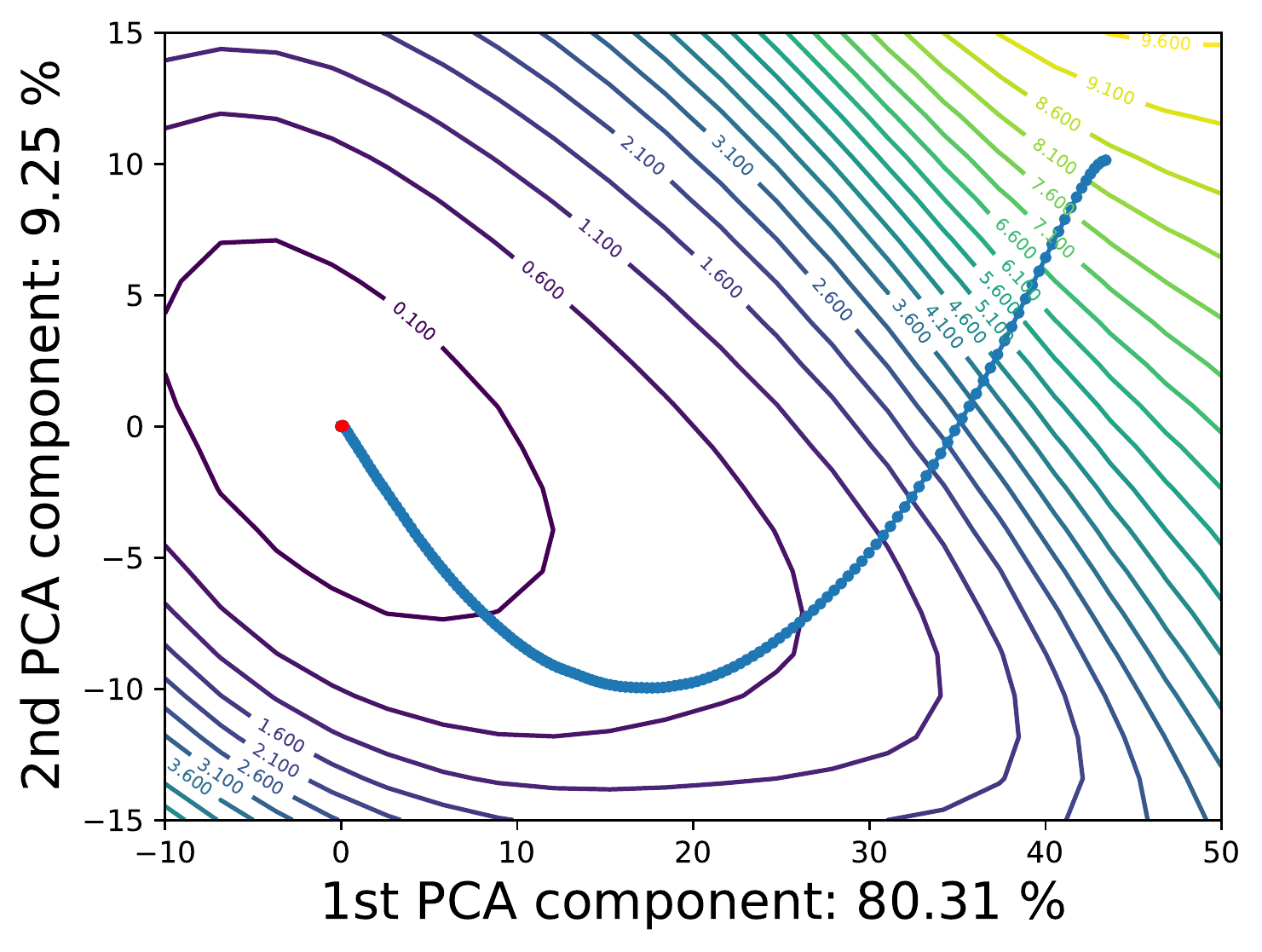}}
\end{tabular}
\caption{Projected learning trajectories use normalized PCA directions for VGG-9.
The left plot in each subfigure uses batch size 128, and the right one uses batch size 8192.}
\label{fig:pca_projections}
\end{figure*}

Let $\theta_i$ denote model parameters at epoch $i,$ and the final parameters after $n$ epochs of training are denoted $\theta_n$.
Given $n$ training epochs, we can apply PCA to the matrix
$M = [\theta_0 - \theta_n; \cdots; \theta_{n-1} - \theta_n],$
and then select the two most explanatory directions.
Optimizer trajectories (blue dots) and loss surfaces along PCA directions are shown in Figure~\ref{fig:pca_projections}.
Epochs where the learning rate was decreased are shown as red dots. On each axis, we measure the amount of variation in the descent path captured by that PCA direction.

At early stages of training, the paths tend to move perpendicular to the contours of the loss surface, i.e., along the gradient directions as one would expect from non-stochastic gradient descent.  The stochasticity becomes fairly pronounced in several plots during the later stages of training.  This is particularly true of the plots that use weight decay and small batches (which leads to more gradient noise, and a more radical departure from deterministic gradient directions).  When weight decay and small batches are used, we see the path turn nearly parallel to the contours and ``orbit'' the solution when the stepsize is large.  When the stepsize is dropped (at the red dot), the effective noise in the system decreases, and we see a kink in the path as the trajectory falls into the nearest local minimizer.

Finally, we can directly observe that the descent path is very low dimensional: between 40\% and 90\% of the variation in the descent paths lies in a space of only 2 dimensions.  The optimization trajectories in Figure~\ref{fig:pca_projections} appear to be dominated by movement in the direction of a nearby attractor.
This low dimensionality is compatible with the observations in Section \ref{sec:exp_different_networks}, where we observed that non-chaotic landscapes are dominated by wide, nearly convex minimizers.

\section{Conclusion}
We presented a
visualization technique that provides insights into the consequences of a
variety of choices facing the neural network practitioner, including network
architecture, optimizer selection, and batch size.
Neural networks have advanced dramatically in recent years, largely on the
back of anecdotal knowledge and theoretical results with complex assumptions.  
For progress to continue to be made, a more general understanding of the structure of neural networks is needed.
Our hope is that effective visualization, when coupled with continued advances in theory, can result in faster
training, simpler models, and better generalization.

\section*{Acknowledgements}
Li, Xu, and Goldstein were supported by the Office of Naval Research (N00014-17-1-2078), DARPA Lifelong Learning Machines (FA8650-18-2-7833), the DARPA YFA program (D18AP00055), and the Sloan Foundation.  Taylor was supported by the ONR (N0001418WX01582), and the DOD HPC Modernization Program.
Studer was supported in part by Xilinx, Inc. and by the US National Science Foundation (NSF) under grants ECCS-1408006, CCF-1535897, CCF-1652065, CNS-1717559, and ECCS-1824379.
\small
\bibliographystyle{plainnat}
\bibliography{reference}

\begin{thebibliography}{42}
\providecommand{\natexlab}[1]{#1}
\providecommand{\url}[1]{\texttt{#1}}
\expandafter\ifx\csname urlstyle\endcsname\relax
  \providecommand{\doi}[1]{doi: #1}\else
  \providecommand{\doi}{doi: \begingroup \urlstyle{rm}\Url}\fi

\bibitem[Balduzzi et~al.(2017)Balduzzi, Frean, Leary, Lewis, Ma, and
  McWilliams]{balduzzi2017shattered}
David Balduzzi, Marcus Frean, Lennox Leary, JP~Lewis, Kurt Wan-Duo Ma, and
  Brian McWilliams.
\newblock The shattered gradients problem: If resnets are the answer, then what
  is the question?
\newblock In \emph{ICML}, 2017.

\bibitem[Blum and Rivest(1989)]{blum1989training}
Avrim Blum and Ronald~L Rivest.
\newblock Training a 3-node neural network is np-complete.
\newblock In \emph{NIPS}, 1989.

\bibitem[Chaudhari et~al.(2017)Chaudhari, Choromanska, Soatto, and
  LeCun]{entropysgd}
Pratik Chaudhari, Anna Choromanska, Stefano Soatto, and Yann LeCun.
\newblock Entropy-sgd: Biasing gradient descent into wide valleys.
\newblock In \emph{ICLR}, 2017.

\bibitem[Choromanska et~al.(2015)Choromanska, Henaff, Mathieu, Arous, and
  LeCun]{choromanska2015loss}
Anna Choromanska, Mikael Henaff, Michael Mathieu, G{\'e}rard~Ben Arous, and
  Yann LeCun.
\newblock The loss surfaces of multilayer networks.
\newblock In \emph{AISTATS}, 2015.

\bibitem[Dauphin et~al.(2014)Dauphin, Pascanu, Gulcehre, Cho, Ganguli, and
  Bengio]{dauphin2014identifying}
Yann~N Dauphin, Razvan Pascanu, Caglar Gulcehre, Kyunghyun Cho, Surya Ganguli,
  and Yoshua Bengio.
\newblock Identifying and attacking the saddle point problem in
  high-dimensional non-convex optimization.
\newblock In \emph{NIPS}, 2014.

\bibitem[De et~al.(2017)De, Yadav, Jacobs, and Goldstein]{de2017automated}
Soham De, Abhay Yadav, David Jacobs, and Tom Goldstein.
\newblock Automated inference with adaptive batches.
\newblock In \emph{AISTATS}, 2017.

\bibitem[Dinh et~al.(2017)Dinh, Pascanu, Bengio, and Bengio]{dinh2017sharp}
Laurent Dinh, Razvan Pascanu, Samy Bengio, and Yoshua Bengio.
\newblock Sharp minima can generalize for deep nets.
\newblock In \emph{ICML}, 2017.

\bibitem[Dziugaite and Roy(2017)]{dziugaite2017computing}
Gintare~Karolina Dziugaite and Daniel~M Roy.
\newblock Computing nonvacuous generalization bounds for deep (stochastic)
  neural networks with many more parameters than training data.
\newblock In \emph{UAI}, 2017.

\bibitem[Freeman and Bruna(2017)]{freeman2016topology}
C~Daniel Freeman and Joan Bruna.
\newblock Topology and geometry of half-rectified network optimization.
\newblock In \emph{ICLR}, 2017.

\bibitem[Gallagher and Downs(2003)]{gallagher2003visualization}
Marcus Gallagher and Tom Downs.
\newblock Visualization of learning in multilayer perceptron networks using
  principal component analysis.
\newblock \emph{IEEE Transactions on Systems, Man, and Cybernetics, Part B
  (Cybernetics)}, 33\penalty0 (1):\penalty0 28--34, 2003.

\bibitem[Glorot and Bengio(2010)]{glorot2010understanding}
Xavier Glorot and Yoshua Bengio.
\newblock Understanding the difficulty of training deep feedforward neural
  networks.
\newblock In \emph{AISTATS}, 2010.

\bibitem[Goldstein and Studer(2016)]{goldstein2016phasemax}
Tom Goldstein and Christoph Studer.
\newblock Phasemax: Convex phase retrieval via basis pursuit.
\newblock \emph{arXiv preprint arXiv:1610.07531}, 2016.

\bibitem[Goodfellow et~al.(2015)Goodfellow, Vinyals, and
  Saxe]{goodfellow2014qualitatively}
Ian~J Goodfellow, Oriol Vinyals, and Andrew~M Saxe.
\newblock Qualitatively characterizing neural network optimization problems.
\newblock In \emph{ICLR}, 2015.

\bibitem[Goyal et~al.(2017)Goyal, Doll{\'a}r, Girshick, Noordhuis, Wesolowski,
  Kyrola, Tulloch, Jia, and He]{goyal2017accurate}
Priya Goyal, Piotr Doll{\'a}r, Ross Girshick, Pieter Noordhuis, Lukasz
  Wesolowski, Aapo Kyrola, Andrew Tulloch, Yangqing Jia, and Kaiming He.
\newblock Accurate, large minibatch sgd: Training imagenet in 1 hour.
\newblock \emph{arXiv preprint arXiv:1706.02677}, 2017.

\bibitem[Haeffele and Vidal(2017)]{haeffele2017global}
Benjamin~D Haeffele and Ren{\'e} Vidal.
\newblock Global optimality in neural network training.
\newblock In \emph{CVPR}, 2017.

\bibitem[Hardt and Ma(2017)]{hardt2016identity}
Moritz Hardt and Tengyu Ma.
\newblock Identity matters in deep learning.
\newblock In \emph{ICLR}, 2017.

\bibitem[He et~al.(2016)He, Zhang, Ren, and Sun]{resnet}
Kaiming He, Xiangyu Zhang, Shaoqing Ren, and Jian Sun.
\newblock {Deep Residual Learning for Image Recognition}.
\newblock In \emph{CVPR}, 2016.

\bibitem[Hochreiter and Schmidhuber(1997)]{hochreiter1997flat}
Sepp Hochreiter and J{\"u}rgen Schmidhuber.
\newblock Flat minima.
\newblock \emph{Neural Computation}, 9\penalty0 (1):\penalty0 1--42, 1997.

\bibitem[Hoffer et~al.(2017)Hoffer, Hubara, and Soudry]{hoffer2017train}
Elad Hoffer, Itay Hubara, and Daniel Soudry.
\newblock Train longer, generalize better: closing the generalization gap in
  large batch training of neural networks.
\newblock \emph{NIPS}, 2017.

\bibitem[Huang et~al.(2017)Huang, Liu, Weinberger, and van~der
  Maaten]{huang2016densely}
Gao Huang, Zhuang Liu, Kilian~Q Weinberger, and Laurens van~der Maaten.
\newblock Densely connected convolutional networks.
\newblock In \emph{CVPR}, 2017.

\bibitem[Im et~al.(2016)Im, Tao, and Branson]{im2016empirical}
Daniel~Jiwoong Im, Michael Tao, and Kristin Branson.
\newblock An empirical analysis of deep network loss surfaces.
\newblock \emph{arXiv preprint arXiv:1612.04010}, 2016.

\bibitem[Ioffe and Szegedy(2015)]{batchnorm}
Sergey Ioffe and Christian Szegedy.
\newblock {Batch Normalization: Accelerating Deep Network Training by Reducing
  Internal Covariate Shift}.
\newblock In \emph{ICML}, 2015.

\bibitem[Kawaguchi et~al.(2017)Kawaguchi, Kaelbling, and
  Bengio]{kawaguchi2017generalization}
Kenji Kawaguchi, Leslie~Pack Kaelbling, and Yoshua Bengio.
\newblock Generalization in deep learning.
\newblock \emph{arXiv preprint arXiv:1710.05468}, 2017.

\bibitem[Keskar et~al.(2017)Keskar, Mudigere, Nocedal, Smelyanskiy, and
  Tang]{keskar2016large}
Nitish~Shirish Keskar, Dheevatsa Mudigere, Jorge Nocedal, Mikhail Smelyanskiy,
  and Ping Tak~Peter Tang.
\newblock On large-batch training for deep learning: Generalization gap and
  sharp minima.
\newblock In \emph{ICLR}, 2017.

\bibitem[Krogh and Hertz(1992)]{weight_decay}
Anders Krogh and John~A Hertz.
\newblock A simple weight decay can improve generalization.
\newblock In \emph{NIPS}, 1992.

\bibitem[Li and Yuan(2017)]{li2017convergence}
Yuanzhi Li and Yang Yuan.
\newblock Convergence analysis of two-layer neural networks with relu
  activation.
\newblock \emph{arXiv preprint arXiv:1705.09886}, 2017.

\bibitem[Liao and Poggio(2017)]{liao2017theory}
Qianli Liao and Tomaso Poggio.
\newblock Theory of deep learning ii: Landscape of the empirical risk in deep
  learning.
\newblock \emph{arXiv preprint arXiv:1703.09833}, 2017.

\bibitem[Lipton(2016)]{lipton2016stuck}
Zachary~C Lipton.
\newblock Stuck in a what? adventures in weight space.
\newblock In \emph{ICLR Workshop}, 2016.

\bibitem[Lorch(2016)]{viztrajectory}
Eliana Lorch.
\newblock Visualizing deep network training trajectories with pca.
\newblock In \emph{ICML Workshop on Visualization for Deep Learning}, 2016.

\bibitem[Neyshabur et~al.(2017)Neyshabur, Bhojanapalli, McAllester, and
  Srebro]{neyshabur2017exploring}
Behnam Neyshabur, Srinadh Bhojanapalli, David McAllester, and Nati Srebro.
\newblock Exploring generalization in deep learning.
\newblock In \emph{NIPS}, 2017.

\bibitem[Nguyen and Hein(2017)]{nguyen2017loss}
Quynh Nguyen and Matthias Hein.
\newblock The loss surface of deep and wide neural networks.
\newblock In \emph{ICML}, 2017.

\bibitem[Safran and Shamir(2016)]{safran2016quality}
Itay Safran and Ohad Shamir.
\newblock On the quality of the initial basin in overspecified neural networks.
\newblock In \emph{ICML}, 2016.

\bibitem[Simonyan and Zisserman(2015)]{vgg}
Karen Simonyan and Andrew Zisserman.
\newblock {Very Deep Convolutional Networks for Large-Scale Image Recognition}.
\newblock In \emph{ICLR}, 2015.

\bibitem[Smith and Topin(2017)]{smith2017exploring}
Leslie~N Smith and Nicholay Topin.
\newblock Exploring loss function topology with cyclical learning rates.
\newblock \emph{arXiv preprint arXiv:1702.04283}, 2017.

\bibitem[Soltanolkotabi et~al.(2017)Soltanolkotabi, Javanmard, and
  Lee]{soltanolkotabi2017theoretical}
Mahdi Soltanolkotabi, Adel Javanmard, and Jason~D Lee.
\newblock Theoretical insights into the optimization landscape of
  over-parameterized shallow neural networks.
\newblock \emph{arXiv preprint arXiv:1707.04926}, 2017.

\bibitem[Soudry and Hoffer(2017)]{soudry2017exponentially}
Daniel Soudry and Elad Hoffer.
\newblock Exponentially vanishing sub-optimal local minima in multilayer neural
  networks.
\newblock \emph{arXiv preprint arXiv:1702.05777}, 2017.

\bibitem[Swirszcz et~al.(2016)Swirszcz, Czarnecki, and
  Pascanu]{swirszcz2016local}
Grzegorz Swirszcz, Wojciech~Marian Czarnecki, and Razvan Pascanu.
\newblock Local minima in training of deep networks.
\newblock \emph{arXiv preprint arXiv:1611.06310}, 2016.

\bibitem[Tian(2017)]{tian2017analytical}
Yuandong Tian.
\newblock An analytical formula of population gradient for two-layered relu
  network and its applications in convergence and critical point analysis.
\newblock In \emph{ICML}, 2017.

\bibitem[Xie et~al.(2017)Xie, Liang, and Song]{xie2017diverse}
Bo~Xie, Yingyu Liang, and Le~Song.
\newblock Diverse neural network learns true target functions.
\newblock In \emph{AISTATS}, 2017.

\bibitem[Yun et~al.(2017)Yun, Sra, and Jadbabaie]{yun2017global}
Chulhee Yun, Suvrit Sra, and Ali Jadbabaie.
\newblock Global optimality conditions for deep neural networks.
\newblock In \emph{ICLR}, 2017.

\bibitem[Zagoruyko and Komodakis(2016)]{zagoruyko2016wide}
Sergey Zagoruyko and Nikos Komodakis.
\newblock Wide residual networks.
\newblock In \emph{BMVC}, 2016.

\bibitem[Zhang et~al.(2017)Zhang, Bengio, Hardt, Recht, and
  Vinyals]{zhang2016understanding}
Chiyuan Zhang, Samy Bengio, Moritz Hardt, Benjamin Recht, and Oriol Vinyals.
\newblock Understanding deep learning requires rethinking generalization.
\newblock In \emph{ICLR}, 2017.

\end{thebibliography}

\clearpage
\begin{center}\Large{\textbf{Visualizing the Loss Landscape of Neural Nets}}\end{center}
\appendix
\section{Comparison of Loss Surfaces}

\subsection{The Change of Weights Norm during Training}

\begin{figure*}[h]
\centering
\begin{tabular}{l}
\hspace{-0.5cm}
\subfigure[SGD, WD=0, epoch]{\includegraphics[width=0.25\linewidth]{figures/weight_norm_curves/{vgg9_sgd_lr=0.1_wd=0.0_save_epoch=1_l2_norm_epoch}.pdf}}
\subfigure[SGD, WD=5e-4, epoch]{\includegraphics[width=0.25\linewidth]{figures/weight_norm_curves/{vgg9_sgd_lr=0.1_wd=0.0005_save_epoch=1_l2_norm_epoch}.pdf}}
\subfigure[Adam, WD=0, epoch]{\includegraphics[width=0.25\linewidth]{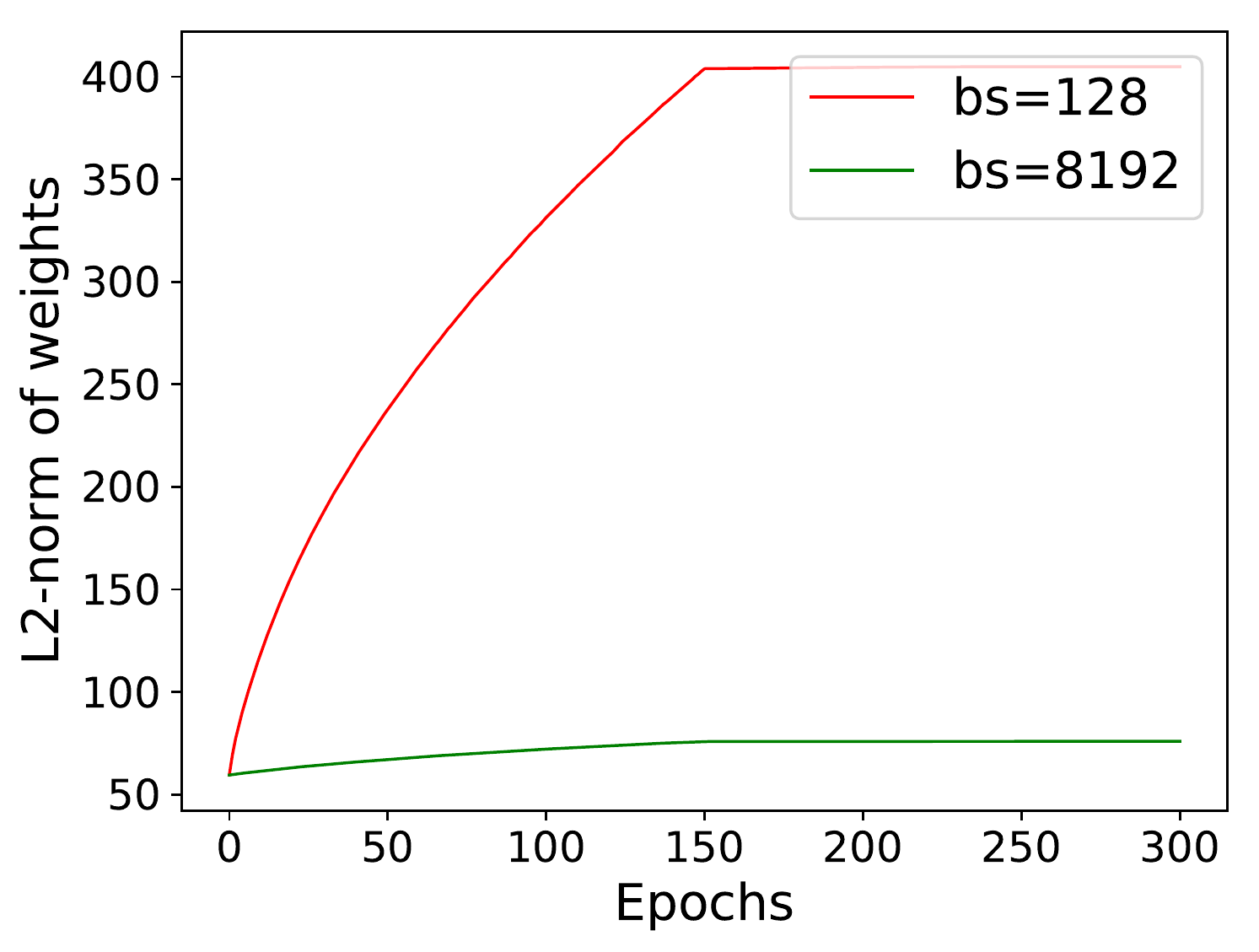}}
\subfigure[Adam, WD=5e-4, epoch]{\includegraphics[width=0.25\linewidth]{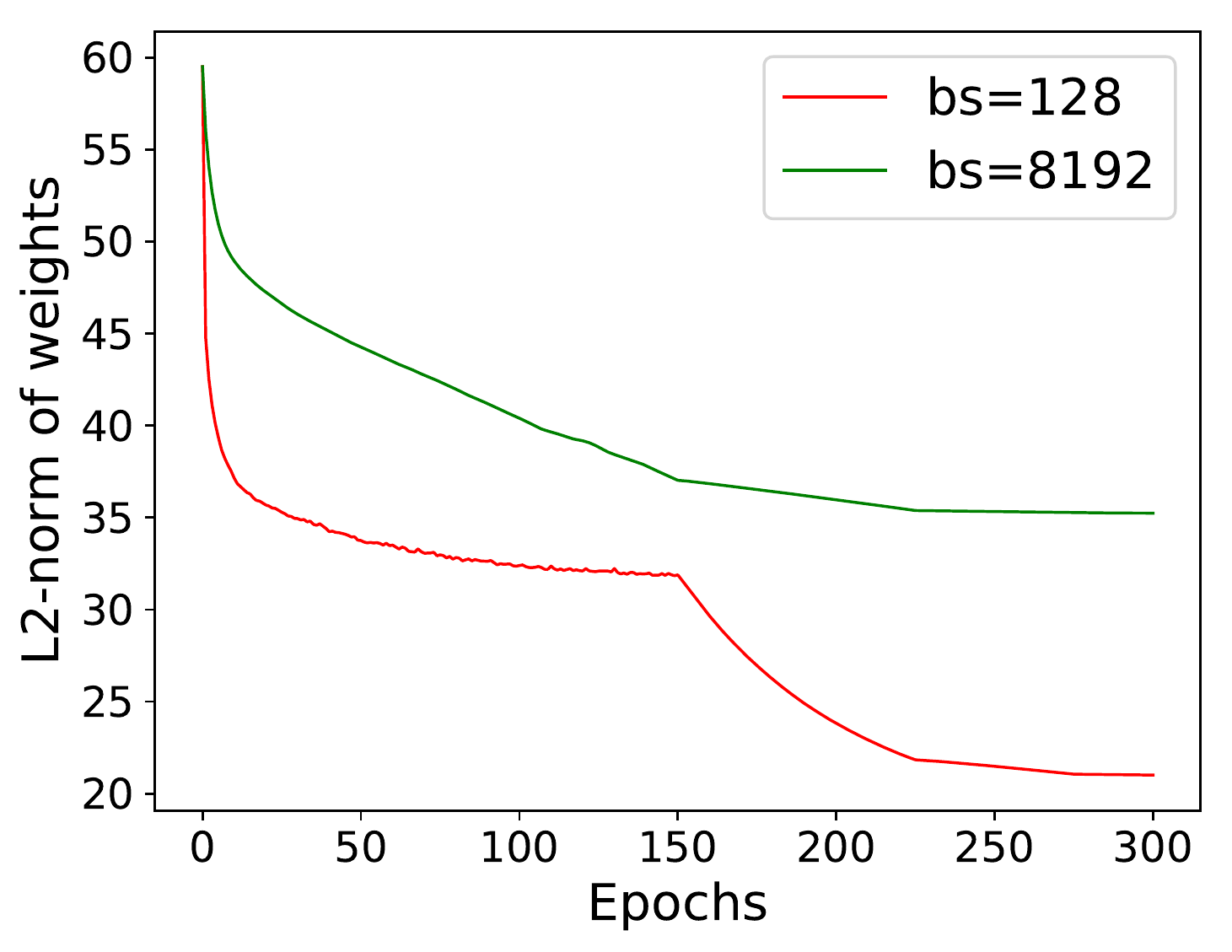}}
\\
\hspace{-0.5cm}
\subfigure[SGD, WD=0, iter]{\includegraphics[width=0.25\linewidth]{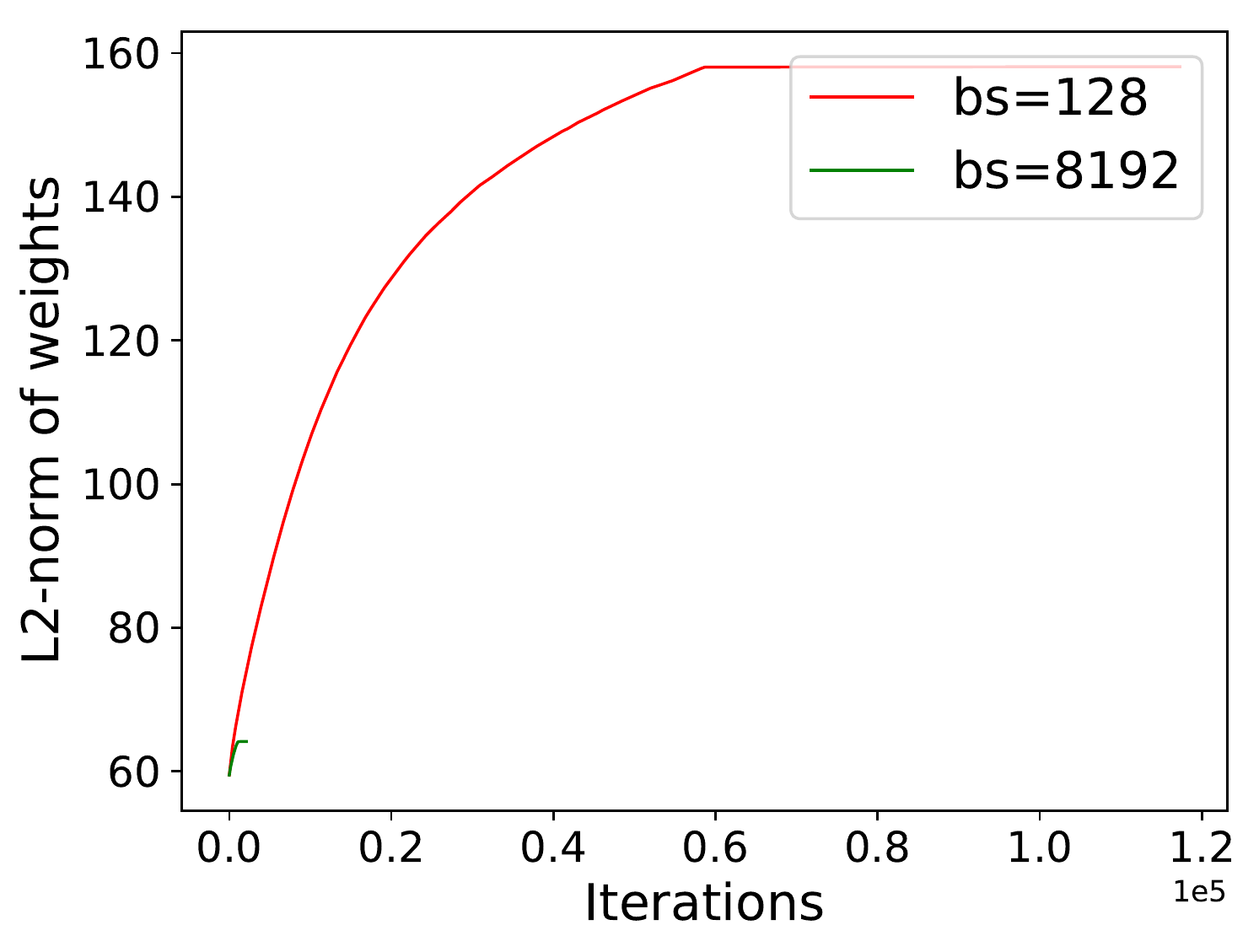}}
\subfigure[SGD, WD=5e-4, iter]{\includegraphics[width=0.25\linewidth]{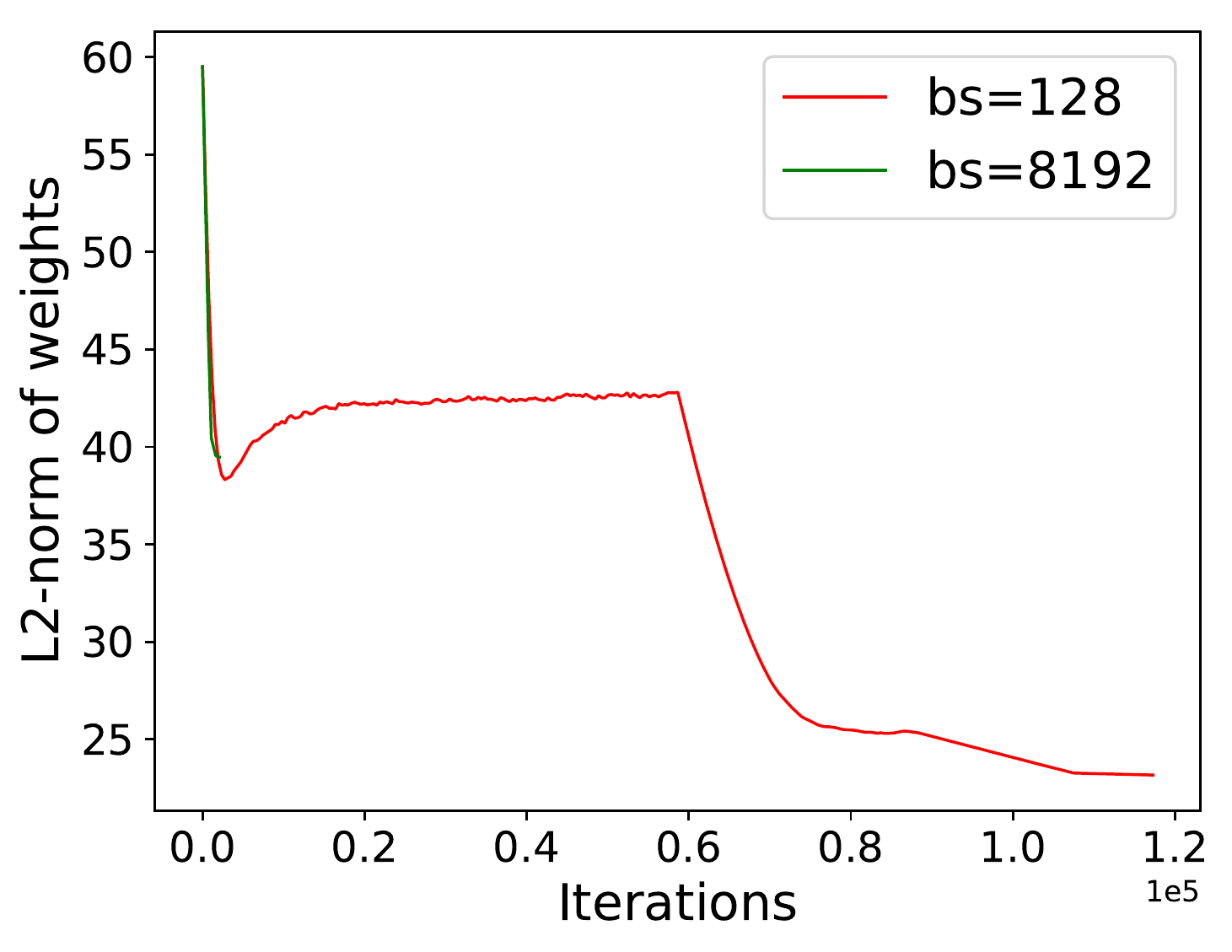}}
\subfigure[Adam, WD=0, iter]{\includegraphics[width=0.25\linewidth]{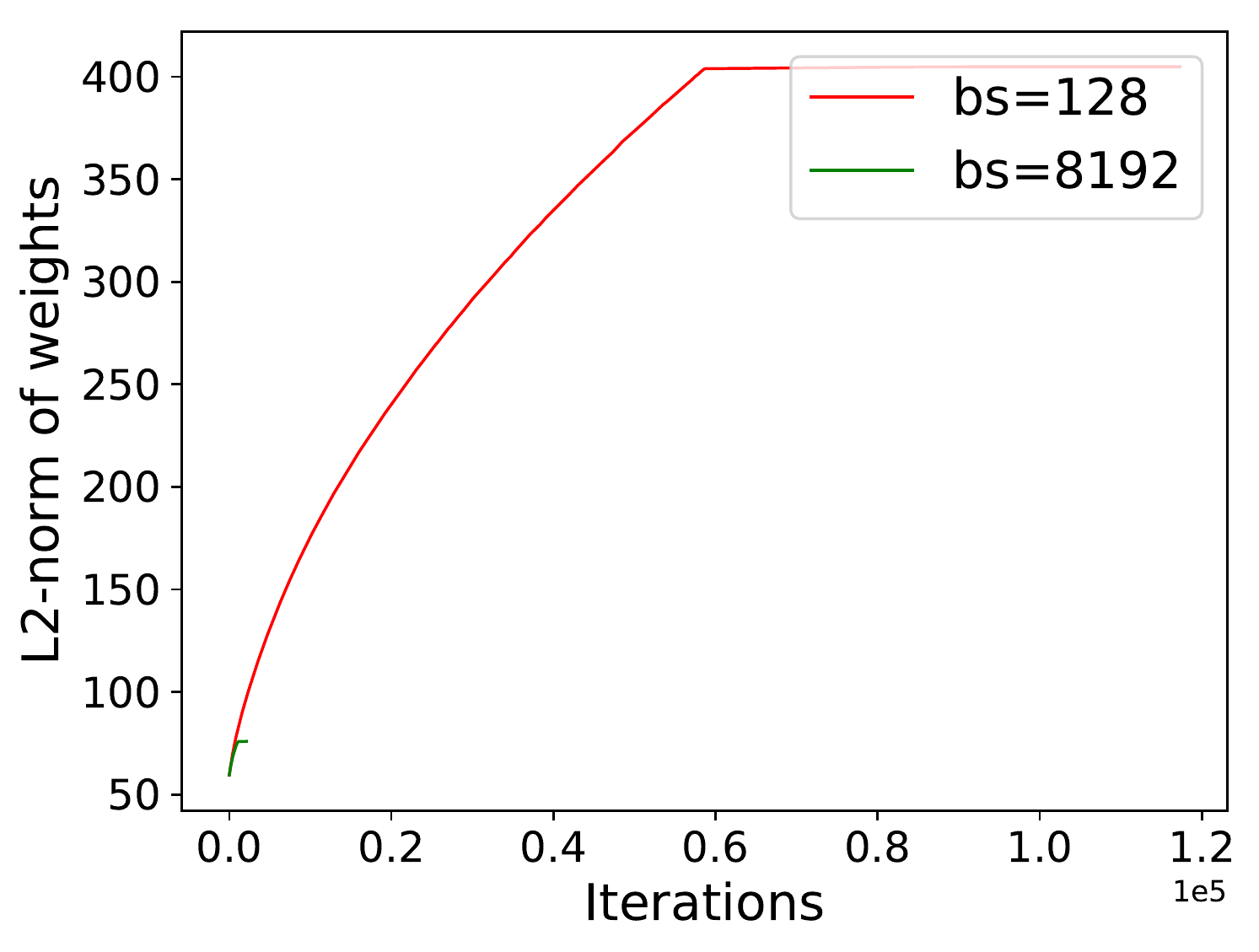}}
\subfigure[Adam, WD=5e-4, iter]{\includegraphics[width=0.25\linewidth]{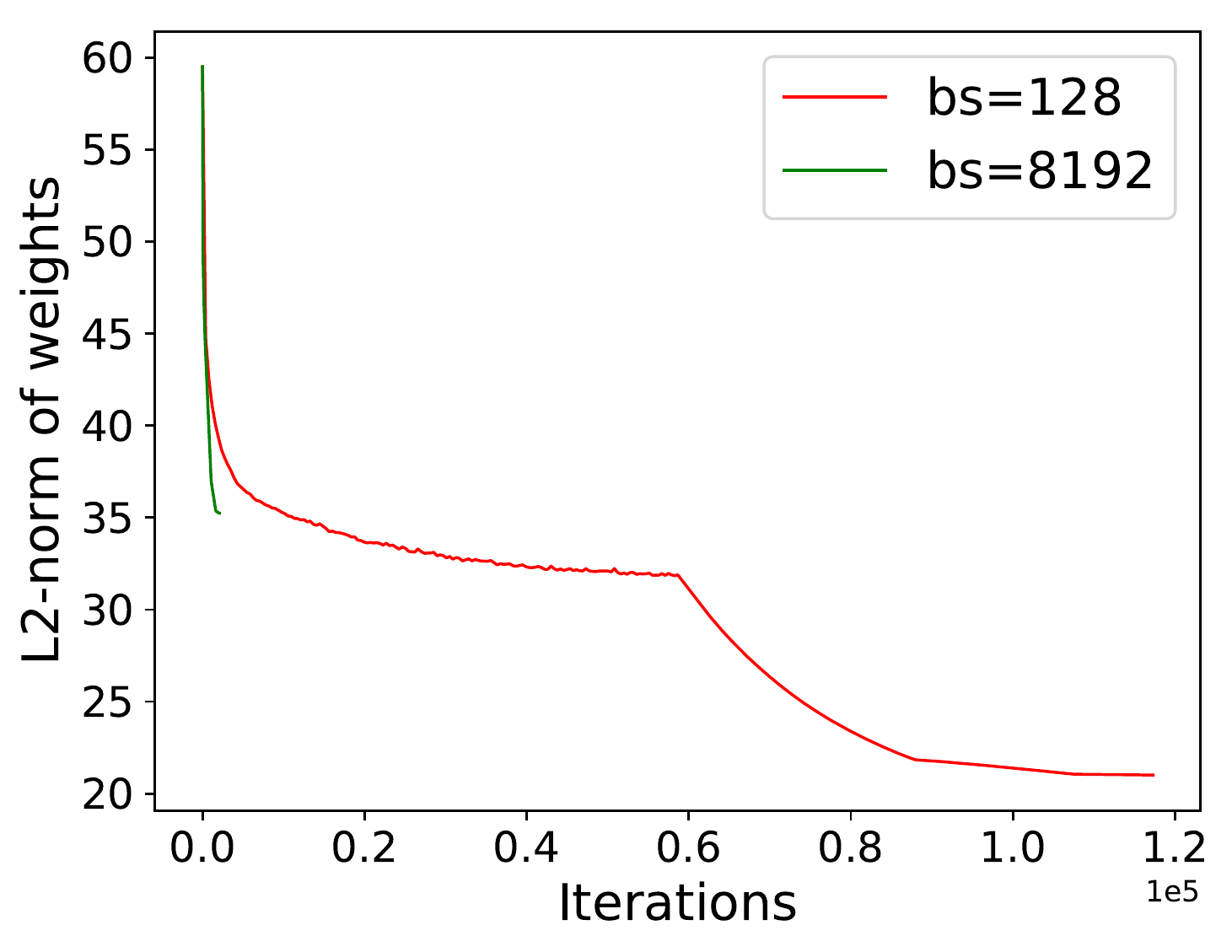}}
\end{tabular}
\caption{The change of weights norm during training for VGG-9.
When weight decay is disabled, the weight norm grows steadily during training without constraints.
When nonzero weight decay is adopted, the weight norm decreases rapidly at the beginning and becomes stable until the learning rate is decayed.
Since we use a fixed number of epochs for different batch sizes, the difference in weight norm change between large-batch and small-batch training is mainly caused by the larger number of updates when a small batch is used.
As shown in the second row, the changes of weight norm are at the same pace for both small and large batch training in terms of iterations.}
\label{fig:weight_norm}
\end{figure*}

Figure~\ref{fig:weight_norm} shows the change of weights norm during training in terms of epochs and iterations.

\subsection{Comparision of Normalization Methods}
\label{sec:compare_norm}
Here we compare several normalization methods for a given random normal direction $d$.
Let $\theta_i$ denote the weights of layer $i$ and $\theta_{i,j}$ represent the $j$-th filter in the $i$-th layer.

\begin{itemize}
\item \textbf{No Normalization}
In this case, the direction $d$ is added to the weights directly without processing.

\item \textbf{Filter Normalization}
The direction $d$ is normalized so that the direction for each filter has the same norm as the corresponding filter in $\theta$,
$$d_{i,j} \gets \frac{d_{i,j}}{\|d_{i,j}\|}\|\theta_{i,j}\|.$$
This is the approach advocated in this article, and is used extensively for plotting loss surfaces.

\item \textbf{Layer Normalization}
The direction $d$ is normalized in the layer level so that the direction for each layer has the same norm as the corresponding layer of $\theta$,
$$d_i \gets \frac{d_i}{\|d_i\|}\|\theta_i\|.$$

\end{itemize}

\begin{figure*}[!h]
\centering
\begin{tabular}{l}
\hspace{-3mm}
\subfigure[SGD, 128, 7.37\%]{\includegraphics[width=0.25\linewidth]{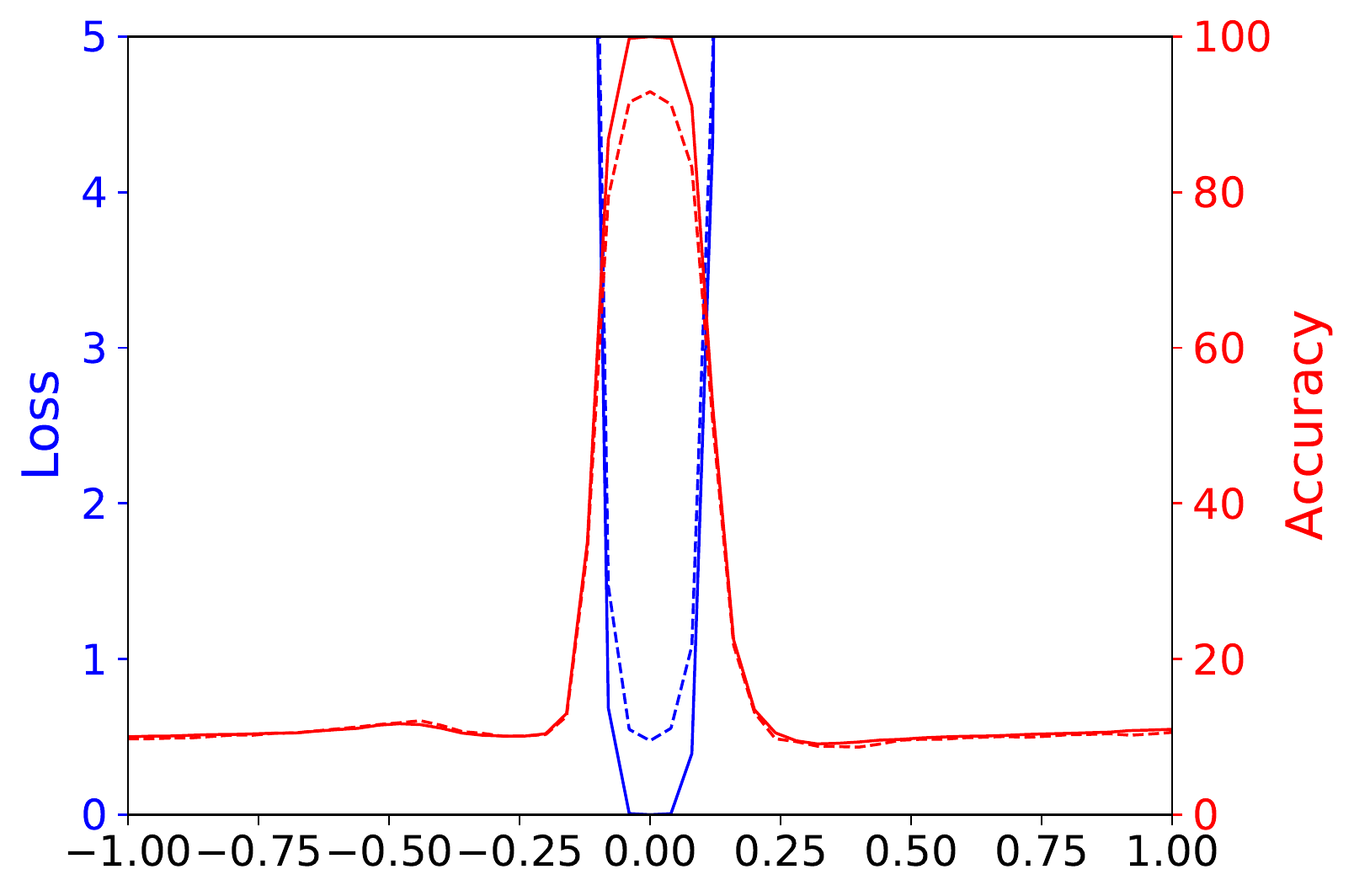}}
\subfigure[SGD, 8192, 11.07\%]{\includegraphics[width=0.25\linewidth]{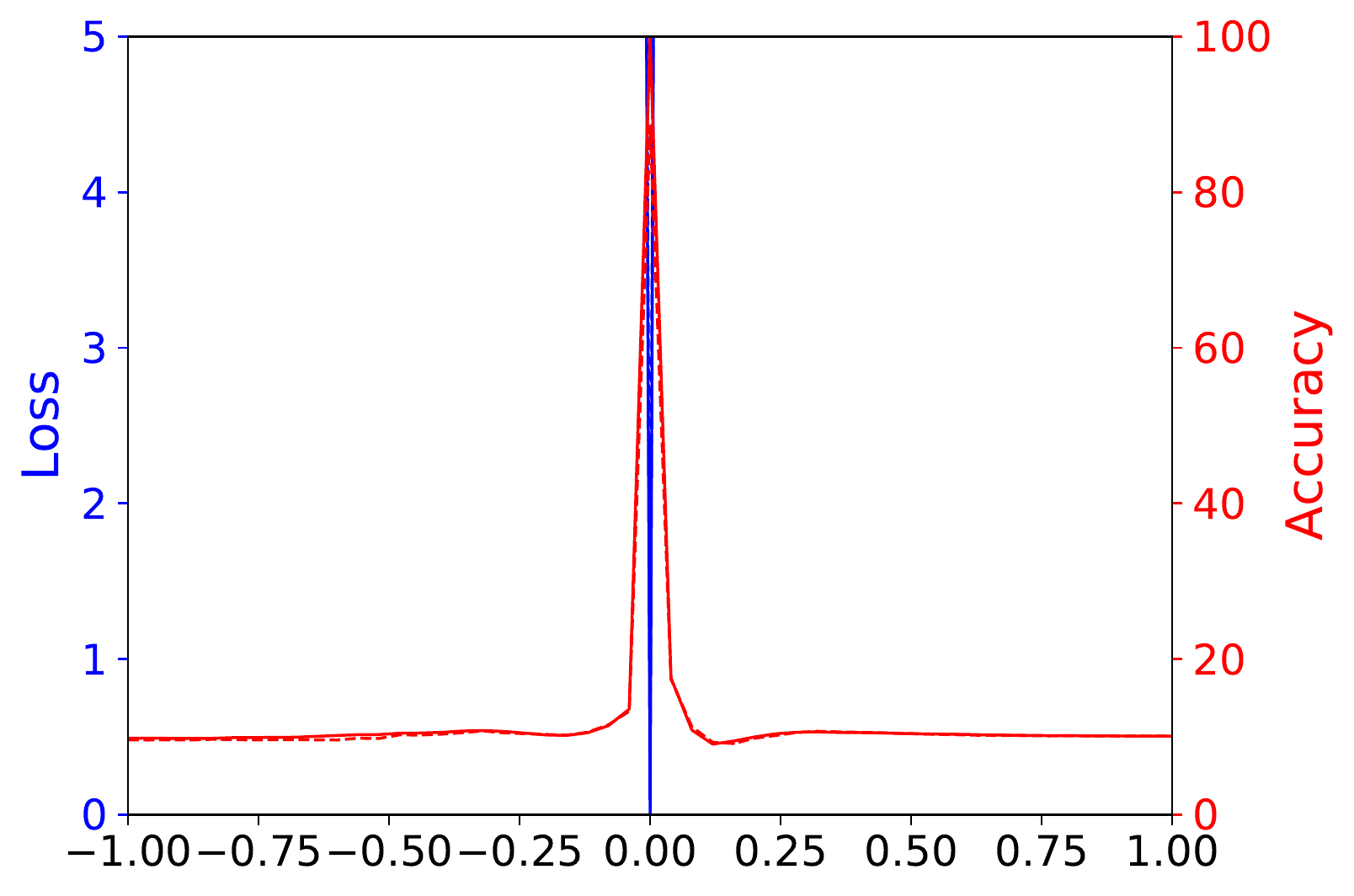}}
\subfigure[Adam, 128, 7.44\%]{\includegraphics[width=0.25\linewidth]{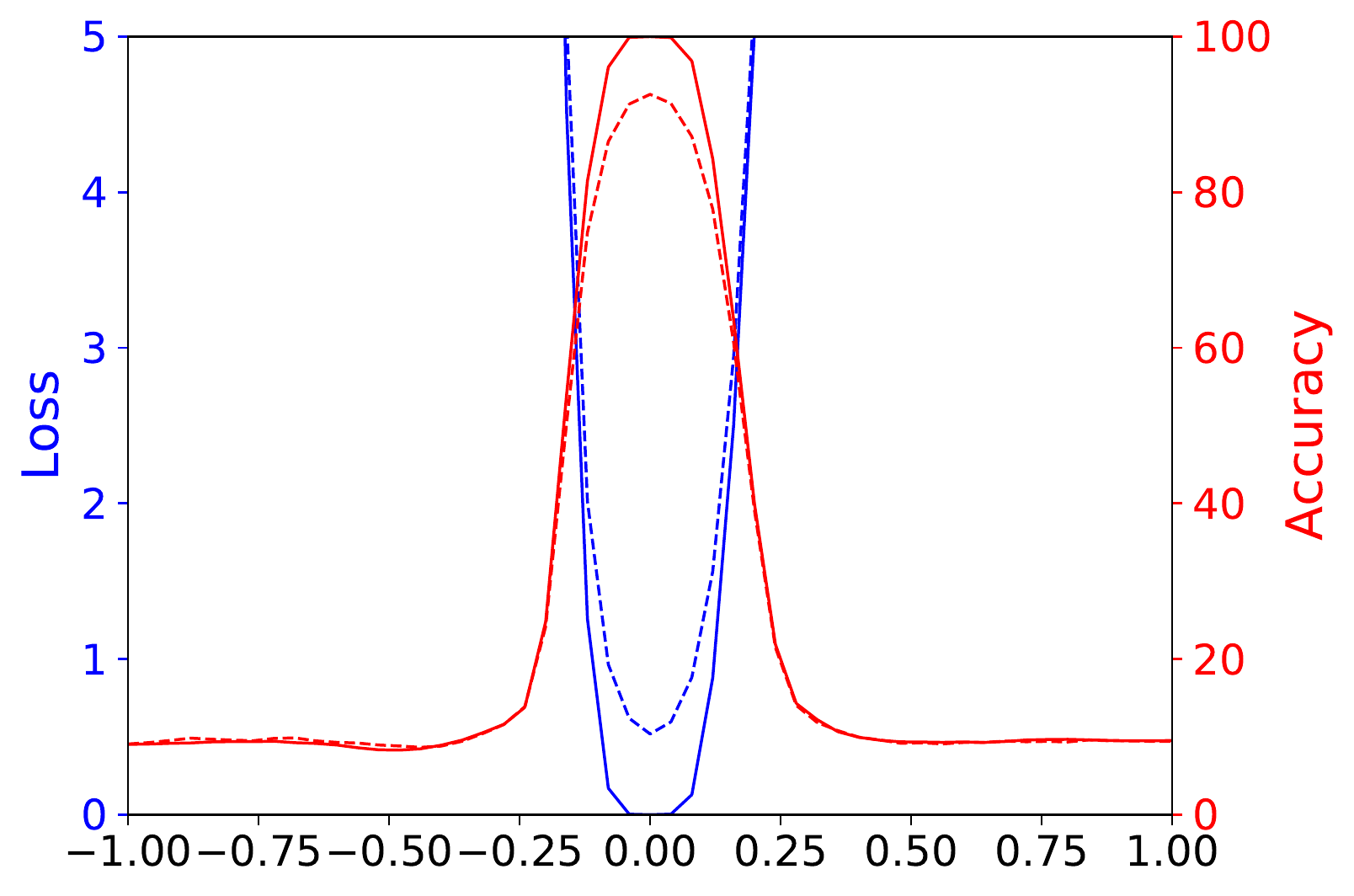}}
\subfigure[Adam, 8192, 10.91\%]{\includegraphics[width=0.25\linewidth]{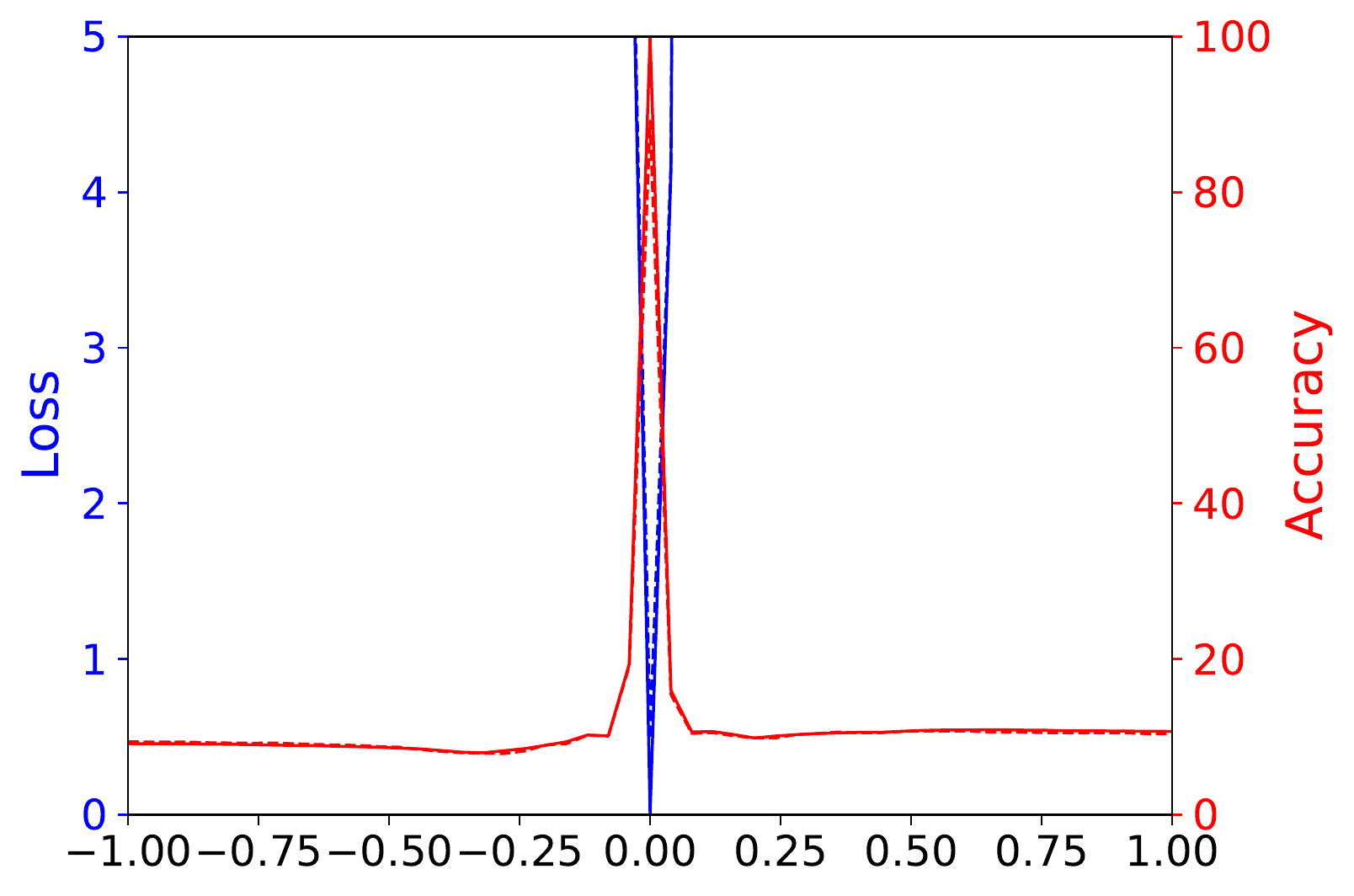}}\\
\hspace{-3mm}
\subfigure[SGD, 128, 6.00\%]{\includegraphics[width=0.25\linewidth]{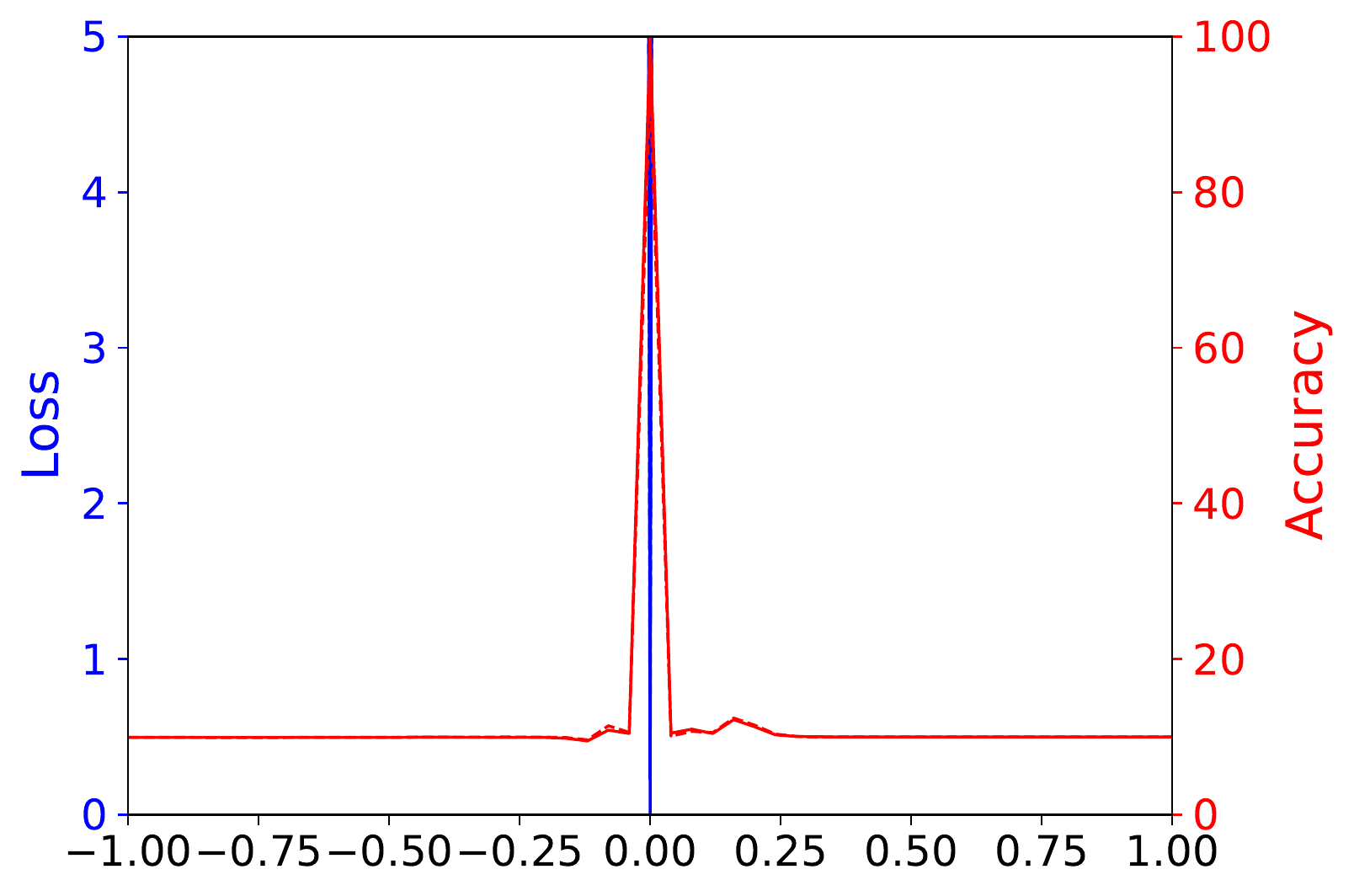}}
\subfigure[SGD, 8192, 10.19\%]{\includegraphics[width=0.25\linewidth]{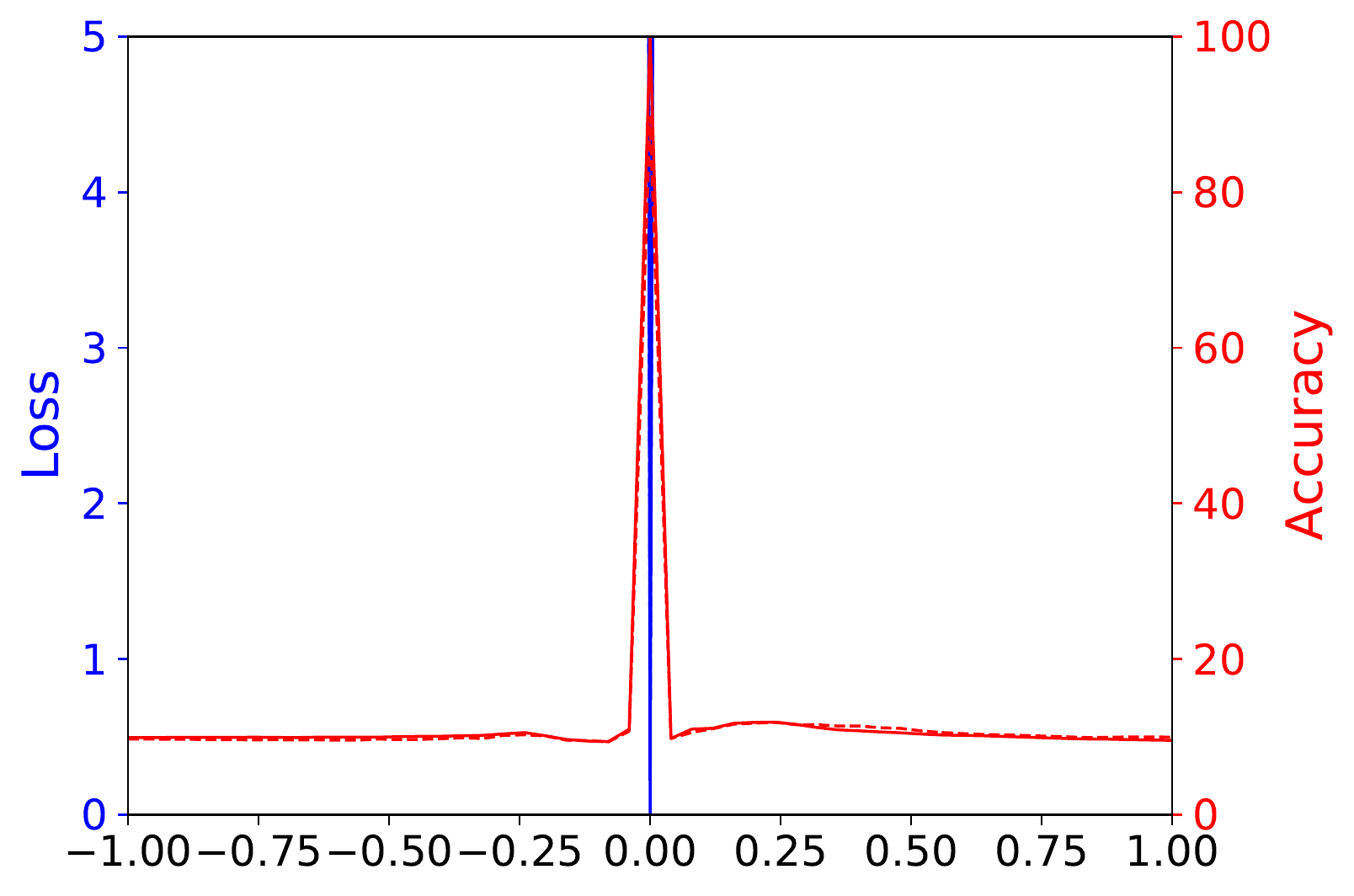}}
\subfigure[Adam, 128, 7.80\%]{\includegraphics[width=0.25\linewidth]{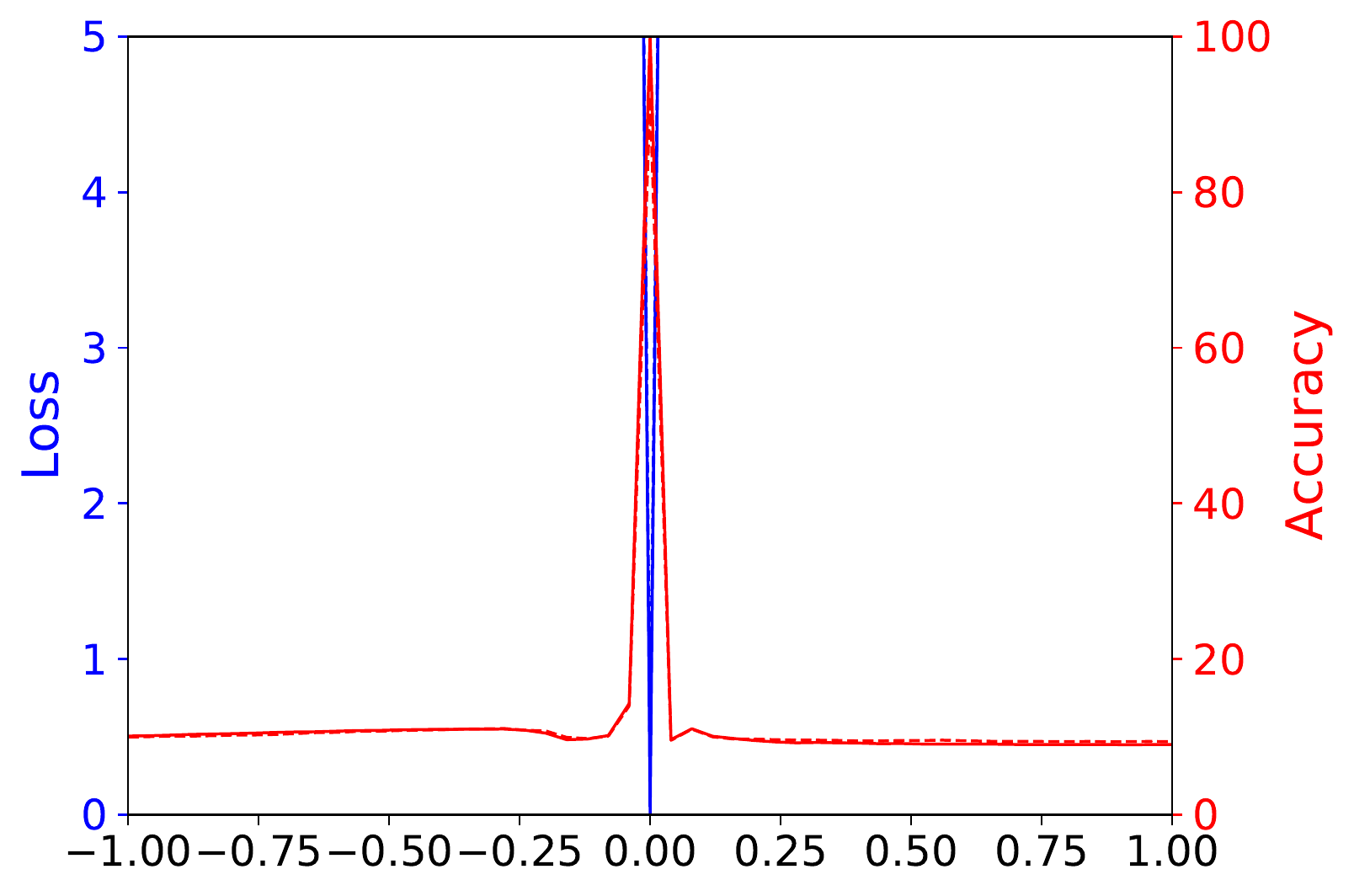}}
\subfigure[Adam, 8192, 9.52\%]{\includegraphics[width=0.25\linewidth]{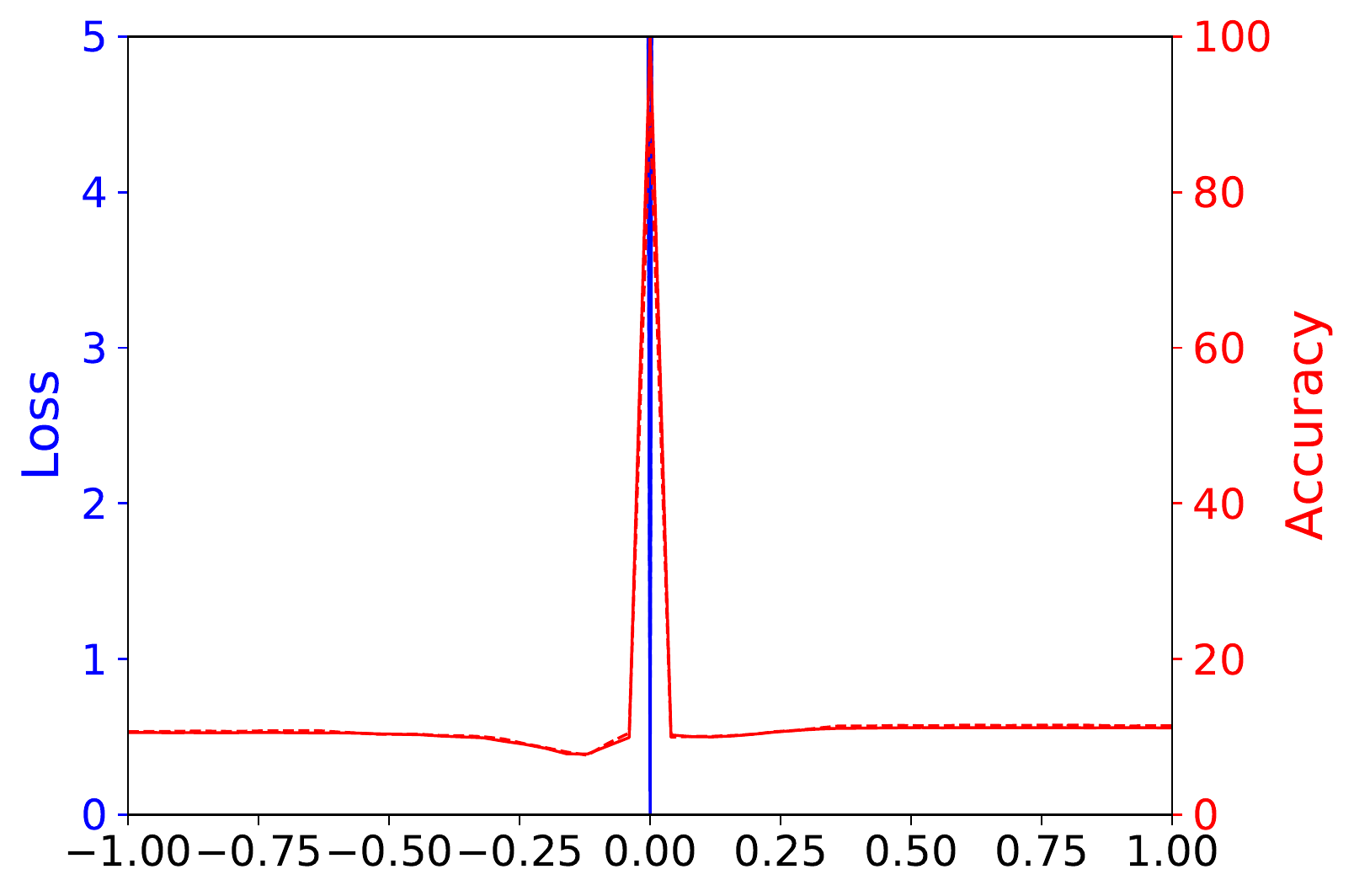}}
\end{tabular}
\caption{1D loss plots for VGG-9 without normalization.
The first row has no weight decay and the second row uses weight decay~0.0005.
}
\label{fig:vgg9_nonoram}
\end{figure*}

\begin{figure*}[!h]
\centering
\begin{tabular}{l}
\hspace{-3mm}
\subfigure[SGD, 128, 7.37\%]{\includegraphics[width=0.25\linewidth]{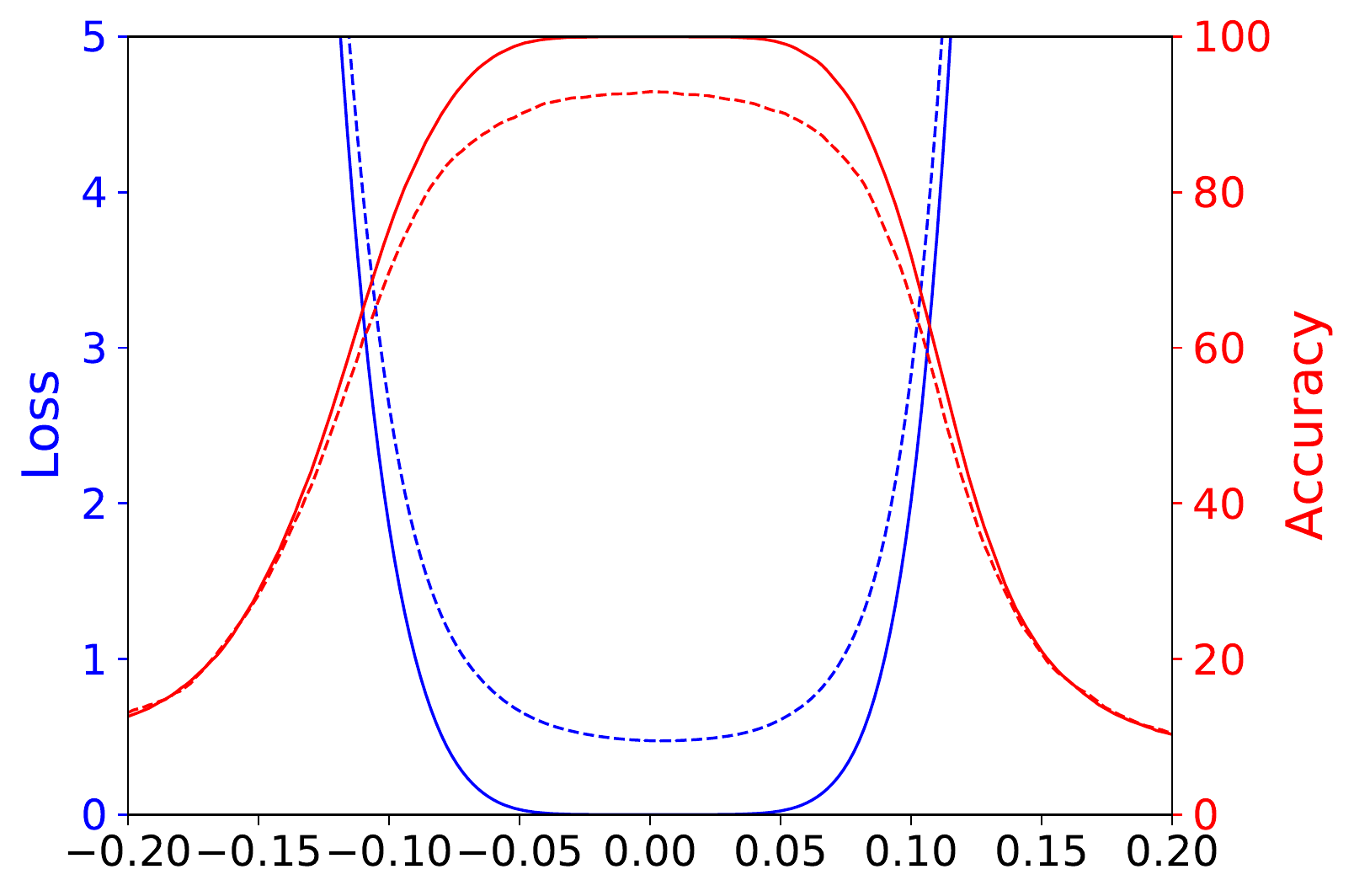}}
\subfigure[SGD, 8192, 11.07\%]{\includegraphics[width=0.25\linewidth]{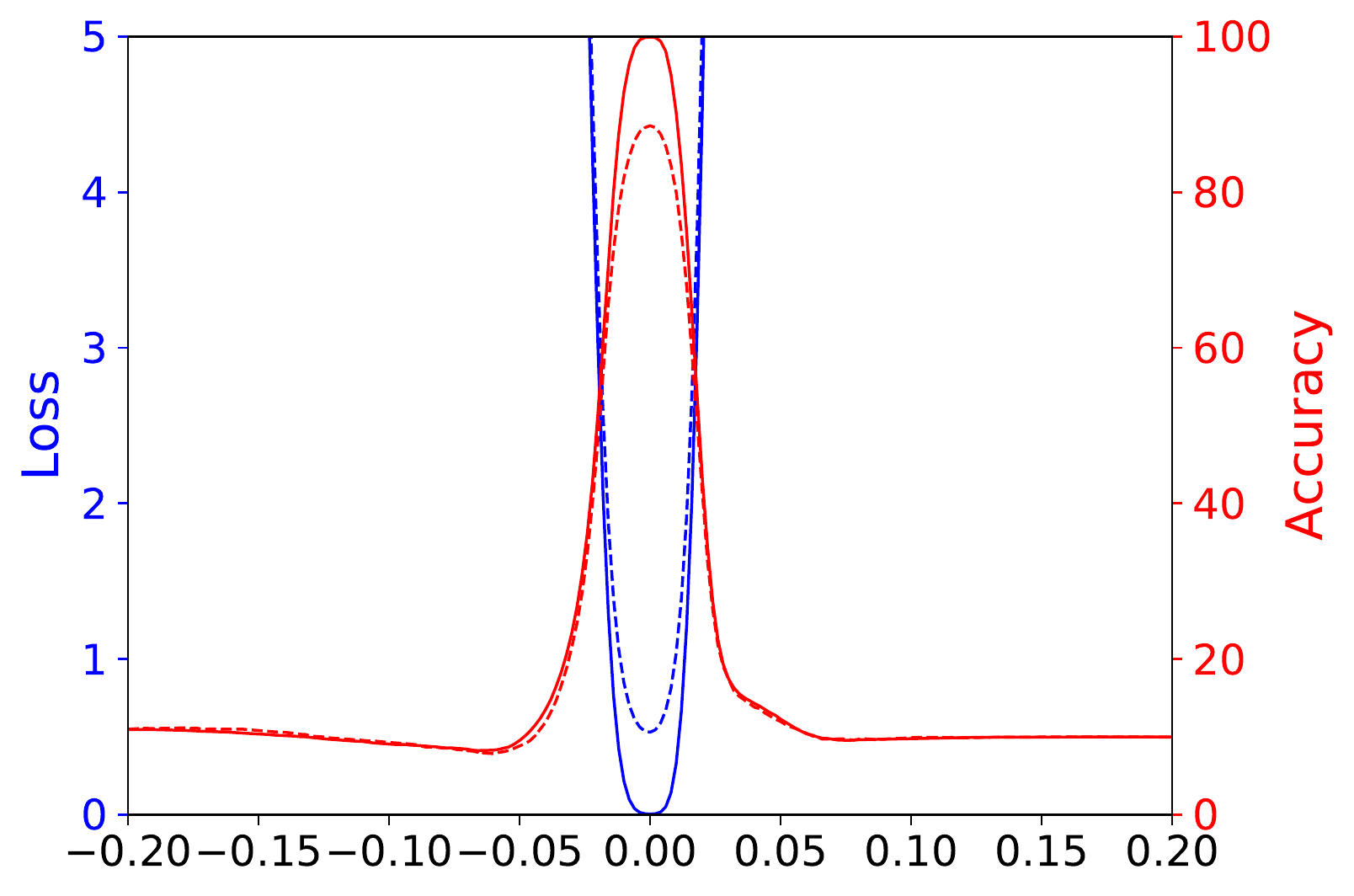}}
\subfigure[Adam, 128, 7.44\%]{\includegraphics[width=0.25\linewidth]{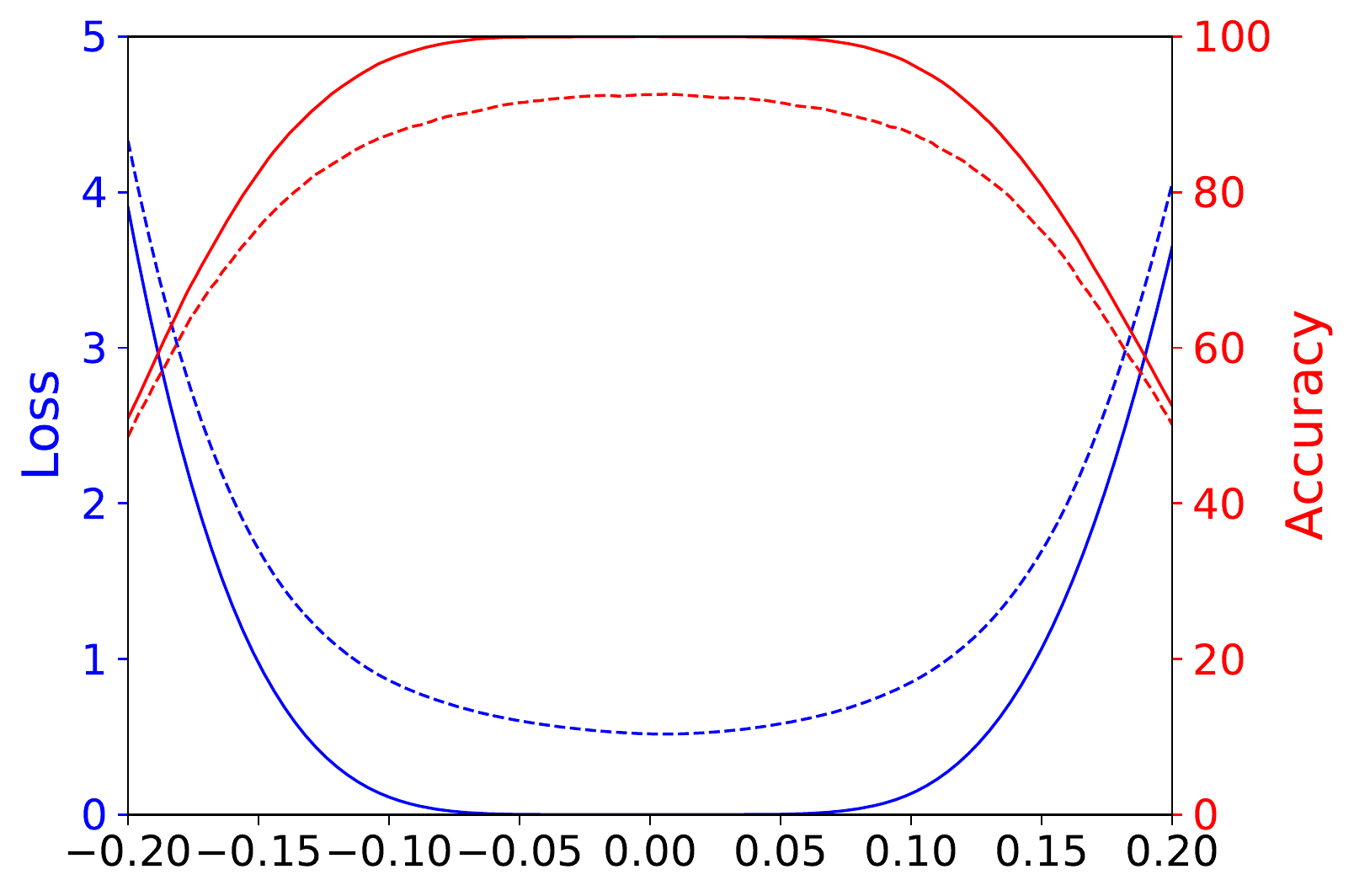}}
\subfigure[Adam, 8192, 10.91\%]{\includegraphics[width=0.25\linewidth]{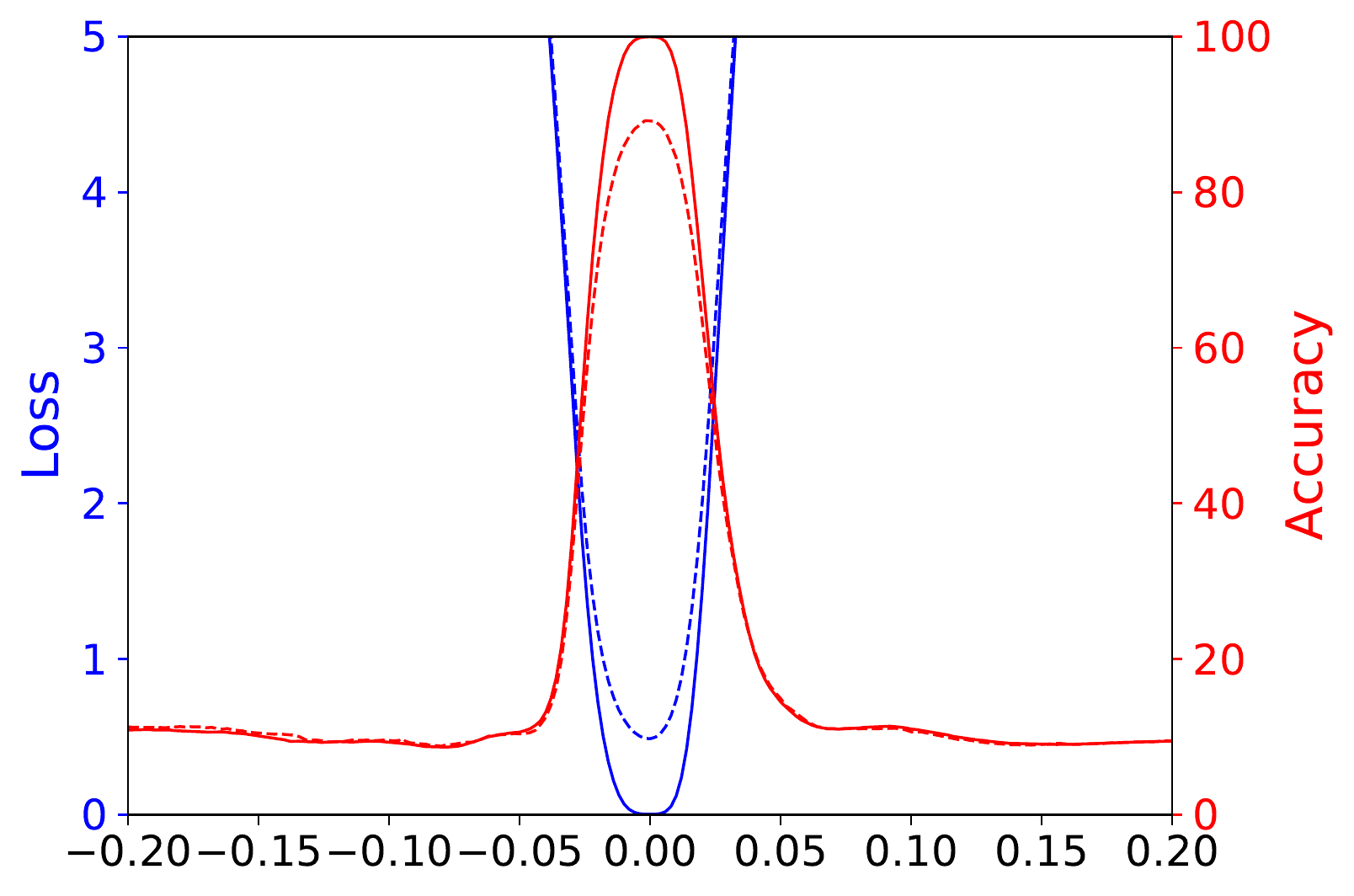}}\\
\hspace{-3mm}
\subfigure[SGD, 128, 6.00\%]{\includegraphics[width=0.25\linewidth]{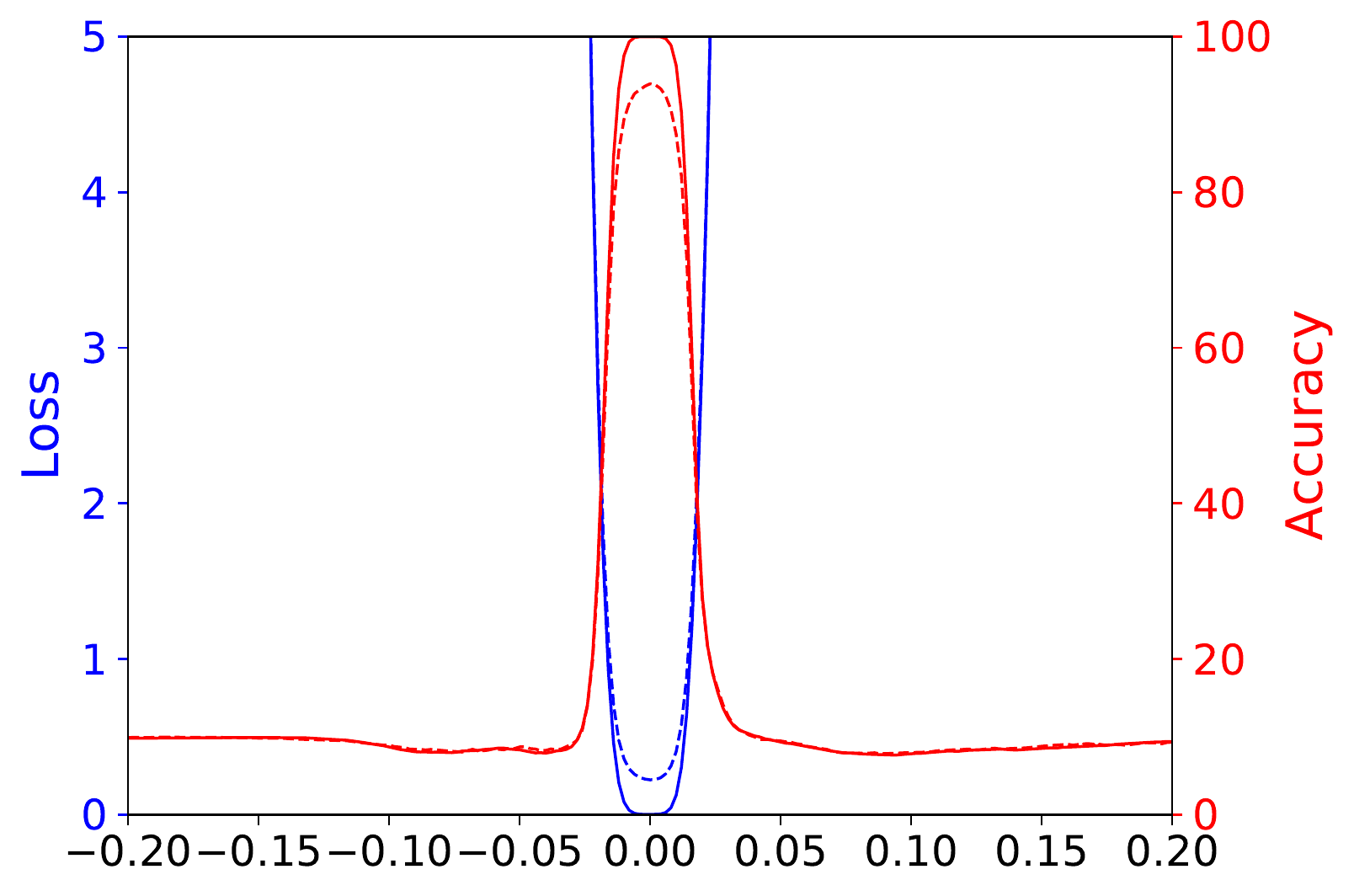}}
\subfigure[SGD, 8192, 10.19\%]{\includegraphics[width=0.25\linewidth]{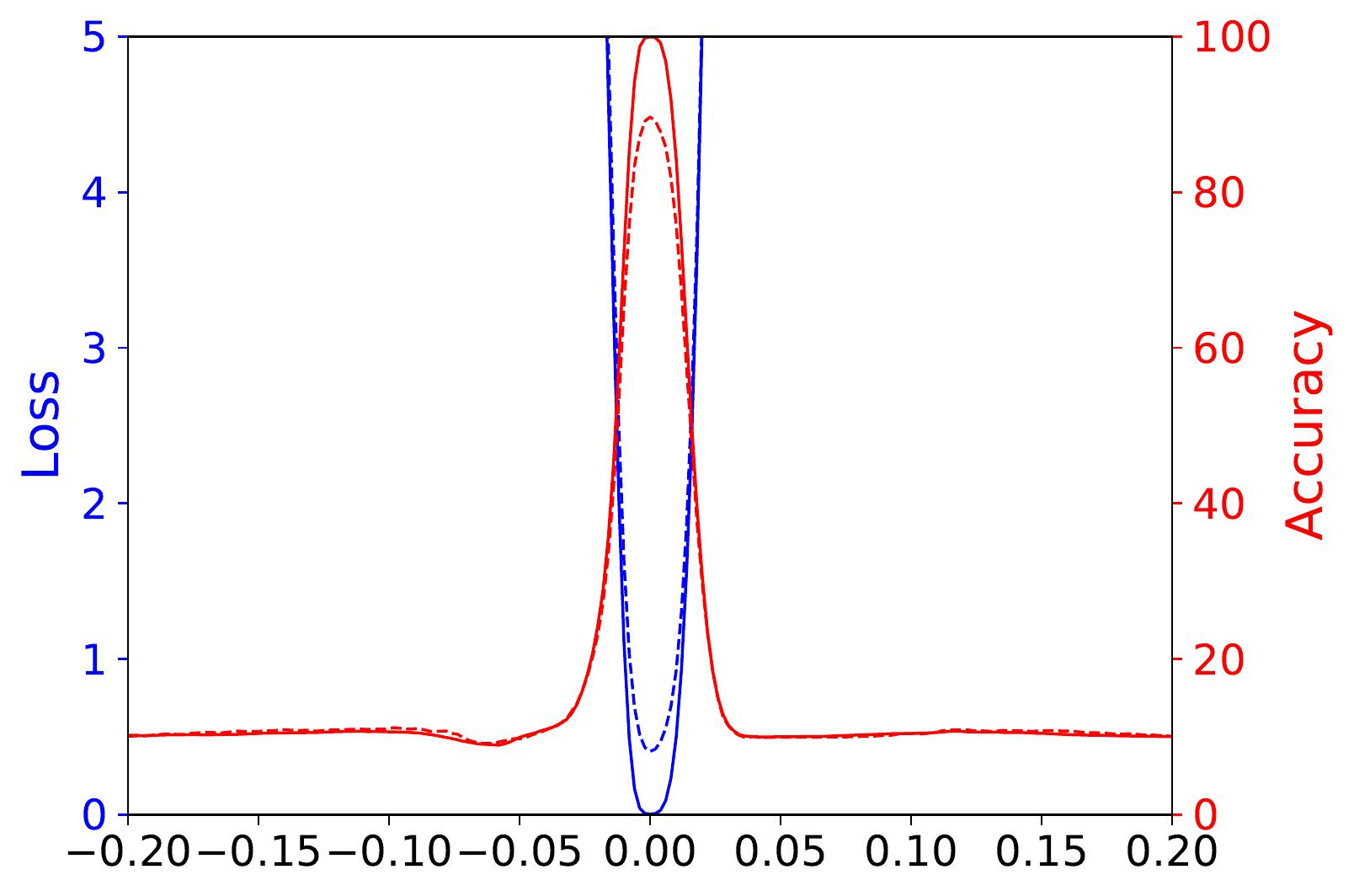}}
\subfigure[Adam, 128, 7.80\%]{\includegraphics[width=0.25\linewidth]{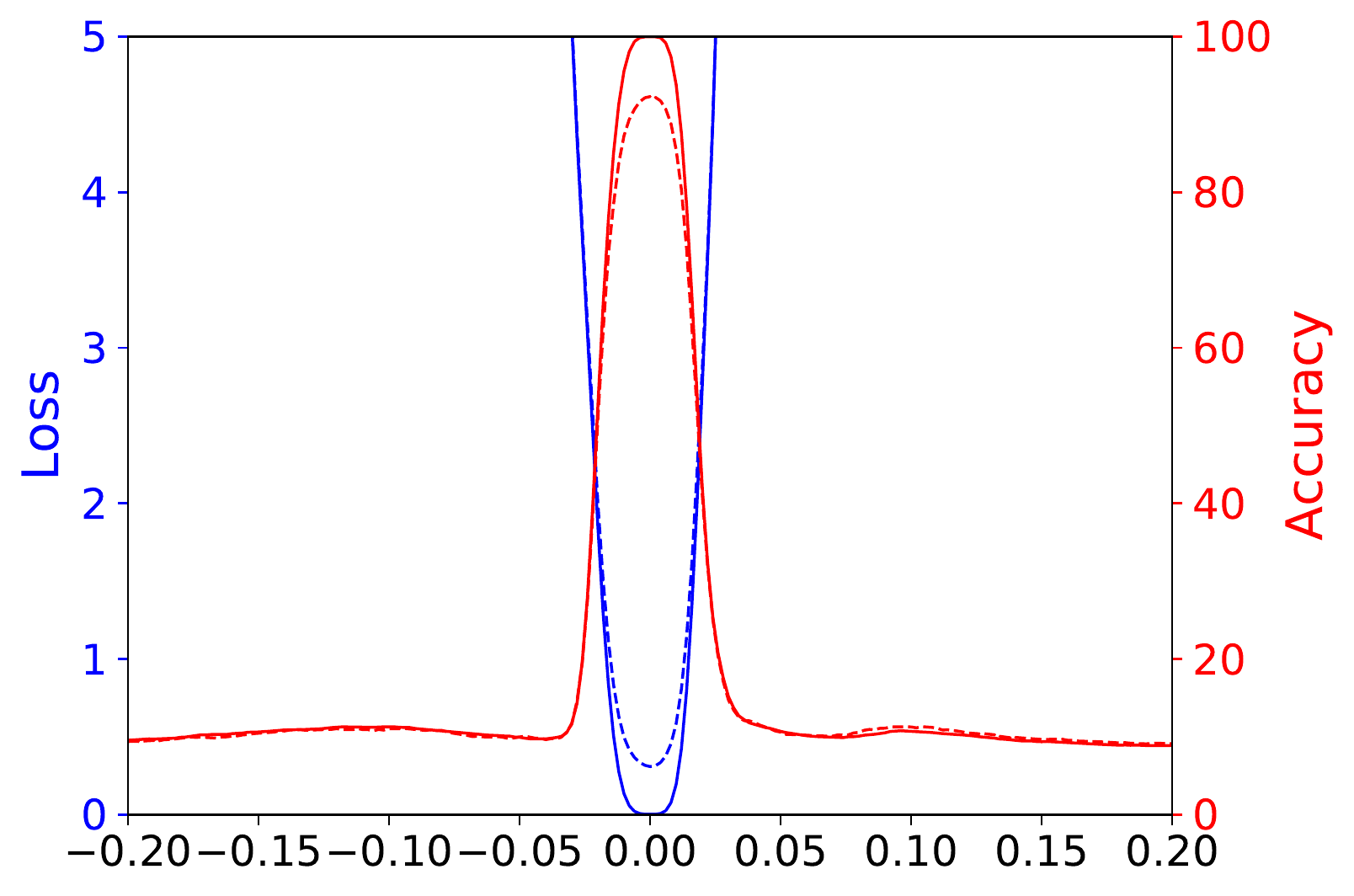}}
\subfigure[Adam, 8192, 9.52\%]{\includegraphics[width=0.25\linewidth]{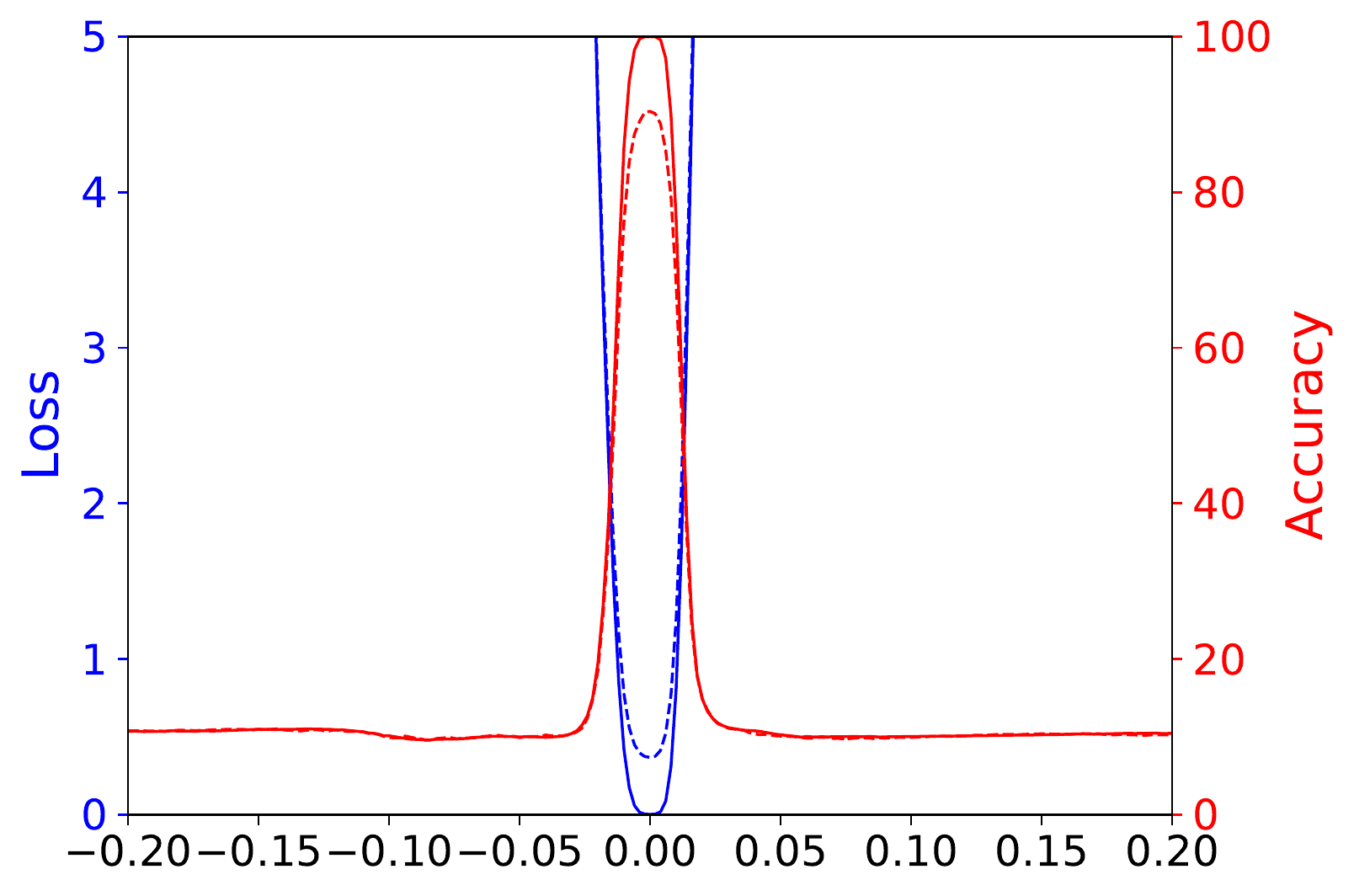}}
\end{tabular}
\caption{Enlarged Figure~\ref{fig:vgg9_nonoram}. The range of the $x$-axis is [-0.2, 0.2] instead of [-1.0, 1.0.
The first row has no weight decay and the second row uses weight decay 0.0005.
The pairs (a, e) and (c, g) show that sharpness of minima does not correlate well with test error.
}
\label{fig:vgg9_nonoram_enlarged}
\end{figure*}

\begin{figure*}[!h]
\centering
\begin{tabular}{l}
\hspace{-3mm}
\subfigure[SGD, 128, 7.37\%]{\includegraphics[width=0.25\linewidth]{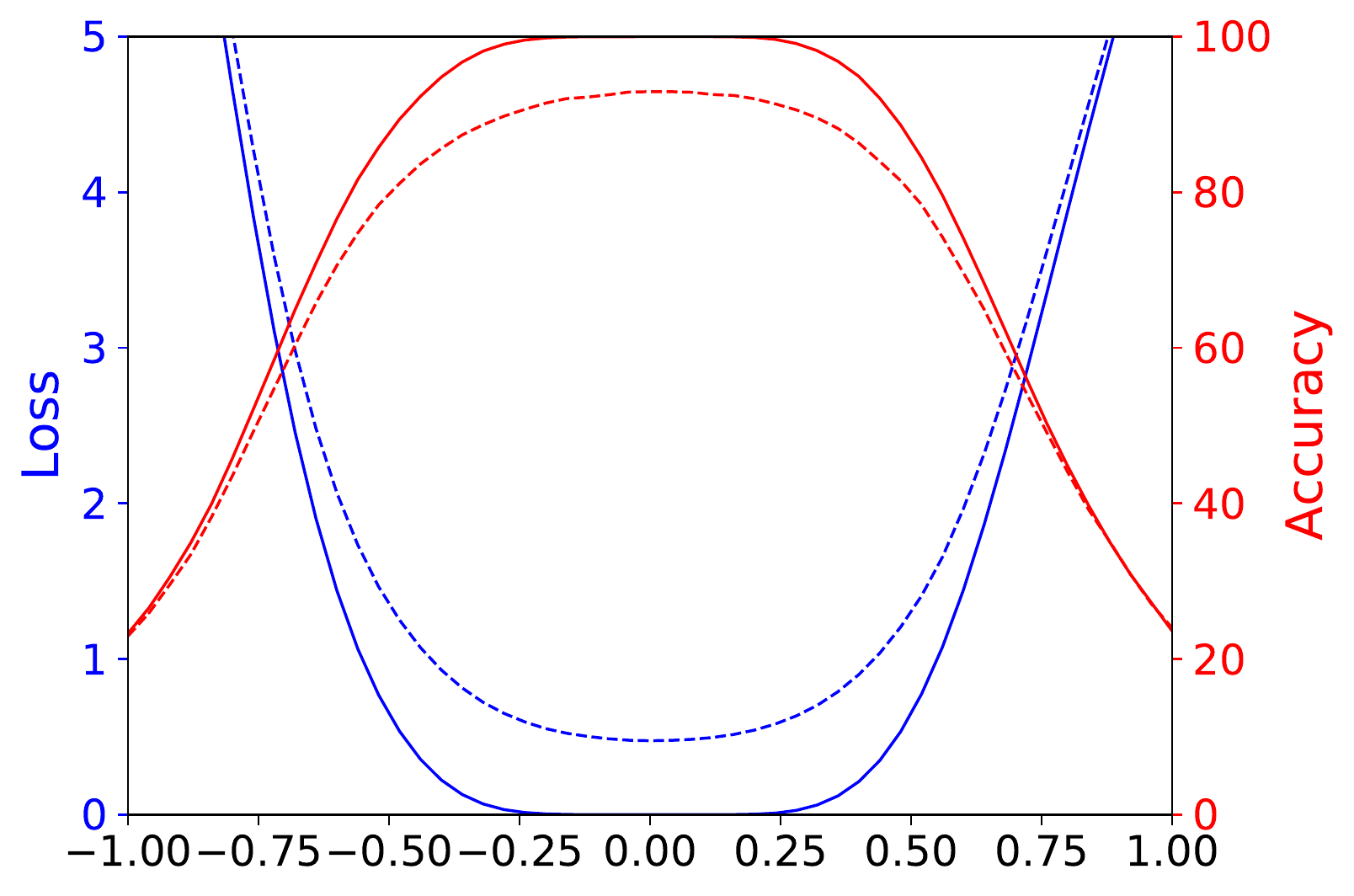}}
\subfigure[SGD, 8192, 11.07\%]{\includegraphics[width=0.25\linewidth]{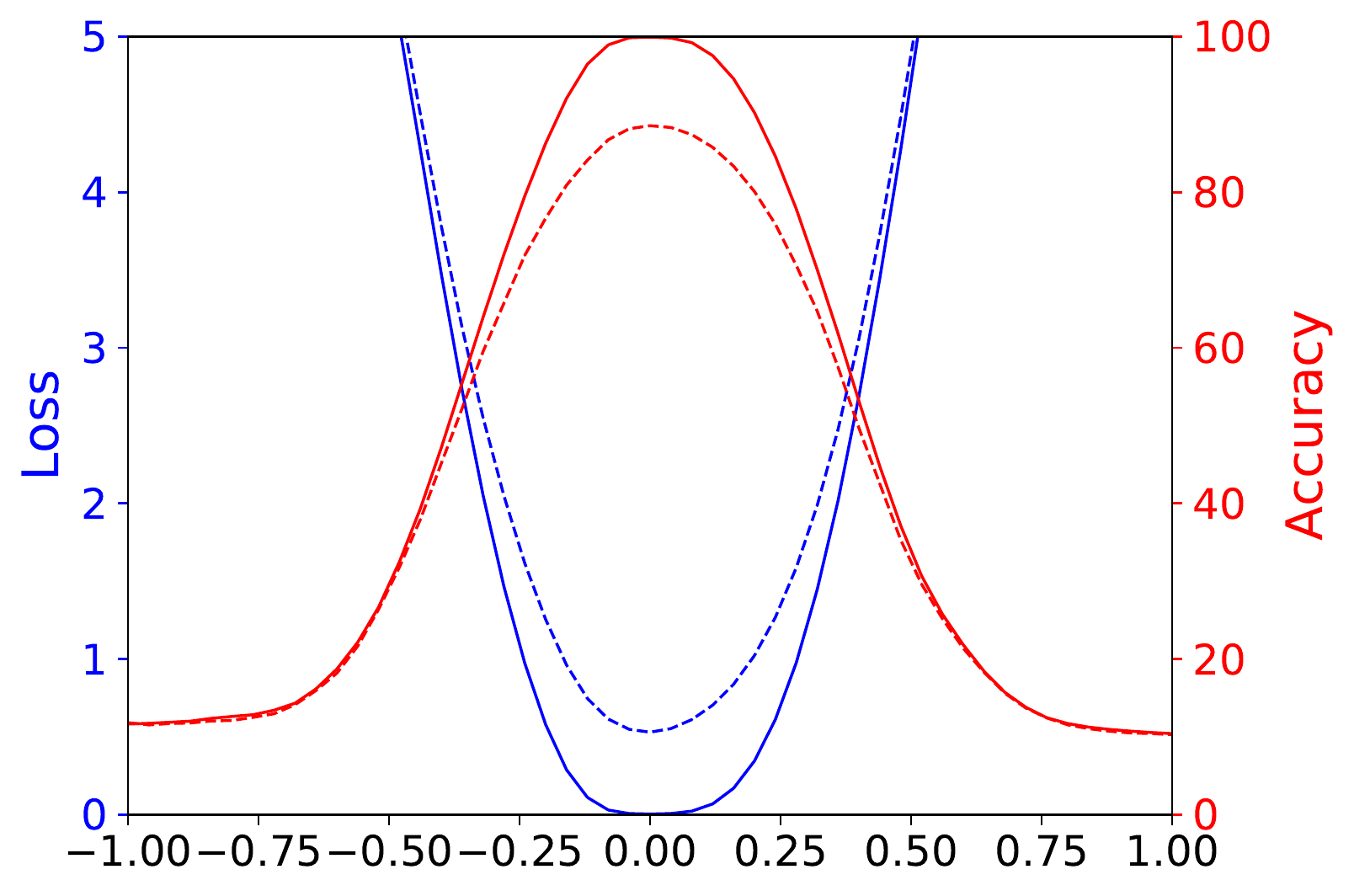}}
\subfigure[Adam, 128, 7.44\%]{\includegraphics[width=0.25\linewidth]{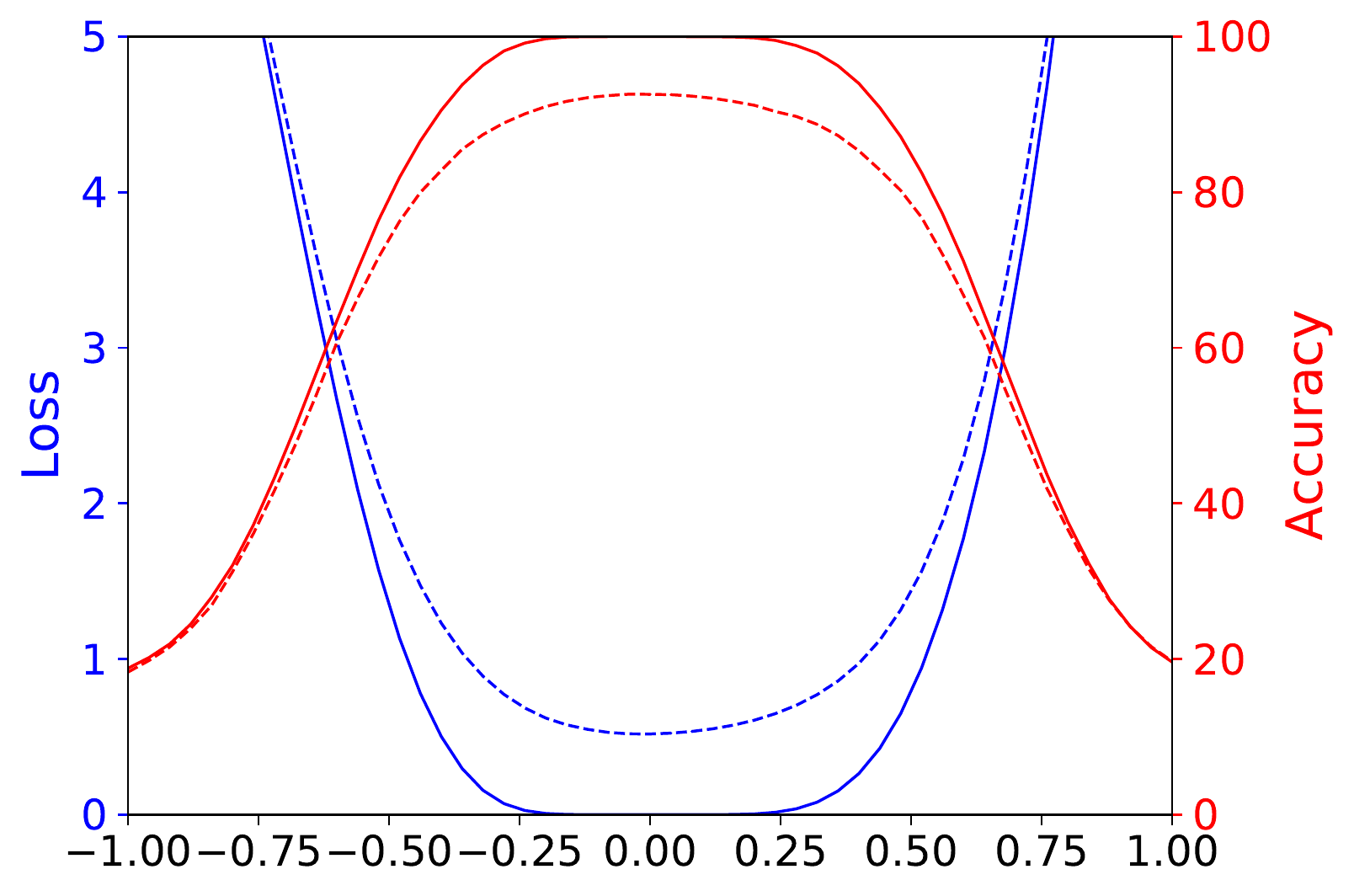}}
\subfigure[Adam, 8192, 10.91\%]{\includegraphics[width=0.25\linewidth]{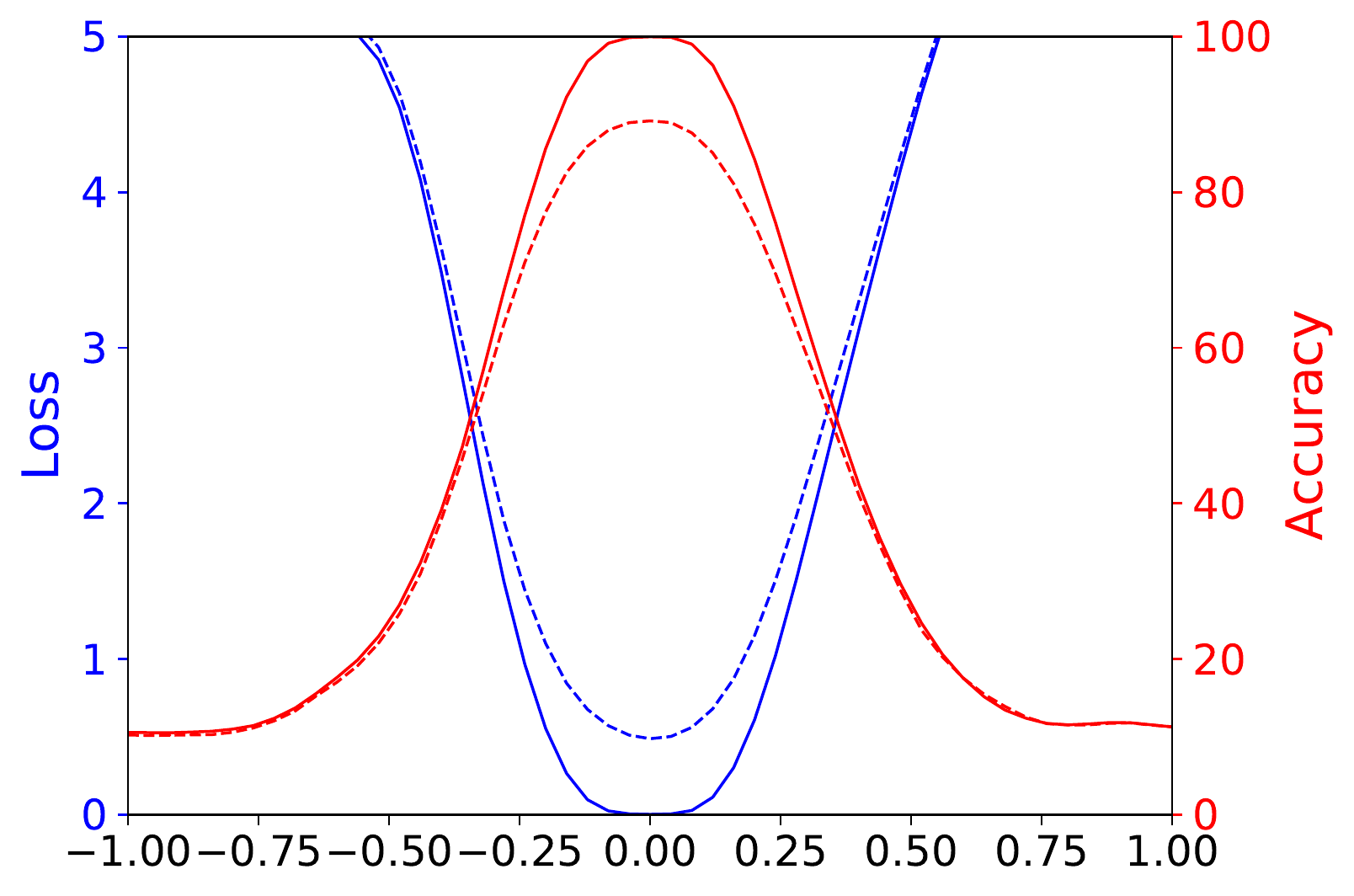}}\\
\hspace{-3mm}
\subfigure[SGD, 128, 6.00\%]{\includegraphics[width=0.25\linewidth]{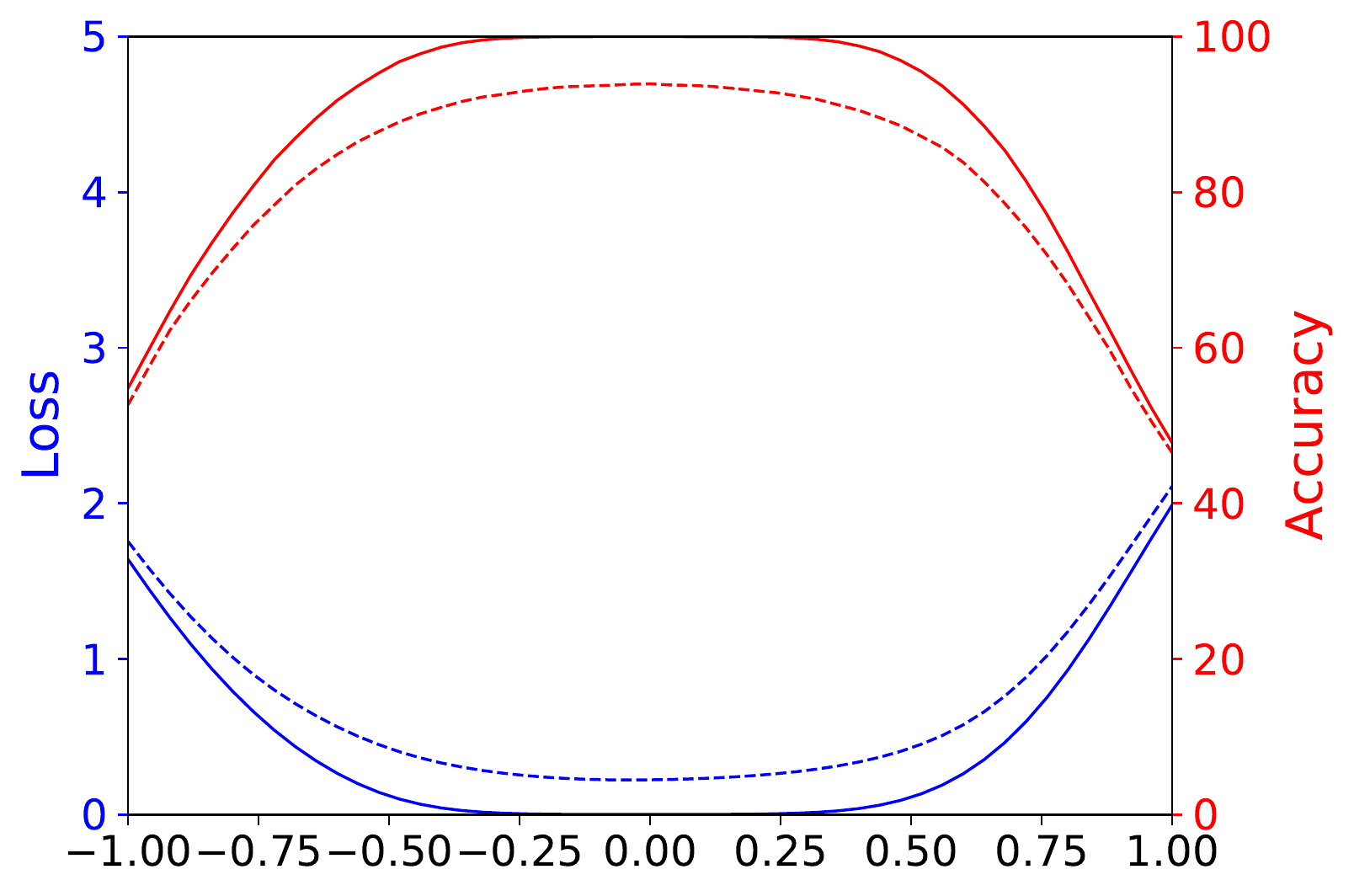}}
\subfigure[SGD, 8192, 10.19\%]{\includegraphics[width=0.25\linewidth]{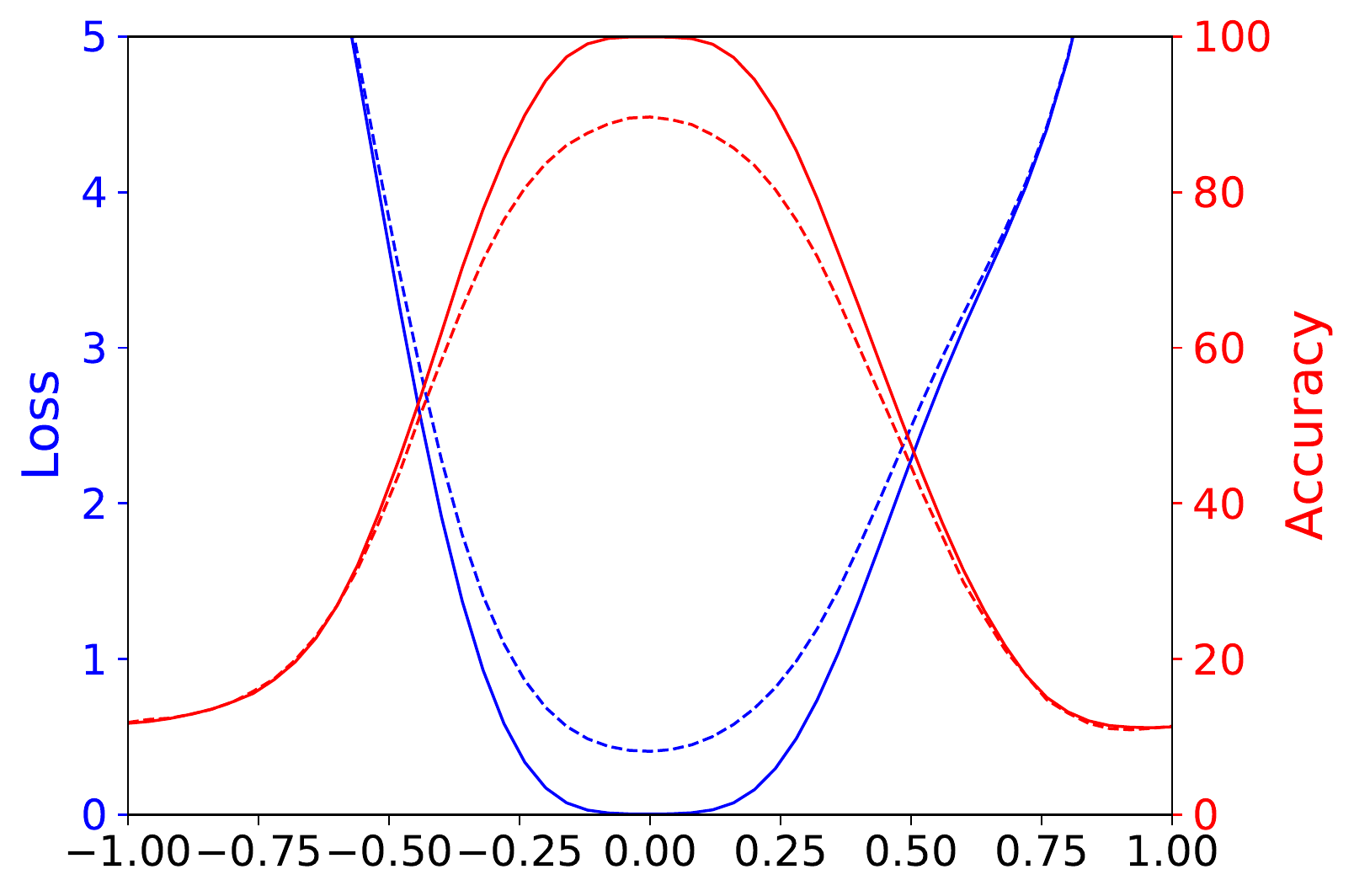}}
\subfigure[Adam, 128, 7.80\%]{\includegraphics[width=0.25\linewidth]{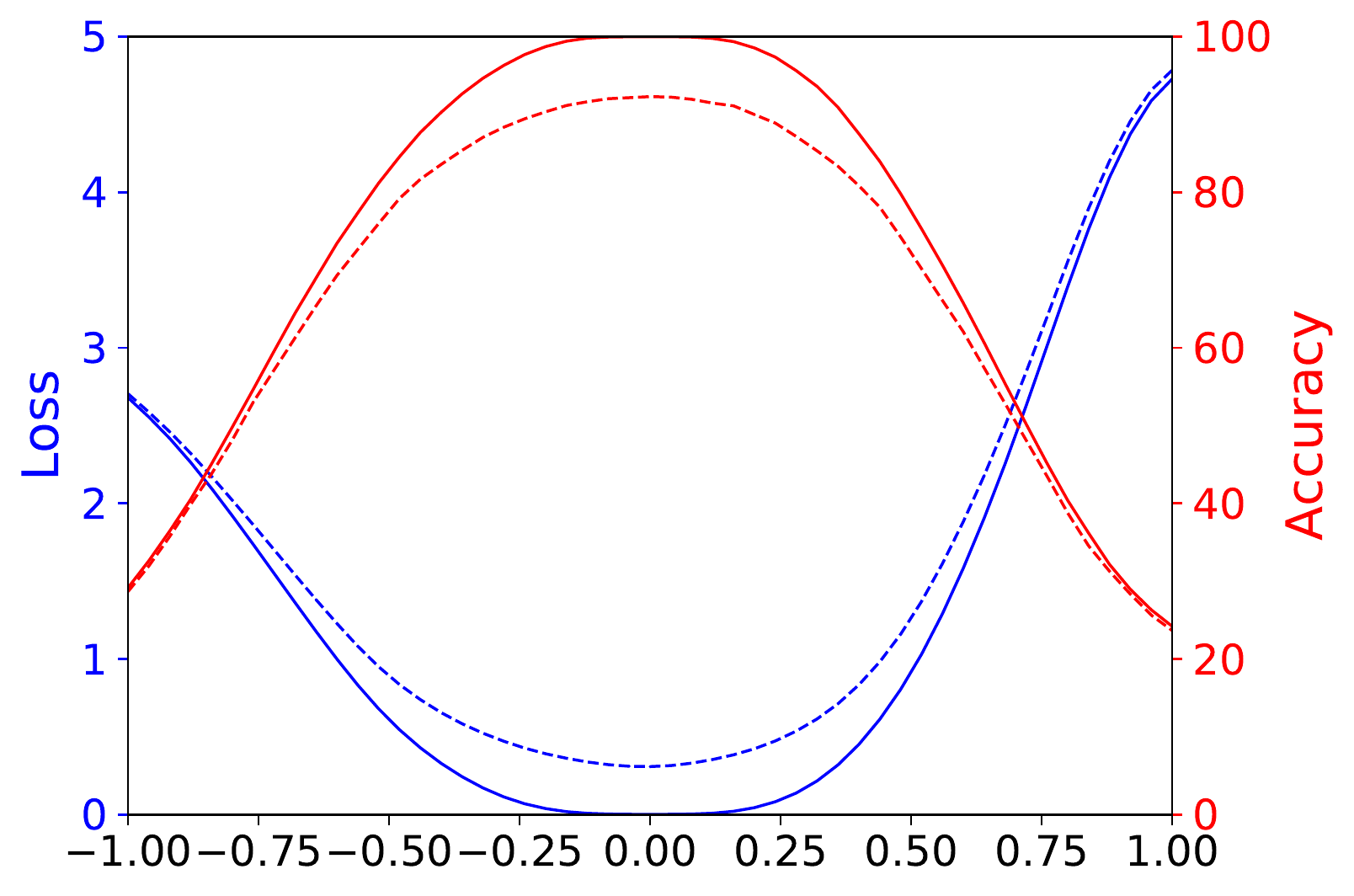}}
\subfigure[Adam, 8192, 9.52\%]{\includegraphics[width=0.25\linewidth]{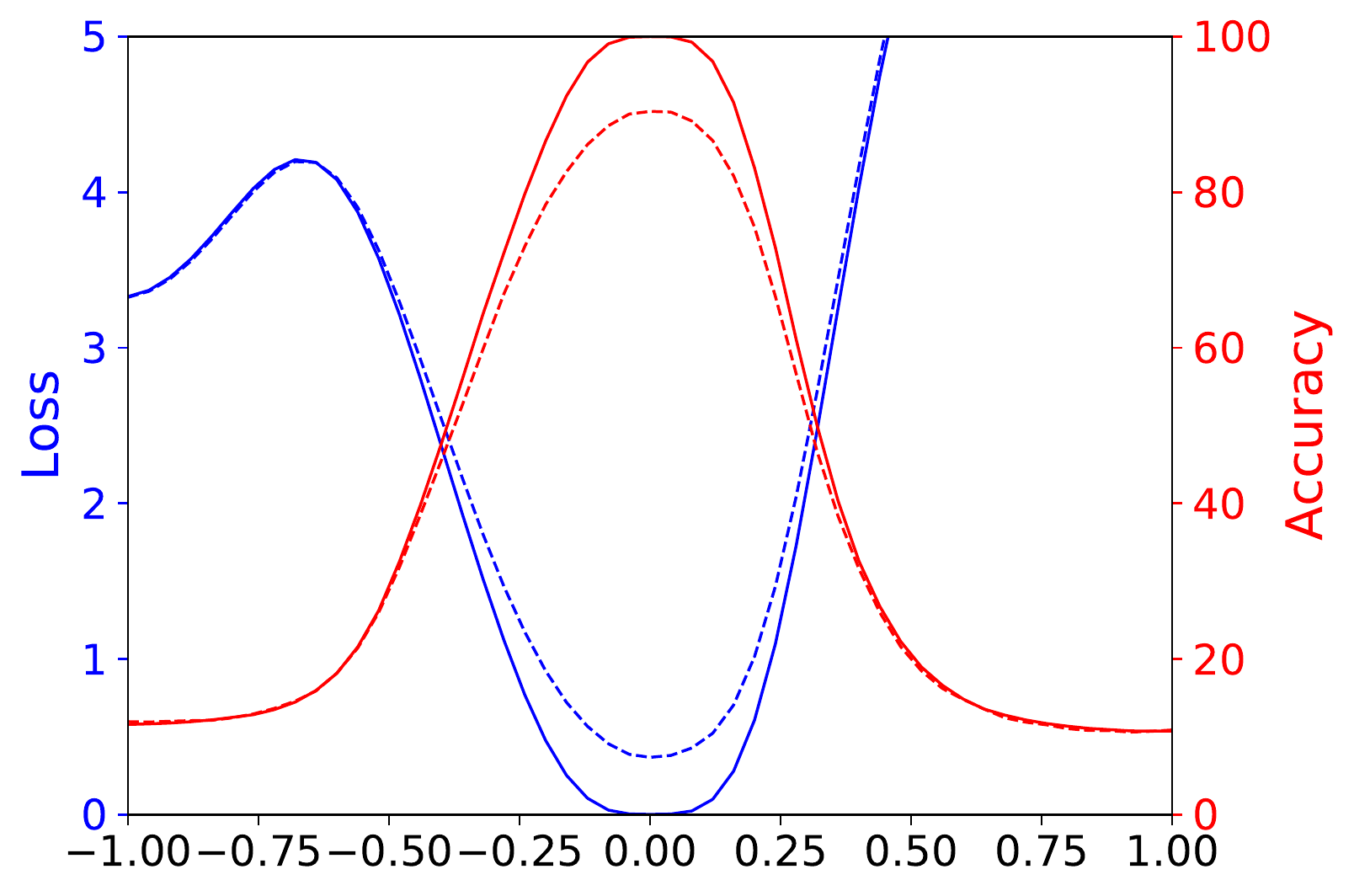}}
\end{tabular}
\caption{1D loss plots for VGG-9 with layer normalization.
The first row has no weight decay and the second row uses weight decay 5e-4.
}
\label{fig:vgg9_layernorm}
\end{figure*}

Figure~\ref{fig:vgg9_nonoram} shows the 1D plots without normalization.
One issue with the non-normalized plots is that the $x$-axis range must be chosen carefully.
Figure~\ref{fig:vgg9_nonoram_enlarged} shows enlarged plots with $[-0.2, 0.2]$ as the range for the $x$-axis.
Without normalization, the plots fail to show consistency between flatness and generalization error.
Here we compare filter normalization with layer normalization.
We find filter normalization is more accurate than layer normalization.
One failing case for layer normalization is shown in Figure~\ref{fig:vgg9_layernorm}, where Figure~\ref{fig:vgg9_layernorm}(g) is flatter than Figure~\ref{fig:vgg9_layernorm}(c), but with worse generalization error.

\subsection{Small-Batch vs Large-Batch for ResNet-56}
\label{sec:resnet56}
\begin{figure*}[!h]
\centering
\begin{tabular}{l}
\hspace{-3mm}
 \subfigure[SGD, WD=0]{\includegraphics[width=0.25\linewidth]{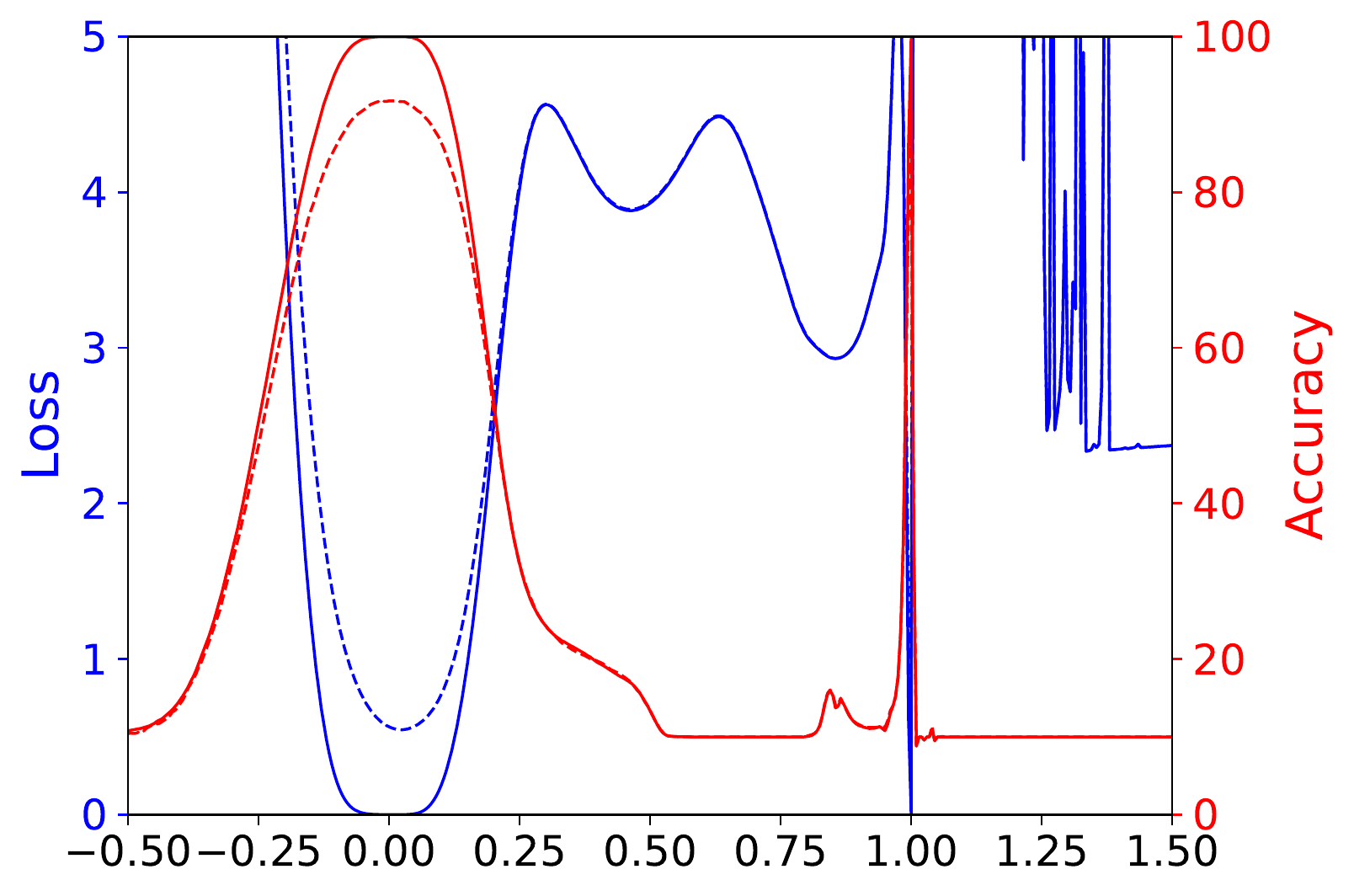}}
 \subfigure[SGD, WD=5e-4]{\includegraphics[width=0.25\linewidth]{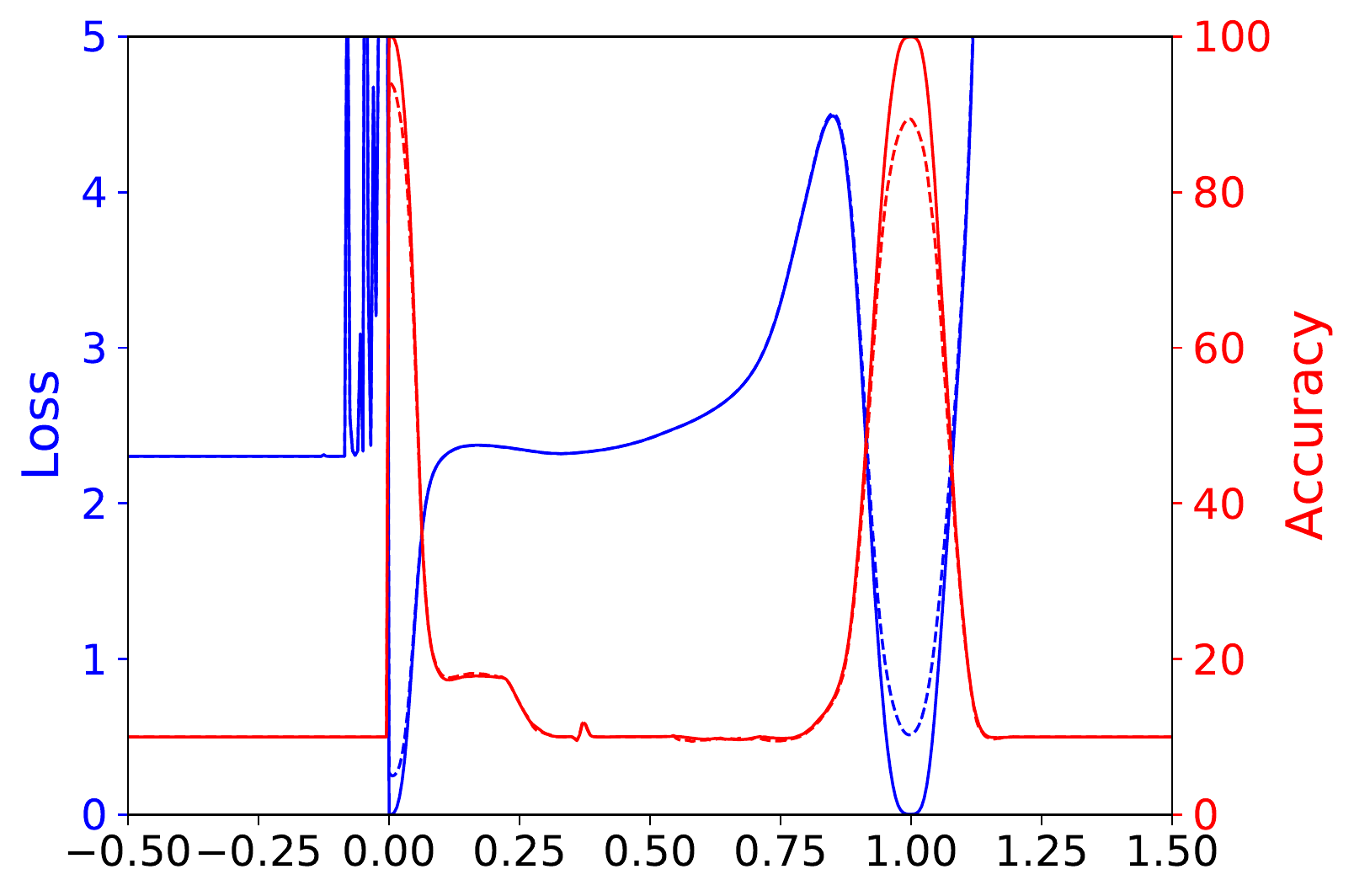}}
\subfigure[Adam, WD=0]{\includegraphics[width=0.25\linewidth]{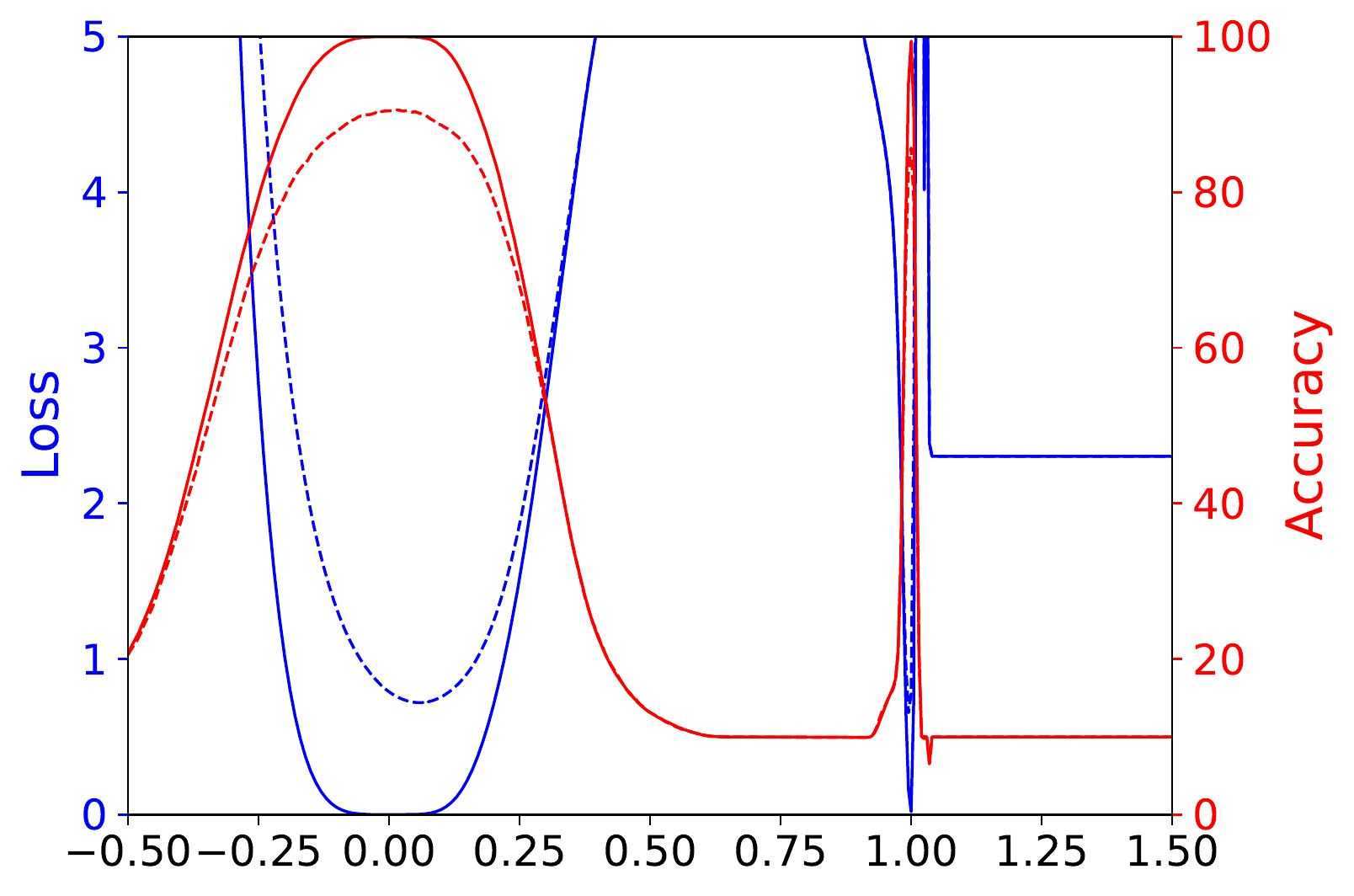}}
 \subfigure[Adam, WD=5e-4]{\includegraphics[width=0.25\linewidth]{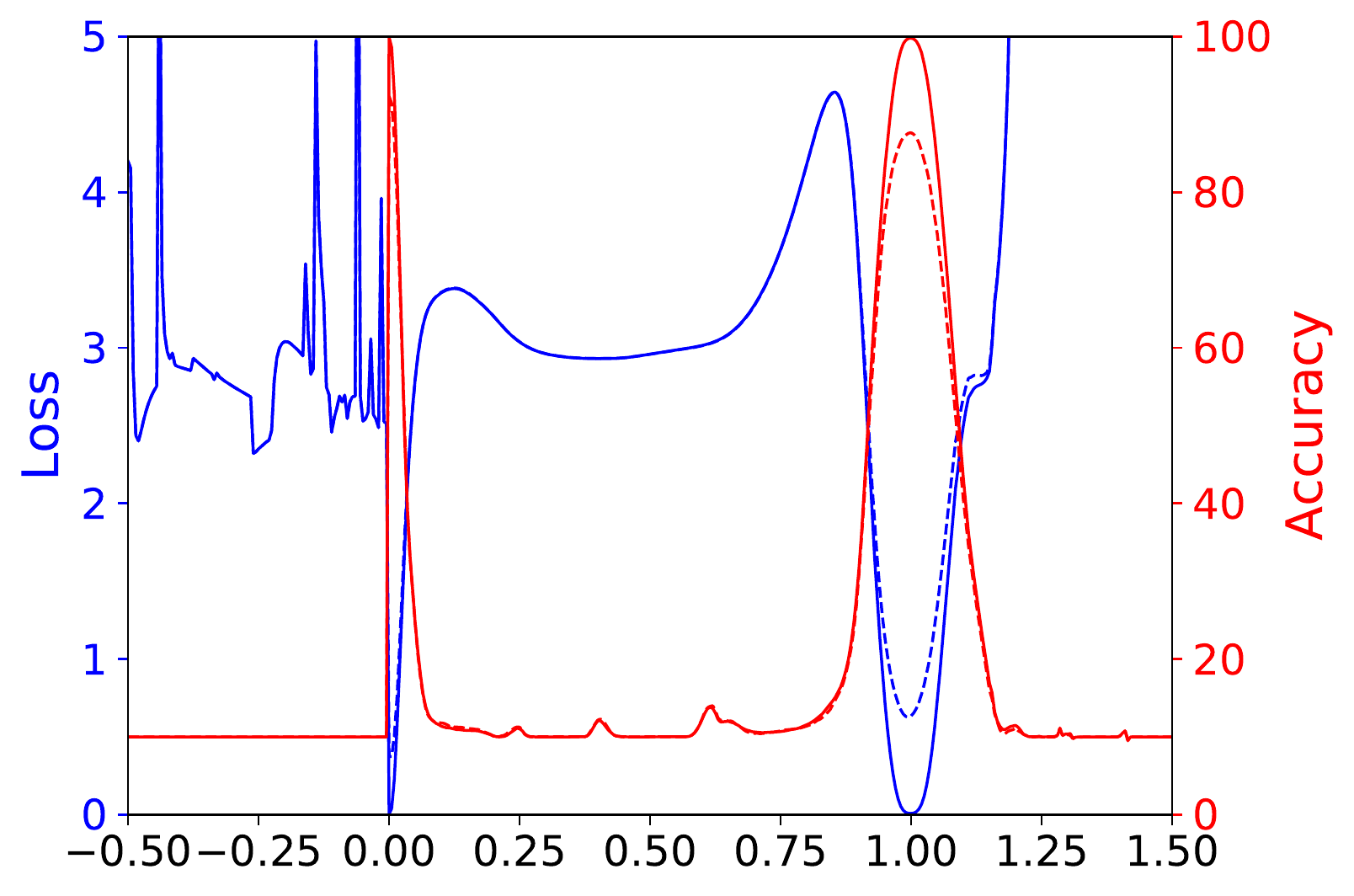}}
\end{tabular}
\caption{1D linear interpolation of solutions obtained by small-batch and large-batch methods for ResNet-56. The blue lines are loss values and the red lines are error.}
\label{fig:weight_decay_batchsize_resnet56}
\vspace{-3mm}
\end{figure*}

\begin{figure*}[!h]
\centering
\begin{tabular}{l}
\hspace{-3mm}
\subfigure[SGD, 128, 8.26\%]{\includegraphics[width=0.25\linewidth]{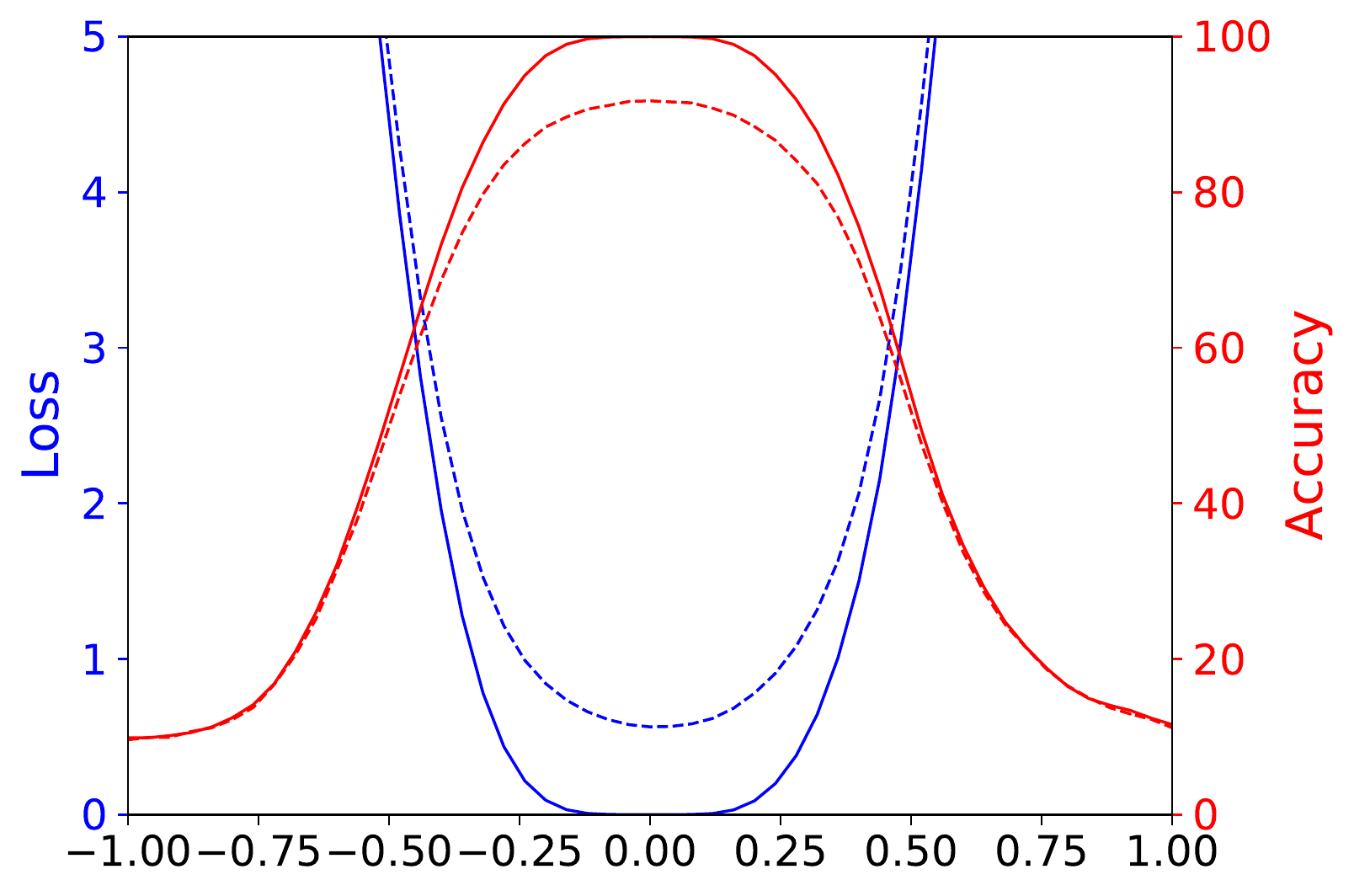}}
\subfigure[SGD, 4096, 13.93\%]{\includegraphics[width=0.25\linewidth]{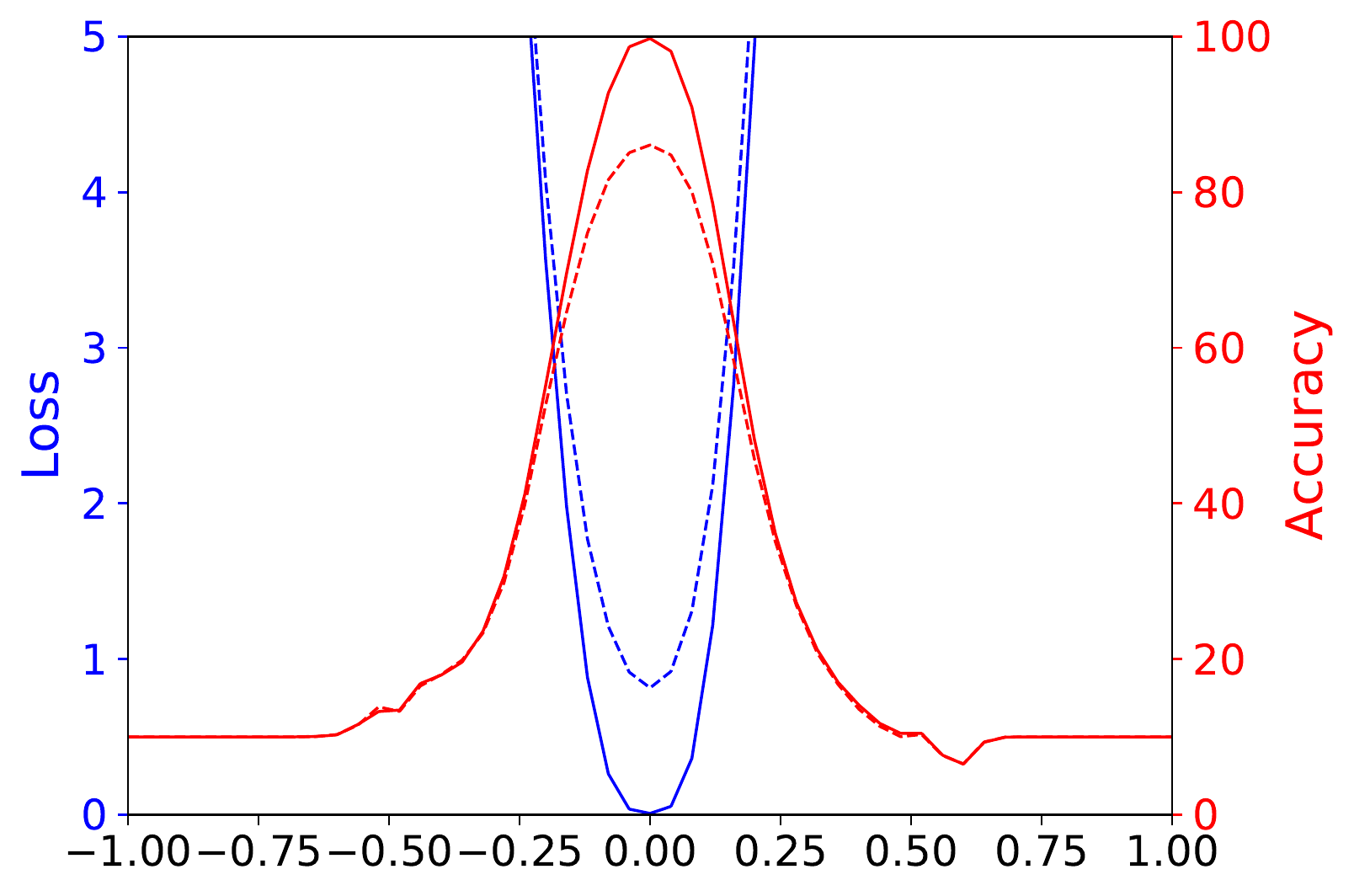}}
\subfigure[Adam, 128, 9.55\%]{\includegraphics[width=0.25\linewidth]{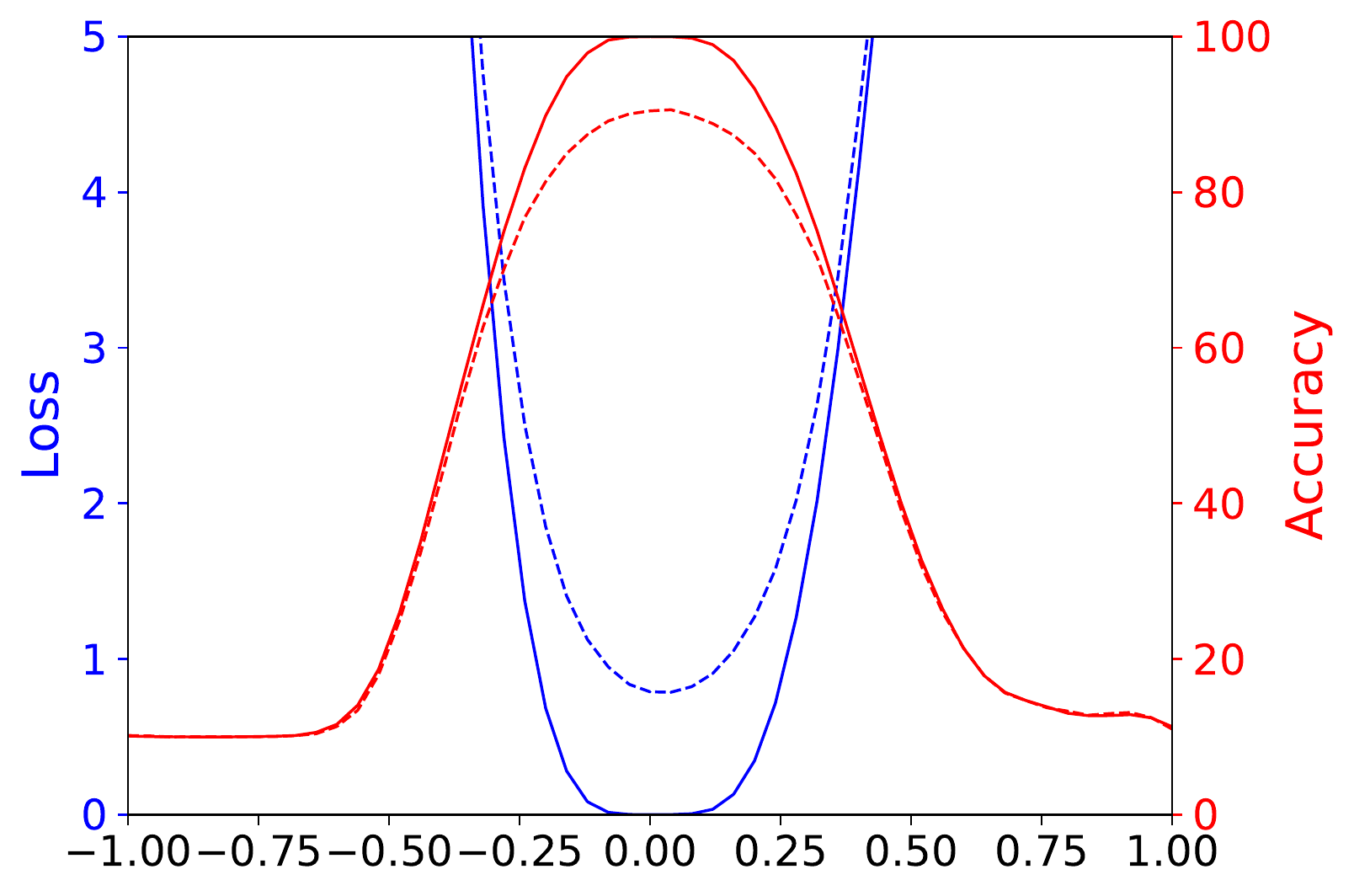}}
\subfigure[Adam, 4096, 14.30\%]{\includegraphics[width=0.25\linewidth]{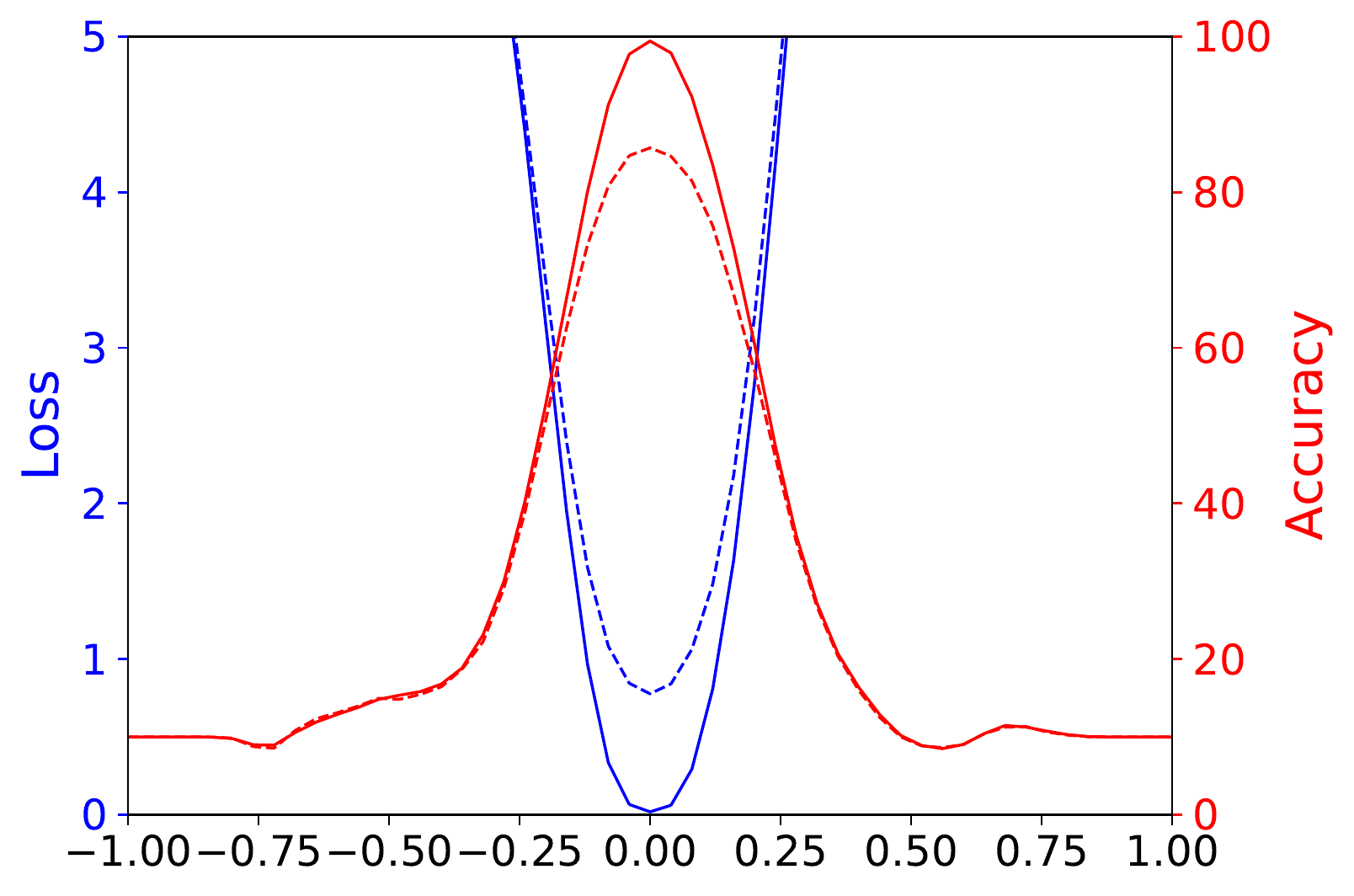}}\\
\hspace{-3mm}
\subfigure[SGD, 128, 5.89\%]{\includegraphics[width=0.25\linewidth]{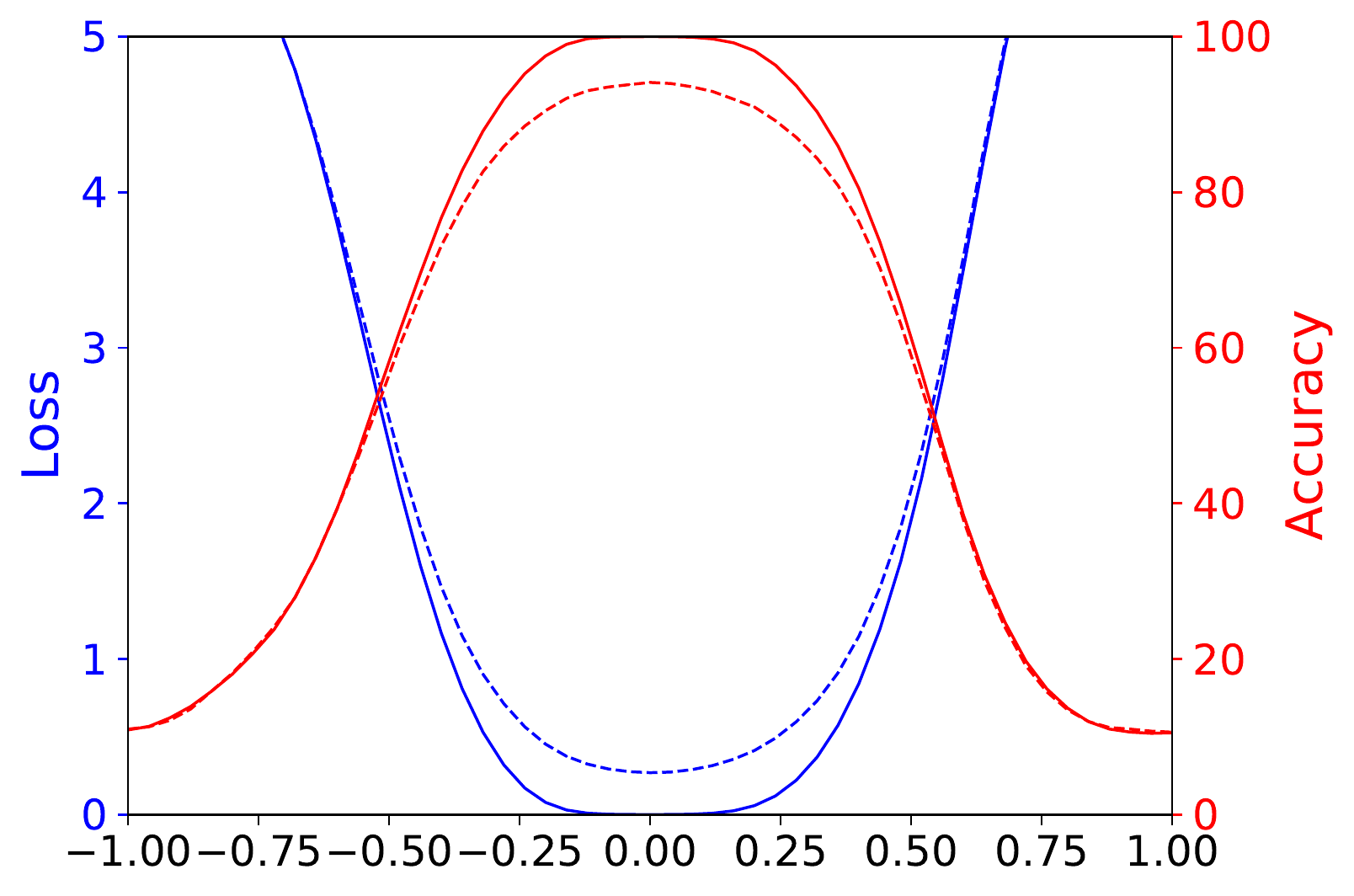}}
\subfigure[SGD, 4096, 10.59\%]{\includegraphics[width=0.25\linewidth]{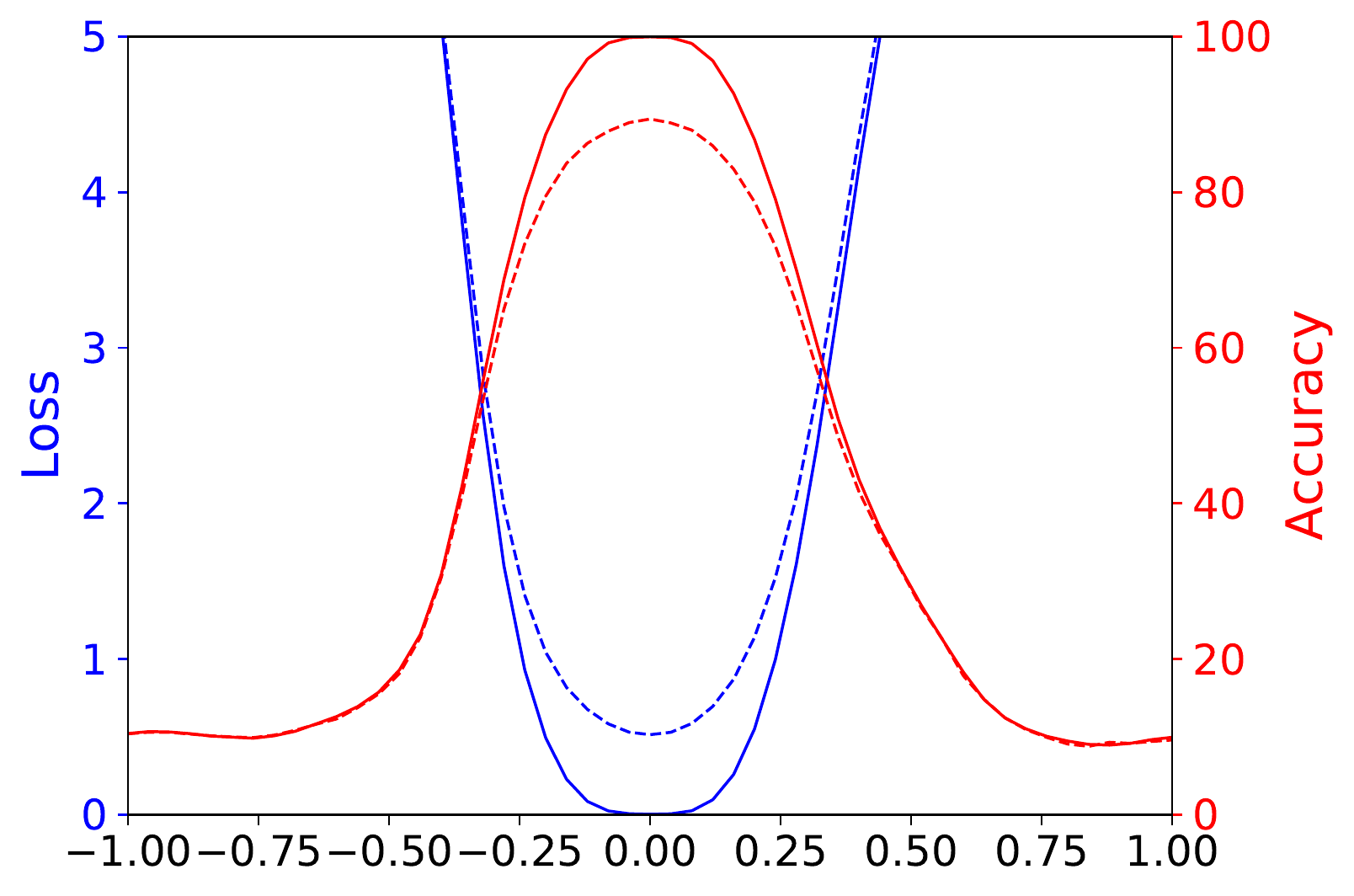}}
\subfigure[Adam, 128, 7.67\%]{\includegraphics[width=0.25\linewidth]{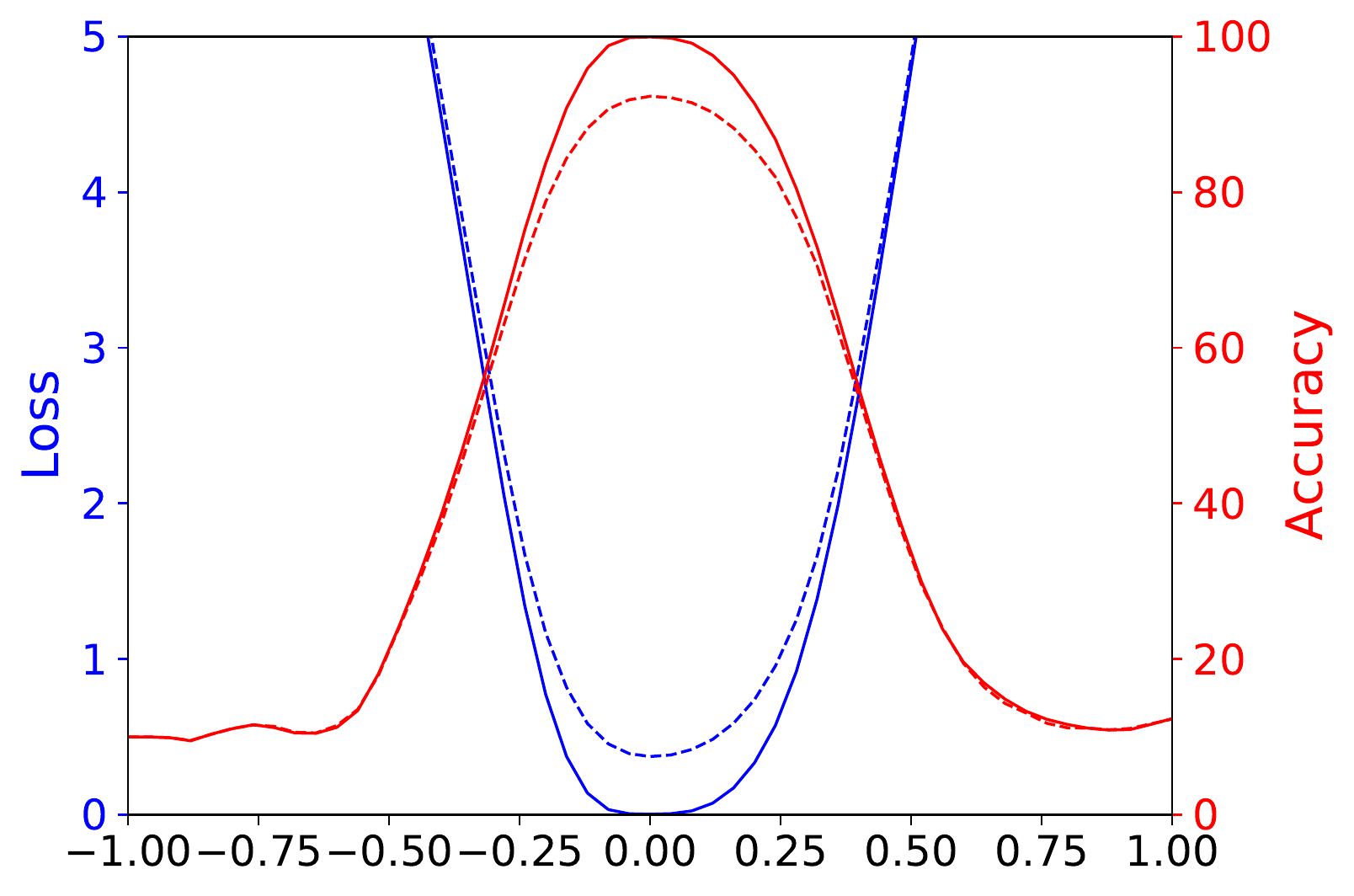}}
\subfigure[Adam, 4096, 12.36\%]{\includegraphics[width=0.25\linewidth]{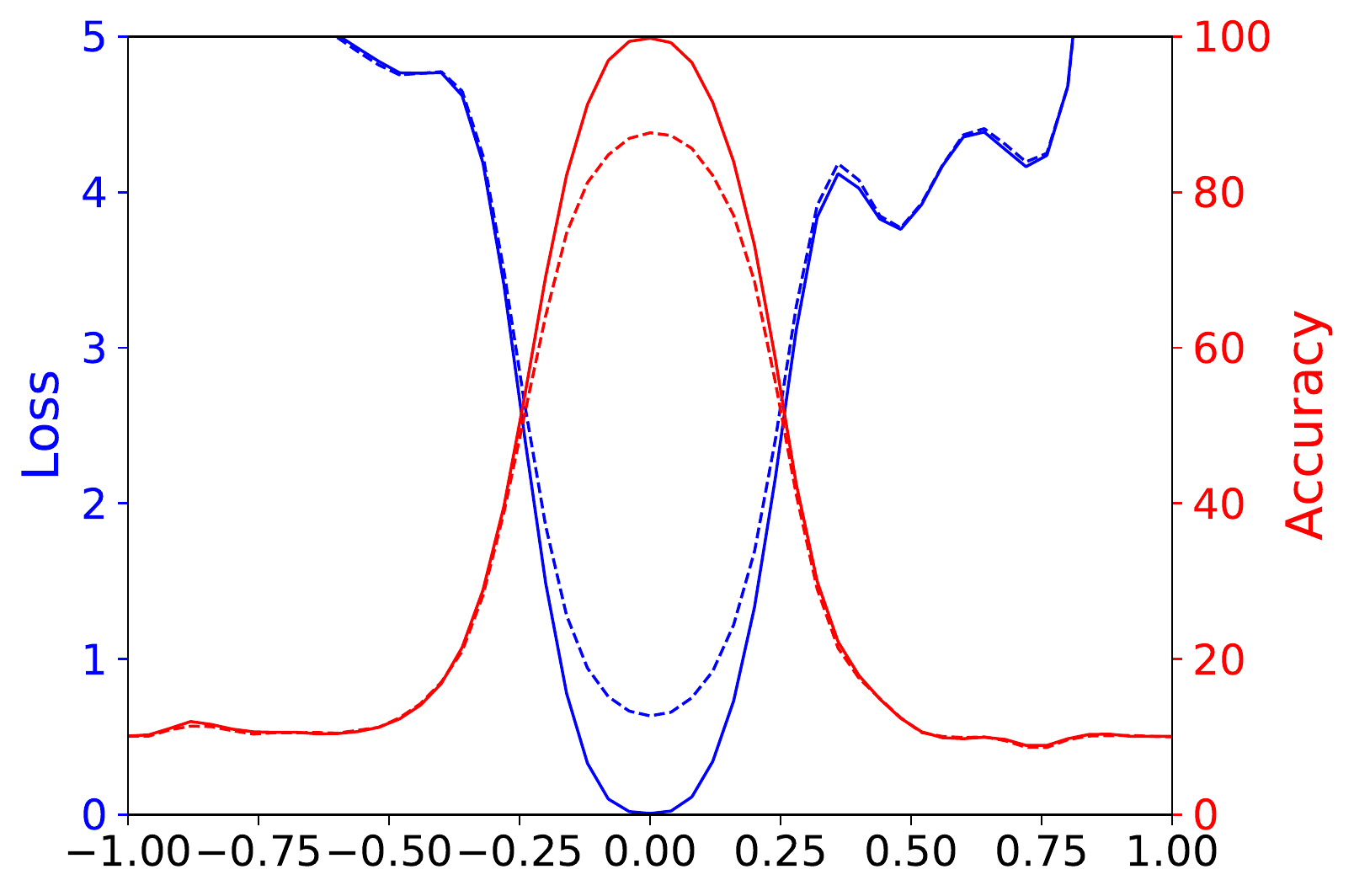}}\\
\hspace{-3mm}
\subfigure[SGD, 128, 8.26\%]{\includegraphics[width=0.25\linewidth]{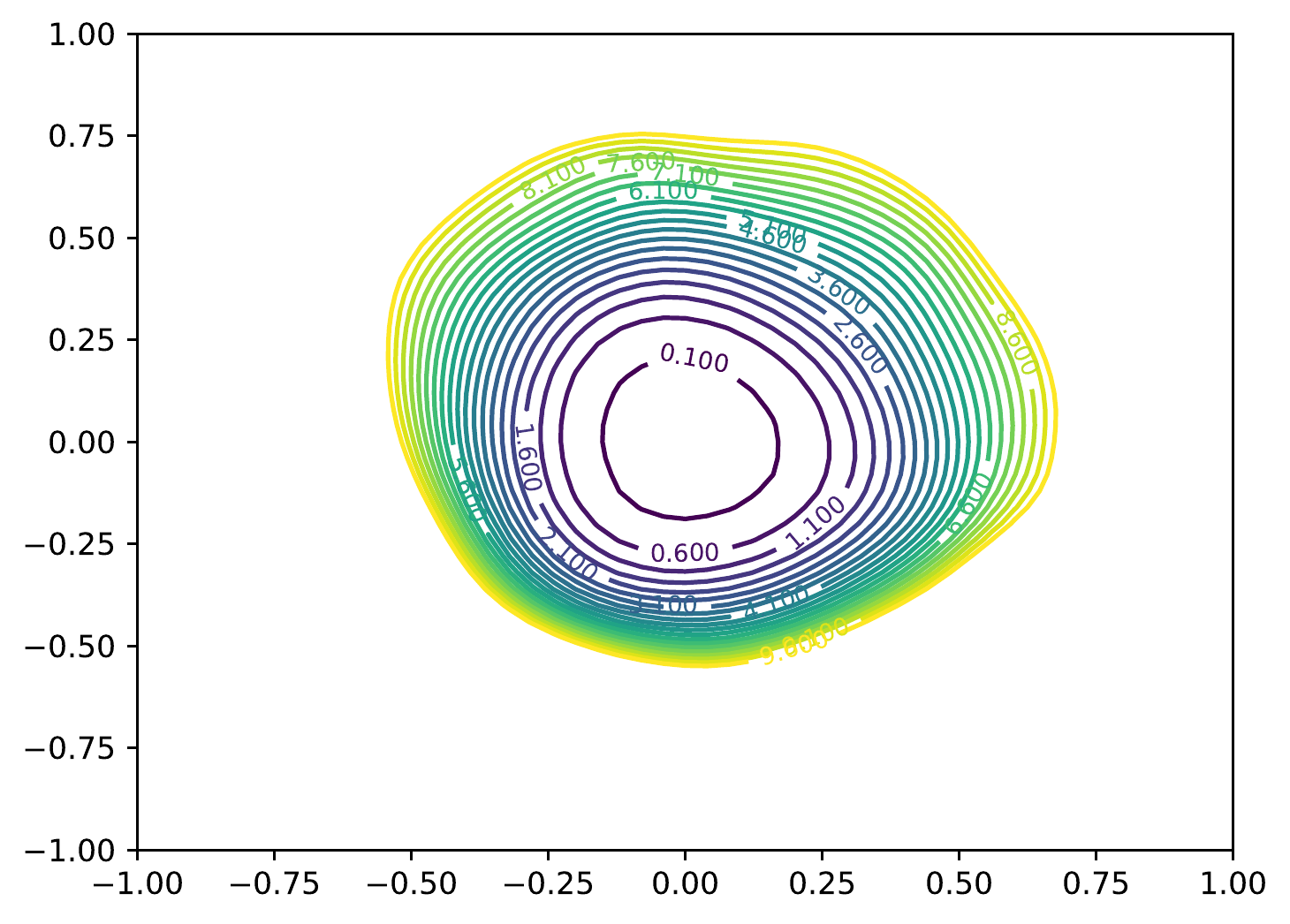}}
\subfigure[SGD, 4096, 13.93\%]{\includegraphics[width=0.25\linewidth]{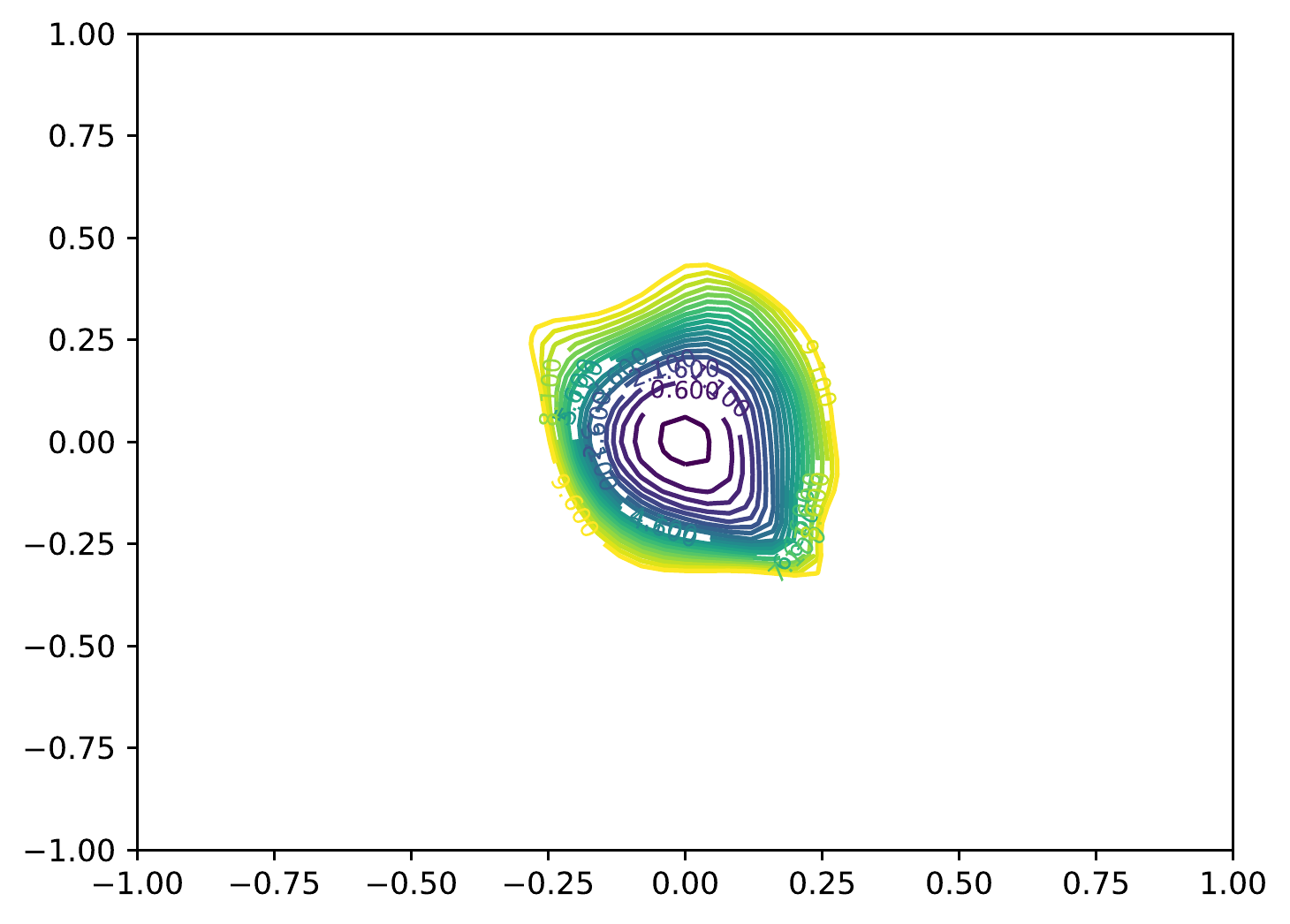}}
\subfigure[Adam, 128, 9.55\%]{\includegraphics[width=0.25\linewidth]{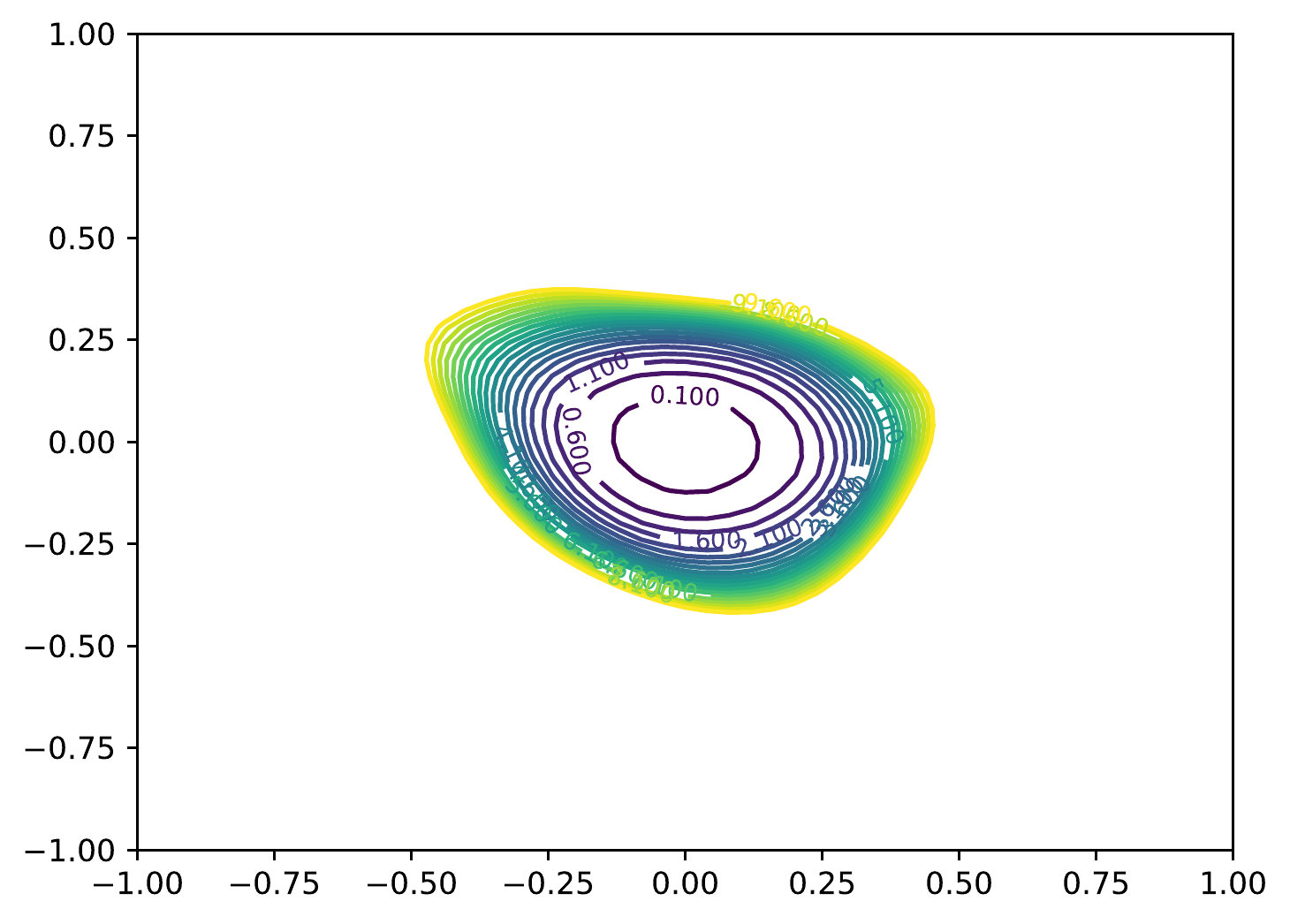}}
\subfigure[Adam, 4096, 14.30\%]{\includegraphics[width=0.25\linewidth]{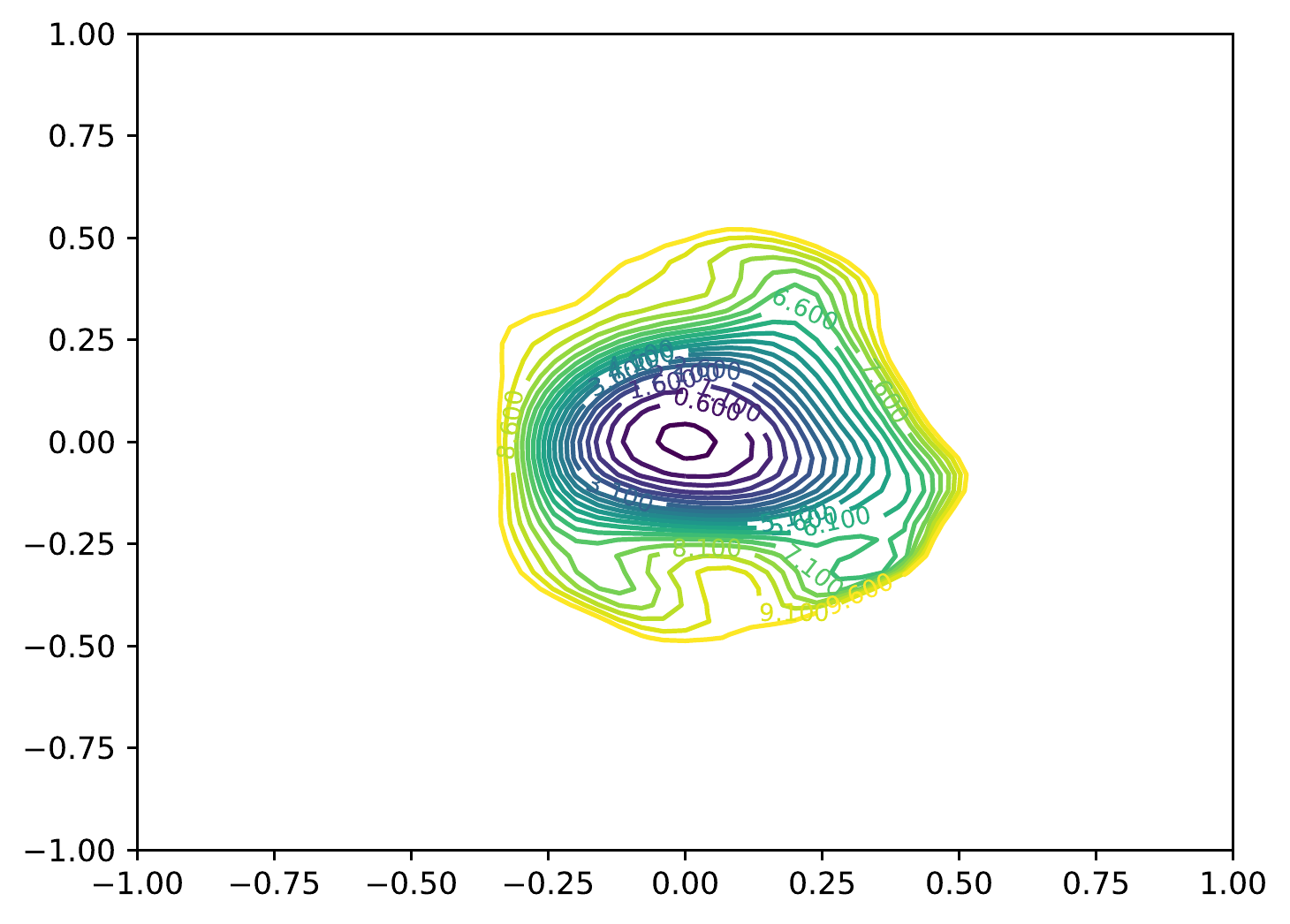}}\\
\hspace{-3mm}
\subfigure[SGD, 128, 5.89\%]{\includegraphics[width=0.25\linewidth]{figures/{resnet56_sgd_lr=0.1_bs=128_wd=0.0005/resnet56_random_-1.0,1.0x-1.0,1.0.h5_2dcontour}.pdf}}
\subfigure[SGD, 4096, 10.59\%]{\includegraphics[width=0.25\linewidth]{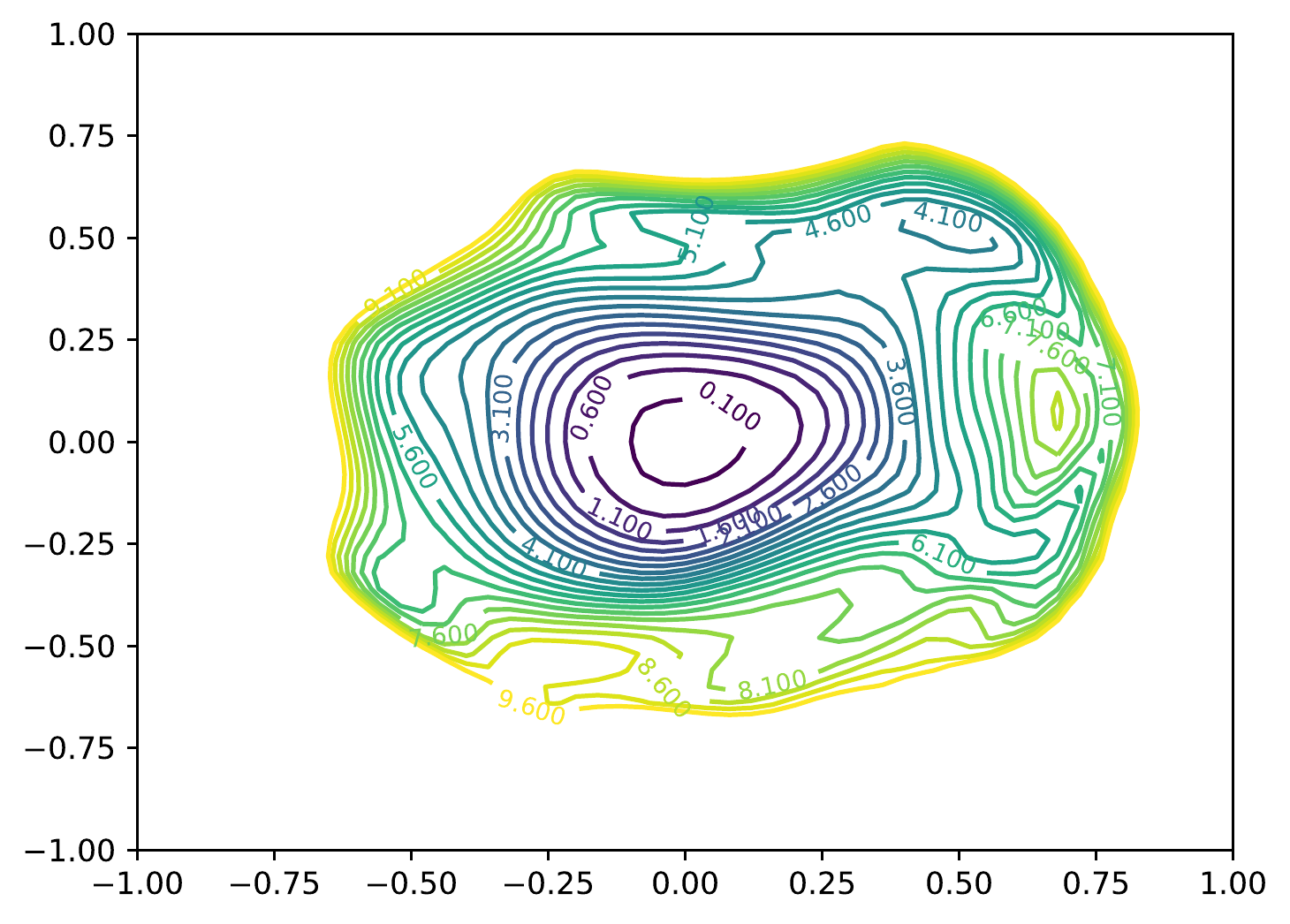}}
\subfigure[Adam, 128, 7.67\%]{\includegraphics[width=0.25\linewidth]{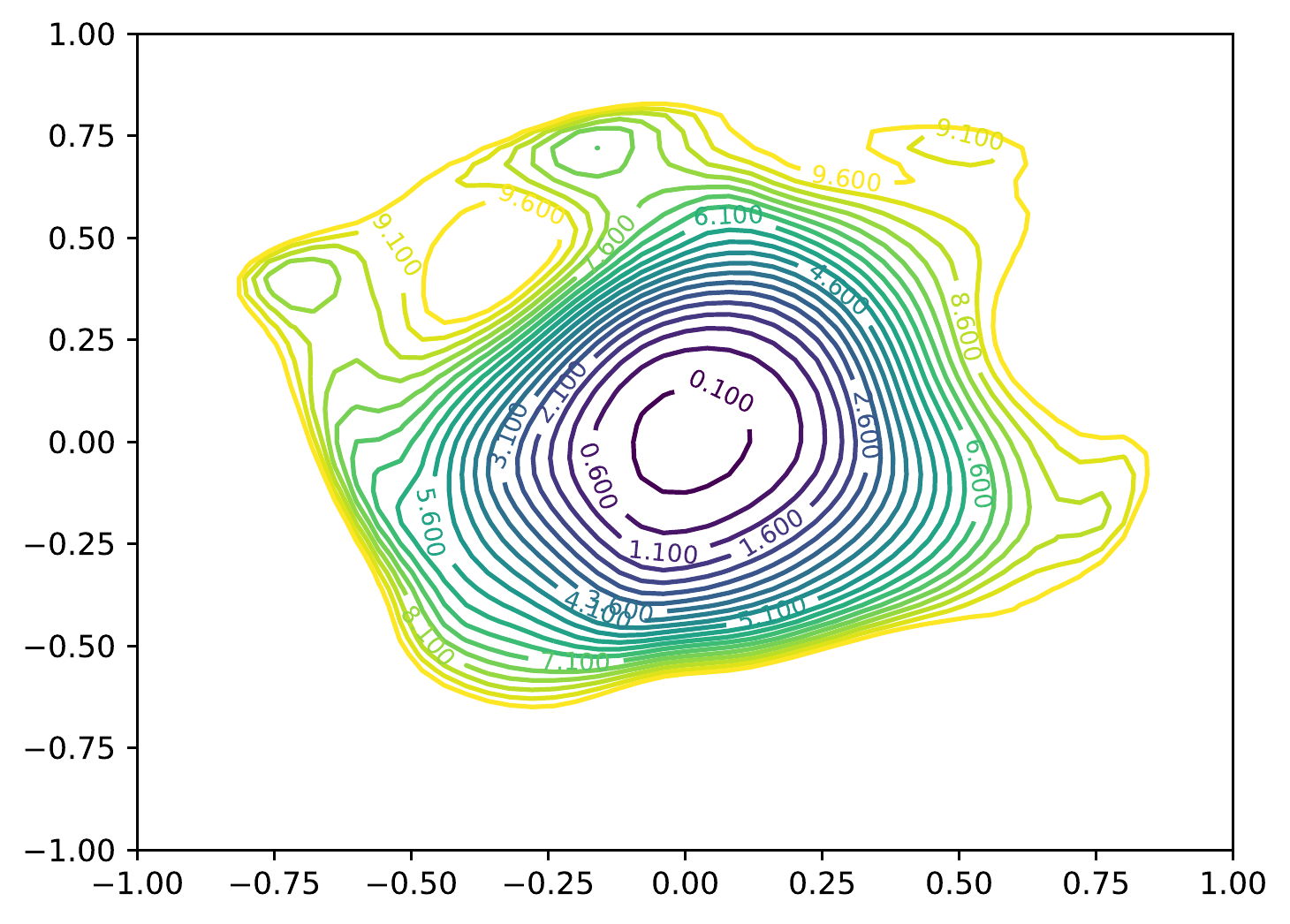}}
\subfigure[Adam, 4096, 12.36\%]{\includegraphics[width=0.25\linewidth]{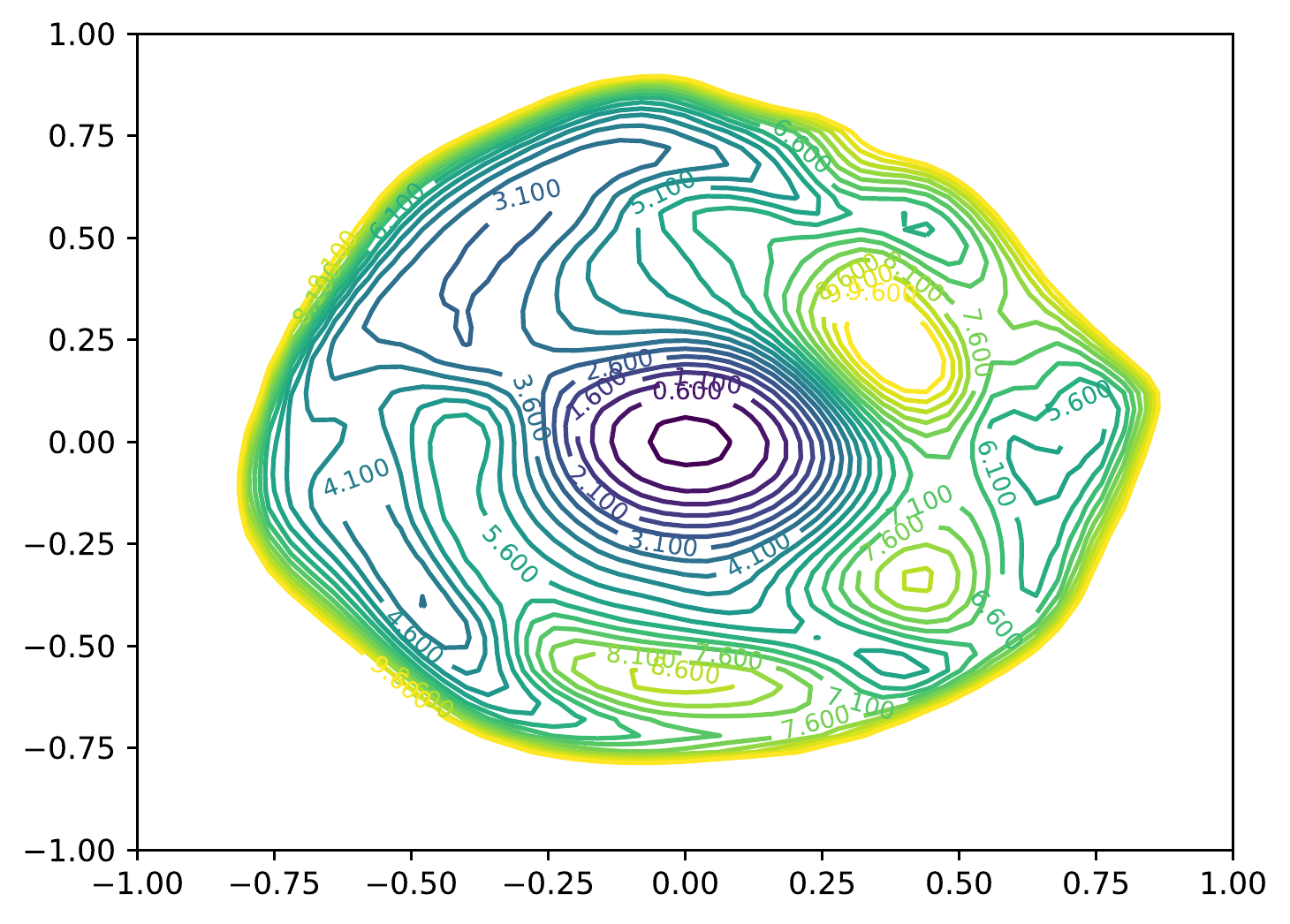}}
\end{tabular}
\caption{1D and 2D visualization of ResNet-56 trained with different optimizer, batch size, and weight decay.
The first and third row uses zero weight decay and the second and fourth row uses 5e-4 weight decay.
}
\label{fig:noramlized_shape_batchsize_resnet56}
\vspace{-3mm}
\end{figure*}

Similar to the observations made in Section~\ref{sec:sharp}, the ``sharp vs flat dilemma" also applies to ResNet-56 as shown in Figure~\ref{fig:weight_decay_batchsize_resnet56}.
The generalization error for each solution is shown in Table \ref{tab:test_acc56}.
The 1D and 2D visualizations with filter normalized directions are shown in Figure~\ref{fig:noramlized_shape_batchsize_resnet56}.

\begin{table}[!h]
\centering
\caption{Test errors for ResNet-56 with different optimizer, batch-size and weight-decay.}
\label{tab:test_acc56}
\begin{tabular}{lcccc}
\toprule
                          & \multicolumn{2}{c}{SGD} & \multicolumn{2}{c}{Adam} \\\cline{2-5}
                          & bs=128     & bs=4096    & bs=128     & bs=4096     \\\cline{2-5}
WD = 0                    & 8.26       & 13.93      & 9.55           &  14.30    \\
WD = 5e-4                 & \textbf{5.89}& \textbf{10.59} & \textbf{7.67} &  \textbf{12.36}           \\
\bottomrule
\end{tabular}
\vspace{-3mm}
\end{table}

\clearpage
\subsection{Repeatability of the Loss Surface Visualization}

\begin{figure*}[!h]
\centering
\begin{tabular}{l}
\hspace{-.3cm}
\subfigure[SGD, 128, 7.37\%]{\includegraphics[width=0.25\linewidth]{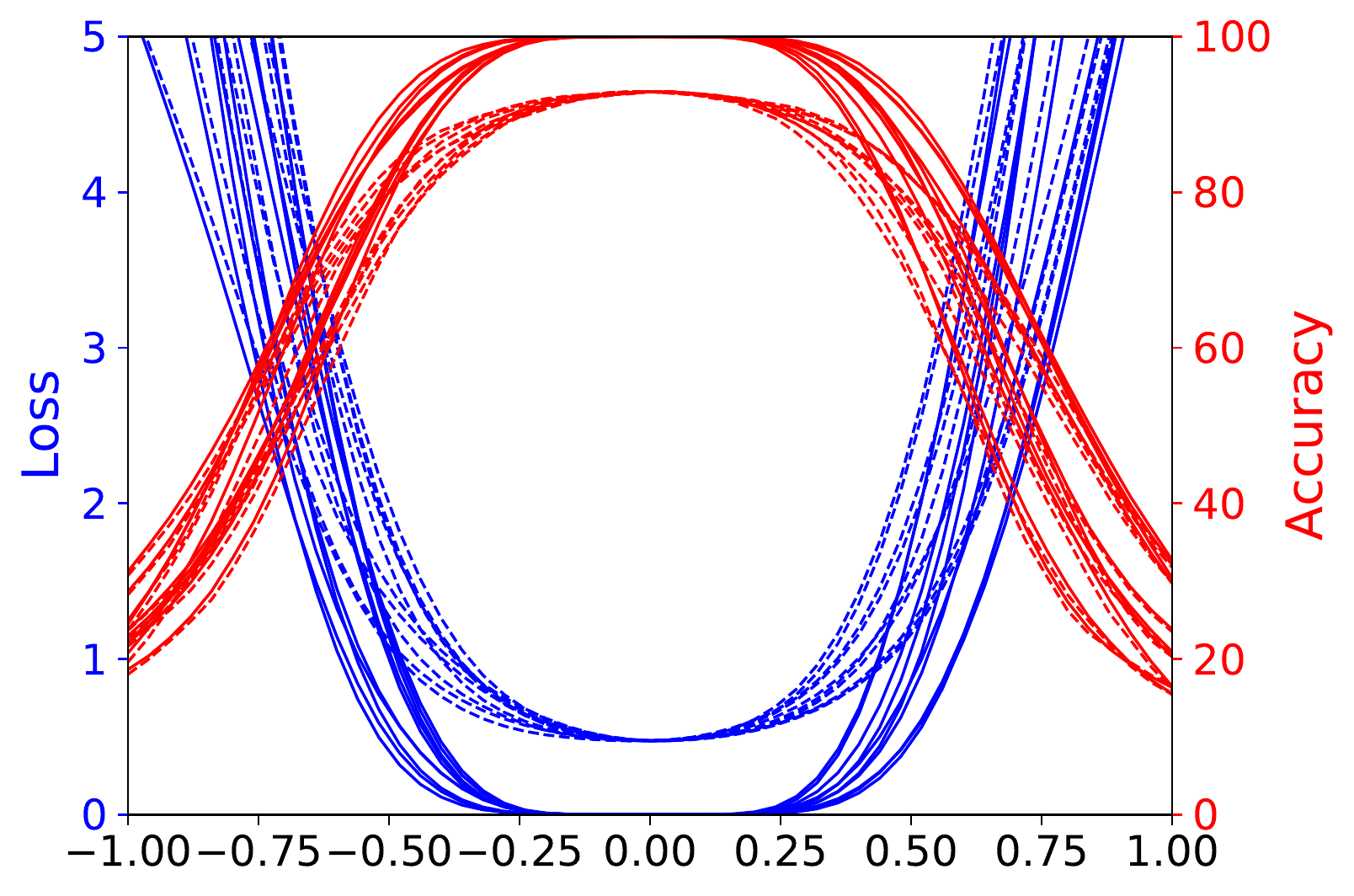}}
\subfigure[SGD, 8192, 11.07\%]{\includegraphics[width=0.25\linewidth]{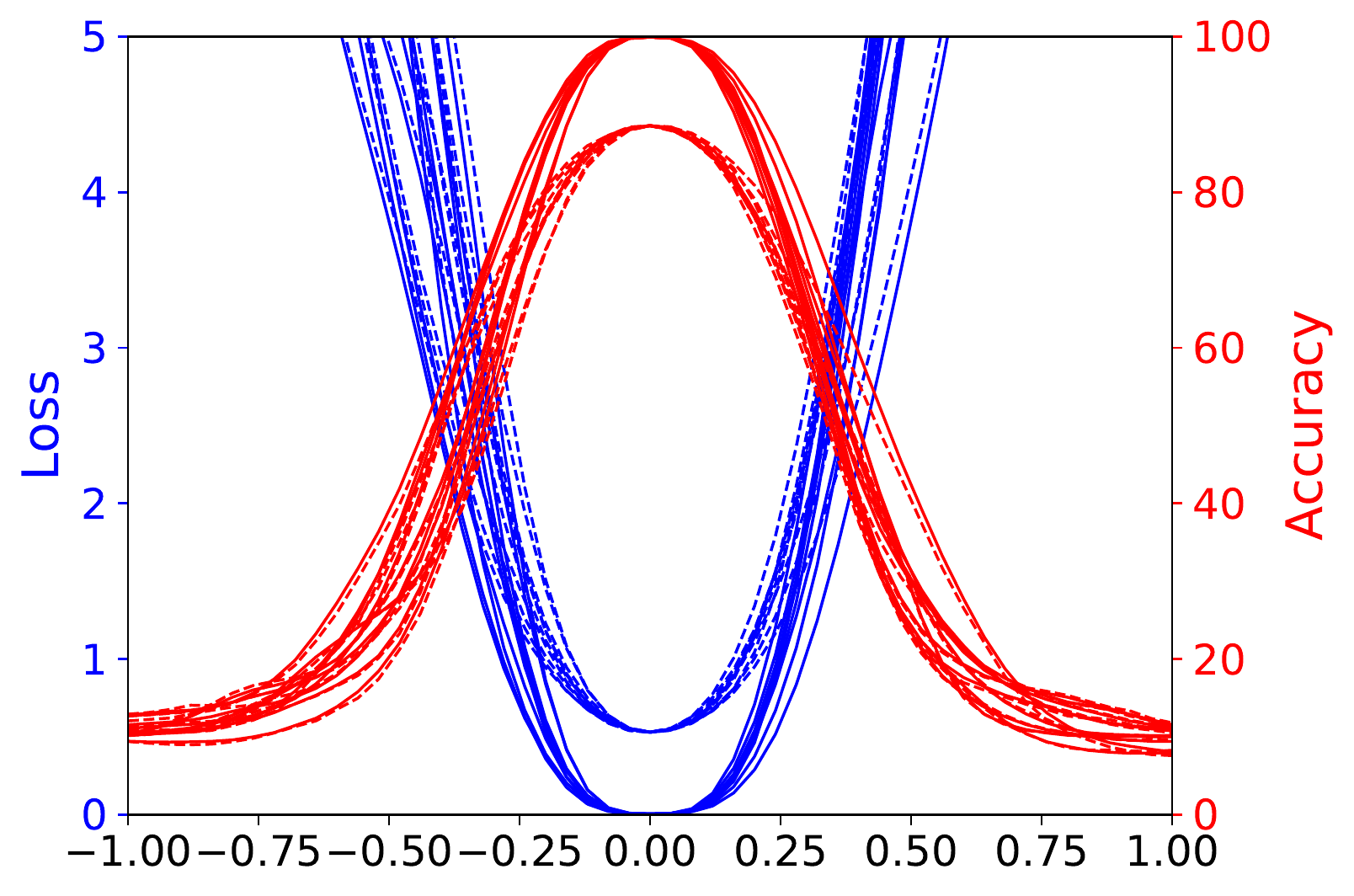}}
\subfigure[Adam, 128, 7.44\%]{\includegraphics[width=0.25\linewidth]{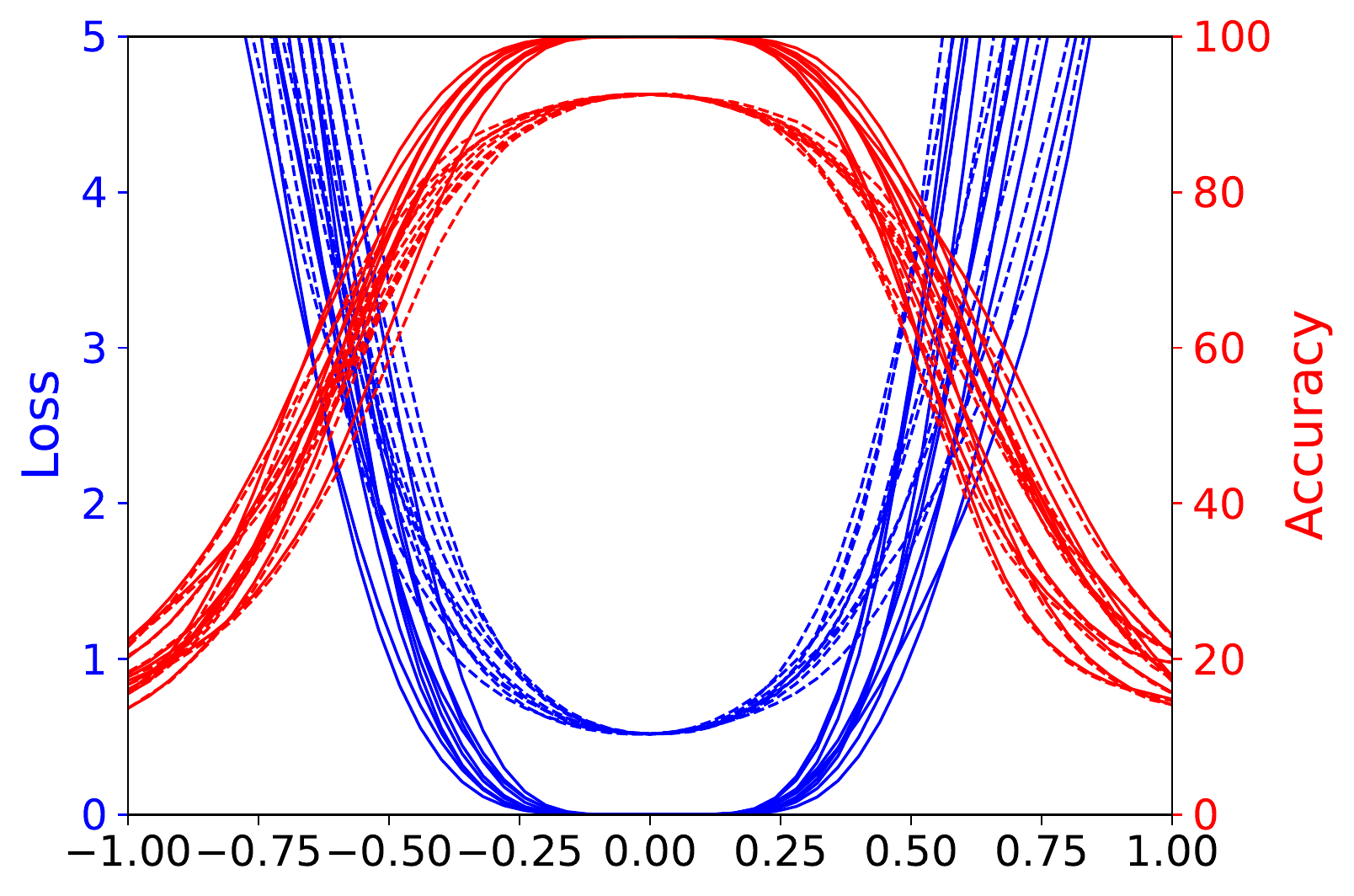}}
\subfigure[Adam, 8192, 10.91\%]{\includegraphics[width=0.25\linewidth]{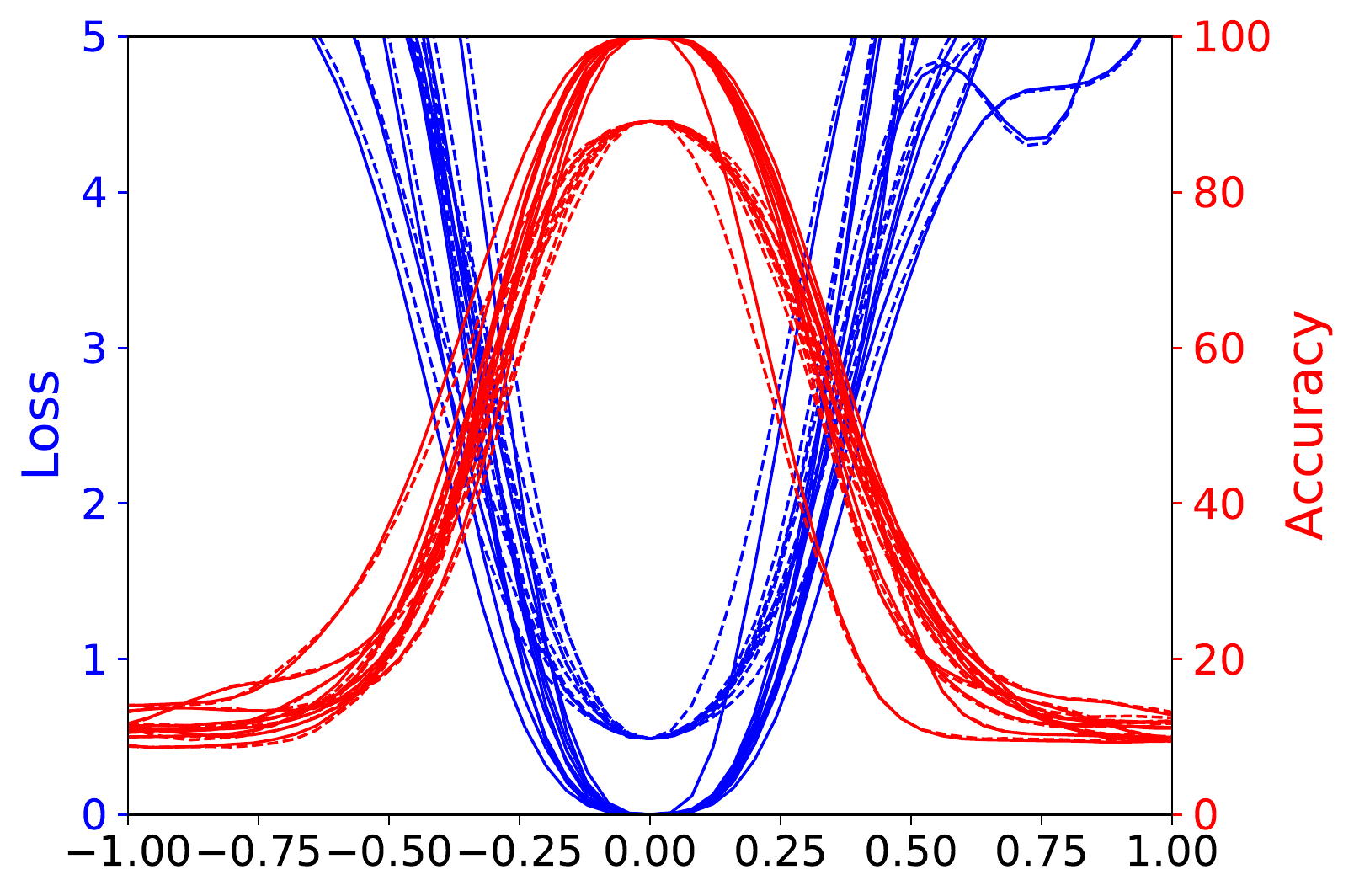}
}\\
\hspace{-.3cm}
\subfigure[SGD, 128, 6.00\%]{\includegraphics[width=0.25\linewidth]{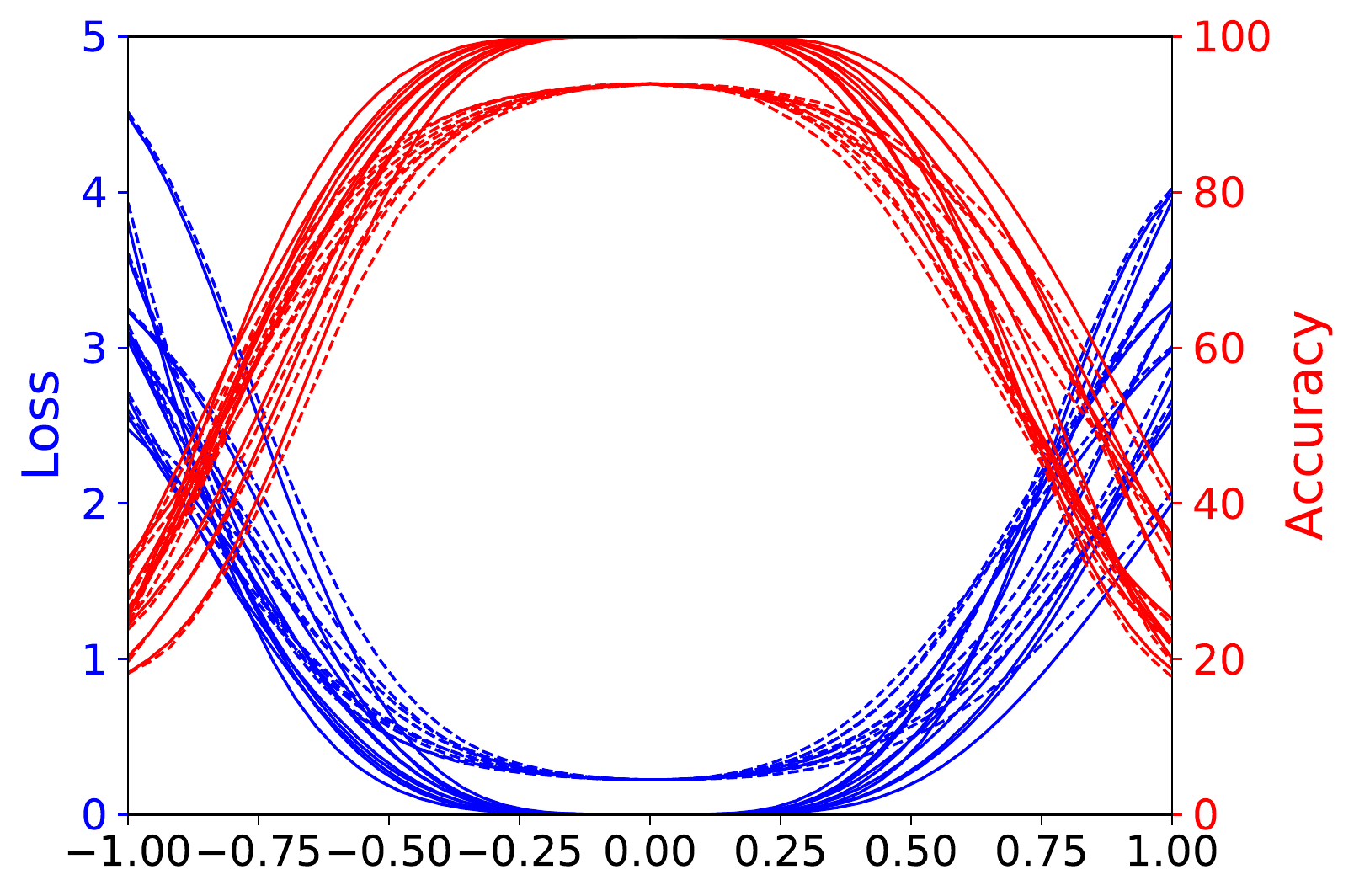}}
\subfigure[SGD, 8192, 10.19\%]{\includegraphics[width=0.25\linewidth]{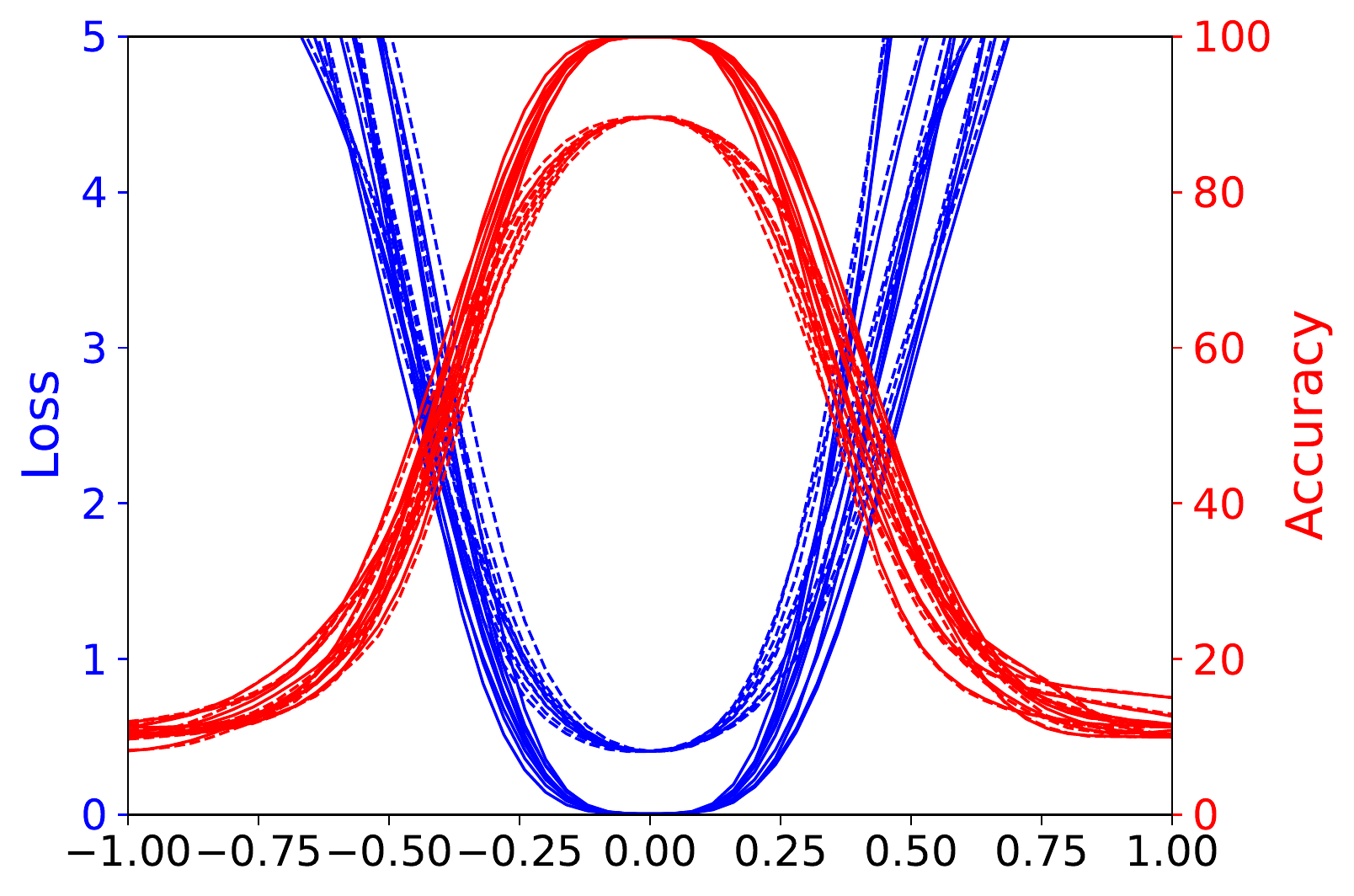}}
\subfigure[Adam, 128, 7.80\%]{\includegraphics[width=0.25\linewidth]{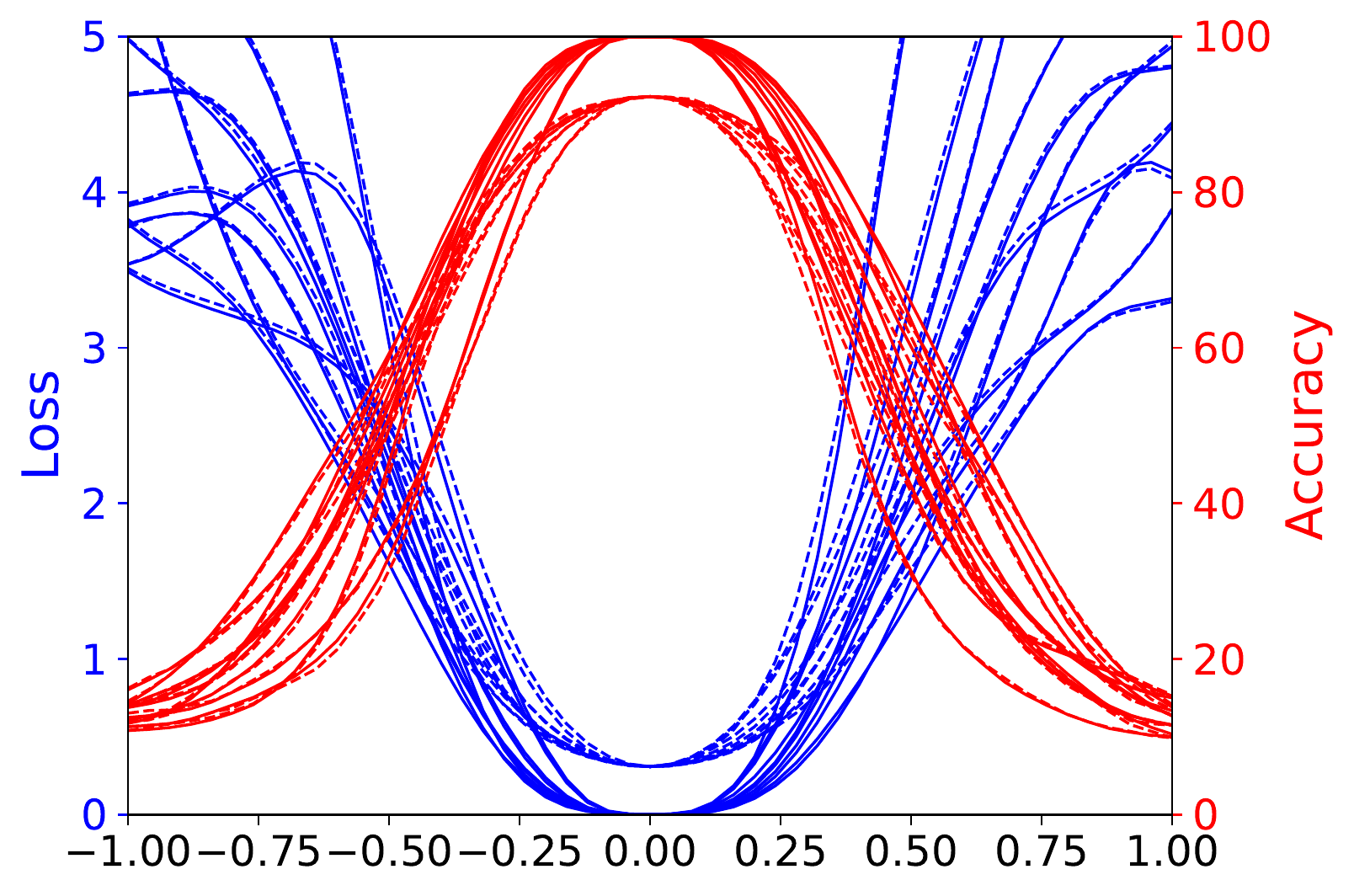}}
\subfigure[Adam, 8192, 9.52\%]{\includegraphics[width=0.25\linewidth]{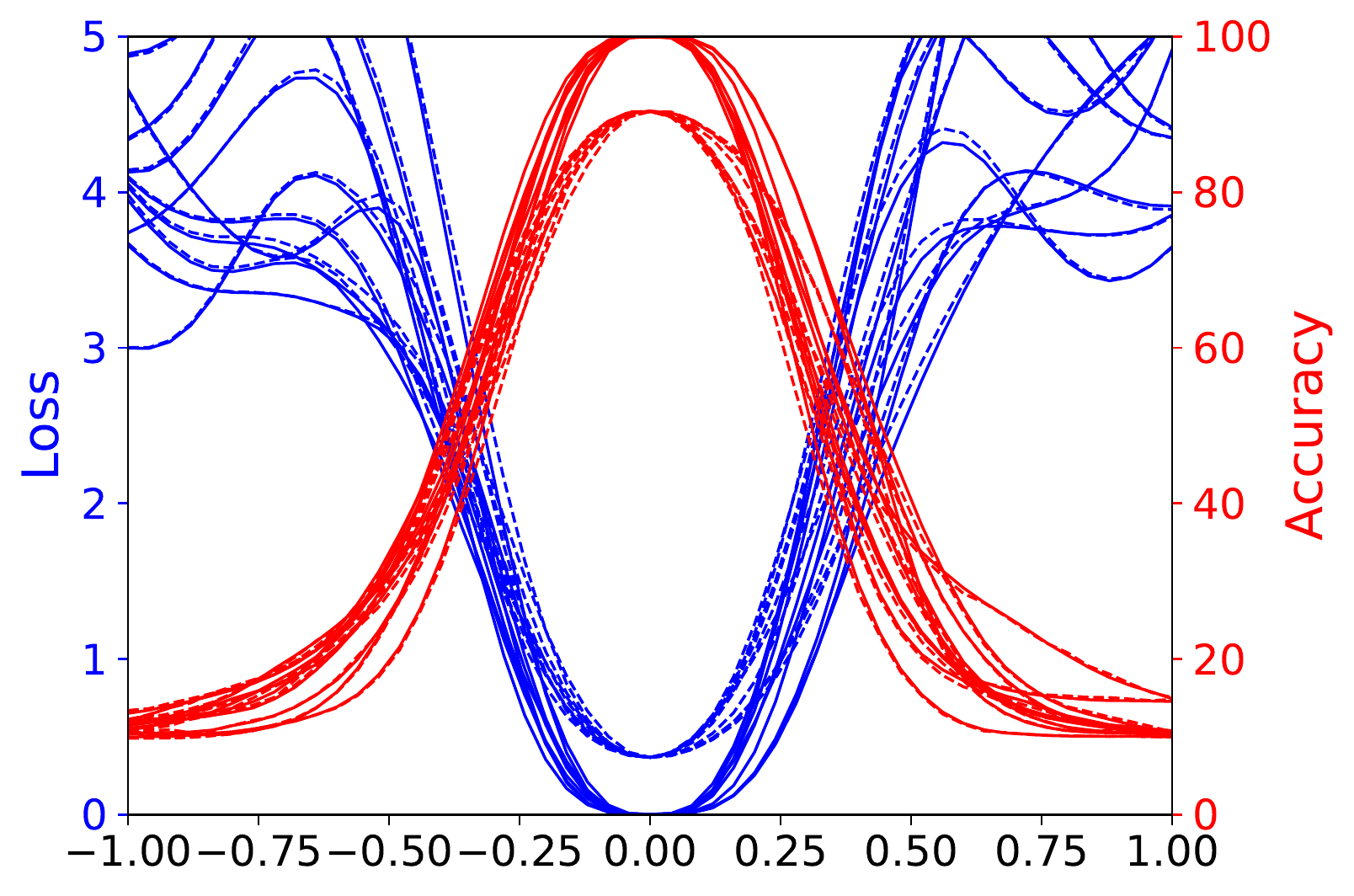}}
\end{tabular}
\caption{Repeatability of the surface plots for VGG-9 with filter normalization.
The shape of minima obtained using 10 different random filter-normalized directions.
}
\label{fig:filter_normalization_repeat}
\end{figure*}

\begin{figure*}[h]
\centering
\hspace{-6mm}
\begin{tabular}{l}
\includegraphics[width=0.245\linewidth]{figures/{resnet56_noshort_sgd_lr=0.1_bs=128_wd=0.0005/resnet56_noshort_random_-1.0,1.0x-1.0,1.0.h5_2dcontour}.pdf}
\includegraphics[width=0.245\linewidth]{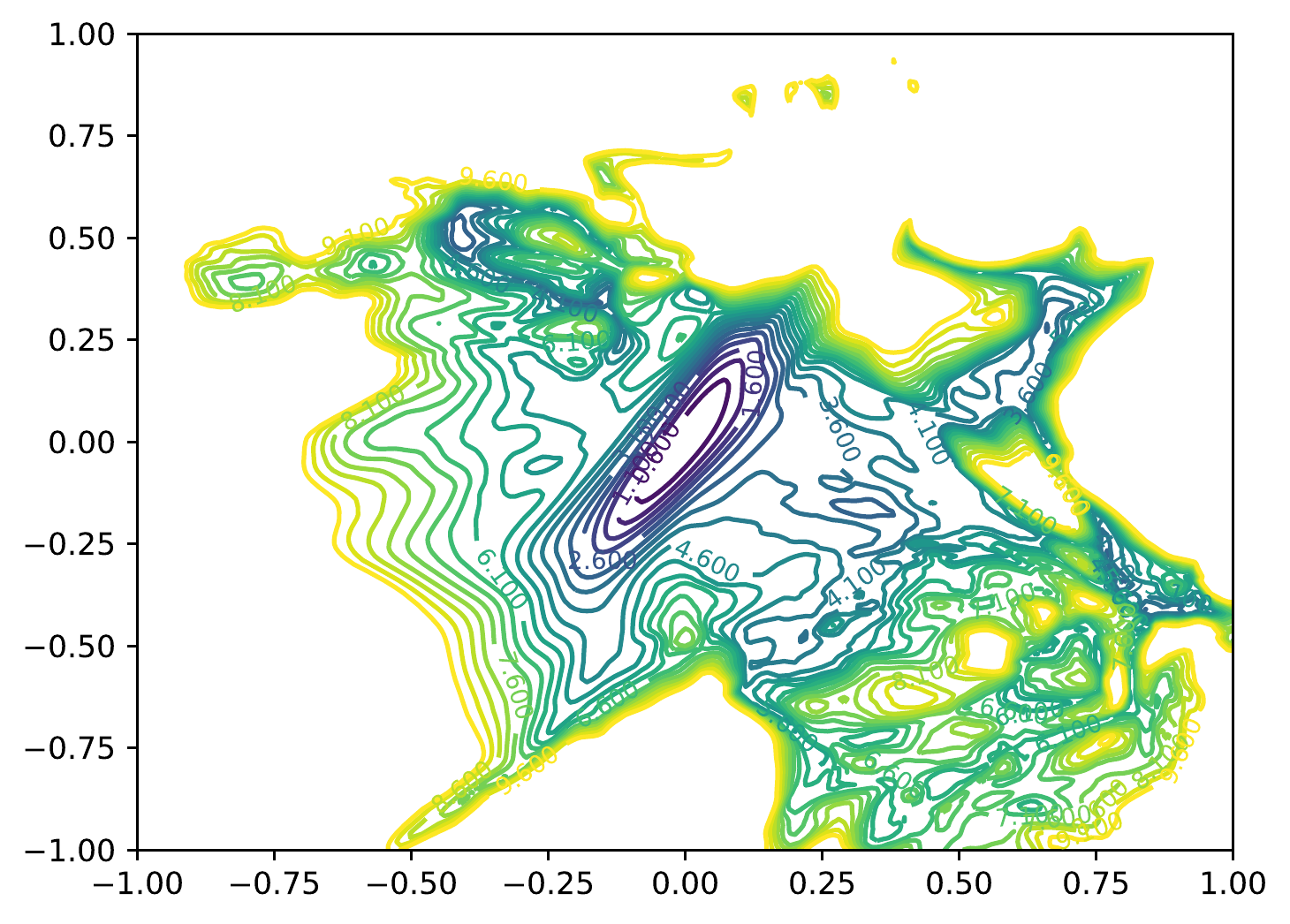}
\includegraphics[width=0.245\linewidth]{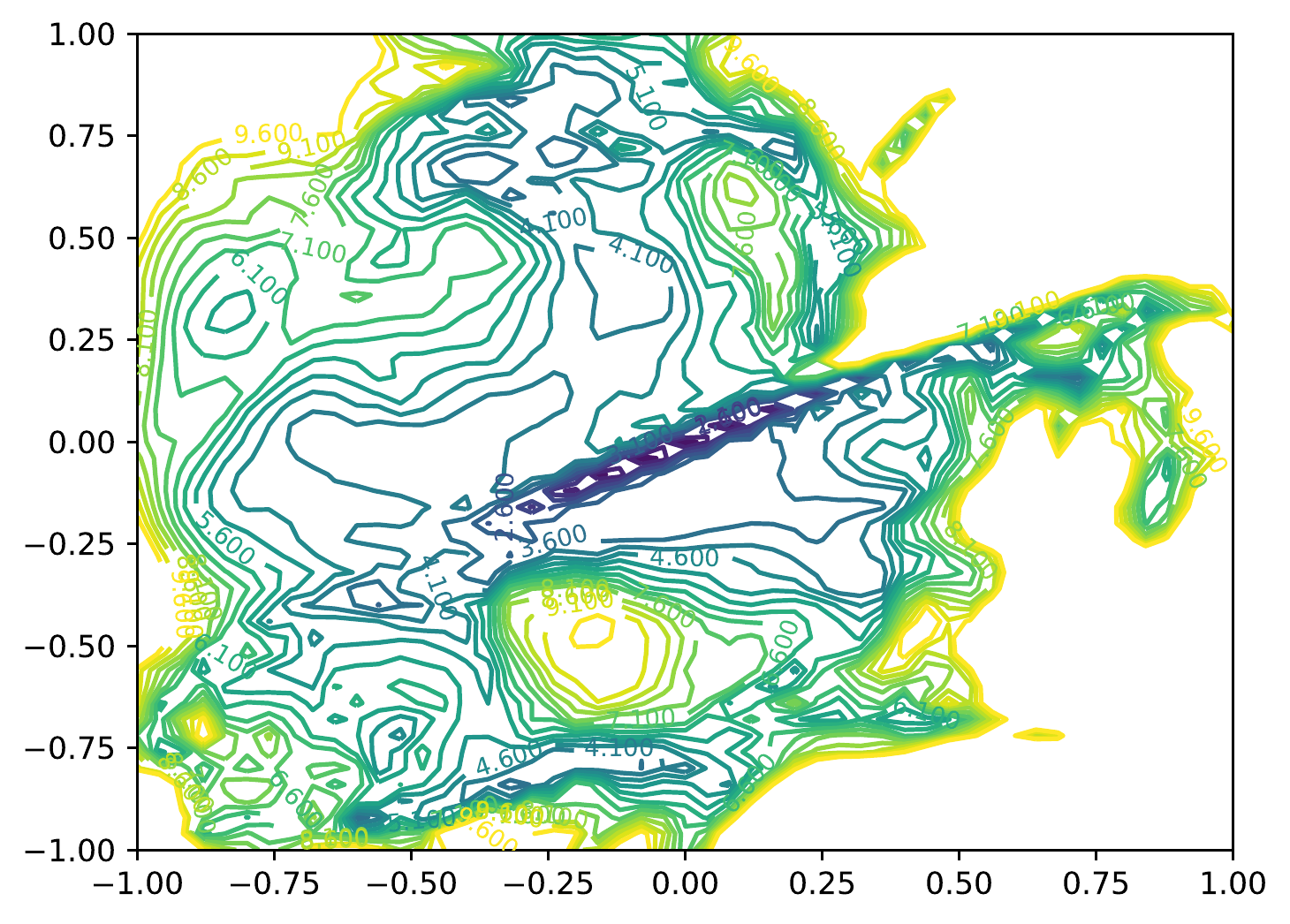}
\includegraphics[width=0.245\linewidth]{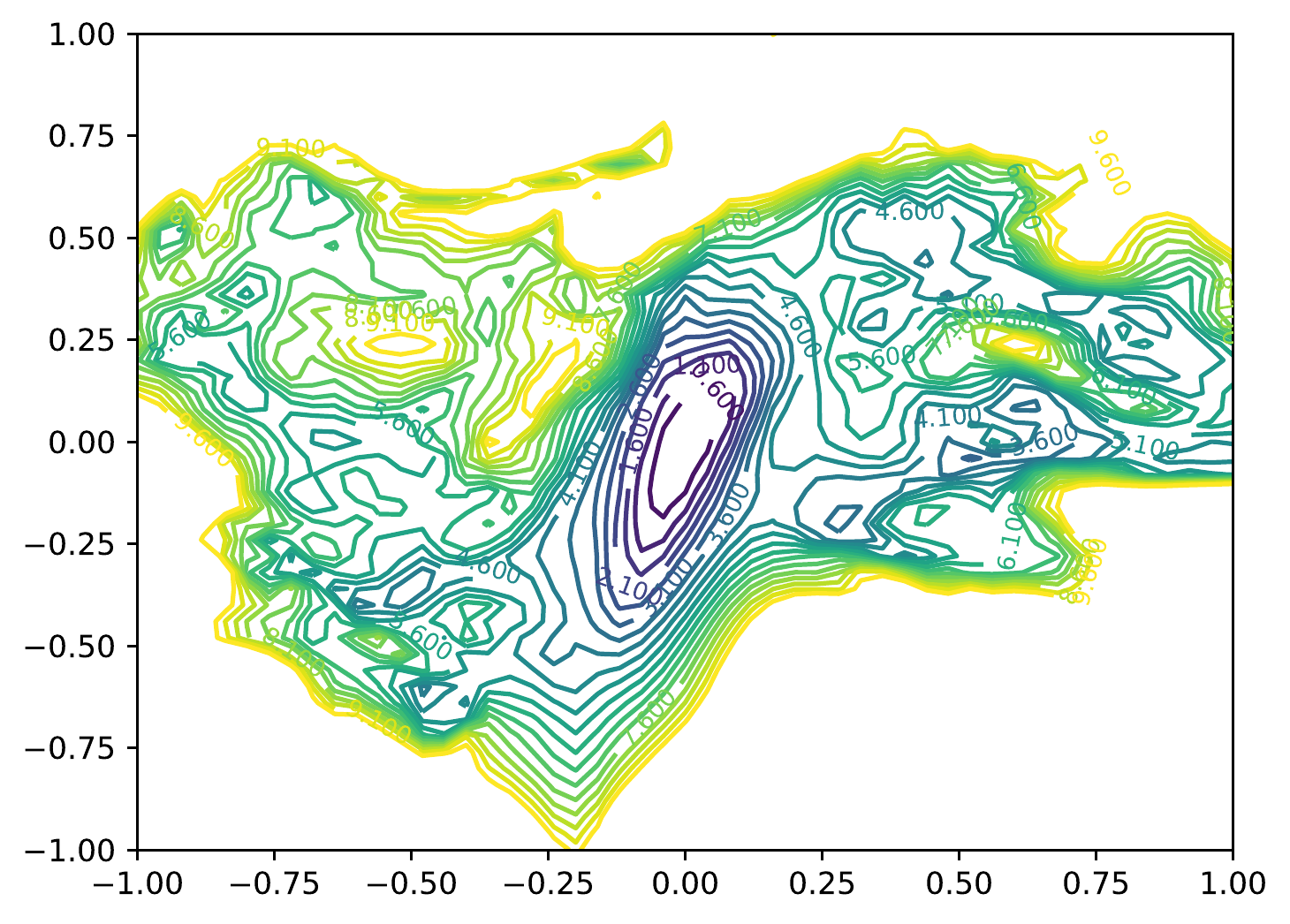}
\end{tabular}
\caption{Repeatability of the 2D surface plots for ResNet-56-noshort. The model is trained with batch size 128, initial learning rate 0.1 and weight decay 5e-4. The final training loss is 0.192, the training error is 6.49 and the test error is 13.31.}
\label{fig:repeatness_2d_resnet56_noshort}
\end{figure*}

Do different random directions produce dramatically different plots?
We plot the 1D loss surface of VGG-9 with 10 random filter-normalized directions.
As shown in Figure~\ref{fig:filter_normalization_repeat}, the plots are very close in shape.
We also repeat the 2D loss surface plots multiple times for ResNet-56-noshort, which has worse generalization error.
As shown in Figure~\ref{fig:repeatness_2d_resnet56_noshort}, there are apparent changes in the loss surface for different plots, however, the qualitative choatic behaviour is quite consistent across plots.

\subsection{Implementation Details}
\paragraph{Computing resources for generating the figures}
Our PyTorch code can be executed in a multiple GPU workstation as well as an HPC with hundreds of GPUs using \texttt{mpi4py}.
The computation time depends on the model's inference speed on the training set, the resolution of the plots, and the number of GPUs.
The resolution for the 1D plots in Figure~\ref{fig:noramlized_shape_batchsize} is 401$\times$401.
The default resolutions used for the 2D contours in Figure~\ref{fig:noramlized_shape_batchsize} and Figure~\ref{fig:depth} is $51\times51$.
We use higher resolutions ($251\times251$) for the ResNet-56-noshort used in Figure~\ref{fig:surfs} to show more details.
For reference, a 2D contour plot of ResNet-56 with a (relatively low) resolution of $51\times51$ will take about 1 hour on a workstation with 4 GPUs (Titan X Pascal or 1080 Ti).

\paragraph{Batch Normalization parameters}
In the 1D linear interpolation methods, the Batch Normalization (BN) parameters including the ``running mean'' and ``running variance'' need to be considered as part of $\theta$. If these parameters are not considered, then it is not possible to reproduce the exact loss values for both minimizers.  In the filter-normalized visualization, the random direction perturbs all weights except batch norm parameters. Note that the filter normalization process removes the effect of weight scaling, and so the batch normalization can be ignored.

\paragraph{The VGG-9 architecture and parameters for Adam}
VGG-9 is a cropped version of VGG-16, which keeps the first 7 Conv layers in VGG-16 with 2 FC layers. A BN layer is added after each conv layer and the first FC layer.
We find VGG-9 is an efficient network with better performance comparing to VGG-16 on CIFAR-10. We use the default values for $\beta_1$, $\beta_2$ and $\epsilon$ in Adam with the same learning rate schedule as used in SGD.

\subsection{Training Curves for VGG-9 and ResNets}
\begin{figure*}[!h]
\centering
\begin{tabular}{l}
\hspace{-5mm}
 \subfigure[SGD, loss values]{\includegraphics[width=0.25\linewidth]{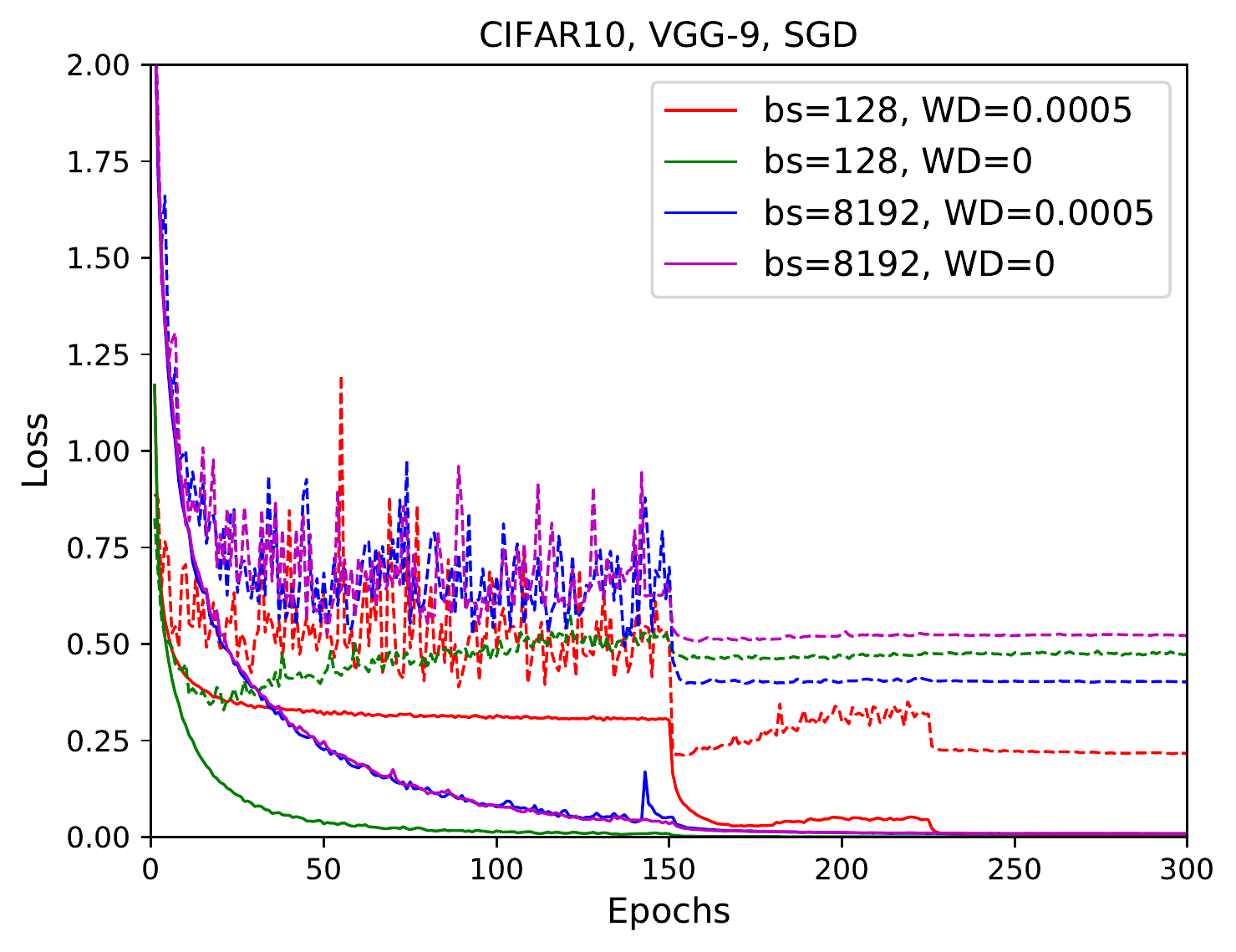}}
 \subfigure[SGD, errors]{\includegraphics[width=0.25\linewidth]{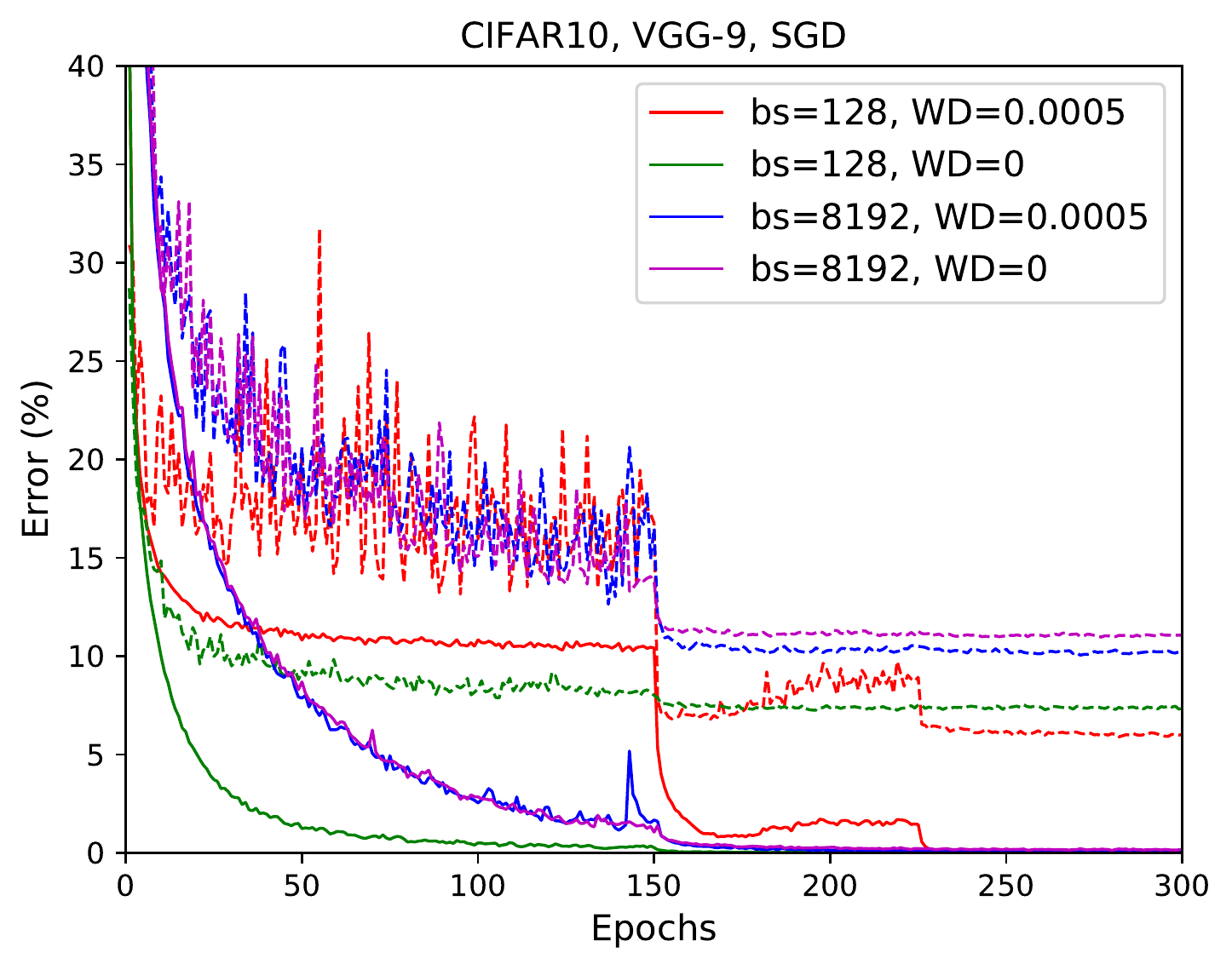}}
 \subfigure[Adam, loss values]{\includegraphics[width=0.25\linewidth]{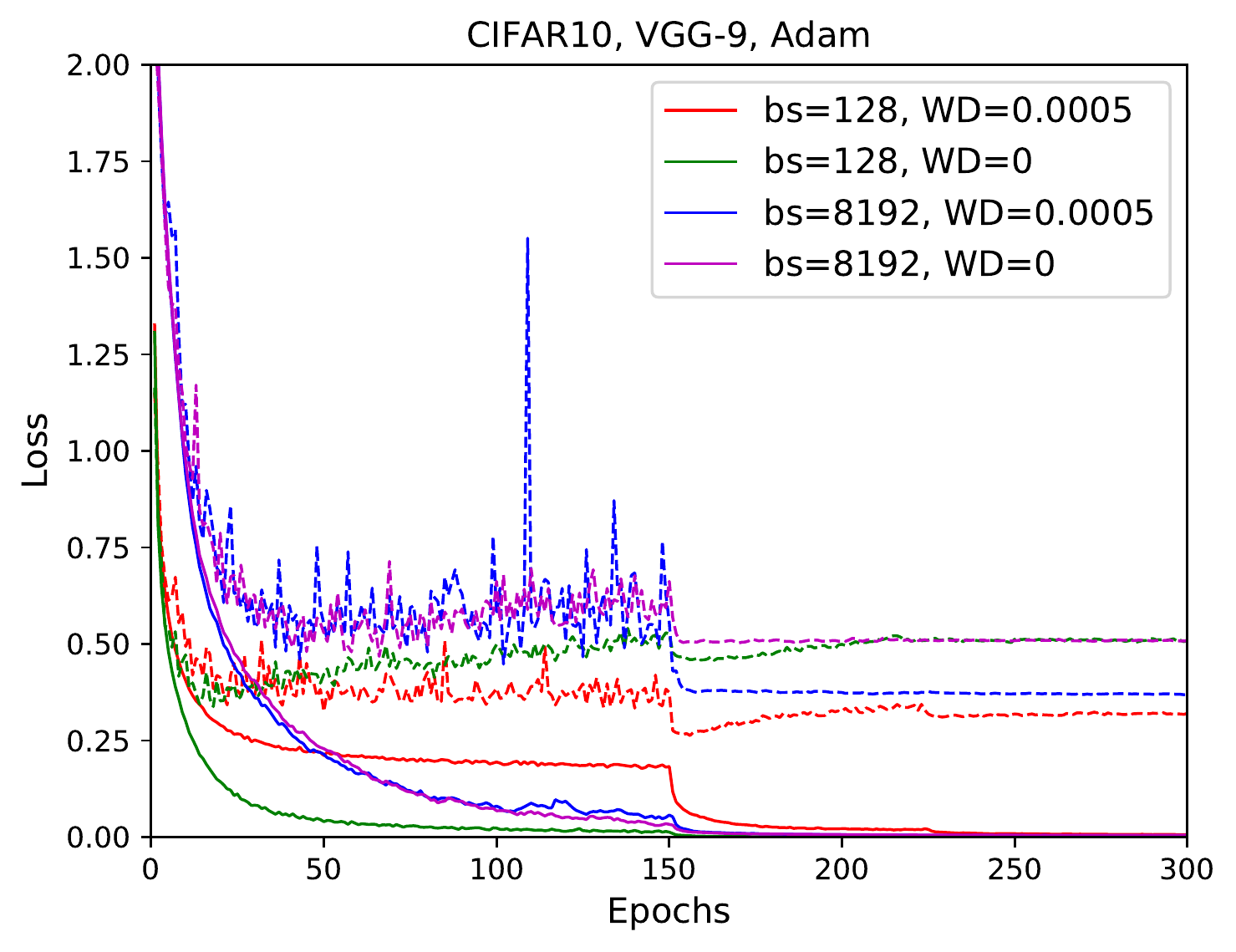}}
 \subfigure[Adam, errors]{\includegraphics[width=0.25\linewidth]{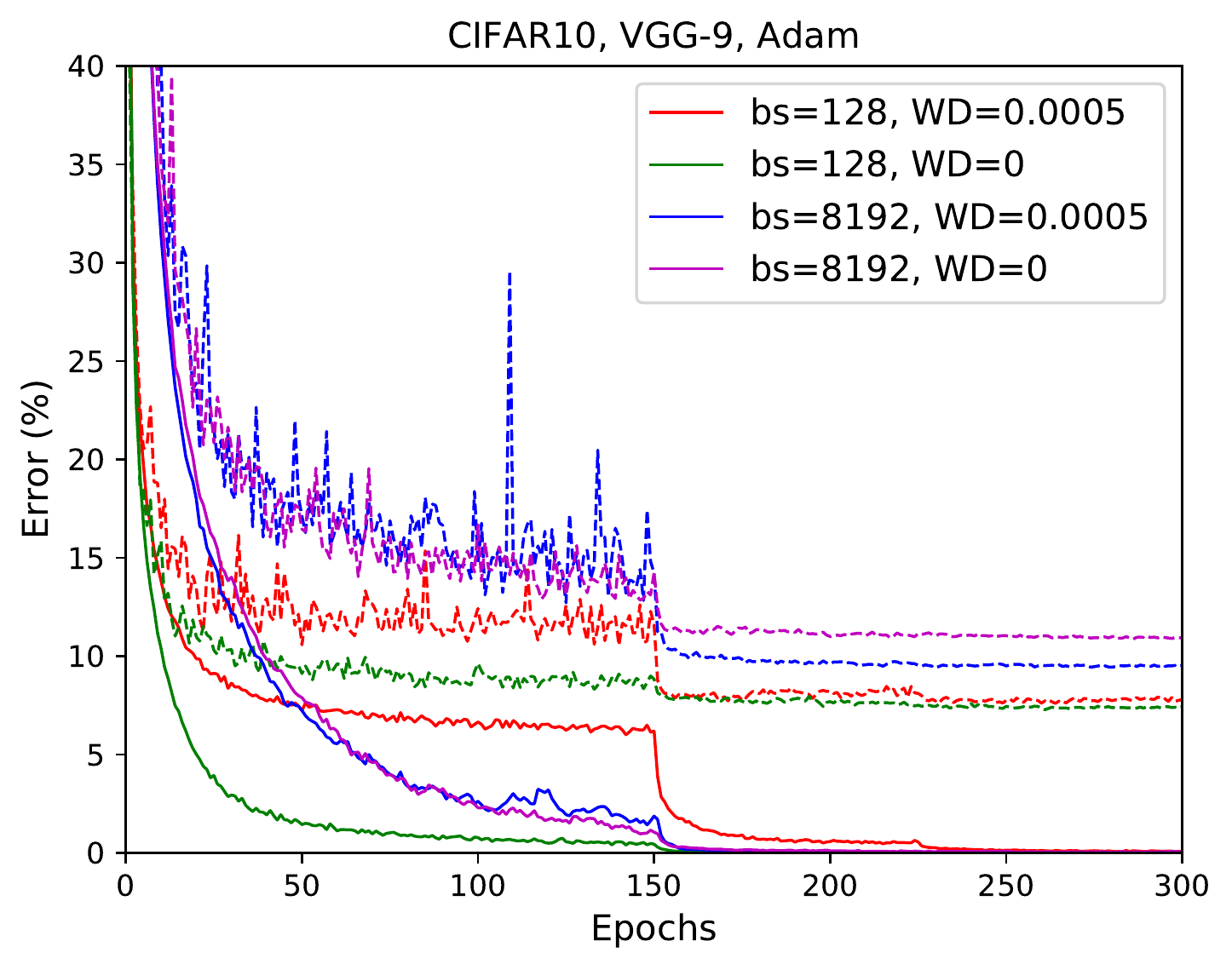}}
\end{tabular}
\caption{Training loss/error curves for VGG-9 with different optimization methods.
Dashed lines are for testing, solid for training.}
\label{fig:loss_curves_vgg9}
\end{figure*}

\begin{figure*}[h]
\centering
\begin{tabular}{l}
\hspace{-5mm}
\subfigure[ResNet-CIFAR]{\includegraphics[width=0.25\linewidth]{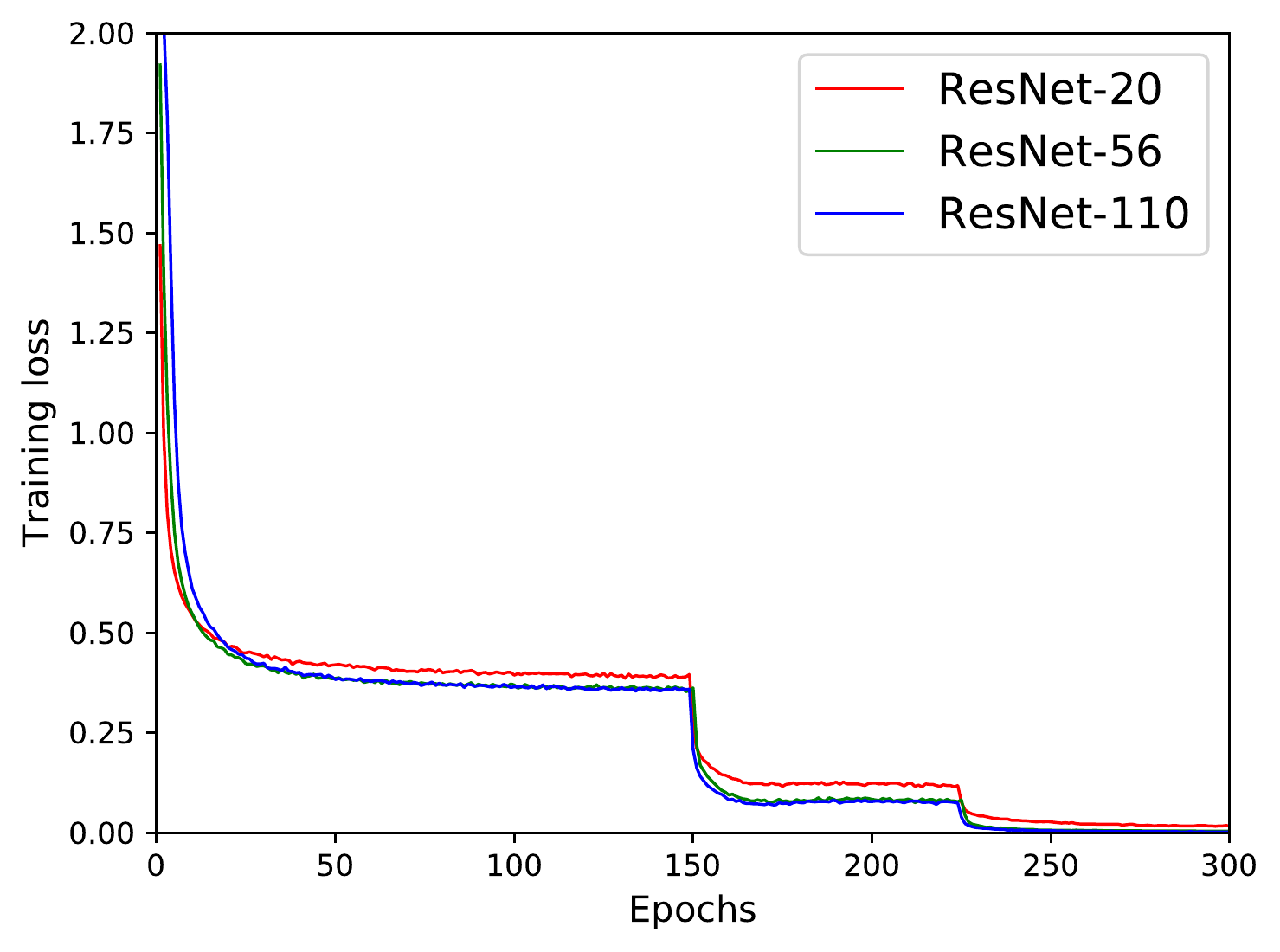}}
\subfigure[ResNet-CIFAR]{\includegraphics[width=0.25\linewidth]{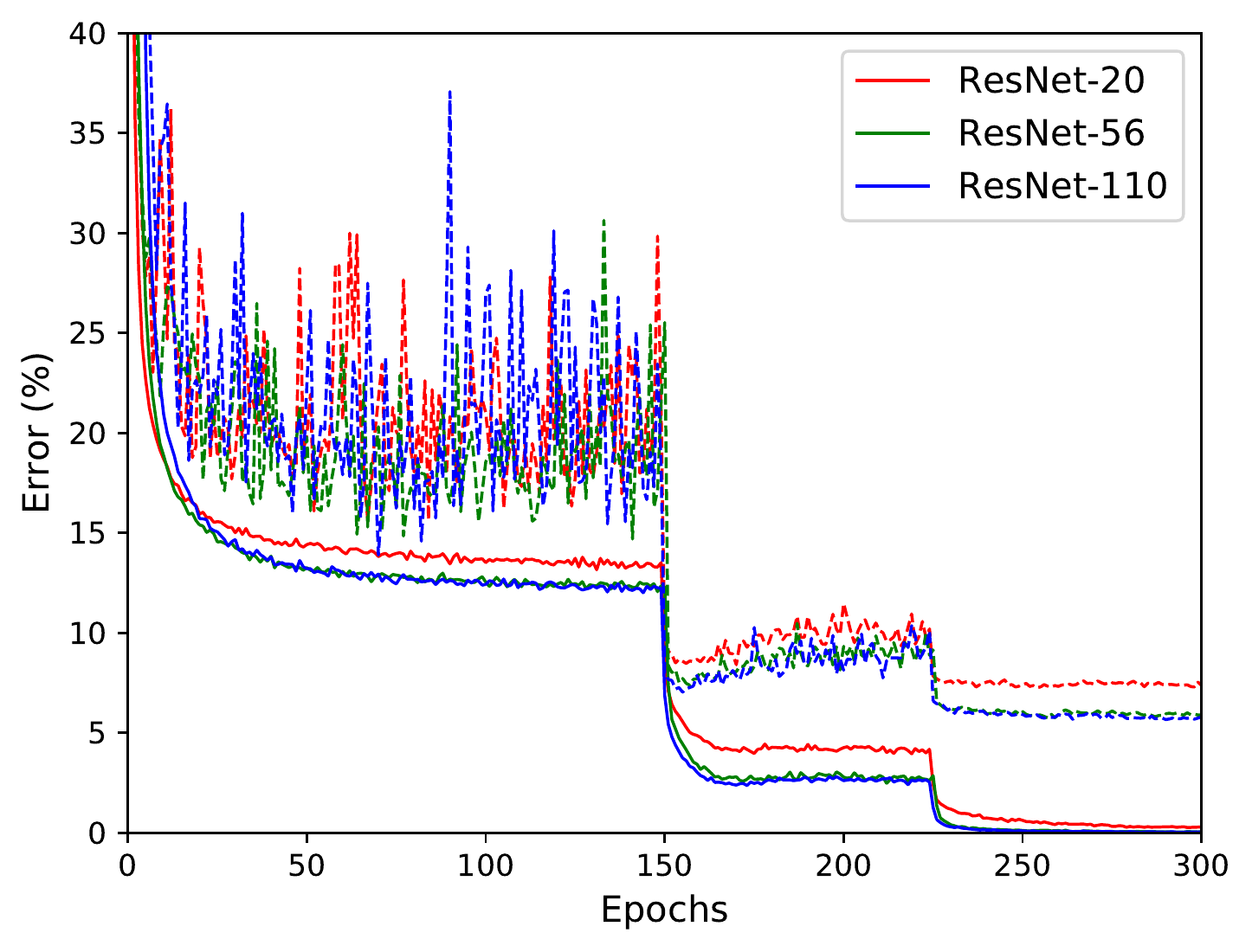}}
\subfigure[ResNet-CIFAR-noshort]{\includegraphics[width=0.25\linewidth]{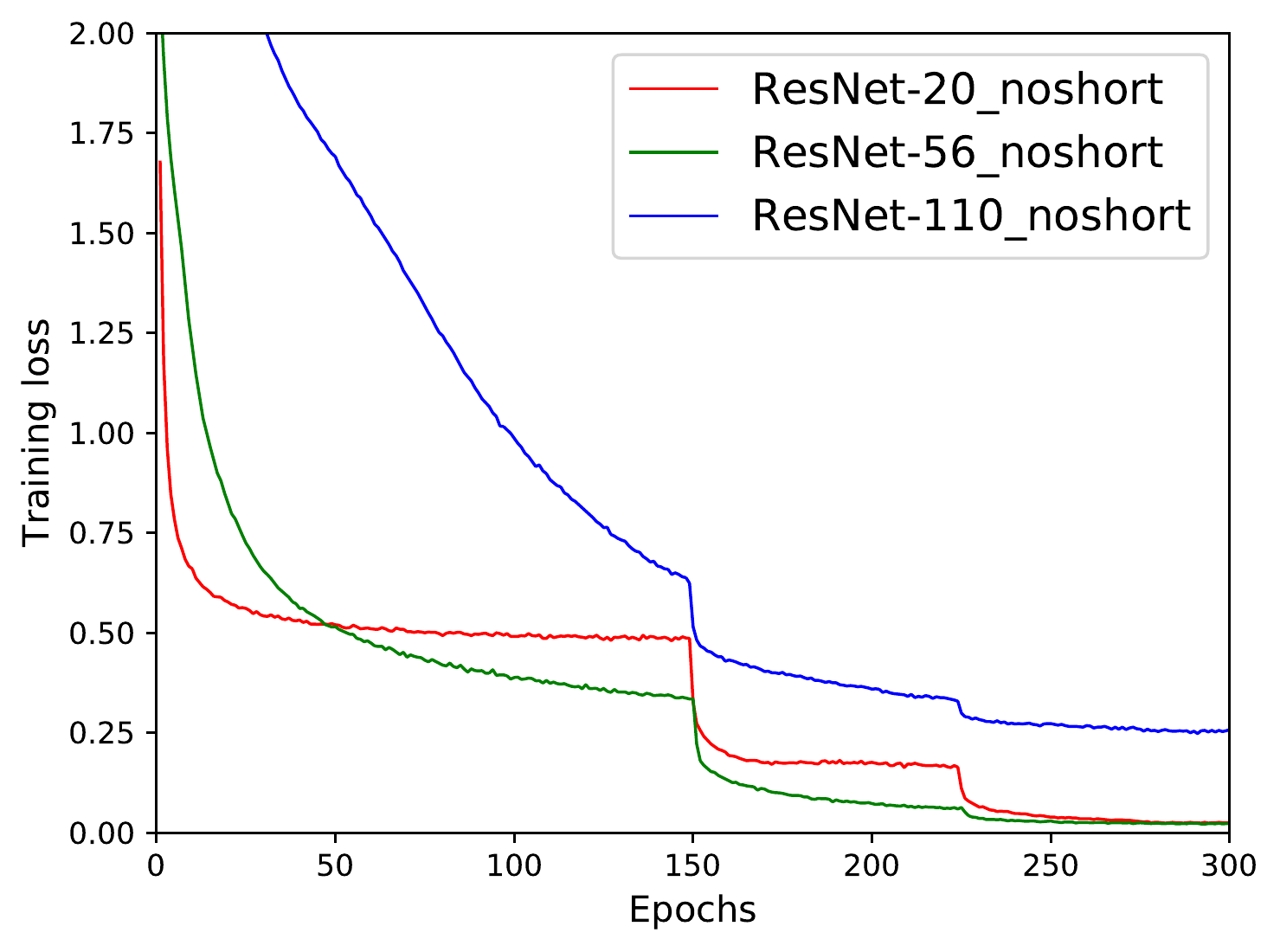}}
\subfigure[ResNet-CIFAR-noshort]{\includegraphics[width=0.25\linewidth]{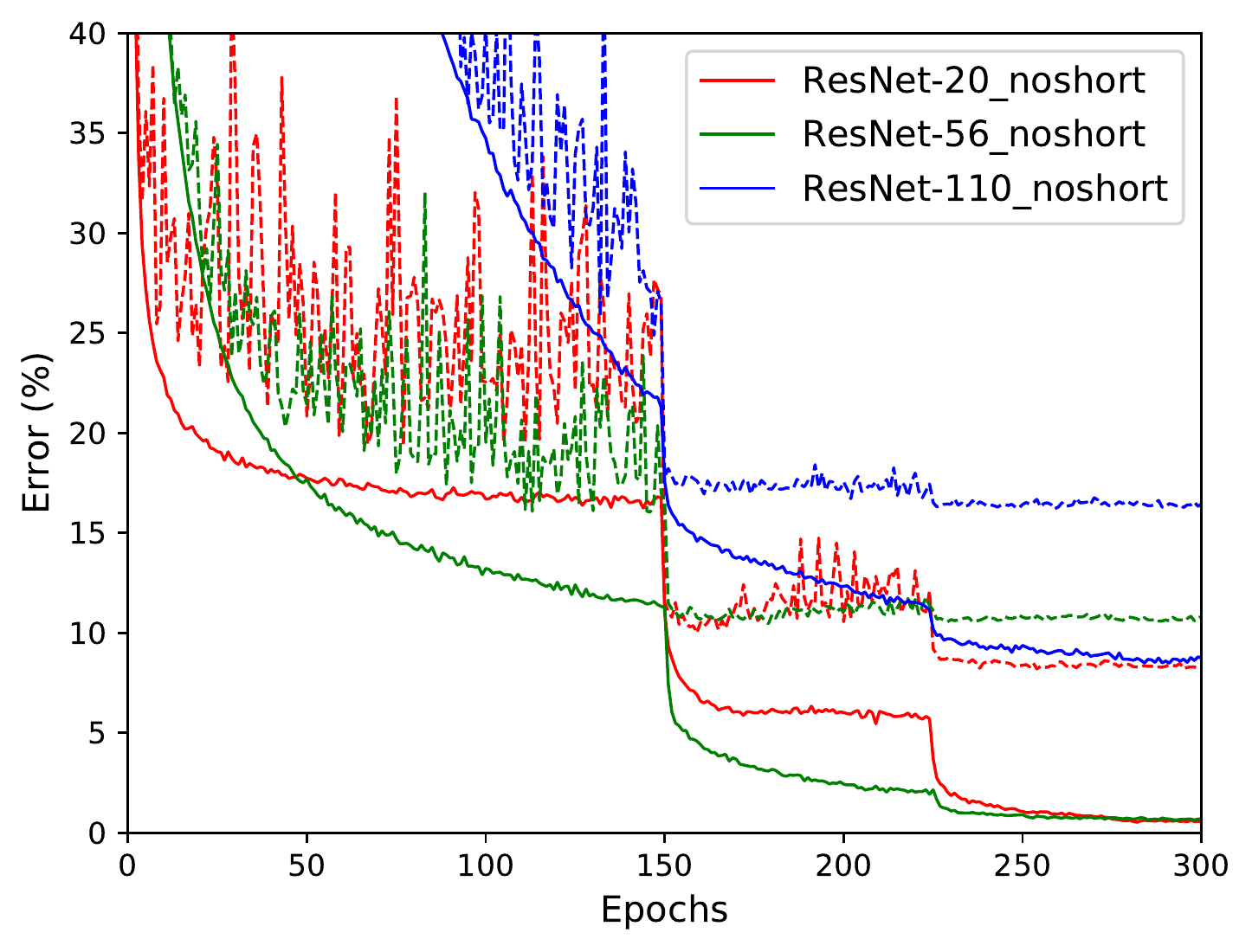}}
\end{tabular}
\caption{Convergence curves for different architectures.}
\label{fig:loss_curves_different architecture}
\end{figure*}

The loss curves for training VGG-9 used in Section~\ref{sec:sharp} are shown in Figure~\ref{fig:loss_curves_vgg9}.
Figure~\ref{fig:loss_curves_different architecture} shows the loss curves and error curves of architectures used in Section~\ref{sec:exp_different_networks} and Table~\ref{tab:architectures} shows the final error and loss values.
The default setting for training is using SGD with Nesterov momentum, batch-size 128, and 0.0005 weight decay for 300 epochs. The default learning rate was initialized at 0.1, and decreased by a factor of 10 at epochs 150, 225 and~275.

\begin{table}[h]
\centering
\caption{Loss values and errors for different architectures trained on CIFAR-10.}
\label{tab:architectures}
\small
\begin{tabular}{lrcccc}
\toprule
                    & init LR  & Training Loss & Training Error & Test Error \\ \midrule
ResNet-20           & 0.1      & 0.017         & 0.286          & ~~7.37          \\
ResNet-20-noshort   & 0.1      & 0.025         & 0.560          & ~~8.18          \\ \hline
ResNet-56           & 0.1      & 0.004         & 0.052          & ~~5.89          \\
ResNet-56-noshort   & 0.1      & 0.192         & 6.494          & 13.31         \\
ResNet-56-noshort   & 0.01     & 0.024         & 0.704          & 10.83         \\\hline
ResNet-110          & 0.1      & 0.002         & 0.042          & ~~5.79          \\
ResNet-110-noshort  & 0.01     & 0.258         & 8.732          & 16.44         \\ 
\bottomrule
\end{tabular}
\end{table}

\end{document}